\def\firstsub{first}%
\def\arxivsub{arxiv}%
\ifx\submissiontype\undefined%
\def\submissiontype{final}%
\fi%
\ifx\submissiontype\firstsub%
\PassOptionsToPackage{twoside}{geometry}
\fi%
\documentclass[titlepage,shield]{hwthesis}
\usepackage{amsthm,graphicx,subfigure,a4wide,exscale,relsize,setspace,lineno}
\usepackage[square,numbers,sort&compress]{natbib}
\usepackage{geometry}
\usepackage[ruled,vlined]{algorithm2e}
\usepackage[nottoc]{tocbibind}
\usepackage[title,titletoc]{appendix}
\usepackage{tikz} % the one true package for figures
\usetikzlibrary{matrix,cd,arrows} % packages commonly used with tikz
\usepackage{mathtools} % who would not want some mathtools
\usepackage{amsmath,amssymb} % for all the mathematical goodness
\usepackage{tensor}   % more options for indices
\usepackage{mathrsfs} % contains fancy math fonts
\usepackage{bbm}      % contains the double-lined letters used for sets (like N) and fields (like R)
\usepackage{booktabs} % pretty tables, see documentation on how to use
\usepackage{pdfpages} % needed to include researchthesissubmission form
\usepackage{enumitem} % more itemize options
\setlist{nosep}        % enumerate-option which reduces the seperation in itemize list (this looks better
\usepackage{lipsum}    % you can delete this. it just gives a textexample for this template				       

\usepackage{imakeidx} % for index
\usepackage{pdfpages}
\usepackage{listings}
\newcommand{\idxKeywordName}{Keyword}

\makeindex[%
name=\idxKeywordName,%
title=\idxKeywordName{} index,
intoc,
]

\lstdefinestyle{myModelSummaryStyle}{
	language=Matlab,
	numbersep=10pt,
	tabsize=2,
	showspaces=false,
	showstringspaces=false
}

\ifx\submissiontype\arxivsub%
\else%
\fi%

\definecolor{MyDarkBlue}{rgb}{0.15,0.25,0.45}

\usepackage{hyperref}
\ifx\submissiontype\arxivsub%
\hypersetup{
hypertexnames=false,
colorlinks=true,
citecolor=MyDarkBlue,
linkcolor=MyDarkBlue,
urlcolor=MyDarkBlue,
pdfauthor={\href{https://jimut123.github.io/}{Jimut Bahan Pal}},                    
pdftitle={Deceiving Computers in Reverse Turing Test through Deep Learning},            % TITLE,
pdfsubject={cs.CV cs.ML},              % AND SUBJECT!
breaklinks=true
}
\else%
\hypersetup{
hypertexnames=false,
colorlinks=true,
citecolor=black,
linkcolor=black,
urlcolor=black,
pdfauthor={\href{https://jimut123.github.io/}{Jimut Bahan Pal}},
pdftitle={Deceiving Computers in Reverse Turing Test through Deep Learning},            % TITLE,
pdfsubject={cs.CV cs.ML},              % AND SUBJECT!
breaklinks=true
}
\fi%

\ifx\submissiontype\arxivsub%
\relax
\else%
\fi%                % the usual command \onehalfspacing uses a weird/its own definition of what one-and-a-half-spacing should mean
\usepackage{macros} % use the macros in macros.sty, optional, edit as convenient; need the abstract-environment inside somewhere however!

\global\hbadness=100000 % use this to temporarily turn off underfull warnings to have a better overview of warnings

\let\oldapp\appendix
\renewcommand{\appendix}{\oldapp\renewcommand*{\chaptername}{\appendixname}}

\begin{document}
\includepdf[scale=1,pages=1]{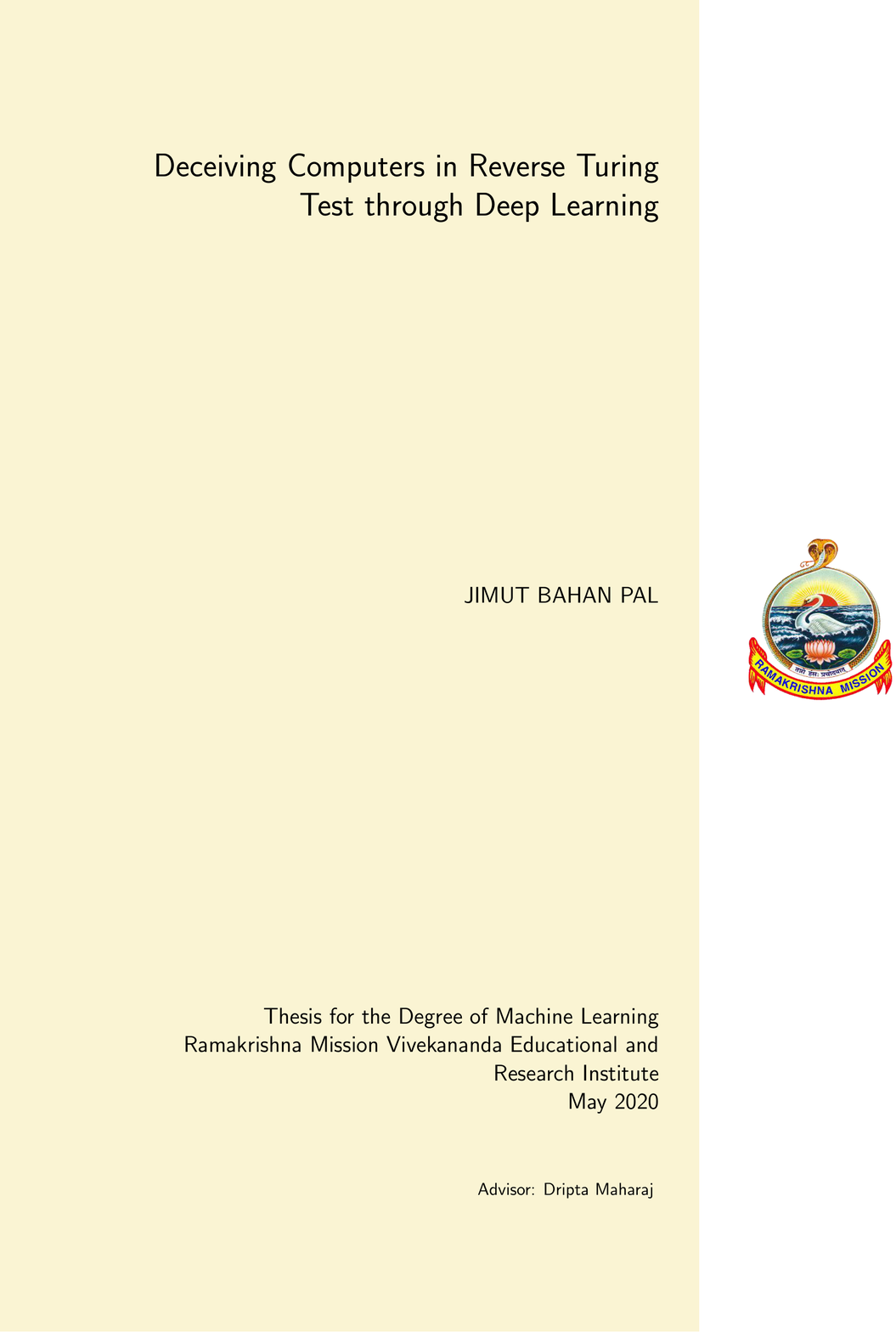}
%%%%%%%%%%%%%%%%%%%%%%%%%%%%%%%%%%%%%%%%%%%%%%%%%%%%%%%%%%%%%%%%%%%%%%%%%
% Make titlepage and top matter
%%%%%%%%%%%%%%%%%%%%%%%%%%%%%%%%%%%%%%%%%%%%%%%%%%%%%%%%%%%%%%%%%%%%%%%%%

\title{Deceiving Computers in Reverse Turing Test through Deep Learning}

%\subtitle{}

\date{\today} % this gives the current date in MONTH, YEAR format - as required by the guidelines

\author{\textbf{\href{https://jimut123.github.io/}{Jimut Bahan Pal}}} % don't forget do change this to your name

\newcommand\supervisor{\textbf{\href{http://www2.eng.ox.ac.uk/civil/efm/people/dripta-sarkar}{Dripta Maharaj}}}

%\newcommand\supervisor{\textbf{Dripta Maharaj}}
%\address{ $^1$Department of Mathematics and the Maxwell Institute for Mathematical Sciences, Heriot-Watt University, Edinburgh, EH14 4AS, Scotland, UK. }

\maketitle
\pagenumbering{gobble}
\ifx\submissiontype\firstsub%
\mbox{}
\fi%

\begin{acknowledgements}
	It is ritual that scholars express their gratitude to their supervisors.  This acknowledgement is very special to me to express my deepest sense of gratitude and pay respect to my supervisor, \supervisor, School of Mathematical Sciences, for his constant encouragement, guidance, supervision, and support throughout the completion of my project. His close scrutiny, constructive criticism, and intellectual insight have immensely helped me in every stage of my work. I would like to thank him for patiently answering my often naive questions related to statistical machine learning. 

I would like to acknowledge the help recieved from Anindya Chowdhury for helping me in data generation and uploading some of the data to public repositories, without which it would have taken much longer time for me to finish this project. I would also like to thank my aunt, Nupur Kundu, who had helped in labelling some of the images from the scrapped JAM dataset. I would also like to mention about the Google Collaboratory platform, which has always been a lab for my experiments, especially this project is completed by the use of their free GPUs.

I would like to thank all the professors of Department of Computer Science for creating an intellectually stimulating environment that enabled me to think more carefully
and methodically about my research than I had ever done before. I would like to thank Swathy Prabhu Maharaj, Tamal Maharaj for the stimulating discussions that he had shared with me during the  Computer Vision course.

I'm  grateful  to  my  father,  Dr.   Jadab  Kumar  Pal,  Deputy  Chief  Executive,  Indian  Statistical Institute, Kolkata for constantly motivating and supporting me to develop this documentation along with the proofreading. I will also mention about my brother Jisnoo Dev Pal and my mother Sumita Pal, for supporting me.

I would like to thank Nilayan Paul for modifying the railway CAPTCHA generator to suit my need. Finally, I acknowledge the help received from all my friends and well-wishers whose constant motivation has promoted the completion of this project.
\end{acknowledgements}

\begin{center}
	\textbf{{\Large Declaration}}
\end{center}
\addcontentsline{toc}{chapter}{\textbf{\normalsize{\emph{Declaration}}}}
\bigskip\bigskip
I hereby certify that the submitted work is my own work, was completed while registered
as a candidate for the degree stated on the Title Page, and I have not obtained a degree elsewhere on the basis of the research presented in this submitted work. Where collaborative research was involved, every effort has been made to indicate this clearly.

\hspace*{15.0mm}
\vspace*{15.0mm}
%\subsubsection*{Supervisor:}
\vspace*{15.0mm}

\hspace*{15.0mm}\begin{minipage}[t]{0.6\textwidth}
	\hrule\medskip
	\textbf{\href{http://www2.eng.ox.ac.uk/civil/efm/people/dripta-sarkar}{Jimut Bahan Pal}},
	  B.Sc.(Hons) \\
	  jpal.cs@gm.rkmvu.ac.in \par
	\today
	
\end{minipage}

\hspace*{12.0cm}
\vspace*{12.0cm}

\begin{center}
	\textbf{{\Large Certification }}
\end{center}
\addcontentsline{toc}{chapter}{\textbf{\normalsize{\emph{Certification}}}}
\bigskip\bigskip
This thesis titled, \textbf{Deceiving computers in Reverse Turing Test through Deep Learning}, submitted by \textbf{\href{https://jimut123.github.io/}{Jimut Bahan Pal}}
as mentioned below has been accepted as satisfactory in
partial fulfillment of the requirements for the degree M.Sc.\ in
Computer Science in \today.

\hspace*{15.0mm}
\vspace*{15.0mm}
\subsubsection*{Supervisor:}
\vspace*{15.0mm}

\hspace*{15.0mm}\begin{minipage}[t]{0.6\textwidth}
	\hrule\medskip
	\textbf{\href{http://www2.eng.ox.ac.uk/civil/efm/people/dripta-sarkar}{Dripta Maharaj}}
	\par  Ph.D from University College Dublin \par
	School of Mathematical Sciences \par 
	Ramakrishna Mission Vivekananda Educational and Research Institute.
	
\end{minipage}

\vspace{\stretch{2}}

\begin{abstract}
It is increasingly becoming difficult for human beings to work on their day to day life without going through the process of reverse Turing test, where the Computers tests the users to be humans or not. Almost every website and service providers today have the process of checking whether their website is being crawled or not by automated bots which could extract valuable information from their site. In the process the bots are getting more intelligent by the use of Deep Learning techniques to decipher those tests and gain unwanted automated access to data while create nuisance by posting spam. Humans spend a considerable amount of time almost every day when trying to decipher CAPTCHAs. The aim of this investigation is to check whether the use of a subset of commonly used CAPTCHAs, known as the text CAPTCHA is a reliable process for verifying their human customers. We mainly focused  on the preprocessing step for every CAPTCHA which converts them in binary intensity and removes the confusion as much as possible and developed various models to correctly label as many CAPTCHAs as possible. We also suggested some ways to improve the process of verifying the humans which makes it easy for humans to solve the existing CAPTCHAs and difficult for bots to do the same.
\end{abstract}
\ifx\submissiontype\firstsub%
\mbox{}
\fi%

\clearpage
\vspace*{12cm}
\hfil
\hspace{2.5cm}
\textit{Dedicated to my parents Dr. Jadab Kr. Pal and Sumita Pal.}
\clearpage
\ifx\submissiontype\firstsub%
\mbox{}
\clearpage
\fi%

\ifx\submissiontype\firstsub%
\mbox{}
\fi%

\ifx\submissiontype\arxivsub%
\relax
\else%
%\includepdf[pages={1}]{researchthesissubmission.pdf}
\ifx\submissiontype\firstsub%
\mbox{}
\fi%
\fi%

\clearpage
\setcounter{page}{1}
\pagenumbering{roman}

\tableofcontents

\listoffigures  % Write out the List of Figures

\listoftables  % Write out the List of Tables

%\listoftables % optional
%\listoffigures % optional

%%%%%%%%%%%%%%%%%%%%%%%%%%%%%%%%%%%%%%%%%%%%%%%%%%%%%%%%%%%%%%%%%%%%%%%%%
% Switch to arabic numbering and input content
%%%%%%%%%%%%%%%%%%%%%%%%%%%%%%%%%%%%%%%%%%%%%%%%%%%%%%%%%%%%%%%%%%%%%%%%%
\clearpage
\pagenumbering{arabic}
\chapter{Introduction}

In a reverse turing test ~\cite{wiki_revtt} the roles between computers and humans has been reversed. Alan Turing ~\cite{10.1093/mind/LIX.236.433,wiki_tt} in 1950, developed the Turing test to tell computers and humans apart via imitation game \index[\idxKeywordName]{imitation game}. The rule of the imitation game is simple, it is played with three people, a man (A), a woman (B) and an interrogator (C). The interrogator stays in a room apart from the other two and the objective of the interrogator is to determine which of the two is man and which is a woman. He may consider them as labels X and Y. The interrogator is allowed to put questions to both of A and B and will get answer in text, which will not introduce any bias towards determining the objective of the interrogator. Turing predicted about 70 years ago that it will be possible to program computers with storage capacity of about 10\textsuperscript{9} states to make them play this imitation game so well that an average interrogator will not have more than 70 percent chance of making the right identification after five minutes of questioning, and here we are in 2020 where we cannot distinguish between a bot and a person in online forums and gaming sites!

\begin{figure}
	\centering
	\fbox{\includegraphics[height=2.5cm,width=0.31\linewidth]{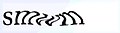}}\vspace{2px}
	\fbox{\includegraphics[height=2.5cm,width=0.31\linewidth]{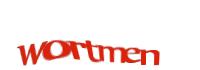}}\vspace{2px}
	\fbox{\includegraphics[height=2.5cm,width=0.31\linewidth]{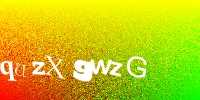}}\vspace{2px}
	\fbox{\includegraphics[height=2.5cm,width=0.31\linewidth]{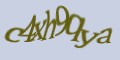}}\vspace{2px}
	\fbox{\includegraphics[height=2.5cm,width=0.31\linewidth]{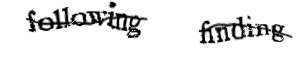}}\vspace{2px}
	\fbox{\includegraphics[height=2.5cm,width=0.31\linewidth]{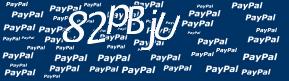}}
	\vspace{2px}
	\fbox{\includegraphics[height=16cm,width=.990\linewidth]{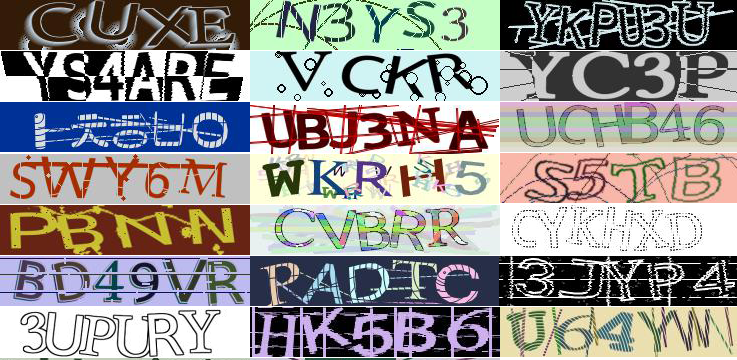}}\vspace{2px}
	\caption{Examples of text \textbf{CAPTCHAs} increasing in difficulty both for Humans and Computers. Image collected from (starting from top) \href{https://en.wikipedia.org/wiki/File:Captcha.jpg}{Wikipedia}, \href{http://jcaptcha.sourceforge.net/}{JCAPTCHA}, \href{http://jcaptcha.sourceforge.net/}{JCAPTCHA},
		\href{http://www.pixprofit.com/}{Pixprofit},
		\href{https://en.wikipedia.org/wiki/File:Modern-captcha.jpg}{Wikipedia},
		~\href{https://registration.paypal.com/\#ProcessorSelectionNode}{PayPal}, 
		and ~\href{http://knpbundles.com/captcha-com/symfony-captcha-bundle}{Knpbundles}~post.}
	\label{text_CAPTCHAS}
\end{figure}

Reverse Turing test deals with computer (here server) acting as an interrogator and determining whether the user is a bot (computer) or a human. Here the computer
is said to pass the Turing test if it always correctly recognizes a human as human and a computer as itself. In today's world there is a necessity to prevent bots from spamming, creating multiple account and jamming the resource of a particular website. There are various ways through which the reverse Turing Test is implemented by websites and service providers to stop illegitimate access of their contents by hackers and spammers. 

\section{CAPTCHA}

One method for performing reverse Turing test is through CAPTCHA \index[\idxKeywordName]{CAPTCHA} which is an acronym for \textbf{C}ompletely \textbf{A}utomated \textbf{P}ublic \textbf{T}uring test to tell \textbf{C}omputers and \textbf{H}umans \textbf{A}part. They are designed to protect online sites from bot attacks which may create automated fake profiles to scam people. CAPTCHAs are designed in such a way that they can be solved by humans easily but not by a computer. This user identification procedure has received criticism from people with disabilities and also those people who thinks their everyday work is being slowed down by solving a typical CAPTCHA for 10 seconds.

\begin{figure} 
	\centering
	\fbox{\includegraphics[height=3.4cm,width=0.479\linewidth]{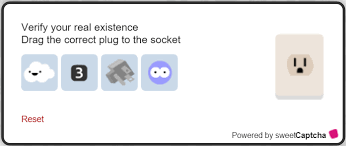}}	\vspace{2px}
	\fbox{\includegraphics[height=3.4cm,width=0.479\linewidth]{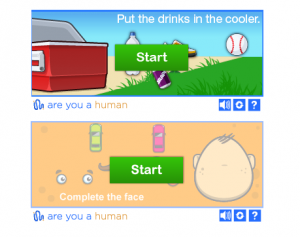}}	\vspace{2px}
	\fbox{\includegraphics[height=3.4cm,width=0.479\linewidth]{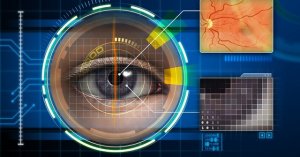}} \vspace{2px}
	\fbox{\includegraphics[height=3.4cm,width=0.479\linewidth]{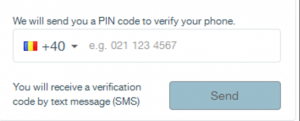}} \vspace{2px}
	\fbox{\includegraphics[height=3cm,width=0.4\linewidth]{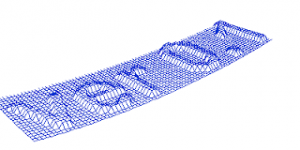}}\fbox{\includegraphics[height=3cm,width=0.16\linewidth]{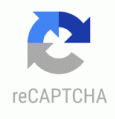}}\fbox{\includegraphics[height=3cm,width=0.4\linewidth]{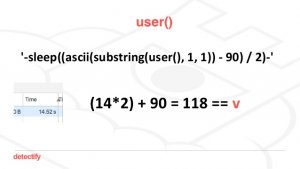}}\vspace{2px}
	\fbox{\includegraphics[height=8cm,width=.475\linewidth]{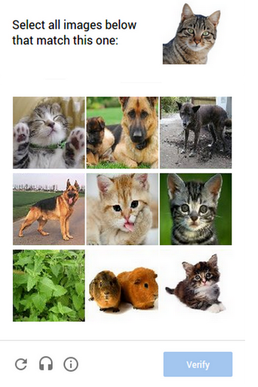}}
	\fbox{\includegraphics[width=0.475\linewidth]{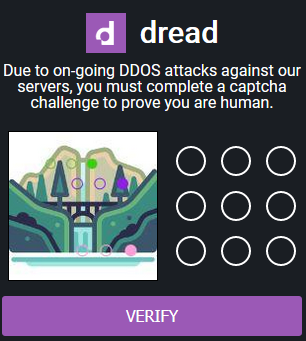}}\vspace{2px}
	\caption{Types of CAPTCHAS. (Picture Courtesey: \href{https://www.iihglobal.com/blog/mostly-used-captcha-examples/}{IIHglobal}and \href{https://en.wikipedia.org/wiki/File:Images_Recaptcha.png}{Wikipedia}).}
	\label{CAPTCHA_types1}
\end{figure}

There are different types of CAPTCHAs, a subset of text CAPTCHAs is shown in Figure ~\ref{text_CAPTCHAS}. We see that the text CAPTCHAs are getting difficult to solve by humans, while the Hackers comes up with more powerful CAPTCHA breaker every day. Automated scripts ~\cite{10.1007/11427896_1} should not be more successful than 1 in 10,000 (0.01\%) for any type of attack, while the human's success rate should approach 90\%. This ensures that humans have to endure less pain in solving CAPTCHAs.

\section{Types of CAPTCHAs}

\begin{figure}
	\centering   	
	\fbox{\includegraphics[height=3.7cm,width=0.475\linewidth]{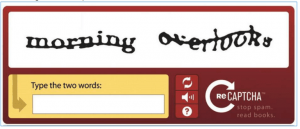}}\vspace{2px}
	\fbox{\includegraphics[height=3.7cm,width=0.475\linewidth]{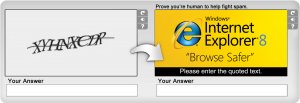}}\vspace{2px}
	\fbox{\includegraphics[height=3.5cm,width=0.50\linewidth]{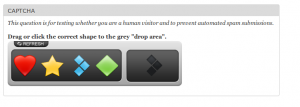}}\vspace{2px}
	\fbox{\includegraphics[height=3.5cm,width=0.46\linewidth]{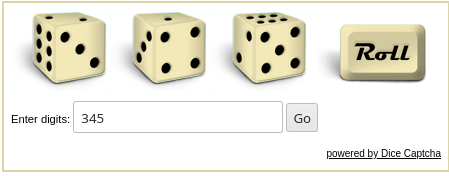}}\vspace{2px}
	\fbox{\includegraphics[height=5.5cm,width=0.475\linewidth]{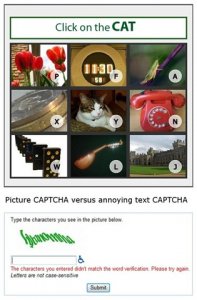}}\vspace{2px}
	\fbox{\includegraphics[height=5.5cm,width=0.475\linewidth]{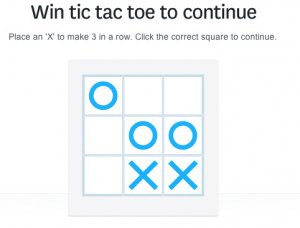}}\vspace{2px}
	\caption{Types of CAPTCHAS. (Picture Courtesey: \href{https://www.iihglobal.com/blog/mostly-used-captcha-examples/}{IIHglobal}, \href{http://dice-captcha.com/demo-dice-captcha.php}, and {Dice CAPTCHA})}
	\label{CAPTCHA_types2}
\end{figure}

There are different types of CAPTCHAs as shown in Figure ~\ref{CAPTCHA_types1} and Figure ~\ref{CAPTCHA_types2}. Sweet CAPTCHA makes you match easy images instead of reading twisted texts. Playthru CAPTCHA replaces the pain of reading texts via fun microgames, where humans get through easily while automated \index[\idxKeywordName]{bots} bots are kept out. Biometric test is the next generation CAPTCHA, that comes with facial and speech recognition based on knowledge set. Text message verification helps the site to verify the actual mobile number of users instead of traditional User IDs and Passwords. Image based CAPTCHA include confident CAPTCHA and Picture identification CAPTCHA which the user needs to recognize and select a set of images which is easy for humans but difficult for bots. Google's reCAPTCHA helps sites to be protected from spambots. There is another type of CAPTCHA which lets users to solve word problems and math problems. Time based CAPTCHA analyses the time required by the user to fill forms and from the analysis it can predict whether the user is a bot or human. There is 3D CAPTCHA which are difficult for bots to decipher but easy for humans. Add based CAPTCHA provides some extra revenue to the site by providing add based text to decipher. There is drag and drop CAPTCHA which is an intuitive way of verifying different humanity tasks which are difficult for robots. Tic-Tac-Toe CAPTCHAs helps to entertain the user by playing mini games.

\begin{figure}
	\centering   	
	\fbox{\includegraphics[height=3.7cm,width=0.485\linewidth]{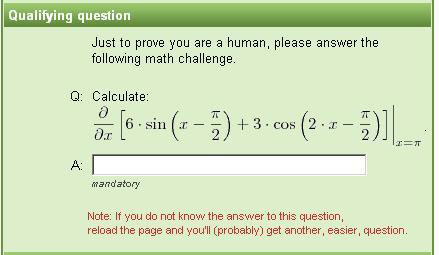}}\vspace{2px}
	\fbox{\includegraphics[height=3.7cm,width=0.475\linewidth]{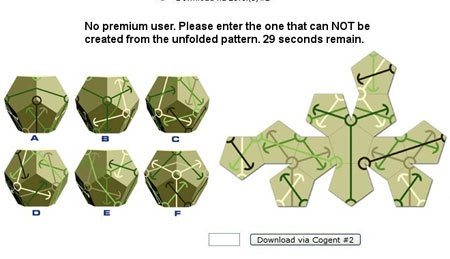}}\vspace{2px}
	\fbox{\includegraphics[height=3.5cm,width=0.50\linewidth]{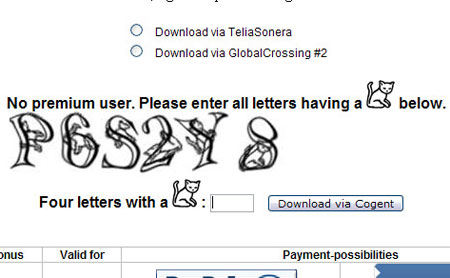}}\vspace{2px}
	\fbox{\includegraphics[height=3.5cm,width=0.46\linewidth]{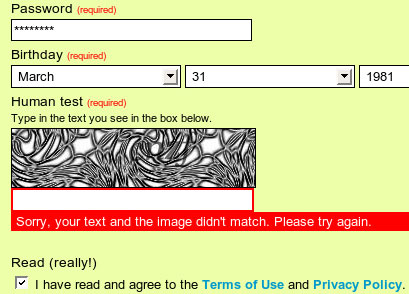}}\vspace{2px}
	\fbox{\includegraphics[height=3.5cm,width=0.475\linewidth]{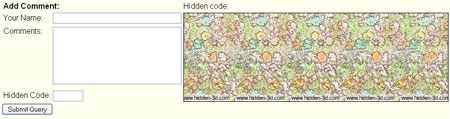}}\vspace{2px}
	\fbox{\includegraphics[height=3.5cm,width=0.475\linewidth]{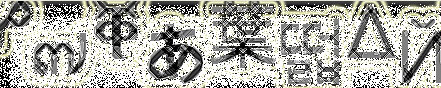}}\vspace{2px}
	
	\fbox{\includegraphics[height=12cm,width=.975\linewidth]{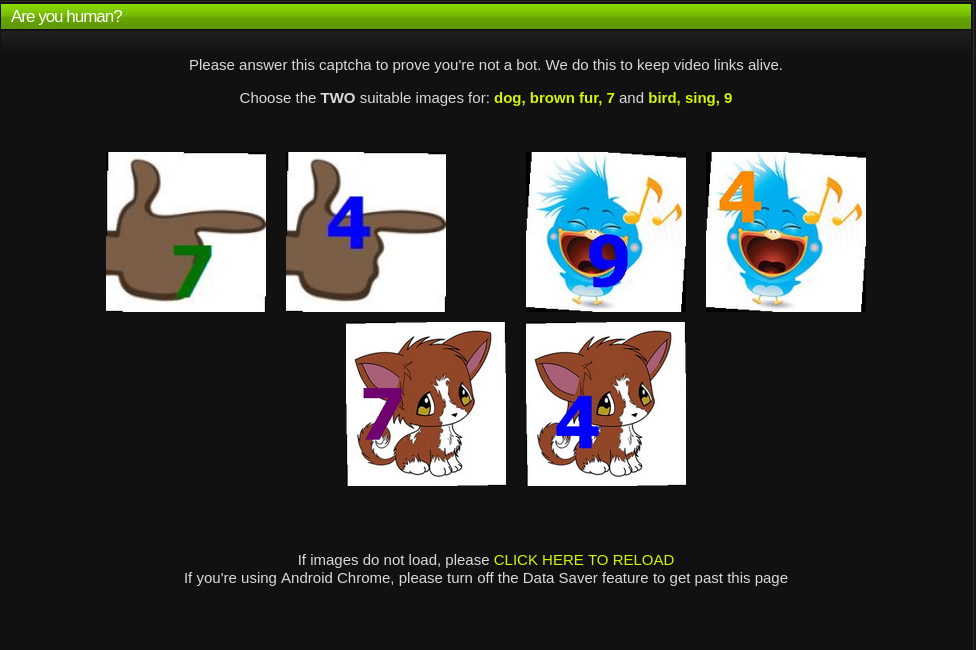}}\vspace{2px}
	
	\caption{Types of CAPTCHAS. (Picture Courtesey: \href{http://www.seosmarty.com/impossible-captcha-it-doesnt-really-matter-if-you-are-human-or-not/}{Seosmarty}, and \href{http://kissanime.ru}{Anime})}
	\label{CAPTCHA_types_HARD}
\end{figure}

CAPTCHAs can even be a little harder to do when there is a time limit to solve CAPTCHA and it may be even difficult for humans. The examples of such type of CAPTCHAs are shown in Figure ~\ref{CAPTCHA_types_HARD}. The first picture has some sort of partial differential equation solver, which is completely unnecessary, and is not designed for naive users. The second figure contains some really sophisticated paper folding problem which also needs some amount of time to solve. CAPTCHAs may be clumsy and undecipherable as shown in the corresponding figures. The second last figure's CAPTCHA is formed from different languages, and would probably be suitable for Cipher Cracking Competitions. The last CAPTCHA is not so difficult for humans, but instead it needs to be solved very quickly and needs a lot of retry to get to the actual website.

\section{Our Work}

From the above discussions we find that there are various types of CAPTCHAs present to stop the nuisance of bots. We have seen that the ways to develop CAPTCHAs are increasing in complexity day by day. This leaves us to one question, is it worth it? We have worked in breaking of a certain subset of CAPTCHAs i.e., text-based image CAPTCHAs. The text-based CAPTCHAs are the backbone of \index[\idxKeywordName]{Optical Character Recognition (OCR)} Optical Character Recognition (OCR) task. These CAPTCHAs are modified day to day in such a way that it becomes hard for humans to understand and much more difficult for computers to decipher. We have tried some of the difficult CAPTCHAs and benchmarked those results, starting from the naive CAPTCHAs. We have also suggested what kind of technologies needs to be introduced to prevent the unauthorized access by bots.

\chapter{About the data}

We have used some of the available open Kaggle data. Now we will discuss about the dataset used in this investigation.

\section{Captcha (4 letter)}

\begin{figure}
	\centering
	\fbox{\includegraphics[height=2cm,width=0.479\linewidth]{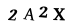}}	
	\fbox{\includegraphics[height=2cm,width=0.479\linewidth]{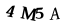}}	
	\caption{Captcha 4 letter dataset available from Kaggle.}
	\label{captcha_4_letter}
\end{figure}

\begin{figure}
	\centering
	\fbox{\includegraphics[height=7cm,width=1\linewidth]{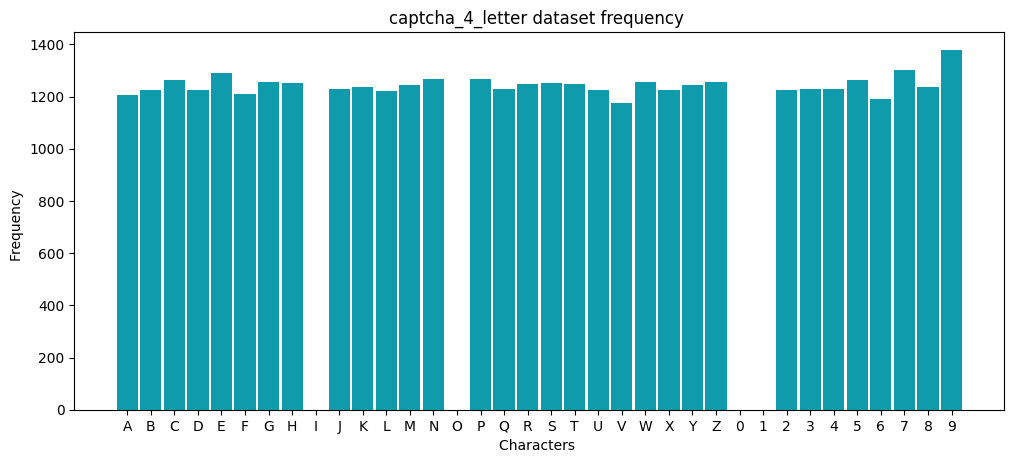}}
	\caption{The initial distribution of the letters for the Captcha (4 letter) dataset.}
	\label{c4l_dist}
\end{figure}

\begin{figure}
	\centering
	\fbox{\includegraphics[height=7cm,width=1\linewidth]{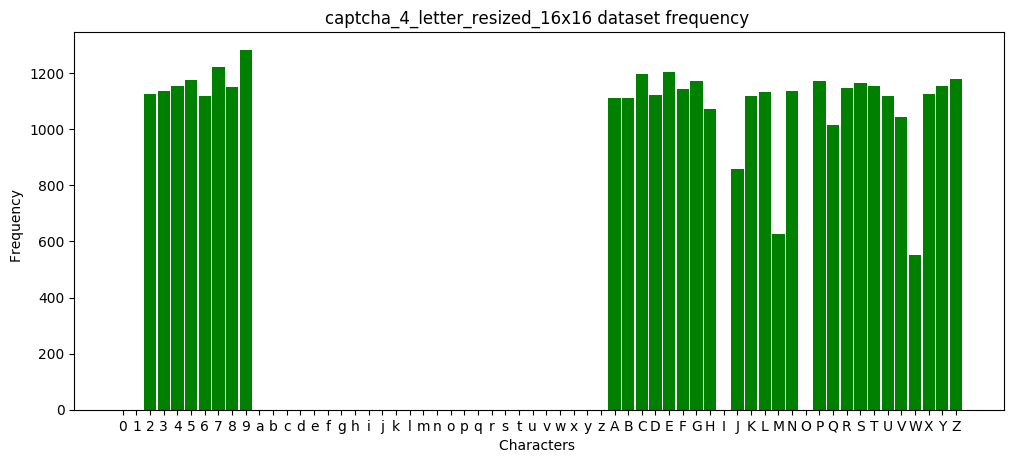}}
	\caption{Separable characters and their frequencies from Captcha 4 letter dataset. }
	\label{c4l_dataset}
\end{figure}

This dataset can be found online on Kaggle (\url{https://www.kaggle.com/genesis16/captcha-4-letter}) and the cleaned version can be found here (\url{https://jimut123.github.io/blogs/CAPTCHA/data/captcha_4_letter.tar.gz}). This dataset contains about 9955 PNG images of 4 letter CAPTCHA, some of which are closely connected as shown in Figure \ref{captcha_4_letter}. The total size of the data is 10.4 M.B.
The images have 3 channels, i.e., they have a Red Green and Blue component, and are of dimension 72x24. We have extracted the separable characters to form a simple dataset containing 32 characters. There were many such characters which were not separable and we have discarded such characters. From Figure \ref{c4l_dataset} we can see that digits 0,1,I and O are missing from this CAPTCHA data. The original distribution of the characters of the dataset is shown in Figure \ref{c4l_dist}.

We have performed various algorithm on this data and created a modified dataset named \index[\idxKeywordName]{c4l-16x16\_550} \textbf{c4l-16x16\_550} using algorithm \ref{ROI_extractor_algo}.

\begin{algorithm}
	\SetAlgoLined
	\KwIn{$list\_of\_files \leftarrow All~the~files~needed~to~be~extracted$}
	\KwResult{Extracted files}
	\For{each\_file in list\_of\_files}
	{
		$img\_file$ $\leftarrow$ read~current~file;\\
		$gray$ $\leftarrow$ convert~image~file~to~gray;\\
		$thresh$ $\leftarrow$ perform~Otsu's~thresholding; ~\cite{1979:ots}\\
		$k$ = 0;\\
		$ROI\_number$ = 0;\\
		\For{$y \leftarrow 0$ \KwTo $img\_file.height()$}{
			$dummy$ = 0 \tcp*{keeps~track~of~separation}
			\For{$x \leftarrow 0$ \KwTo $img\_file.width()$}
			{
				\If{$thresh[x][y] == 0$}
				{
					$dummy$ = 1 \tcp*{separation~found~hence~extract~character}
					Level 2 extracted Ref \textbf{Figure.~\ref{algo_1_diag}}.;\\
				}
			}
			\If{$dummy$ == 1 {\&}{\&} $k$ == 1 }
			{
				$start\_col$ $\leftarrow$ $y$;\\
				$k$ $\leftarrow$ $k$ + 1;\\
			}
			\If{$dummy$ == 0 {\&}{\&} $k$ $>$ 2 }
			{
				$end\_col$ $\leftarrow$ y;\\
				$k1$ $\leftarrow$ 0;\\
				$extract\_img$ $\leftarrow$ $thresh[:,start\_col - 2: end\_col + 1].copy()$
			}
			
			\For{$x1 \leftarrow 0$ \KwTo $extract\_img.width()$}{
				$dummy1$ = 0 \tcp*{keeps~track~of~separation}
				\For{$y1 \leftarrow 0$ \KwTo $extract\_img.height()$}
				{
					\If{$extract\_img[x1][y1] == 0$}
					{
						$dummy1$ = 1 \tcp*{separation~found~hence~extract~character}
					}
				}
				\If{$dummy1$ == 1 {\&}{\&} $k1$ == 1 }
				{
					$start\_row$ $\leftarrow$ $x1$;\\
					$k1$ $\leftarrow$ $k1$ + 1;\\
				}
				\If{$dummy1$ == 0 {\&}{\&} $k1$ $>$ 2 }
				{
					$end\_row$ $\leftarrow$ $x1$;\\
					Level 3 extracted Ref \textbf{Figure.~\ref{algo_1_diag}}.;\\
					$ROI$ = $extract\_img[start\_row-2:end\_row+1,:].copy()$;\\
					\textbf{Save the extracted region of interst}
				}
				
			}
		}
	}
	
	\caption{Region of Interest Extractor}
	\label{ROI_extractor_algo}
\end{algorithm}

\begin{figure}
	\centering
	\includegraphics[width=0.75\linewidth]{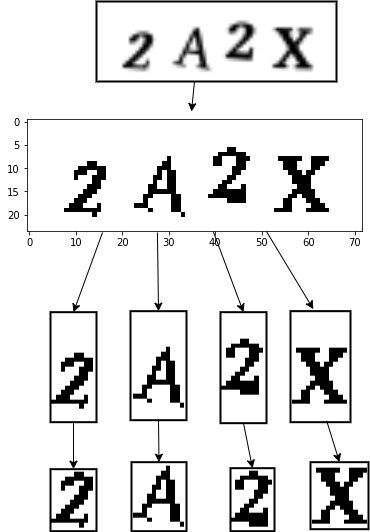}
	\caption{The extraction of separable characters using algorithm ~\ref{ROI_extractor_algo} for creation of c4l-16x16\_550 dataset.}
	\label{algo_1_diag}
\end{figure}

\section{Creation of c4l-16x16\_550 dataset}

\begin{figure}
	\centering
	\fbox{\includegraphics[height=2.6cm,width=0.2\linewidth]{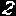}}		\vspace{2px}
	\fbox{\includegraphics[height=2.6cm,width=0.2\linewidth]{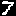}}	\vspace{2px}
	\fbox{\includegraphics[height=2.6cm,width=0.2\linewidth]{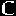}}	\vspace{2px}
	\fbox{\includegraphics[height=2.6cm,width=0.2\linewidth]{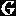}}	\vspace{2px}
	\caption{Sample of c4l-16x16\_550 dataset }
	\label{c4l-16x16_550}
\end{figure}

This is the modified version of the data created by extracting each image and performing \index[\idxKeywordName]{Otsu's thresholding} Otsu's ~\cite{1979:ots} thresholding on the extracted image as shown in Figure ~\ref{c4l-16x16_550}. The dataset can be obtained from here (\url{https://jimut123.github.io/blogs/CAPTCHA/data/c4l-16x16_550.tar.gz}). Otsu's thresholding is an unsupervised and nonparametric method for automatic threshold selection which utilizes the zeroth and first order cumulative moments ~\cite{1979:ots} of gray level histogram.

\begin{figure}
	\centering
	\fbox{\includegraphics[width=0.95\linewidth]{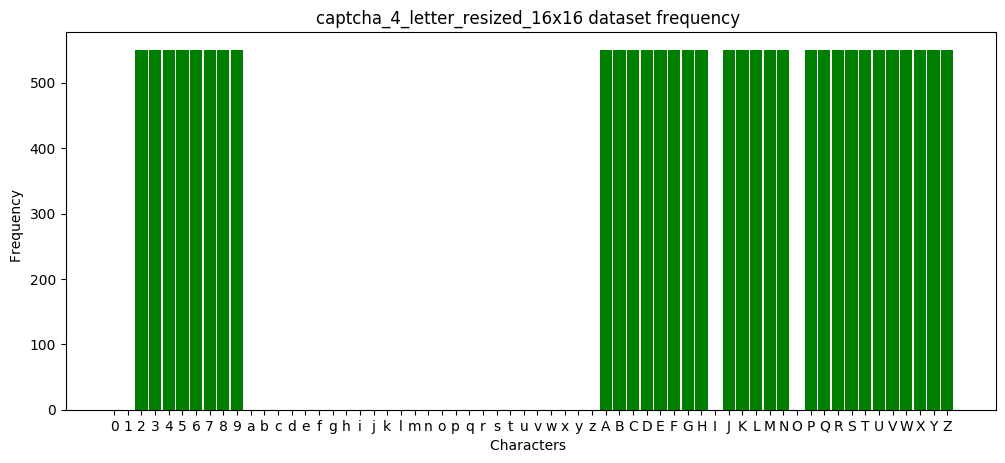}}	\vspace{2px}
	\caption{Frequency of characters obtained from c4l-16x16\_550 dataset }
	\label{c4l-16x16_550_freq}
\end{figure}

Now the dataset contains 32 characters of dimension 16x16 as shown in Figure \ref{c4l-16x16_550_freq}. This dataset can be used for performing benchmarks just like MNIST dataset. The real challenge here is that there are 550 images for each character, and one need to get a good model with such less data.

\section{t-SNE cluster for c4l-16x16\_550 dataset}

t-Distributed Stochastic Neighbor Embedding (t-SNE) visualizes high

\begin{figure}
	\centering
	\includegraphics[width=0.75\linewidth]{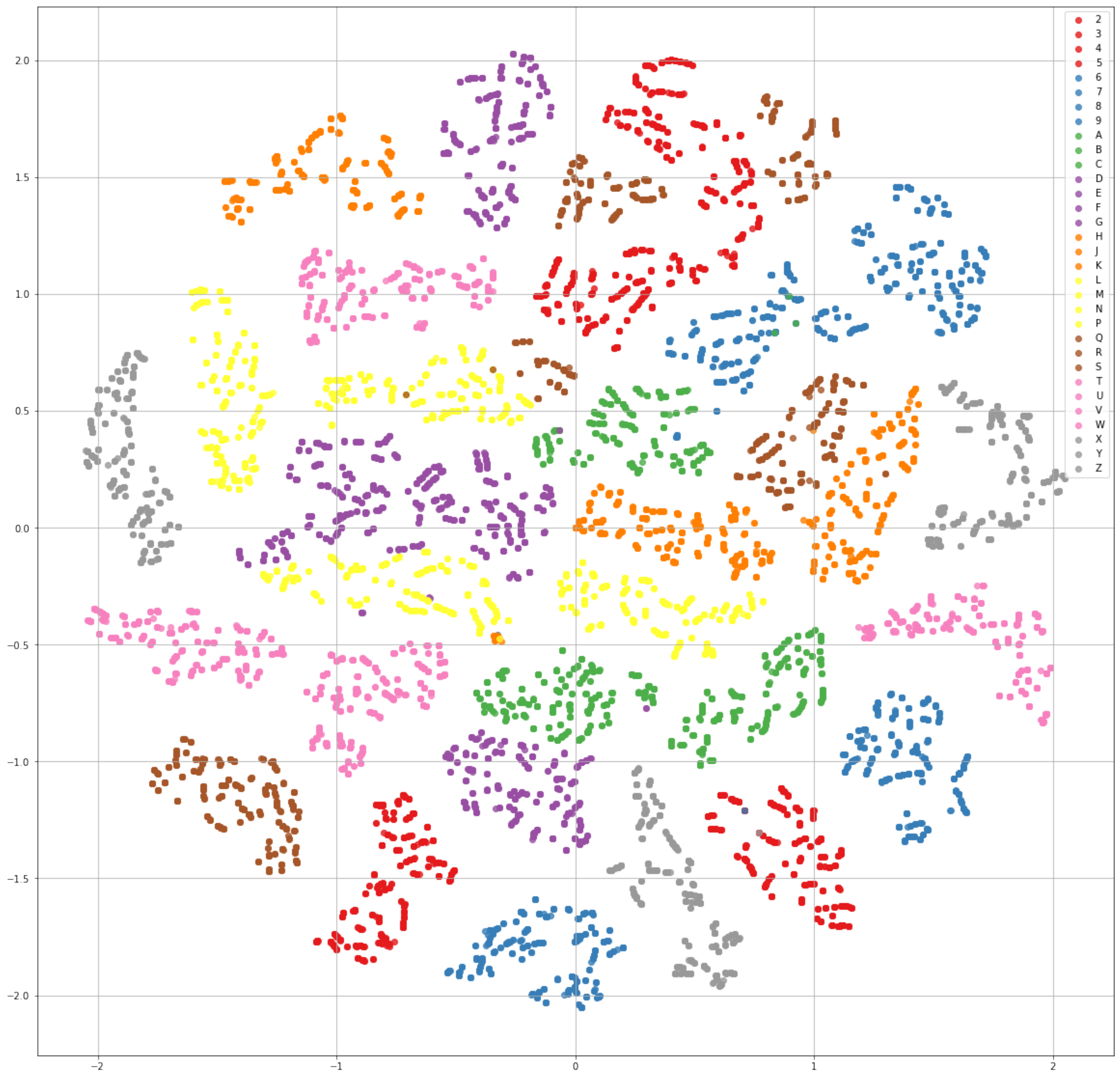}
	\caption{t-SNE cluster of c4l-16x16\_550 dataset in 2D.}
	\label{t-SNE_c4l-16x16_550_2D_simple}
\end{figure}

\noindent%
dimensional data by giving each datapoint a location in three or two-dimensional map. This technique is much easier to optimize, and gives better visualizations by reducing the tendency to crowd points together in the center of the map. \index[\idxKeywordName]{Stochastic Neighbor Embedding (SNE)} Stochastic Neighbor Embedding (SNE) converts the high dimensional Euclidean distance between datapoints into conditional probabilities ~\cite{vanDerMaaten2008} that represent similarities. The similarity between two datapoints $x_{j}$ and $x_{i}$ is the conditional probability $p_{j|i}$, that $x_{i}$ would pick neighbor as $x_{j}$ if neighbors were picked in proportion to their probability density under a Gaussian centered at $x_{i}$. So, from the nature of the Gaussian distribution we can say that for nearby points $p_{j|i}$ is relatively high, whereas for widely spread datapoints $p_{j|i}$ will be infinitely small.

\begin{equation}
	p_{j|i} = \frac{ \exp(-||x_{i}-x_{j}||\textsuperscript{2}/2\sigma_{i}\textsuperscript{2})}
	{\sum_{k\neq i}^{}\exp(-||x_{i}-x_{k}||\textsuperscript{2}/2\sigma_{i}\textsuperscript{2})}
\end{equation}

Here $\sigma_{i}$ is the variance of the Gaussian that is centered on datapoint $x_{i}$. We can do the same for $y_{i}$ and $y_{j}$, the low dimensional counterparts ~\cite{vanDerMaaten2008} of the high dimensional datapoints $x_{i}$ and $x_{j}$, which we denote by $q_{j|i}$.

\begin{equation}
q_{j|i} = \frac{ \exp(-||y_{i}-y_{j}||\textsuperscript{2})}
{\sum_{k\neq i}^{}\exp(-||y_{i}-y_{k}||\textsuperscript{2})}
\end{equation}

If the map points $y_{i}$ and $y_{j}$ correctly model the similarity between the high dimensional data points, i.e., $x_{i}$ and $x_{j}$ then both the conditional probabilities $p_{j|i}$ and $q_{j|i}$ will be equal. This modelling is done through \index[\idxKeywordName]{Kullback-Leiber divergence} Kullback-Leiber divergence ~\cite{vanDerMaaten2008} and SNE minimizes the sum of these overall datapoints using gradient descent method.

The cost function C is given by

\begin{equation}
C = \sum_{i}^{} KL(P_{i}||Q_{i}) = \sum_{i}^{}\sum_{j}^{}p_{j|i} \log
\frac{p_{j|i}}{q_{j|i}}
\end{equation}

Here, $P_{i}$ represents the conditional probability distribution over all other data points given data points $x_{i}$, and $Q_{i}$ represents the conditional probability distribution over all map points given map point $y_{i}$.

\begin{figure}
	\centering
	\includegraphics[width=0.75\linewidth]{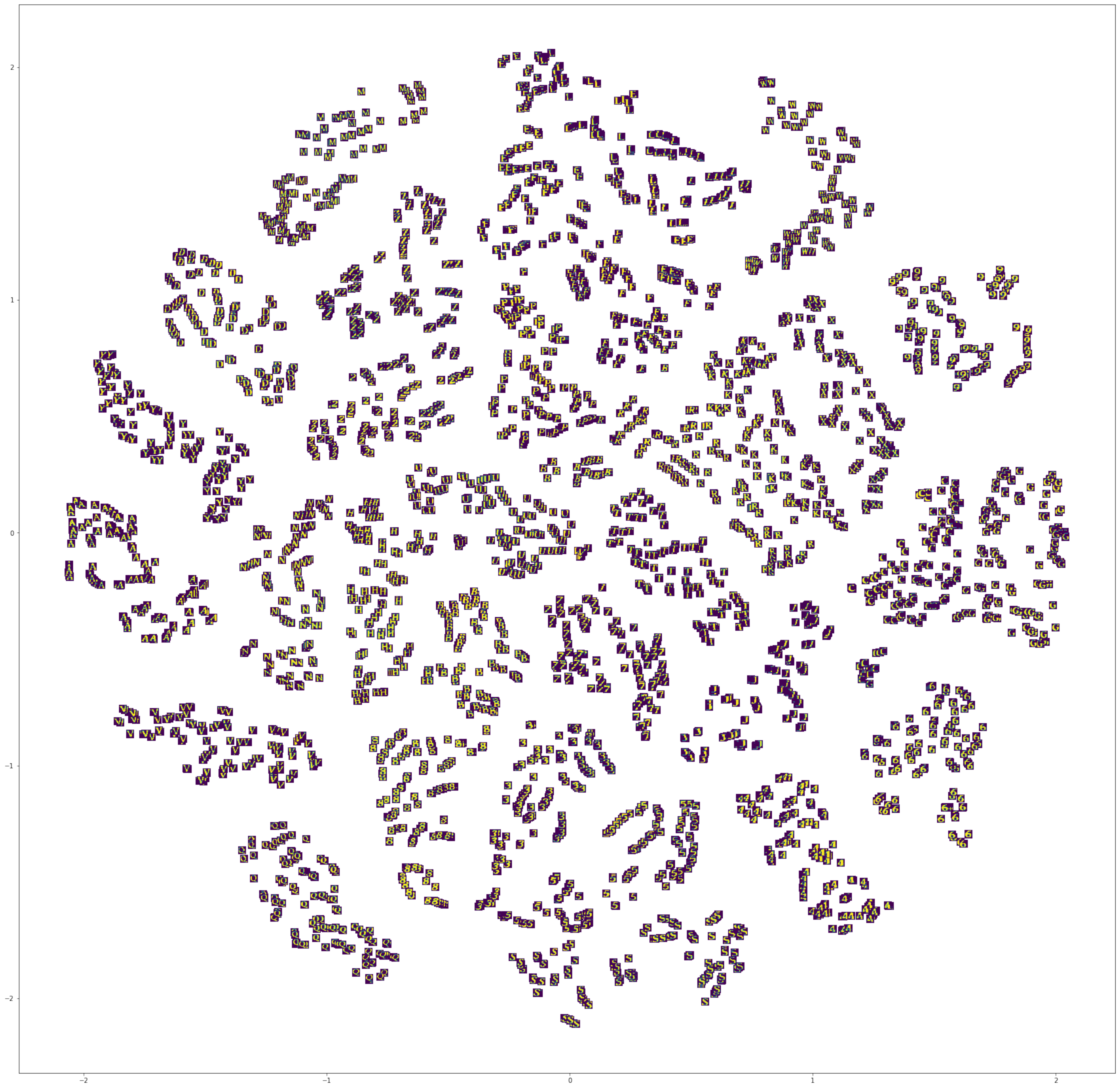}
	\caption{t-SNE cluster of c4l-16x16\_550 dataset in 2D with actual images.}
	\label{t-SNE_c4l-16x16_550_2D}
\end{figure}

\begin{figure}
	\centering
	\includegraphics[width=0.75\linewidth]{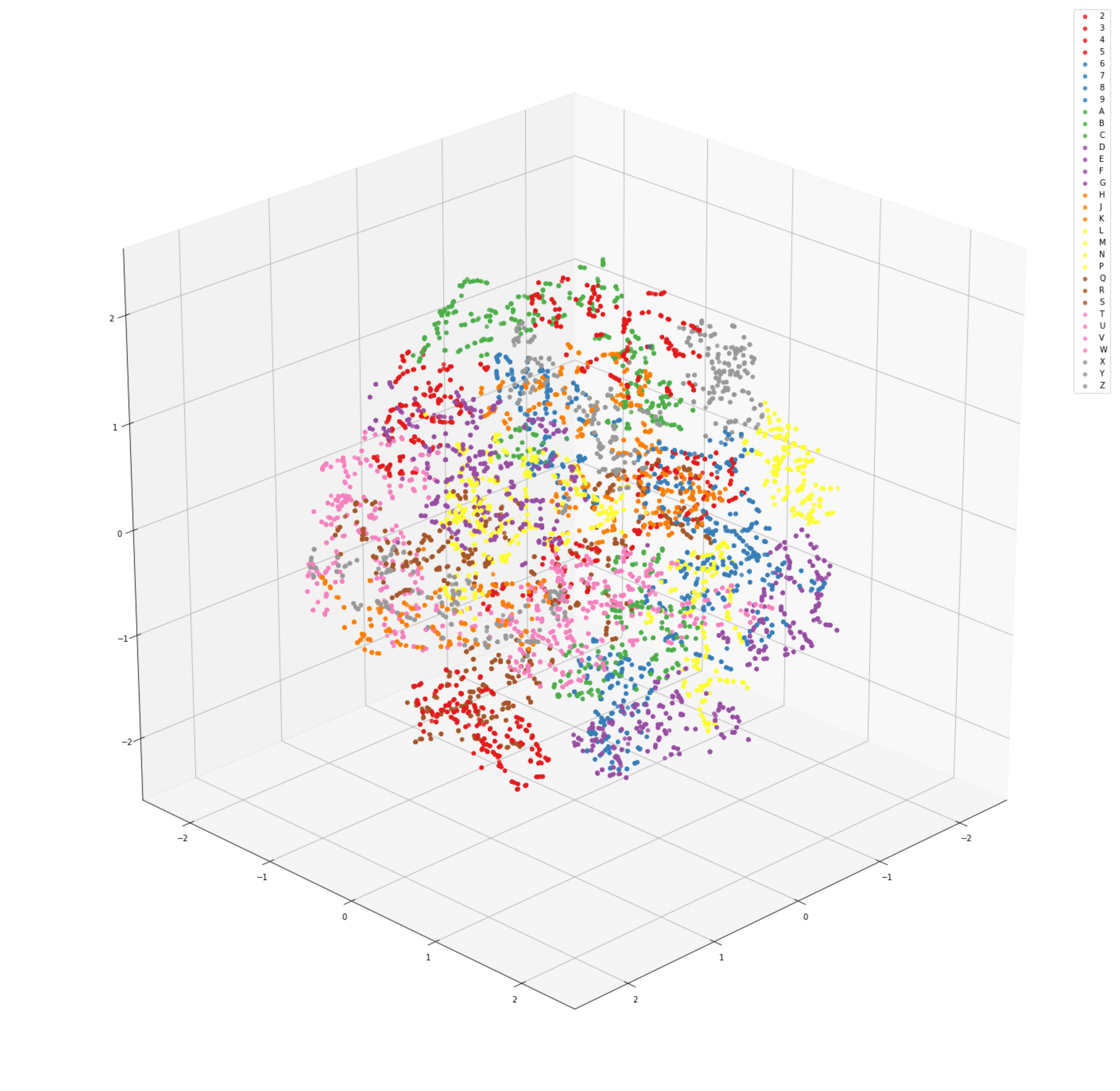}
	\caption{t-SNE cluster of c4l-16x16\_550 dataset in 3D.}
	\label{t-SNE_c4l-16x16_550_3D}
\end{figure}

\index[\idxKeywordName]{t-SNE}t-SNE uses symmetrized version of the SNE cost function with gradient descent which are easier to optimize and a Student-t distribution ~\cite{vanDerMaaten2008} rather than a Gaussian. Figure \ref{t-SNE_c4l-16x16_550_2D_simple} shows the visualisation of the data in 2-dimensional space, similarly Figure \ref{t-SNE_c4l-16x16_550_2D} shows the same visualization but with the actual character that forms the part of the visualization. Figure  ~\ref{t-SNE_c4l-16x16_550_3D} shows the 3-dimensional visualization of the same datapoints.

\section{JAM CAPTCHA dataset}

\begin{figure}
	\centering
	\fbox{\includegraphics[width=1\linewidth]{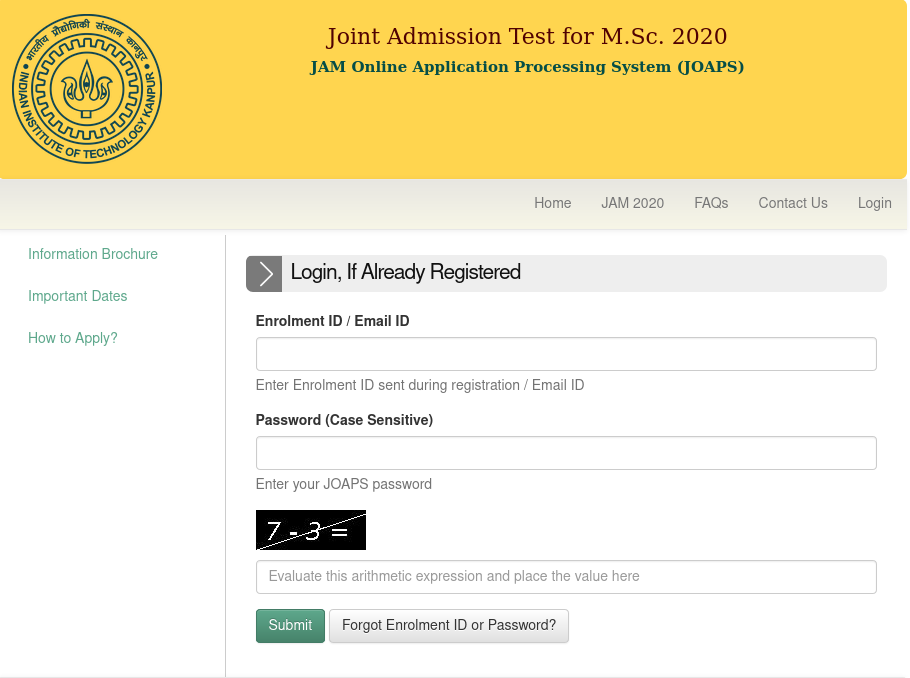}}
	\caption{JAM website containing simple CAPTCHA. (Source: \url{https://joaps.iitk.ac.in/}) }
	\label{captcha_website_JAM}
\end{figure}

\begin{figure}
	\centering
	\fbox{\includegraphics[width=0.45\linewidth]{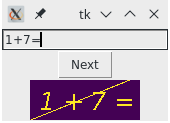}}
	\caption{A simple labeller to label the CAPTCHA. }
	\label{CAPTCHA_labeller_JAM}
\end{figure}

We found this CAPTCHA on IIT-JAM's website as shown in Figure \ref{captcha_website_JAM}. This is an interesting problem to solve since the CAPTCHA has a strike-though in it, to create confusion for machines to classify this CAPTCHA. We thought to ourselves whether this is really a good CAPTCHA? We created a dataset by scraping almost 10,000 of CAPTCHAs and labelling only 308 of them using a tool we created as shown in Figure \ref{CAPTCHA_labeller_JAM}. This tool can be used to label any kind of CAPTCHA without manually selecting rename every time. The initial size of the CAPTCHA was 100x40x3 with the three channels (R,G,B). We needed to modify this data to convert to grayscale since this CAPTCHA doesn't really makes sense to contain 3 channels. The only color that it had was white in black background. We also applied a certain thresholding so that the image turns to pure binary file with just one channel.

\begin{figure}
	\centering
	\includegraphics[height=24cm,width=1\linewidth]{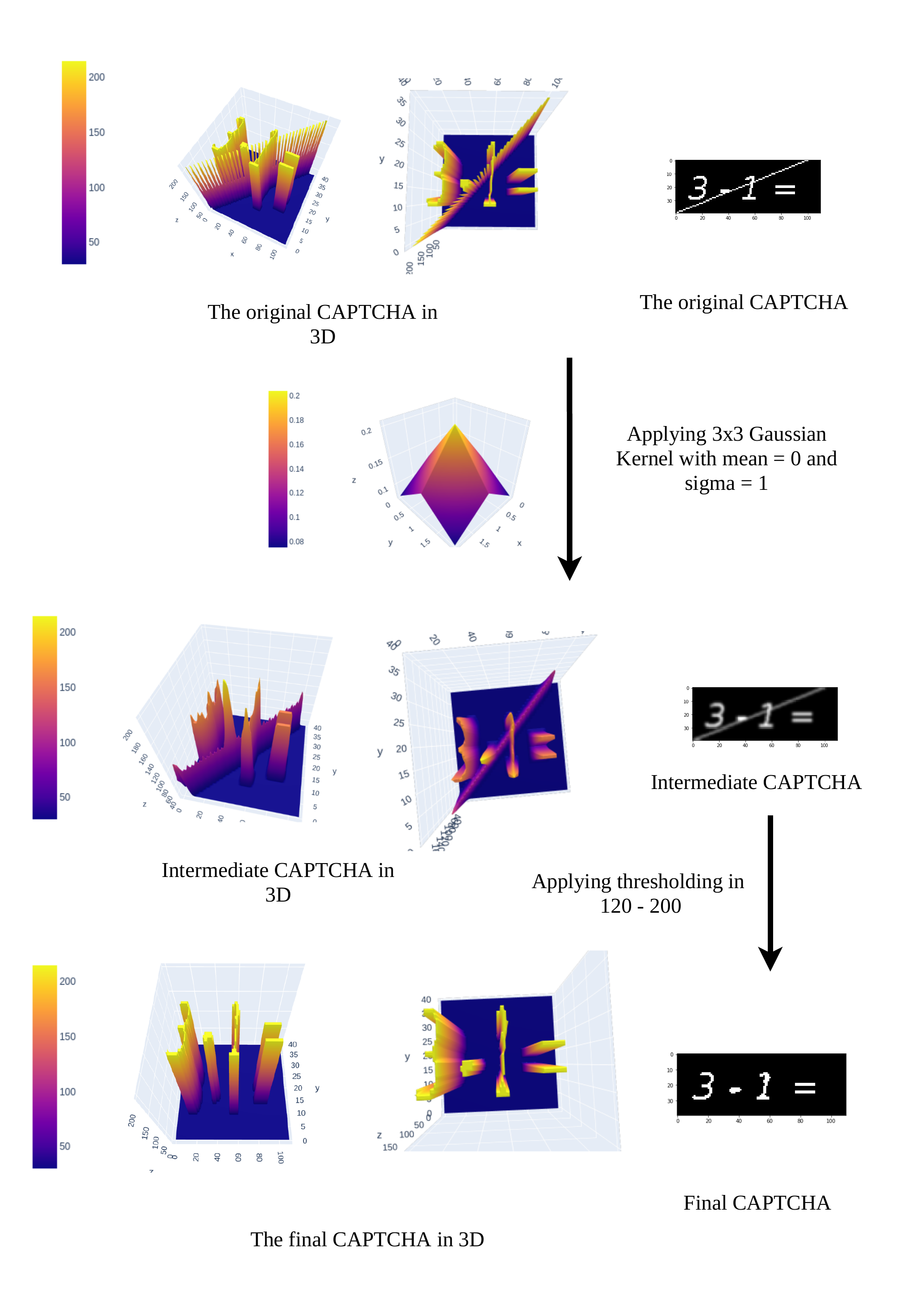}
	\caption{Removal of lines from the JAM CAPTCHA.}
	\label{removing_line}
\end{figure}

Since we want to try out a simple model on this CAPTCHA, we needed to remove the strike-through in it. We found that this CAPTCHA has very similar kind of font style with no change in them. We thought that it can be easily solved with the help of $k$-Nearest Neighbor (KNN) algorithm, if we can somehow remove the noise and partition the characters with the segmentation algorithm \ref{ROI_extractor_algo}. The removal of the line was easy. The procedure for removal of lines for each image is shown in Figure \ref{removing_line}. First, we take the converted thresholded CAPTCHA and pass it through a gaussian filter \cite{pal2020deeper} with $\sigma$ = 1 and $\mu$ = 0. We do this because the strike is thinner than the bold numbers. This will make the strike lower in intensity when seen in 3D. After this we will apply a threshold, i.e., every value of intensity that is less than 120 will turn to 0 and the values above 200 will turn to 255 and hence that will be the final resultant thresholded CAPTCHA as shown in Figure \ref{removing_line} below.

\begin{figure}
	\centering
	\includegraphics[width=0.75\linewidth]{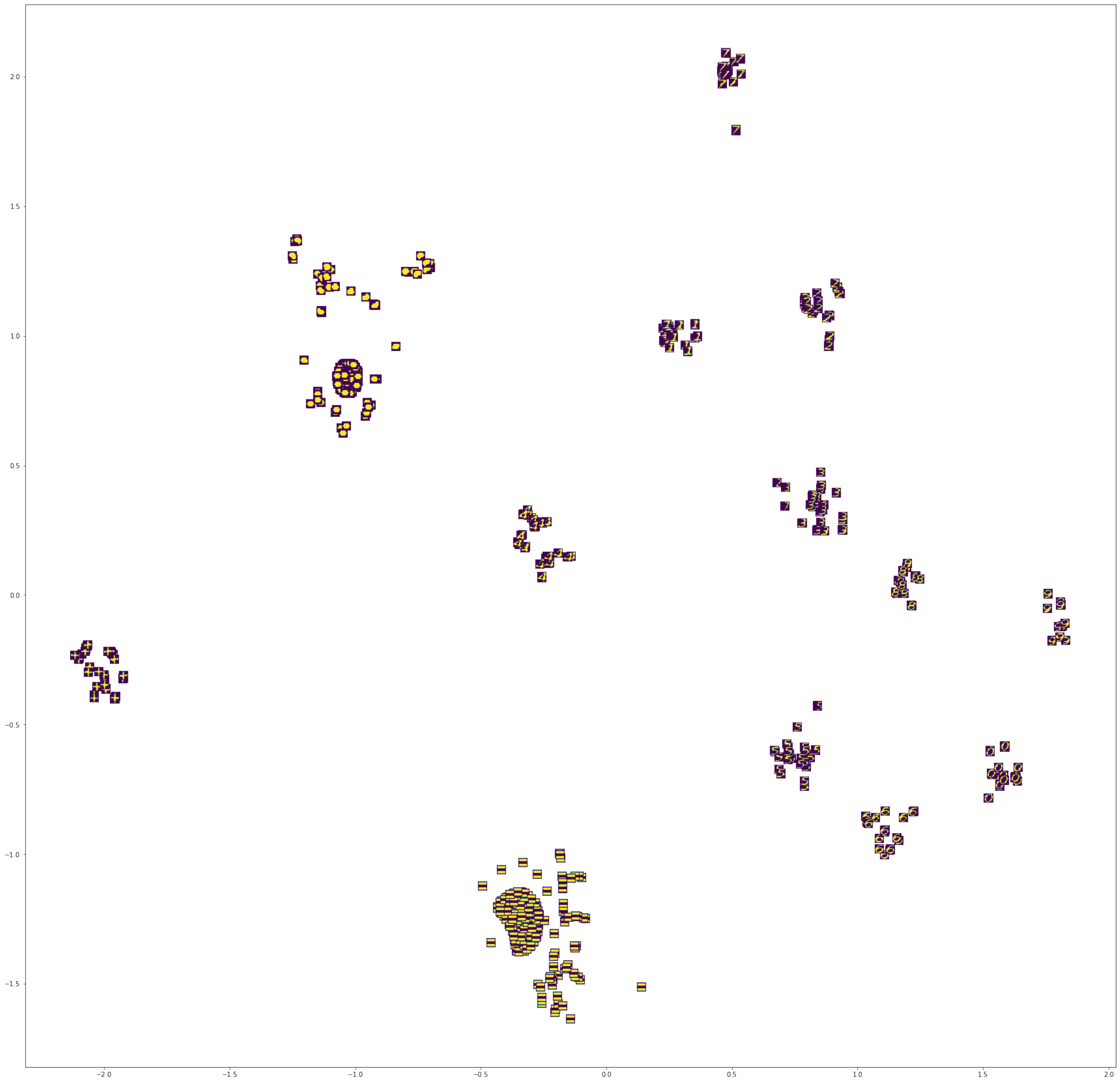}
	\caption{t-SNE cluster of resized\_JAM dataset in 2D with actual images.}
	\label{t-SNE_plot_JAM_letters_2D}
\end{figure}

\section{Classifying digits through k-NN with the resized\_JAM dataset}

\begin{figure}
	\centering
	\includegraphics[width=1\linewidth]{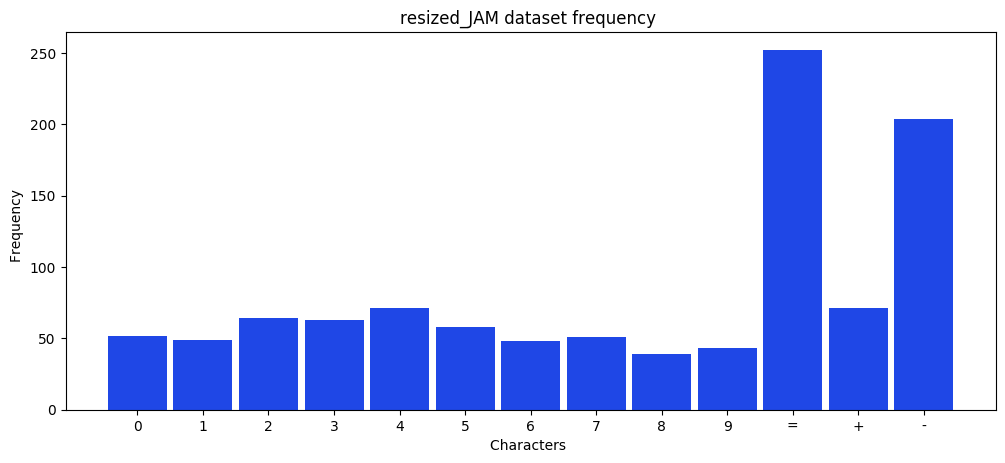}
	\caption{Distribution of resized\_JAM CAPTCHA dataset.}
	\label{dist_JAM_CAPTCHA}
\end{figure}

\begin{figure}
	\centering
	\includegraphics[width=0.5\linewidth]{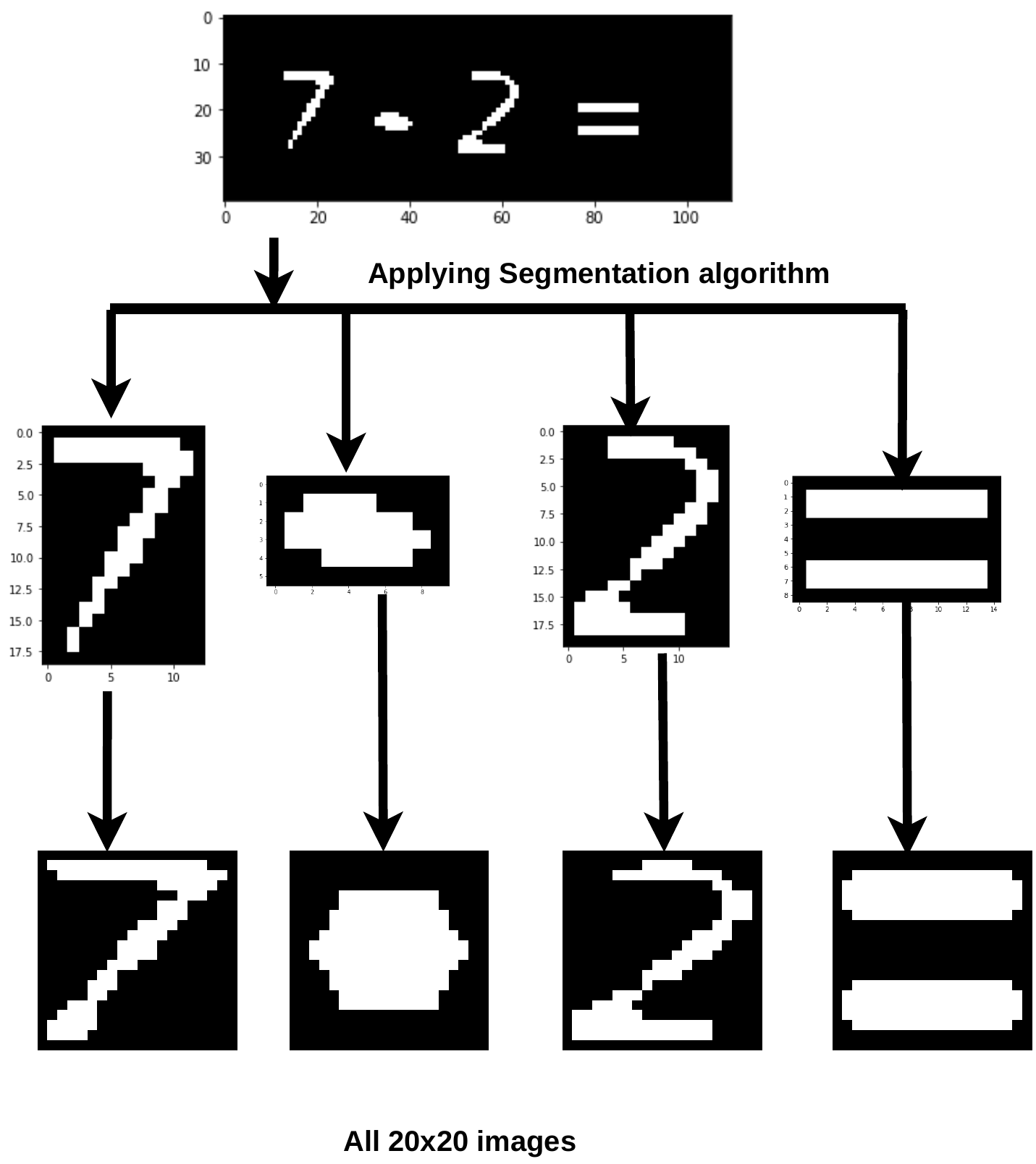}
	\caption{Segmentation of JAM CAPTCHA.}
	\label{segmentation_JAM_CAPTCHA}
\end{figure}

For training in \index[\idxKeywordName]{$k$-NN} $k$-NN  we created the resized\_JAM dataset. This was achieved by applying algorithm \ref{ROI_extractor_algo} in the final image. The process of segmentation is shown in Figure \ref{segmentation_JAM_CAPTCHA}. The segmentation resulted in uneven sized images of each of the characters present in the image. Then we resized each of them in 20x20 images for creating the resized\_JAM dataset. The dataset can be found here (\url{https://jimut123.github.io/blogs/CAPTCHA/data/resized_JAM.tar.gz}). The distribution of characters of the dataset is shown in Figure \ref{dist_JAM_CAPTCHA}.

\begin{figure}
	\centering
	\includegraphics[width=0.75\linewidth]{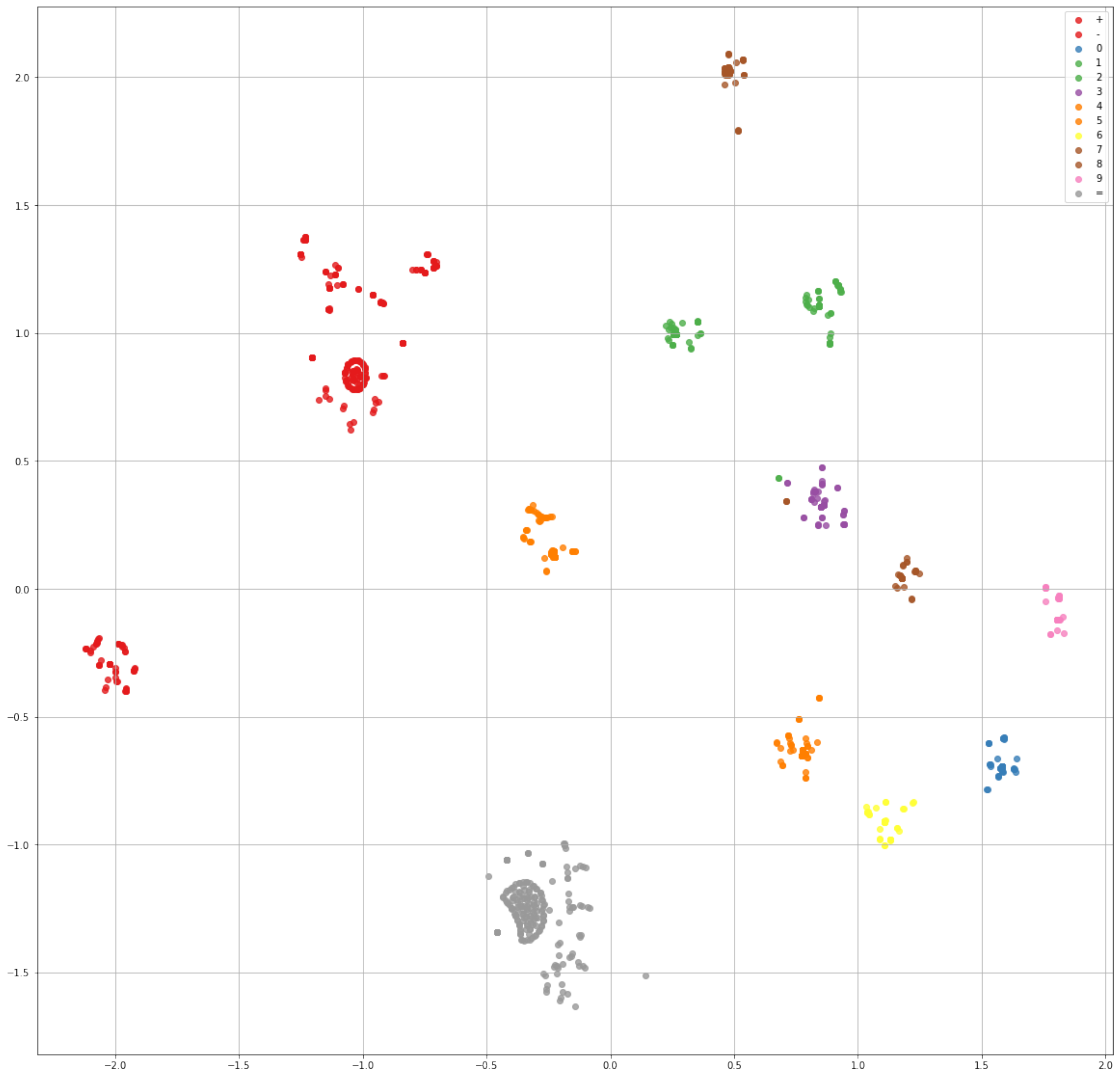}
	\caption{t-SNE cluster of resized\_JAM dataset in 2D.}
	\label{t-SNE_plot_JAM_2D}
\end{figure}

\begin{figure}
	\centering
	\includegraphics[width=0.75\linewidth]{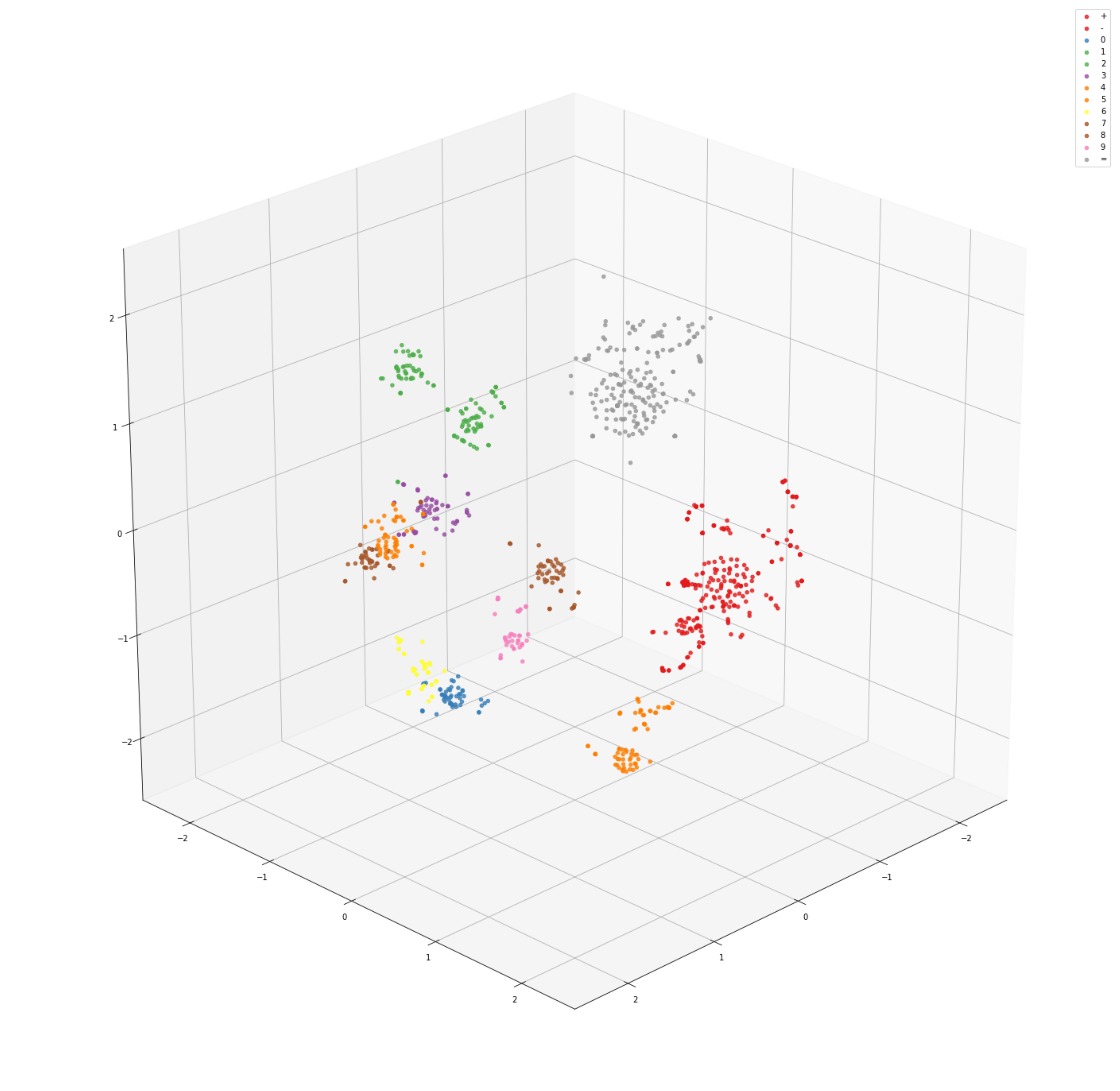}
	\caption{t-SNE cluster of resized\_JAM dataset in 3D.}
	\label{t-SNE_plot_JAM_3D}
\end{figure}

We see that there is a relatively uneven distribution in the case of this dataset. This will not create any major problem. Since the data is relatively low, so we can use the $k$-NN classifier. The t-SNE plot of the data in 2D is shown in Figure \ref{t-SNE_plot_JAM_2D}. The same plot with the actual images is shown in Figure \ref{t-SNE_plot_JAM_letters_2D}. The 3D plot is shown in Figure \ref{t-SNE_plot_JAM_3D}.

\begin{figure}
	\centering
	\fbox{\includegraphics[height=7cm,width=1\linewidth]{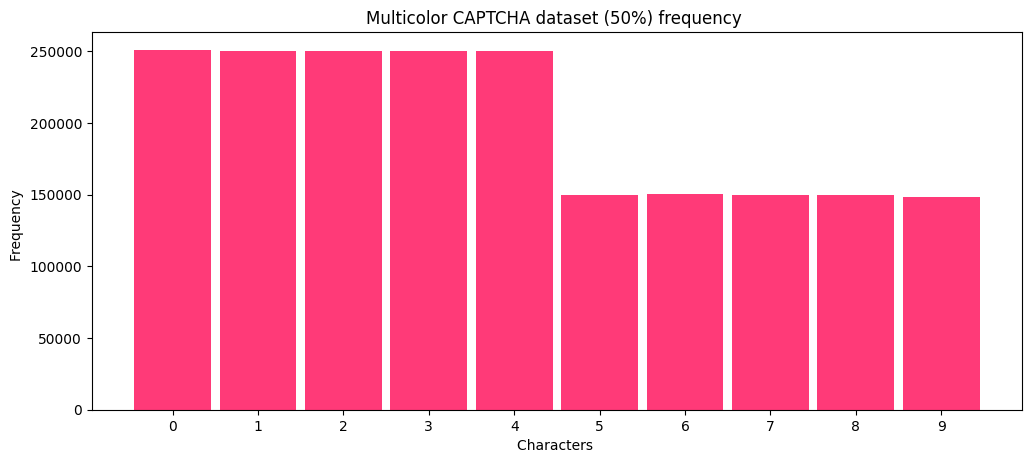}}
	\caption{The initial distribution of the letters for 50\% of Multicolor CAPTCHA.}
	\label{dist_mc}
\end{figure}

\section{CAPTCHA-version-2}

\begin{figure}
	\centering
	\fbox{\includegraphics[height=7cm,width=1\linewidth]{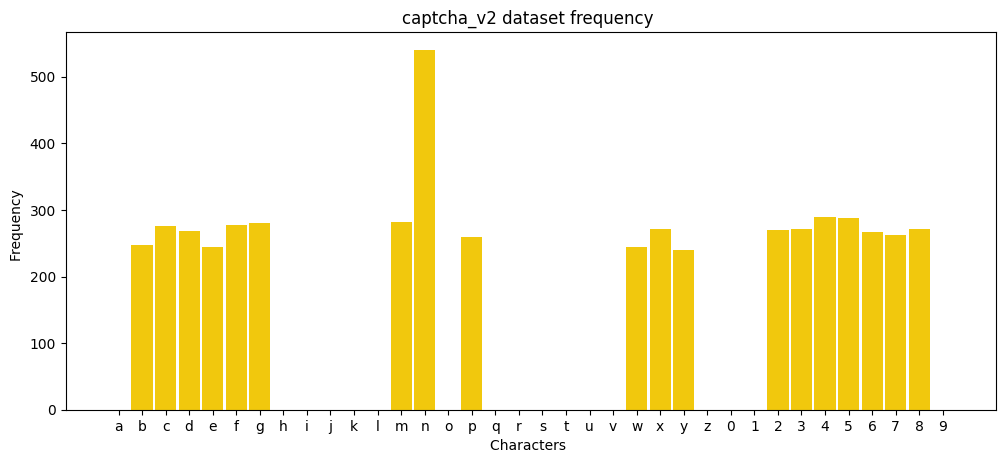}}
	\caption{The initial distribution of the letters for the CAPTCHA-version-2 dataset.}
	\label{captcha_v2_dist}
\end{figure}

\begin{figure}
	\centering
	\fbox{\includegraphics[width=0.455\linewidth]{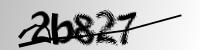}}		\vspace{6px}
	\fbox{\includegraphics[width=0.455\linewidth]{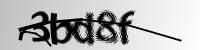}}	\vspace{2px}
	
	\caption{Sample of CAPTCHA-version-2 dataset.}
	\label{captcha-v2}
\end{figure}

This dataset can also be found online on Kaggle (\url{https://www.kaggle.com/fournierp/captcha-version-2-images}). The cleaned version can be found here (\url{https://jimut123.github.io/blogs/CAPTCHA/data/captcha_v2.tar.gz}).
There are two sets of folders each containing 1070 images, the total size of the dataset is about 21.5 MB. The images have 3 channels, i.e., they have a Red Green and Blue component, and are of dimension 200x50. The dataset comprises of small letter character ASCII characters (i.e., a-z) and digits (0-9). It is 5 letter CAPTCHA and the images are cluttered to create confusion as shown in Figure \ref{captcha-v2}. The initial distribution of the characters of the dataset is shown in Figure \ref{captcha_v2_dist}, which is almost uniform except the letter \textbf{n}.

\begin{figure}
	\centering
	\fbox{\includegraphics[width=0.455\linewidth]{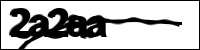}}		\vspace{6px}
	\fbox{\includegraphics[width=0.455\linewidth]{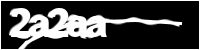}}	\vspace{2px}
	\caption{Sample of 100000-labeled-captchas dataset.}
	\label{captcha-1L}
\end{figure}

\section{100000-labeled-captchas}

\begin{figure}
	\centering
	\fbox{\includegraphics[height=7cm,width=1\linewidth]{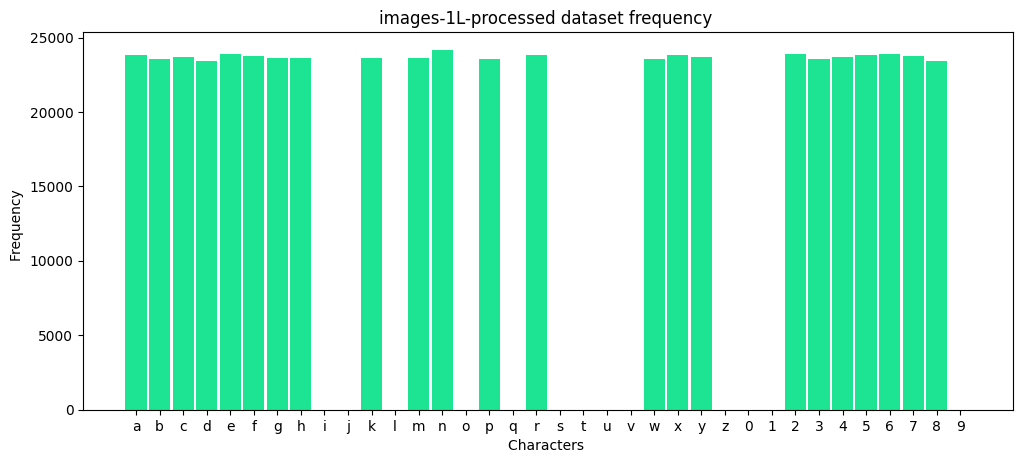}}
	\caption{The initial distribution of the letters for the 100000-labeled-captchas.}
	\label{images-1L-processed}
\end{figure}

This dataset can also be found online on Kaggle \url{https://www.kaggle.com/digdeepbro/100000-labeled-captchas}. The cleaned version can be found here on figshare (\url{https://dx.doi.org/10.6084/m9.figshare.12046881.v1}). The dataset is about 215 MB on figshare. The original dataset found on Kaggle contained only the alpha values, our processed version contains binary images by applying Otsu's thresholding as shown in Figure \ref{captcha-1L} (right). It is necessary to convert the dataset to binary value for getting better predictions and we also need to inverse the colours, since white is represented by intensity value of 255 on grayscale and black by 0. For training it is necessary to take the most prominent pattern to be higher value and less important patterns to be smaller for the network to learn faster. It is also uploaded to figshare because we are taking the advantage of Google's free resource called Collaboratory here, which provides free GPU and 25GB of RAM on cloud, since Kaggle doesn't allow to directly download files to remote server by performing \emph{wget}, we needed to reupload the cleaned version on figshare. This ensures that we can reuse the code without further modification and share the colab notebook with anybody. We could also upload the contents in drive, but from my earlier experience, I have found that the GPU doesn't work on such data after sometime since the server has to relocate the old data every time, on the other hand by this method, it is downloading from figshare with about 25MB/s speed and storing it on temporary storage so that we can perform fast computations easily. The initial distribution of the characters of the dataset is shown in Figure \ref{images-1L-processed}, which is almost uniform.

\begin{figure}
	\centering
	\fbox{\includegraphics[height=7cm,width=1\linewidth]{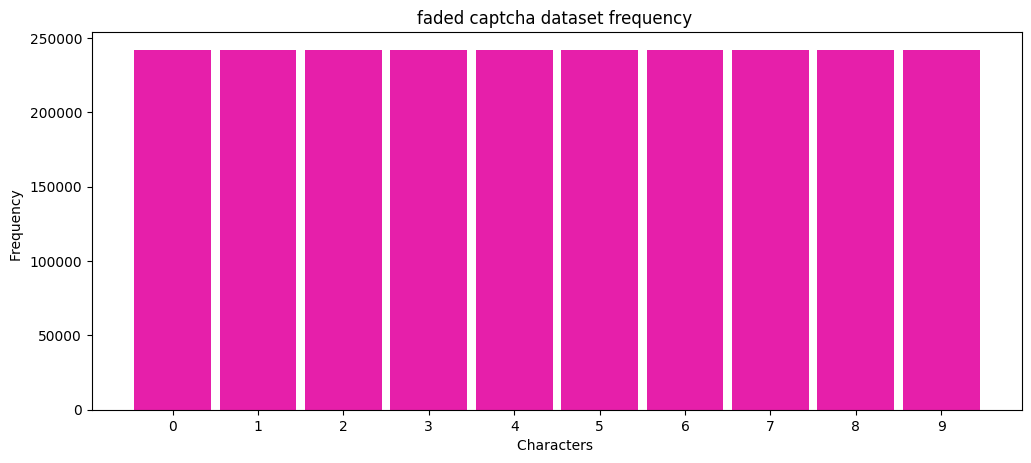}}
	\caption{The initial distribution of the letters for the Faded CAPTCHA.}
	\label{dist_faded}
\end{figure}

\section{Faded CAPTCHA}

\begin{figure}
	\centering
	\includegraphics[width=0.455\linewidth]{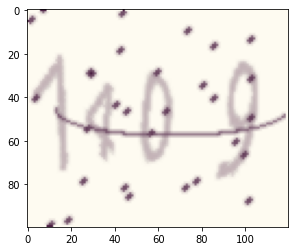}	
	\caption{Sample of the faded dataset.}
	\label{faded_Sample}
\end{figure}

We have found an open sourced CAPTCHA data generator and created a dataset called Faded CAPTCHA dataset. The original source code for the generator can be found here \url{https://github.com/JackonYang/captcha-tensorflow}.
The faded CAPTCHA dataset comprises of about 604800 files of dimension 100x120x3 for the three channels. The charcters present in this CAPTCHA data is from 0-9. A sample of the data is shown in Figure \ref{faded_Sample}. The distribution of the characters of faded CAPTCHA dataset is shown in Figure \ref{dist_faded}. This shows that the CAPTCHA has a uniform distribution of characters which will help it generalise on the test dataset better. The dataset comprises of about 10 folders and the link to each of them is given in the Appendix section. The total size of the dataset is about 3.7 GiB.

\section{Railway CAPTCHA}

\begin{figure}
	\centering
	\fbox{\includegraphics[width=0.455\linewidth]{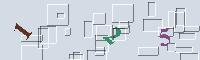}}\vspace{6px}\hspace{2px}
	\fbox{\includegraphics[width=0.455\linewidth]{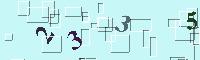}}\vspace{6px}
	\fbox{\includegraphics[width=0.455\linewidth]{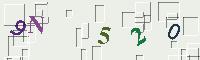}}\vspace{6px}\hspace{2px}	
	\fbox{\includegraphics[width=0.455\linewidth]{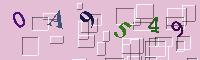}}\vspace{6px}
	\fbox{\includegraphics[width=0.455\linewidth]{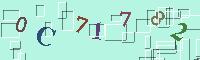}}
	
	\caption{Sample of 100000-labeled generated Railway CAPTCHA dataset.}
	\label{railway-captcha}
\end{figure}

This dataset is a created synthetically by getting hold of the source code from GitHub. The original repository can be found here \url{https://github.com/JasonLiTW/simple-railway-captcha-solver}. We have modified this CAPTCHA generator a bit and created our own dataset comprising of 100K images each for 3,4,5,6, and 7 letter CAPTCHAs as shown in Figure \ref{railway-captcha}. The dataset for 3,4,5,6, and 7 letter CAPTCHAs can be found from \url{https://dx.doi.org/10.6084/m9.figshare.12045249.v1}, \url{https://dx.doi.org/10.6084/m9.figshare.12045288.v1}, \url{https://dx.doi.org/10.6084/m9.figshare.12045294.v1}, \url{https://dx.doi.org/10.6084/m9.figshare.12045657.v1} and \url{https://dx.doi.org/10.6084/m9.figshare.12053937.v1} respectively.

The authors of the original data generator claims that this CAPTCHA can be found to be active on Taiwan railway booking website. The CAPTCHA is generated by applying random transformation to images formed by \emph{Courier New-Bold} and \emph{Times New Roman-Bold} fonts.

\section{Circle CAPTCHA}

\begin{figure}
	\centering
	\includegraphics[width=0.455\linewidth]{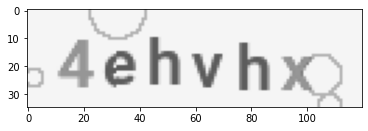}	
	\caption{Sample of the Circle CAPTCHA dataset.}
	\label{cc_sample}
\end{figure}

\begin{figure}
	\centering
	\fbox{\includegraphics[height=7cm,width=1\linewidth]{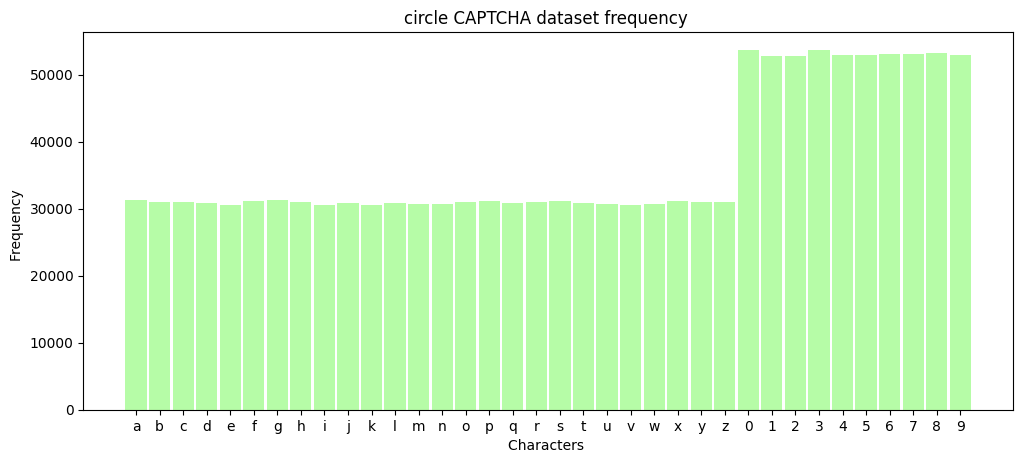}}
	\caption{The initial distribution of the letters for the Circle CAPTCHA dataset.}
	\label{dist_circle}
\end{figure}

We found this type of CAPTCHA from \url{https://github.com/py-radicz/captcha_gen}. We modified a bit, and created our own dataset. The dataset can be found from here \url{https://figshare.com/articles/captcha_gen/12286766}. A sample of the circular CAPTCHA is shown in Figure \ref{cc_sample}. The distribution of the letters of the Circle CAPTCHA can be shown in Figure \ref{dist_circle}. This shows that the circle CAPTCHA consists of all the smaller case (a-z) letters and numbers (0-9) as label in its contents. The dataset comprises of 437.58 MB data, which is of dimension 35x120x3 (for the three channels). The dataset comprises of about 222K samples.

\section{The Sphinx CAPTCHA}

\begin{figure}
	\centering
	\includegraphics[width=0.455\linewidth]{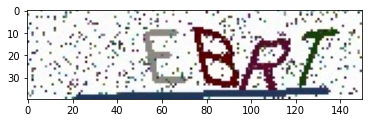}	
	\caption{Sample of the sphinx CAPTCHA dataset.}
	\label{sphinx_sample}
\end{figure}

\begin{figure}
	\centering
	\fbox{\includegraphics[height=7cm,width=1\linewidth]{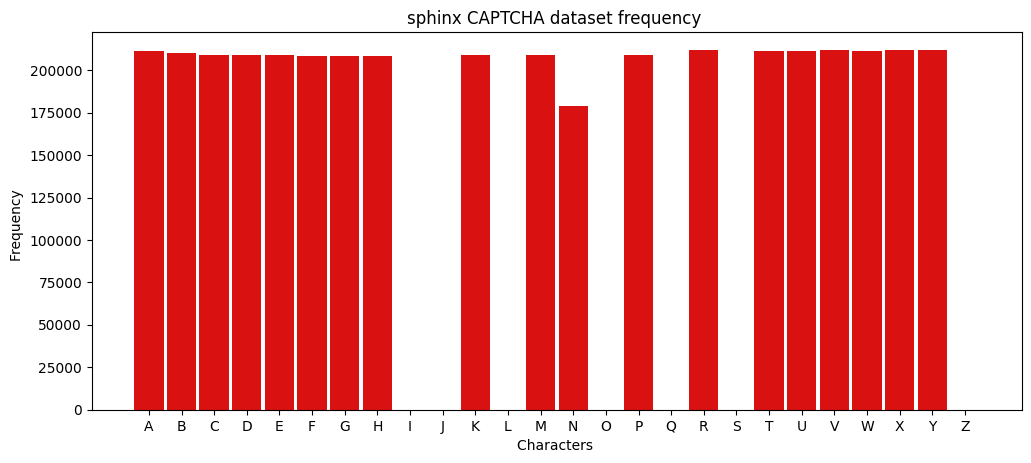}}
	\caption{The initial distribution of the letters for the sphinx CAPTCHA dataset.}
	\label{dist_sphinx}
\end{figure}

We found the open source code for this type of CAPTCHA from \url{https://github.com/davidpalves/sphinx-captcha}. One sample of the Sphinx CAPTCHA is shown in Figure \ref{sphinx_sample}. The sphinx CAPTCHA is a relatively difficult CAPTCHA because it has a line (strikethrough through it), with relatively same thickness of the letters and noise of different color all over it. The distribution of the letters from the sphinx captcha dataset can be found in Figure \ref{dist_sphinx}. The images of the sphinx CAPTCHA dataset have a dimension of about 368x123x3 for the three channels. There are about 990K images which is close to 1M. The link to the figshare version of the dataset can be found in the appendix section. The dataset comprises of 34 parts of tar.gz all uploaded in figshare. The whole size of the dataset is about 12.2 GiB. This dataset contains only some of the characters from A-Z range, with some missing characters.

\section{Fish Eye CAPTCHA }

\begin{figure}
	\centering
	\includegraphics[width=0.455\linewidth]{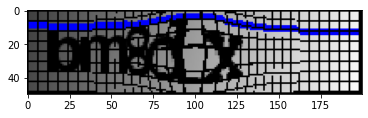}	
	\caption{Sample of the fish eye CAPTCHA dataset.}
	\label{fisheye_Sample}
\end{figure}

\begin{figure}
	\centering
	\fbox{\includegraphics[height=7cm,width=1\linewidth]{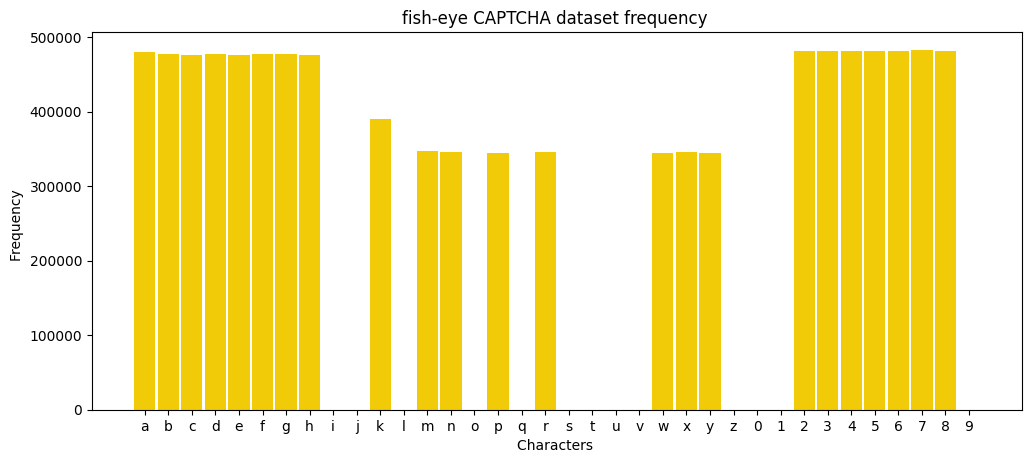}}
	\caption{The initial distribution of the letters for the fish eye CAPTCHA.}
	\label{dist_fisheye}
\end{figure}

We found the Java source code from GitHub, for this type of CAPTCHA and created our own flexible fish eye CAPTCHA generator. One sample of the fish eye type CAPTCHA is shown in Figure \ref{fisheye_Sample}. We can see there is a lot of variety in this type of CAPTCHA. This type of CAPTCHA poses a real challenge to machine learning systems. There is grid, a line and bold letters which is occluded by lines of different colors. The deep learning model needs to learn each of the representation of the CAPTCHA dataset. There is a total of 2 million sample in the generated dataset. The dataset can be found from this url (\url{https://figshare.com/articles/Fish_Eye/12235946}). Each sample of the data has a dimension of about 50x200x3 for the three channels. The data distribution of the characters for this dataset can be shown in Figure \ref{dist_fisheye}.

\section{Mini CAPTCHA}

\begin{figure}
	\centering
	\includegraphics[width=0.455\linewidth]{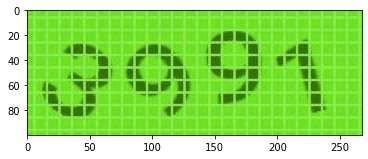}	
	\caption{Sample of the mini CAPTCHA dataset.}
	\label{mini_captcha}
\end{figure}

\begin{figure}
	\centering
	\fbox{\includegraphics[height=7cm,width=1\linewidth]{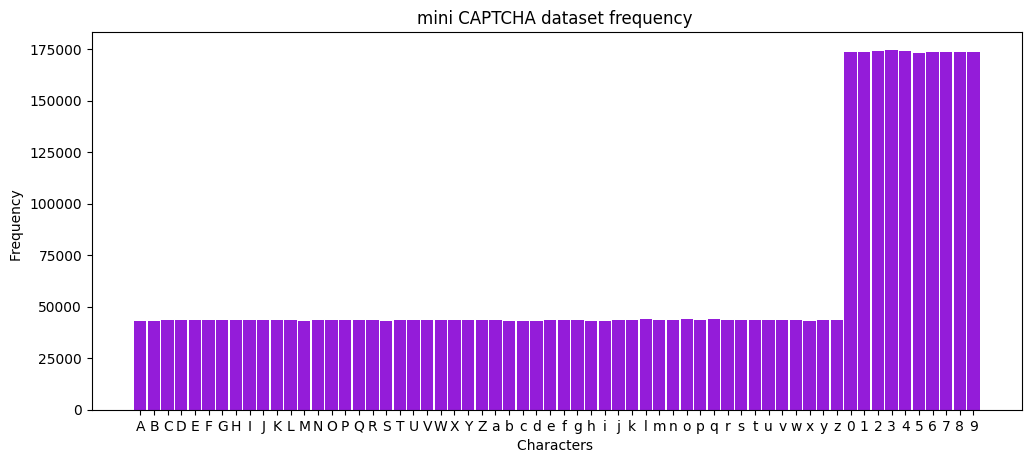}}
	\caption{The initial distribution of the letters for the mini CAPTCHA.}
	\label{mini_captcha_dist}
\end{figure}

We found the open source code from \url{https://github.com/imanhpr/miniCaptcha}, and modified a bit to generate the actual CAPTCHA generator. The dataset has about 1 million samples of size 368x159x3 (for three channels). A sample from the dataset is shown in Figure \ref{mini_captcha}. The distribution of the letters of the dataset is shown in Figure \ref{mini_captcha_dist}. There are a variety of colours that is present in the CAPTCHA dataset. The dataset can be found from here \url{https://figshare.com/articles/Mini_Captcha/12286697}. The distribution shows that there are all the letters present in the English alphabet, both upper case [A-Z] and lower case [a-z] along with digits [0-9], which makes it challenging task for computers to classify about 62 classes all at once.

\section{Multicolor CAPTCHA}

\begin{figure}
	\centering
	\includegraphics[width=0.475\linewidth]{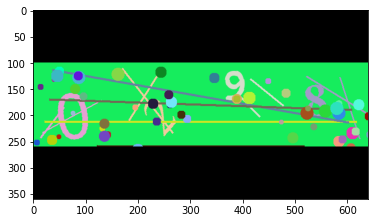}
	\includegraphics[width=0.475\linewidth]{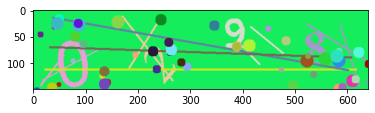}
	\caption{Sample of multicolor CAPTCHA dataset.}
	\label{multicolor}
\end{figure}

We found the original source code for this type of CAPTCHA on this link \url{https://github.com/J-Rios/multicolor_captcha_generator}. We have generated about 1 Million CAPTCHA data, but used only 50\% of the original dataset. The link for the dataset can be found from the appendix section. We have used a difficulty level of 4 for the generation of the data. The highest difficulty is level 5, which has got more different types of colors, but is similar to this type of data. The actual problem is that these CAPTCHAs are slow to generate. A sample of the CAPTCHAs generated are shown in Figure \ref{multicolor}. There is extra back background that is present in the original version of the dataset, which needs to be cropped as shown in the former figure. The dimension of the original uncropped image is 375x223x3 which has black background as clutter. This CAPTCHA contains only numbers i.e., 0-9 and is comprised of 4 letter labels, which is standard. The cropped version has a dimension of about 375x113x3 which is pretty good for training. The size of the ten pieces of tar.gz file is about 12.7 GB. The dataset is hosted in figshare and is available for training.

\chapter{Using Captcha-4-letter dataset (99.87 \% accuracy)}

\section{Preprocessing}

\begin{figure}
	\centering
	\fbox{\includegraphics[width=0.5\linewidth]{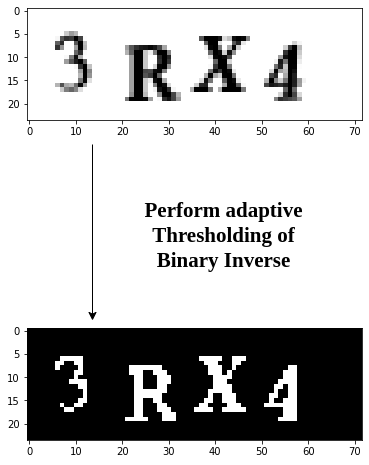}}
	\caption{Performing adaptive thresholding of Binary Inverse before passing it to the model.}
	\label{captcha-4-letter-thresh}
\end{figure}

We have converted the raw images to binary images, i.e., which have only two values, 0 and 255 by performing adaptive thresholding as shown in Figure \ref{captcha-4-letter-thresh}. When there is binary value, we can scale them down to 0 and 1 by dividing it with 255.0, so that the network is able to learn well from the data. This is one of the simplest types of CAPTCHA we have investigated with. The only challenge here is it has some connected letters which are hard to segment via any naive algorithm, so we have used a model to learn to predict all the letters at once. The data consists of about 9955 images of dimension 24x72x3. There is a total of 34 letters that are present in this CAPTCHA. Those letters includes all the capital letters and the digits.

\section{Naive Model for Captcha-4-letter dataset}

\begin{figure} 
	\centering
	\fbox{\includegraphics[height=24cm,width=0.95\linewidth]{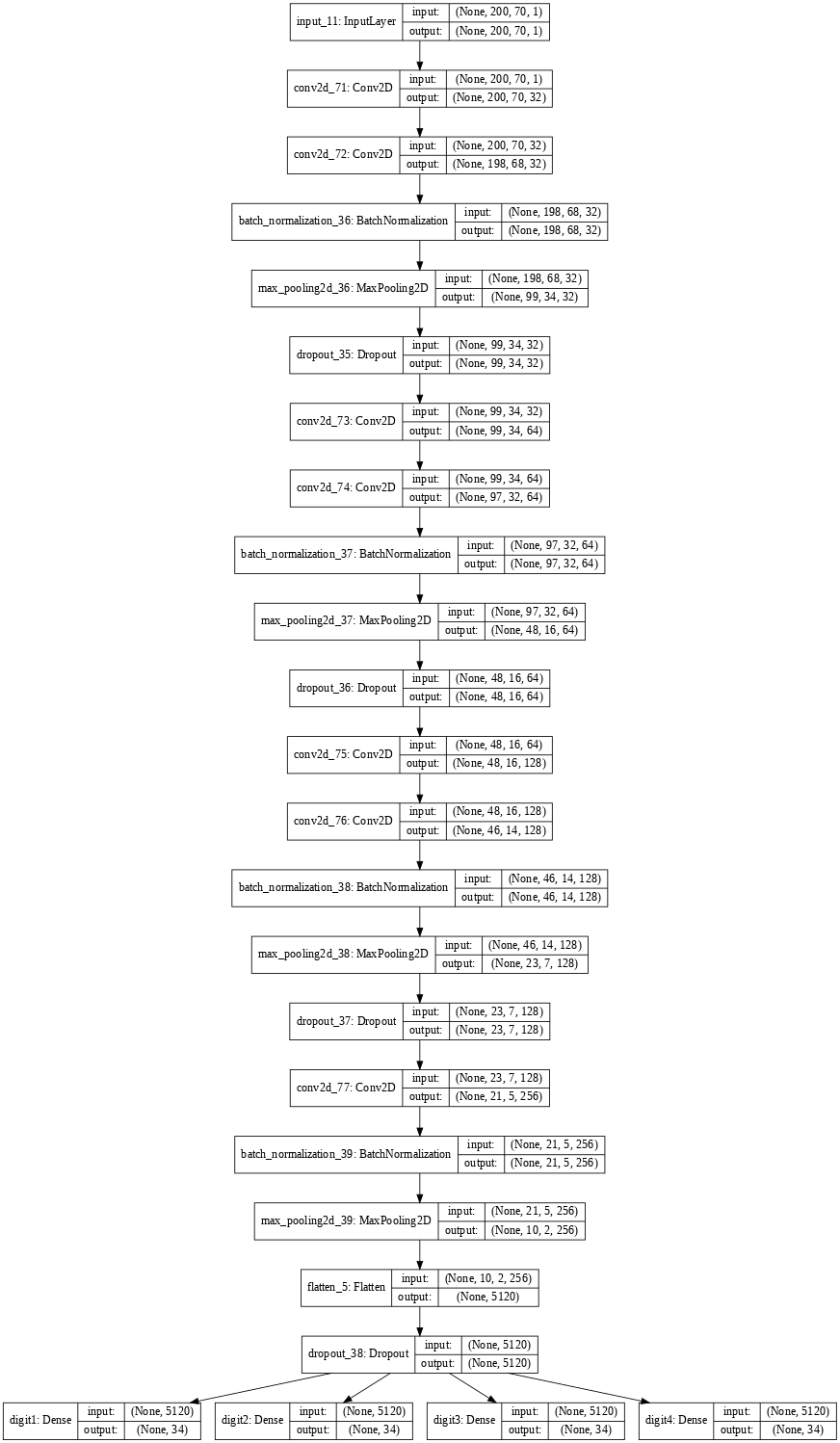}}	
	\caption{Our model for predicting the labels of the 4 letter CAPTCHA dataset.}
	\label{model_c4l_1}
\end{figure}

\begin{figure}
	\centering
	\includegraphics[height=4cm,width=0.985\linewidth]{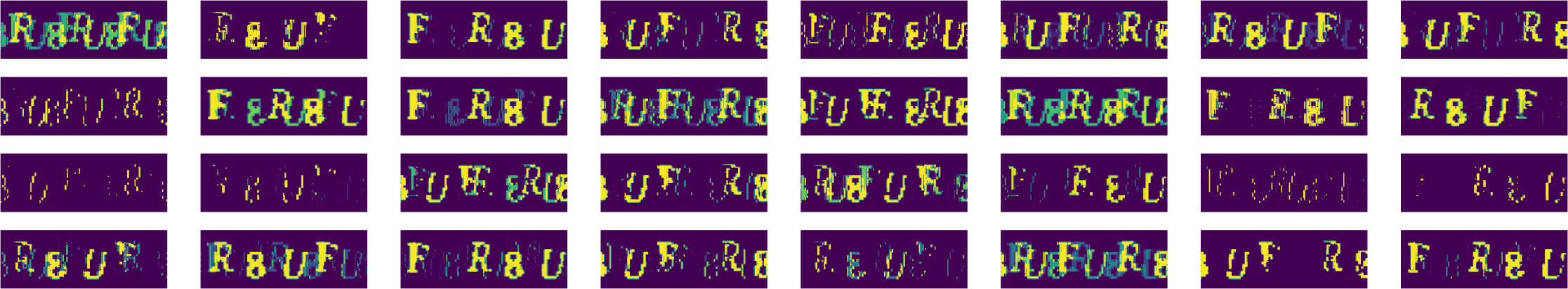}
	\includegraphics[height=3.9cm,width=0.5\linewidth]{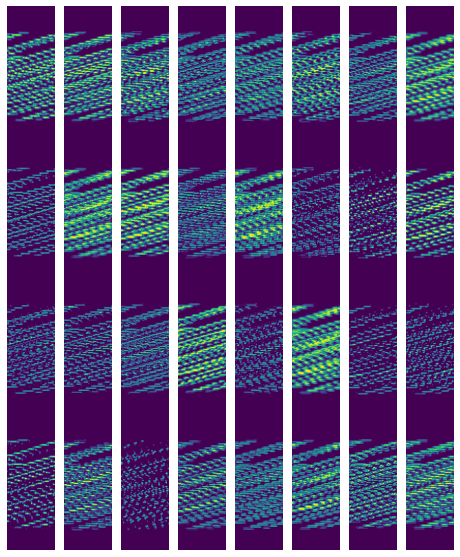}\includegraphics[height=3.9cm,width=0.5\linewidth]{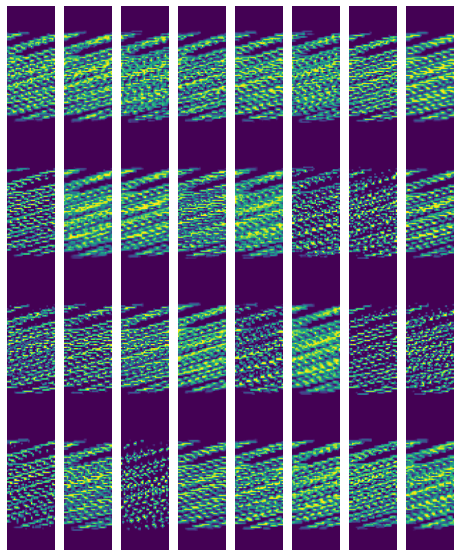}\vspace{2px}
	\includegraphics[height=3.9cm,width=0.5\linewidth]{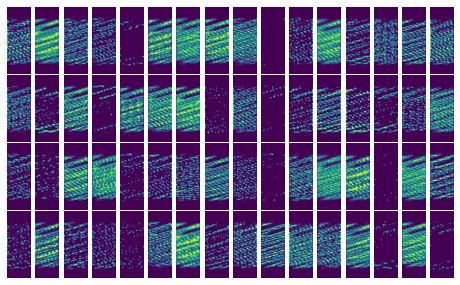}\includegraphics[height=3.9cm,width=0.5\linewidth]{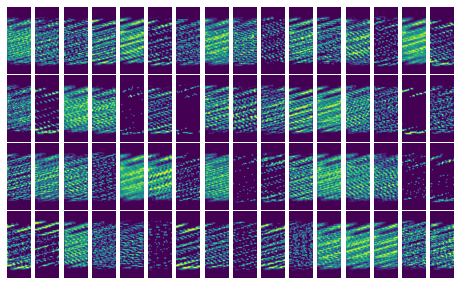}\vspace{2px}
	\includegraphics[height=3.9cm,width=0.5\linewidth]{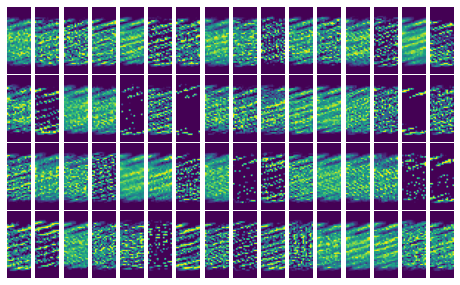}\includegraphics[height=4cm,width=0.5\linewidth]{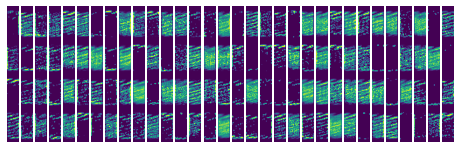}\vspace{2px}
	\includegraphics[height=3.9cm,width=0.5\linewidth]{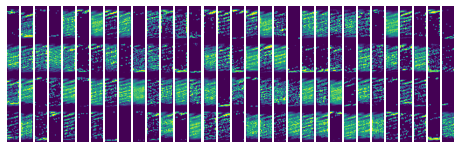}\includegraphics[height=3.9cm,width=0.5\linewidth]{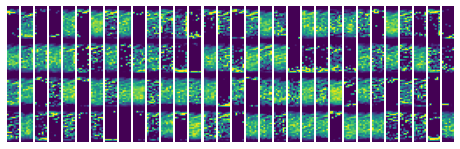}
	\includegraphics[height=3.9cm,width=0.5\linewidth]{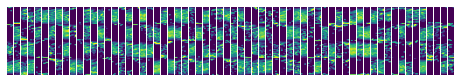}\includegraphics[height=3.9cm,width=0.5\linewidth]{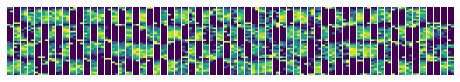}
	\caption{The visualizations of the features in a fully trained model when passed a CAPTCHA of FR8U, for Figure \ref{model_c4l_1}.}
	\label{CNN_c4l_layers_model1}
\end{figure}

\begin{figure}
	\centering
	\includegraphics[width=1\linewidth]{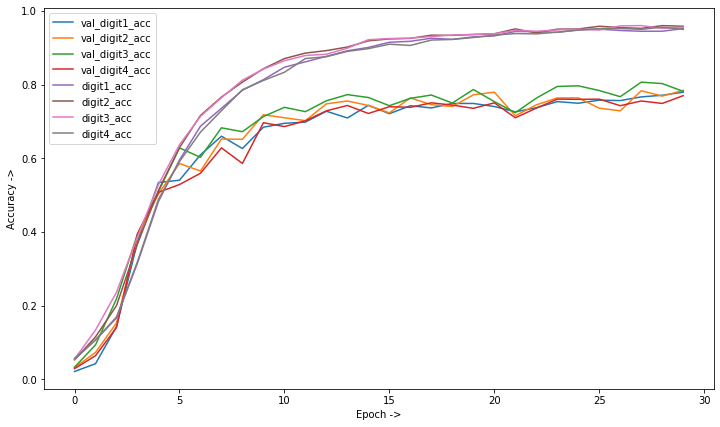}
	\caption{Plot for the training accuracy of our model.}
	\label{c4l-train-acc}
\end{figure}

\begin{figure}
	\centering
	\includegraphics[width=1\linewidth]{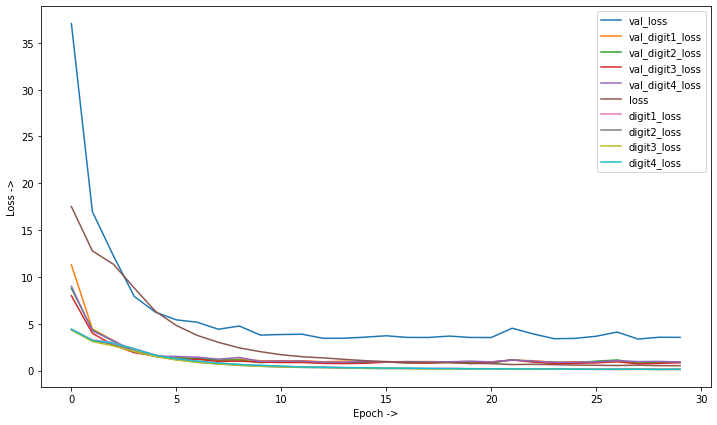}
	\caption{Plot for the training loss of our model.}
	\label{c4l-train-loss}
\end{figure}

The summary of the model is shown in Figure \ref{captcha-4-letter-modelsummary}, which is generated by\\ $model.summary()$. The model is trained quite fast and after a total of 30 epochs, we get the accuracy of 77\% as shown in Figure \ref{captcha-4-letter-epochs}. It takes about 9 seconds per epochs and there are a total of 30 epochs, which takes about 4.5 minutes to train the whole dataset, which is pretty good. The intermediate graph of the model is shown in Figure \ref{model_c4l_1}. There is a total of 34 label so, this is a multi-label classification, we can see that the model has 4 outgoing nodes at the end for this task. The intermediate layers of the model that is learnt is shown in Figure \ref{CNN_c4l_layers_model1}. We first tried this naive model and the plots for the accuracy and the loss obtained from the model is shown in Figure \ref{c4l-train-acc} and Figure \ref{c4l-train-loss}. After we have completed the training, we have tested on some random unseen data, and many were misclassified as shown in Figure \ref{pred_v1_model}. The trained model can be retrieved from this url (\url{https://jimut123.github.io/blogs/CAPTCHA/models/captcha_4_letter_our_model.h5}).

% https://tex.stackexchange.com/questions/180222/how-to-change-font-size-for-specific-lstlisting

% https://docs.opencv.org/master/d7/d1b/group__imgproc__misc.html#ggaa9e58d2860d4afa658ef70a9b1115576a19120b1a11d8067576cc24f4d2f03754

\begin{figure}
	\centering
	{%
		\lstset{frame=single,basicstyle=\scriptsize,style=myModelSummaryStyle}
		\centering
		\begin{lstlisting}
		
		__________________________________________________________________________________________________
		Layer (type)                    Output Shape         Param #     Connected to                     
		==================================================
		input_11 (InputLayer)           (None, 200, 70, 1)   0                                            
		__________________________________________________________________________________________________
		conv2d_71 (Conv2D)              (None, 200, 70, 32)  320         input_11[0][0]                   
		__________________________________________________________________________________________________
		conv2d_72 (Conv2D)              (None, 198, 68, 32)  9248        conv2d_71[0][0]                  
		__________________________________________________________________________________________________
		batch_normalization_36 (BatchNo (None, 198, 68, 32)  128         conv2d_72[0][0]                  
		__________________________________________________________________________________________________
		max_pooling2d_36 (MaxPooling2D) (None, 99, 34, 32)   0           batch_normalization_36[0][0]     
		__________________________________________________________________________________________________
		dropout_35 (Dropout)            (None, 99, 34, 32)   0           max_pooling2d_36[0][0]           
		__________________________________________________________________________________________________
		conv2d_73 (Conv2D)              (None, 99, 34, 64)   18496       dropout_35[0][0]                 
		__________________________________________________________________________________________________
		conv2d_74 (Conv2D)              (None, 97, 32, 64)   36928       conv2d_73[0][0]                  
		__________________________________________________________________________________________________
		batch_normalization_37 (BatchNo (None, 97, 32, 64)   256         conv2d_74[0][0]                  
		__________________________________________________________________________________________________
		max_pooling2d_37 (MaxPooling2D) (None, 48, 16, 64)   0           batch_normalization_37[0][0]     
		__________________________________________________________________________________________________
		dropout_36 (Dropout)            (None, 48, 16, 64)   0           max_pooling2d_37[0][0]           
		__________________________________________________________________________________________________
		conv2d_75 (Conv2D)              (None, 48, 16, 128)  73856       dropout_36[0][0]                 
		__________________________________________________________________________________________________
		conv2d_76 (Conv2D)              (None, 46, 14, 128)  147584      conv2d_75[0][0]                  
		__________________________________________________________________________________________________
		batch_normalization_38 (BatchNo (None, 46, 14, 128)  512         conv2d_76[0][0]                  
		__________________________________________________________________________________________________
		max_pooling2d_38 (MaxPooling2D) (None, 23, 7, 128)   0           batch_normalization_38[0][0]     
		__________________________________________________________________________________________________
		dropout_37 (Dropout)            (None, 23, 7, 128)   0           max_pooling2d_38[0][0]           
		__________________________________________________________________________________________________
		conv2d_77 (Conv2D)              (None, 21, 5, 256)   295168      dropout_37[0][0]                 
		__________________________________________________________________________________________________
		batch_normalization_39 (BatchNo (None, 21, 5, 256)   1024        conv2d_77[0][0]                  
		__________________________________________________________________________________________________
		max_pooling2d_39 (MaxPooling2D) (None, 10, 2, 256)   0           batch_normalization_39[0][0]     
		__________________________________________________________________________________________________
		flatten_5 (Flatten)             (None, 5120)         0           max_pooling2d_39[0][0]           
		__________________________________________________________________________________________________
		dropout_38 (Dropout)            (None, 5120)         0           flatten_5[0][0]                  
		__________________________________________________________________________________________________
		digit1 (Dense)                  (None, 34)           174114      dropout_38[0][0]                 
		__________________________________________________________________________________________________
		digit2 (Dense)                  (None, 34)           174114      dropout_38[0][0]                 
		__________________________________________________________________________________________________
		digit3 (Dense)                  (None, 34)           174114      dropout_38[0][0]                 
		__________________________________________________________________________________________________
		digit4 (Dense)                  (None, 34)           174114      dropout_38[0][0]                 
		==================================================
		Total params: 1,279,976
		Trainable params: 1,279,016
		Non-trainable params: 960
		__________________________________________________________________________________________________
		\end{lstlisting}
	}
	\caption{Summary of the model for \ref{model_c4l_1}.}
	\label{captcha-4-letter-modelsummary}
\end{figure}

\begin{figure}
	\centering
	{%
		\lstset{frame=single,basicstyle=\scriptsize,style=myModelSummaryStyle}
		\begin{lstlisting}
			Epoch 30/30
			7200/7200 [==============================] - 9s 1ms/step - loss: 0.5274 - digit1_loss: 0.1373 - digit2_loss: 0.1242 - digit3_loss: 0.1226 - digit4_loss: 0.1434 - digit1_acc: 0.9519 - digit2_acc: 0.9583 - digit3_acc: 0.9558 - digit4_acc: 0.9497 - val_loss: 3.5470 - val_digit1_loss: 0.9227 - val_digit2_loss: 0.8700 - val_digit3_loss: 0.8337 - val_digit4_loss: 0.9206 - val_digit1_acc: 0.7794 - val_digit2_acc: 0.7844 - val_digit3_acc: 0.7817 - val_digit4_acc: 0.7700
		\end{lstlisting}
	}
	\caption{The summary of 30 epochs for the model \ref{model_c4l_1}.}
	\label{captcha-4-letter-epochs}
\end{figure}

\begin{figure}
	\centering
	\includegraphics[width=1\linewidth]{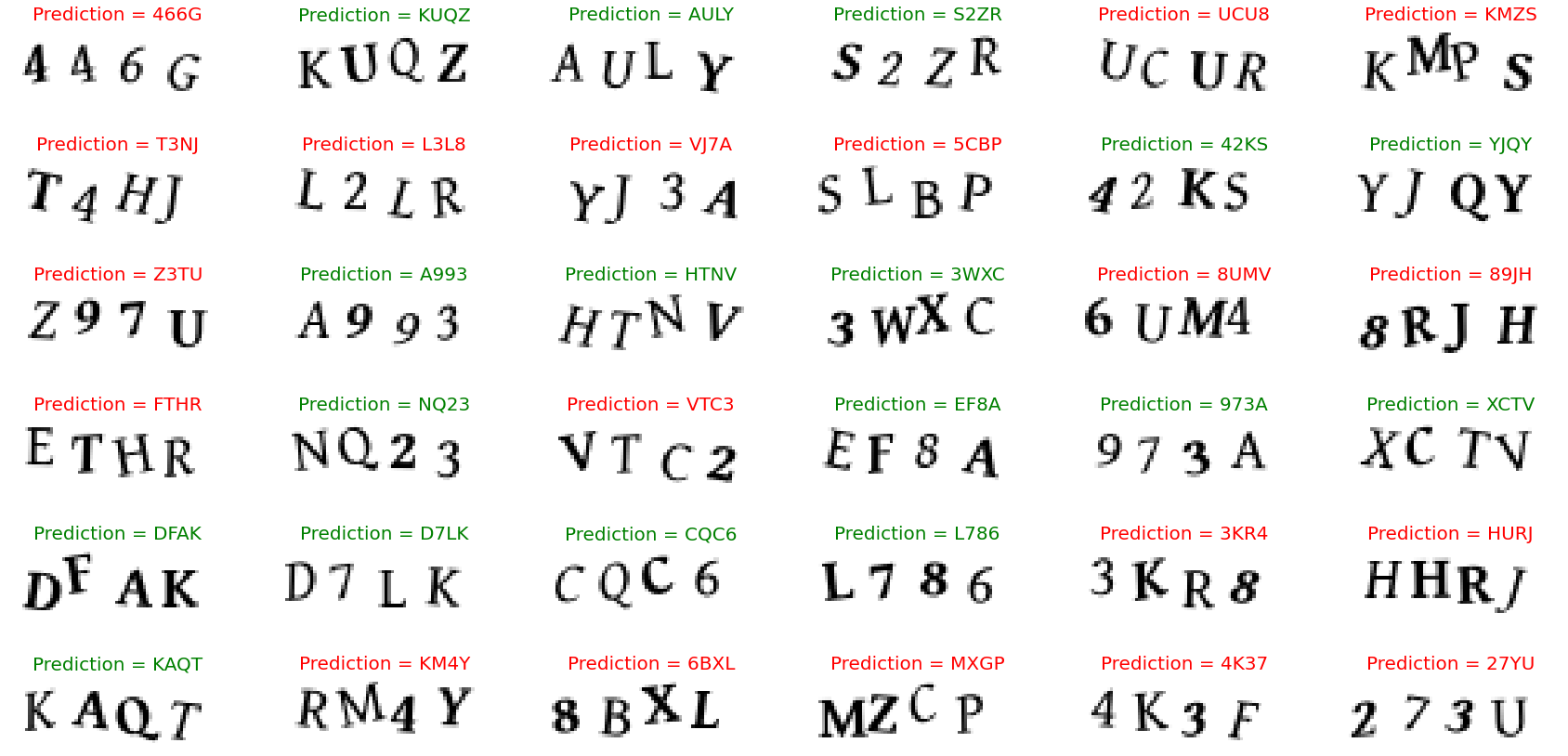}
	\caption{Prediction of the model on unseen data.}
	\label{pred_unseen}
\end{figure}

\section{A better Model for Captcha-4-letter dataset}

\begin{figure} 
	\centering
	\fbox{\includegraphics[height=15cm,width=1\linewidth]{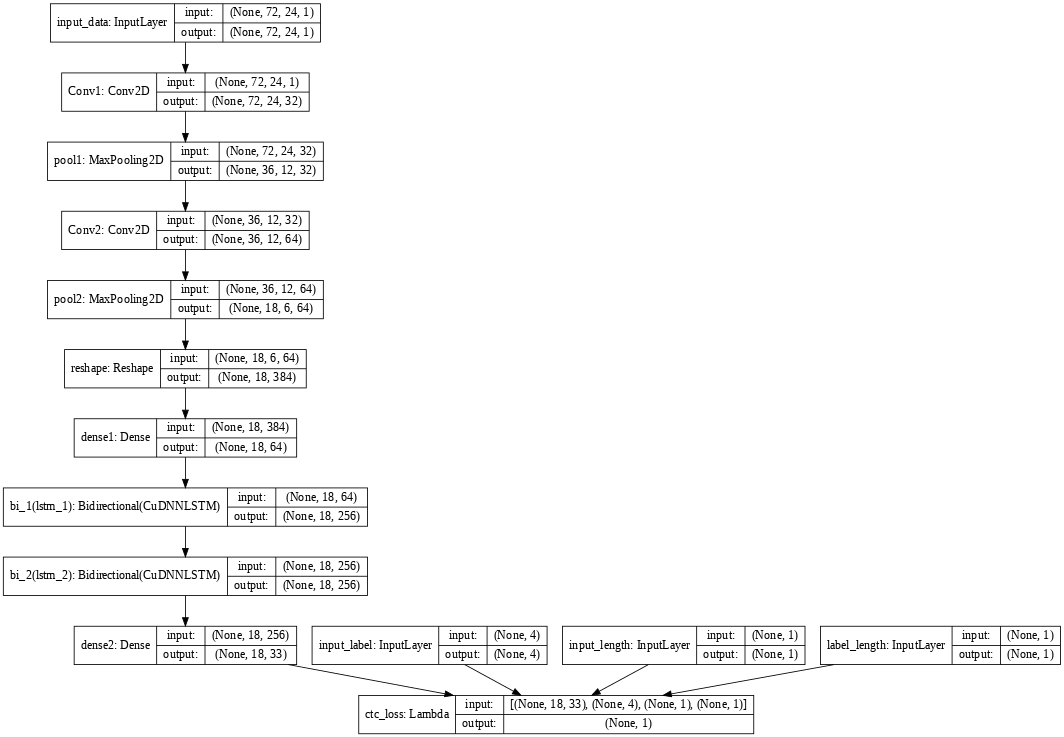}}	
	\caption{Version 2 of the model for identifying individual characters.}
	\label{model_c4l_2}
\end{figure}

\begin{figure}
	\centering
	\includegraphics[width=1\linewidth]{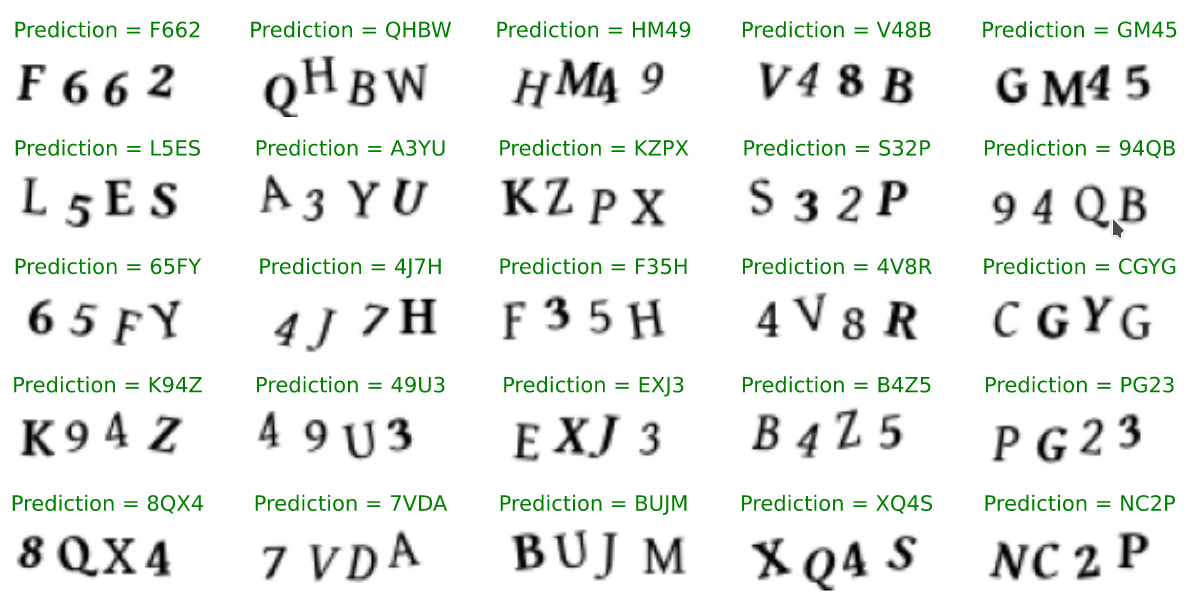}
	\caption{Prediction of the new model (\ref{model_c4l_2}) on unseen data.}
	\label{pred_v1_model}
\end{figure}

\begin{figure}
	\centering
	\includegraphics[width=1\linewidth]{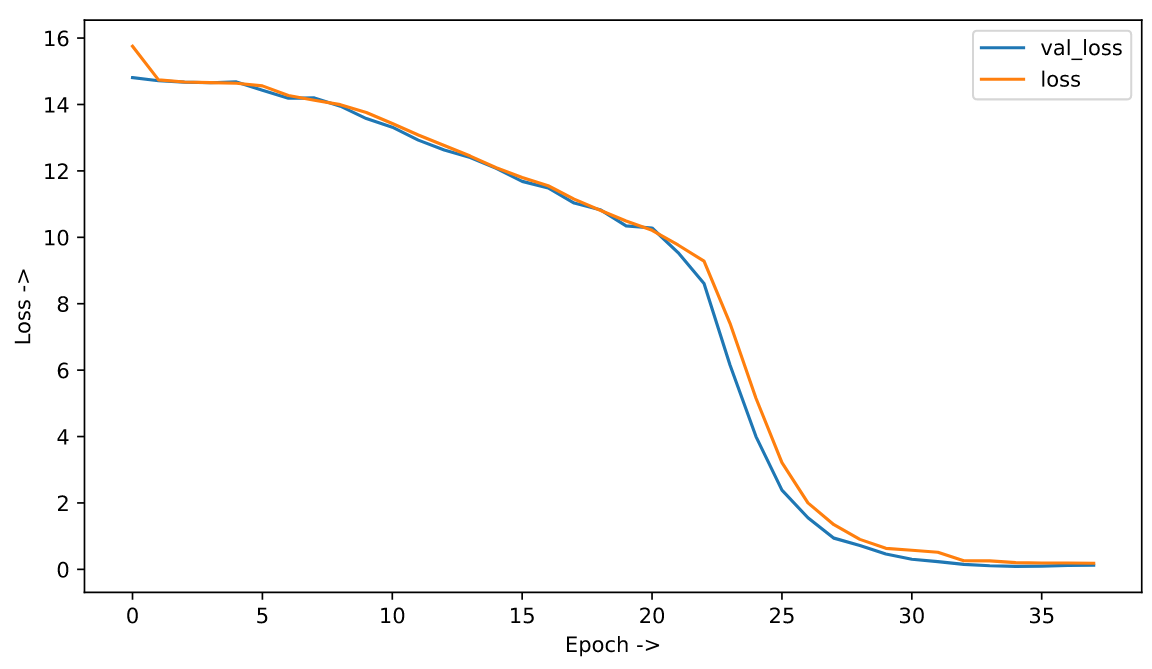}
	\caption{Loss of the new model (\ref{model_c4l_2}).}
	\label{loss_v1_model}
\end{figure}

After a lot of fine tuning the hyperparameters, we found that we cannot improve much on the above model. We then came up with a new model as shown in Figure \ref{model_c4l_2_summary}. The model took about 4.43 minutes in total of 38 epochs which gave an accuracy of 99.87\% in total. There was careful design of TensorFlow callbacks which led to early stopping as shown in Figure \ref{captcha-4-letter-model2-epochs}. The graph of the same model is shown in Figure \ref{model_c4l_2}. The trained model can be retrieved from this url (\url{https://jimut123.github.io/blogs/CAPTCHA/models/model_captcha_4_letter_v1.h5}).  We test the accuracy of the model on some real-world unseen data as shown in Figure \ref{pred_v1_model}, which shows that it operates in a human level accuracy with all of the predictions correct. The loss of the model is shown in Figure \ref{loss_v1_model}.

\begin{figure}
	\centering
	
	{%
		\lstset{frame=single,basicstyle=\scriptsize,style=myModelSummaryStyle}
		\centering
		\begin{lstlisting}
			__________________________________________________________________________________________________
			Layer (type)                    Output Shape         Param #     Connected to                     
			==================================================
			input_data (InputLayer)         (None, 72, 24, 1)    0                                            
			__________________________________________________________________________________________________
			Conv1 (Conv2D)                  (None, 72, 24, 32)   320         input_data[0][0]                 
			__________________________________________________________________________________________________
			pool1 (MaxPooling2D)            (None, 36, 12, 32)   0           Conv1[0][0]                      
			__________________________________________________________________________________________________
			Conv2 (Conv2D)                  (None, 36, 12, 64)   18496       pool1[0][0]                      
			__________________________________________________________________________________________________
			pool2 (MaxPooling2D)            (None, 18, 6, 64)    0           Conv2[0][0]                      
			__________________________________________________________________________________________________
			reshape (Reshape)               (None, 18, 384)      0           pool2[0][0]                      
			__________________________________________________________________________________________________
			dense1 (Dense)                  (None, 18, 64)       24640       reshape[0][0]                    
			__________________________________________________________________________________________________
			bi_1 (Bidirectional)            (None, 18, 256)      198656      dense1[0][0]                     
			__________________________________________________________________________________________________
			bi_2 (Bidirectional)            (None, 18, 256)      395264      bi_1[0][0]                       
			__________________________________________________________________________________________________
			dense2 (Dense)                  (None, 18, 33)       8481        bi_2[0][0]                       
			__________________________________________________________________________________________________
			input_label (InputLayer)        (None, 4)            0                                            
			__________________________________________________________________________________________________
			input_length (InputLayer)       (None, 1)            0                                            
			__________________________________________________________________________________________________
			label_length (InputLayer)       (None, 1)            0                                            
			__________________________________________________________________________________________________
			ctc_loss (Lambda)               (None, 1)            0           dense2[0][0]                     
			input_label[0][0]                
			input_length[0][0]               
			label_length[0][0]               
			==================================================
			Total params: 645,857
			Trainable params: 645,857
			Non-trainable params: 0
			__________________________________________________________________________________________________
		\end{lstlisting}
	}
	\caption{Summary of the Version 2 of the Model}
	\label{model_c4l_2_summary}
\end{figure}

\begin{figure}
	\centering
	{%
		\lstset{frame=single,basicstyle=\scriptsize,style=myModelSummaryStyle}
		\begin{lstlisting}
			Epoch 38/50
			279/279 [==============================] - 7s 26ms/step - loss: 0.1842 - val_loss: 0.1233
		\end{lstlisting}
	}
	\caption{The epochs for version 2 of the model. }
	\label{captcha-4-letter-model2-epochs}
\end{figure}

\begin{figure}
	\centering
	\includegraphics[width=0.5\linewidth]{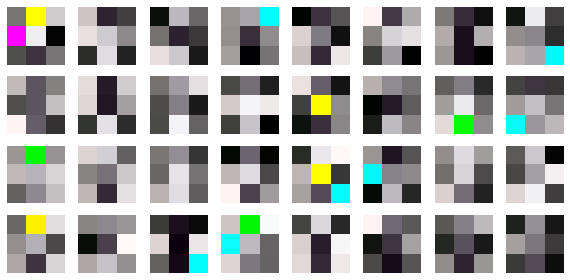}
	\caption{The 3x3 kernels (filter) learned by the model (\ref{model_c4l_2}) for layer 1.}
	\label{c4l-filter-learned}
\end{figure}

\begin{figure}
	\centering
	\includegraphics[width=0.85\linewidth]{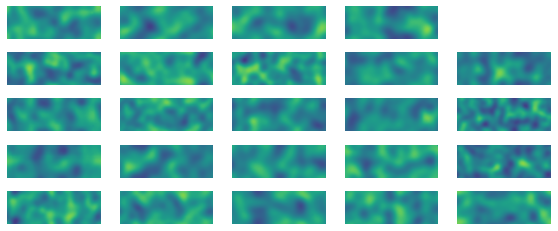}
	\caption{The weights learned by the model (\ref{model_c4l_2}) for layers 1 to 24.}
	\label{c4l-weights-learned}
\end{figure}

In Figure \ref{c4l-filter-learned} we see the 3x3 kernels that are being learned by the layer 1 of the network for the Model \ref{model_c4l_2}. The visualization of the weights learned by the same model can be seen in Figure \ref{c4l-weights-learned}.

\begin{figure}
	\centering
	\includegraphics[width=0.85\linewidth]{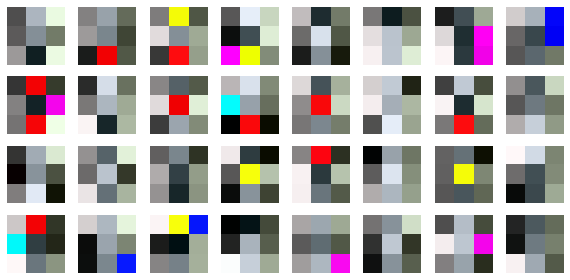}
	\caption{The 3x3 kernels (filter) learned by the model (\ref{model_1Lc_1}) for layer 1.}
	\label{1Lc-filter-learned}
\end{figure}

\begin{figure}
	\centering
	\includegraphics[width=0.85\linewidth]{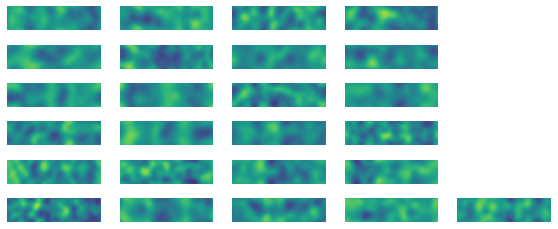}
	\caption{The weights learned by the model (\ref{model_1Lc_1}) for layers 1 to 26.}
	\label{1Lc-weights-learned}
\end{figure}

\chapter{Using c4l\_16x16\_550 dataset (99.91 \% accuracy)}

\section{Model for c4l\_16x16\_550 dataset}

We have used a modified version of CIFAR 10 like model for predicting the segmented images. The summary of the architecture of the model is shown below

{%
	\centering
	\begin{lstlisting}
	
	_________________________________________________________________
	Layer (type)                 Output Shape              Param #   
	=================================================================
	conv2d_1 (Conv2D)            (None, 16, 16, 32)        320       
	_________________________________________________________________
	activation_1 (Activation)    (None, 16, 16, 32)        0         
	_________________________________________________________________
	conv2d_2 (Conv2D)            (None, 14, 14, 32)        9248      
	_________________________________________________________________
	activation_2 (Activation)    (None, 14, 14, 32)        0         
	_________________________________________________________________
	max_pooling2d_1 (MaxPooling2 (None, 7, 7, 32)          0         
	_________________________________________________________________
	dropout_1 (Dropout)          (None, 7, 7, 32)          0         
	_________________________________________________________________
	conv2d_3 (Conv2D)            (None, 7, 7, 64)          18496     
	_________________________________________________________________
	activation_3 (Activation)    (None, 7, 7, 64)          0         
	_________________________________________________________________
	conv2d_4 (Conv2D)            (None, 5, 5, 64)          36928     
	_________________________________________________________________
	activation_4 (Activation)    (None, 5, 5, 64)          0         
	_________________________________________________________________
	max_pooling2d_2 (MaxPooling2 (None, 2, 2, 64)          0         
	_________________________________________________________________
	dropout_2 (Dropout)          (None, 2, 2, 64)          0         
	_________________________________________________________________
	flatten_1 (Flatten)          (None, 256)               0         
	_________________________________________________________________
	dense_1 (Dense)              (None, 512)               131584    
	_________________________________________________________________
	activation_5 (Activation)    (None, 512)               0         
	_________________________________________________________________
	dropout_3 (Dropout)          (None, 512)               0         
	_________________________________________________________________
	dense_2 (Dense)              (None, 32)                16416     
	_________________________________________________________________
	activation_6 (Activation)    (None, 32)                0         
	=================================================================
	Total params: 212,992
	Trainable params: 212,992
	Non-trainable params: 0
	_________________________________________________________________
		
	\end{lstlisting}
}

\begin{figure} 
	\centering
	\fbox{\includegraphics[height=24cm,width=0.55\linewidth]{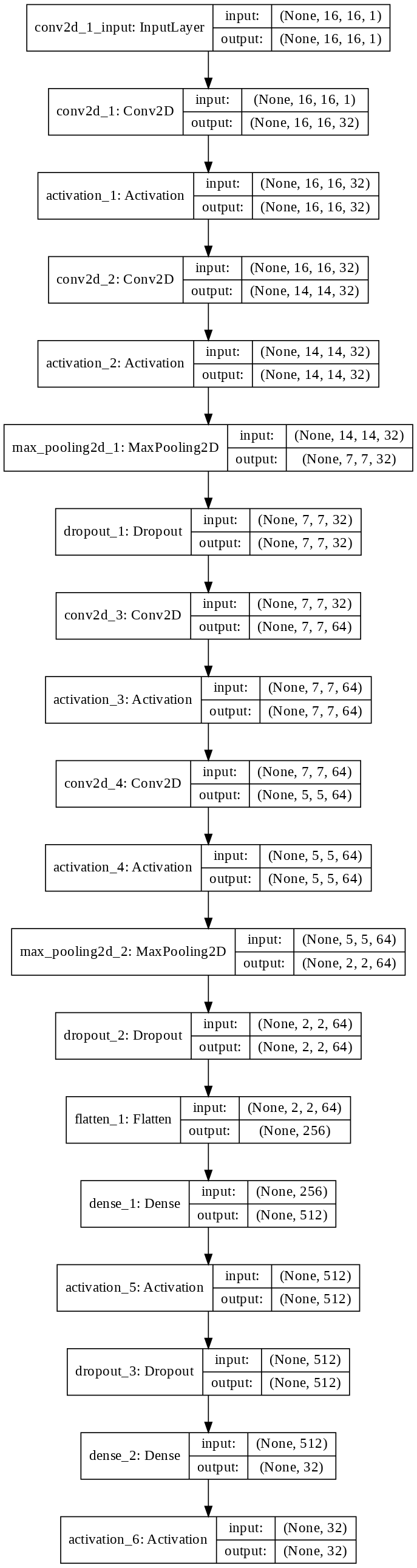}}	
	\caption{CIFAR 10 like model for identifying individual characters.}
	\label{model_segmented_c4l_16x16_550}
\end{figure}

After about 20 epochs, we got an accuracy of 99.91 \% on the validation split. We find that the training of the individual split characters is very fast due to its small size and uniform distribution. It took a total of about 0.6 minutes to train on the whole dataset. We further note that the accuracy is beyond the model used in Figure \ref{model_c4l_2}. This operates in human level accuracy. It might be due to overfitting but this works well in general.

\begin{figure} 
	\centering
	\begin{lstlisting}
	Epoch 20/20
	14105/14105 [==============================] - 2s 172us/step - loss: 0.0257 - acc: 0.9950 - val_loss: 0.0140 - val_acc: 0.9991
	\end{lstlisting}
	\caption{epochs.}
\label{model_epochs_c4l_16x16_550}
\end{figure}

The visualization of the architecture of the model is shown in Figure ~\ref{model_segmented_c4l_16x16_550}.

\begin{figure} 
	\centering
	\includegraphics[width=0.75\linewidth]{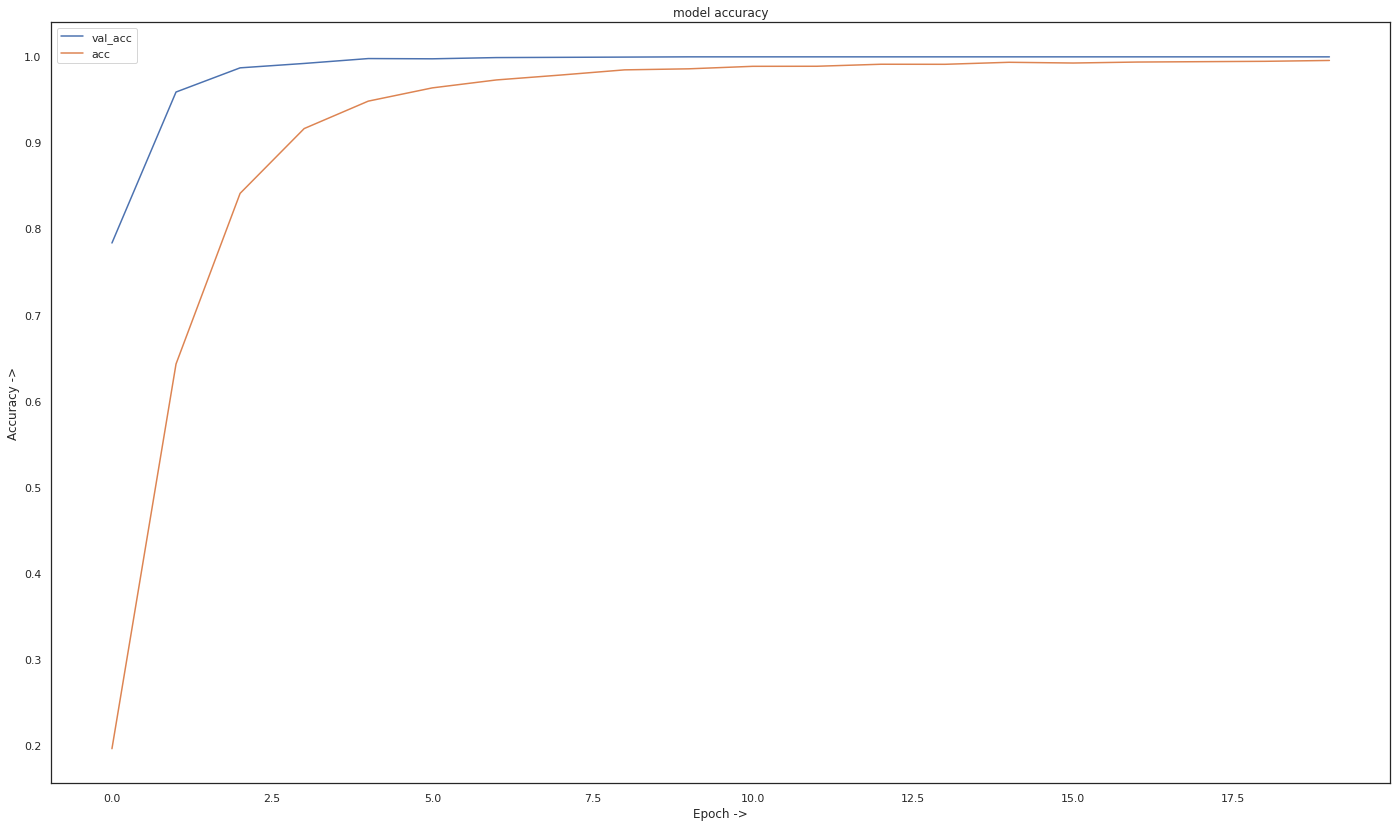}
	\caption{Accuracy of the model as shown in Figure ~\ref{model_segmented_c4l_16x16_550}}
	\label{model_accuracy_segmented_c4l_16x16_550}
\end{figure}

\begin{figure}
	\centering
	\includegraphics[width=0.75\linewidth]{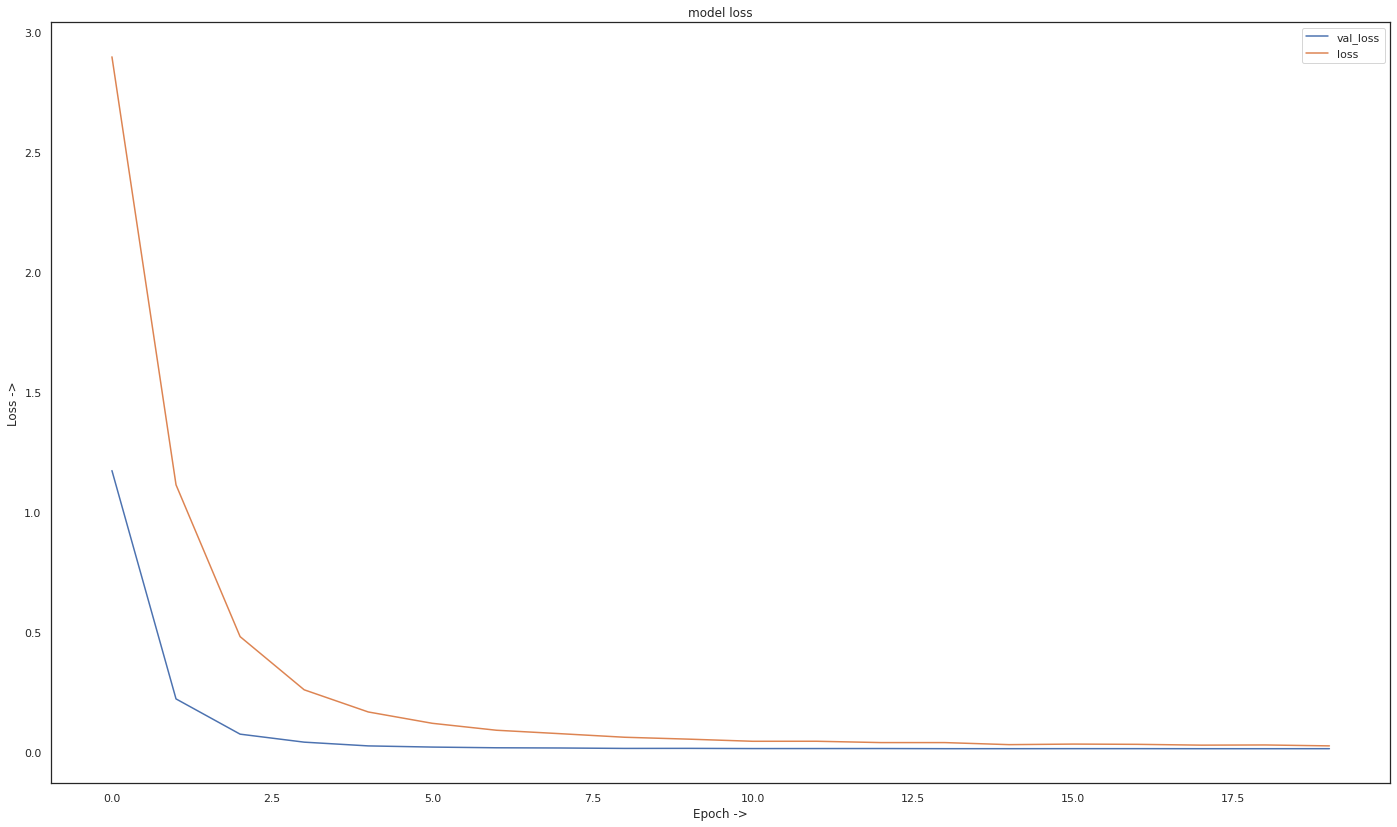}
	\caption{Accuracy of the model as shown in Figure ~\ref{model_segmented_c4l_16x16_550}}
	\label{model_loss_segmented_c4l_16x16_550}
\end{figure}

The graph of the accuracy of the model is shown in Figure ~\ref{model_accuracy_segmented_c4l_16x16_550} and the loss of the model is shown in Figure ~\ref{model_loss_segmented_c4l_16x16_550}. The trained model can be retrieved from this url (\url{https://jimut123.github.io/blogs/CAPTCHA/models/CNN_c4l-16x16_550.h5}).

\begin{figure}
	\centering
	\includegraphics[width=0.75\linewidth]{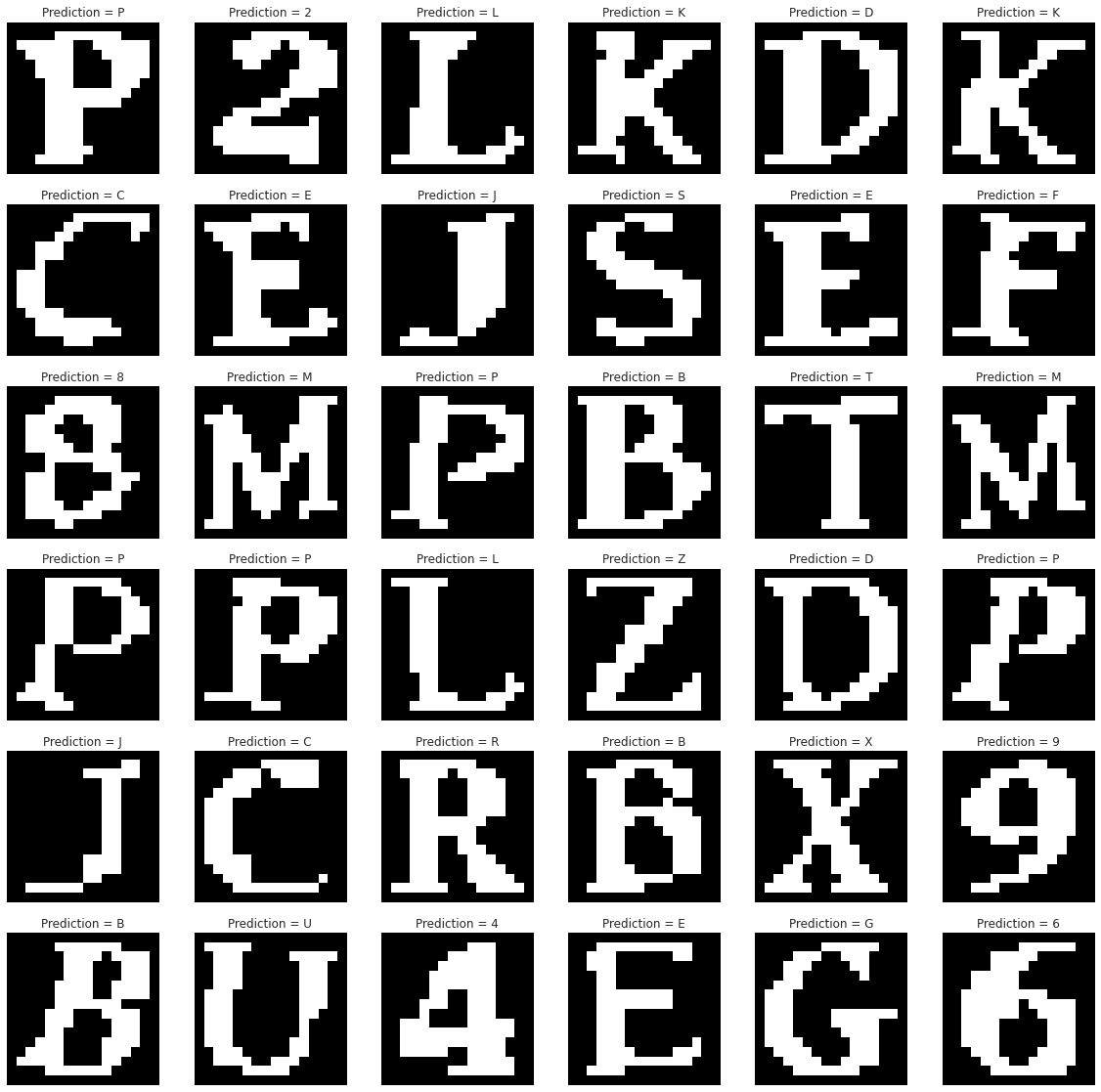}	
	\caption{The prediction of the model from the individual characters}
	\label{CNN_c4l_16x16_550_prediction}
\end{figure}

We can see that the prediction of the model on individual character is almost with 100\% accuracy as shown in Figure \ref{CNN_c4l_16x16_550_prediction}.

\begin{figure}
	\centering
	\includegraphics[width=1\linewidth]{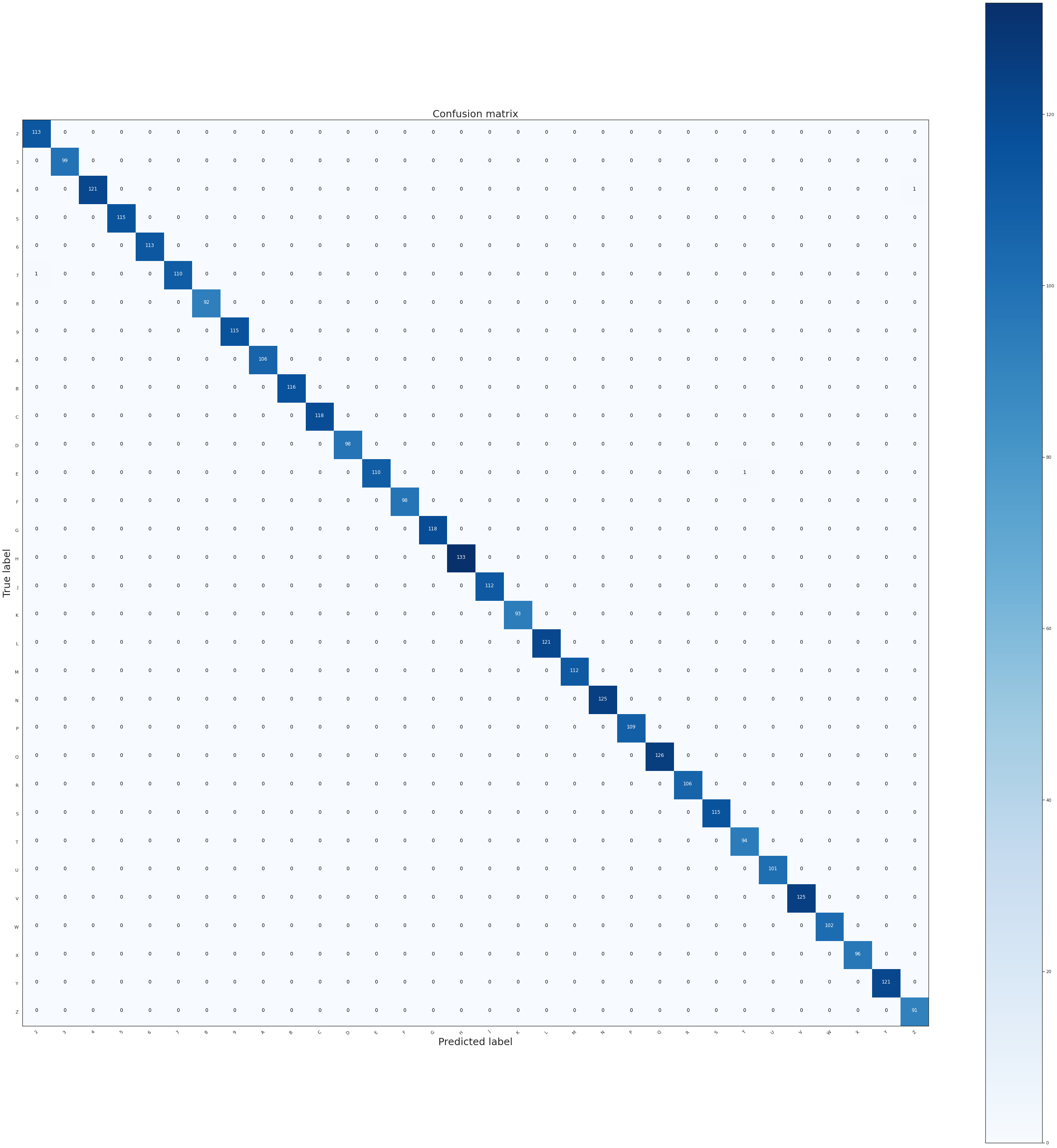}
	\caption{The confusion matrix of the individual characters}
	\label{CNN_c4l_16x16_550_confusion_matrix}
\end{figure}

The confusion matrix for individual character predicted is shown in Figure \ref{CNN_c4l_16x16_550_confusion_matrix}.

\begin{figure}
	\centering
	\includegraphics[height=3.45cm,width=0.5\linewidth]{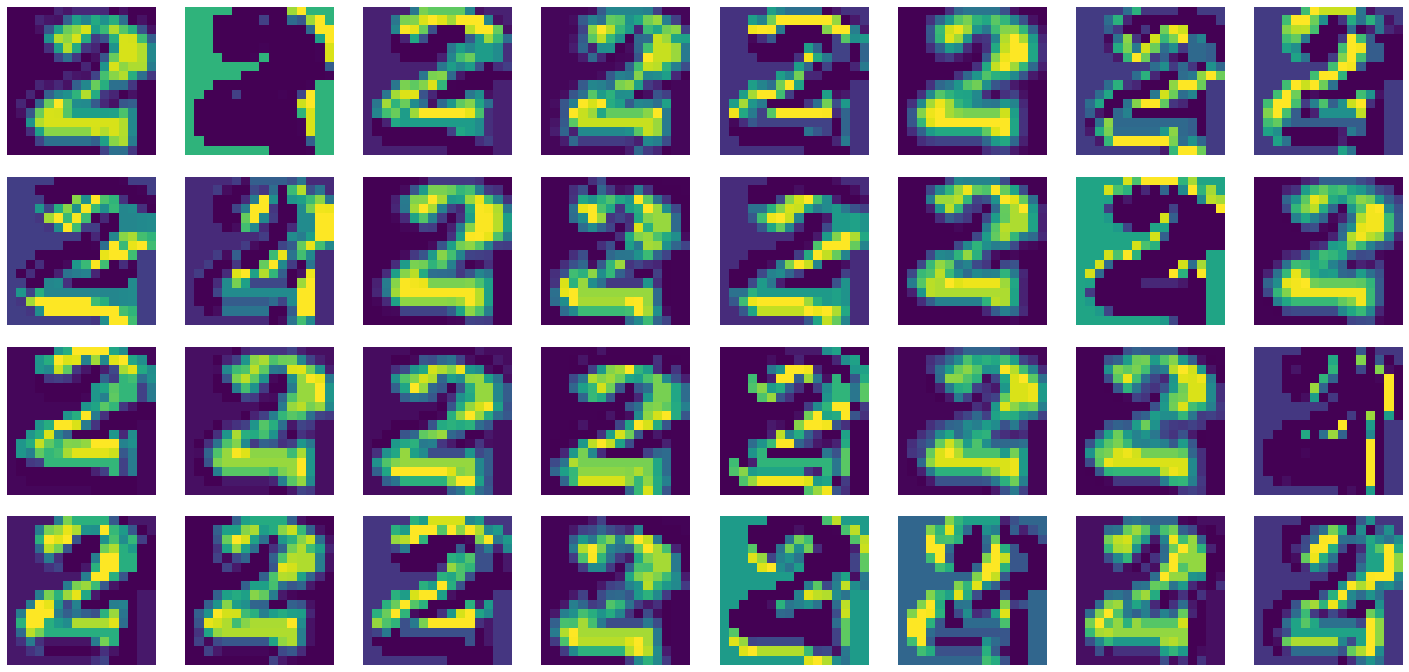}\includegraphics[height=3.45cm,width=0.5\linewidth]{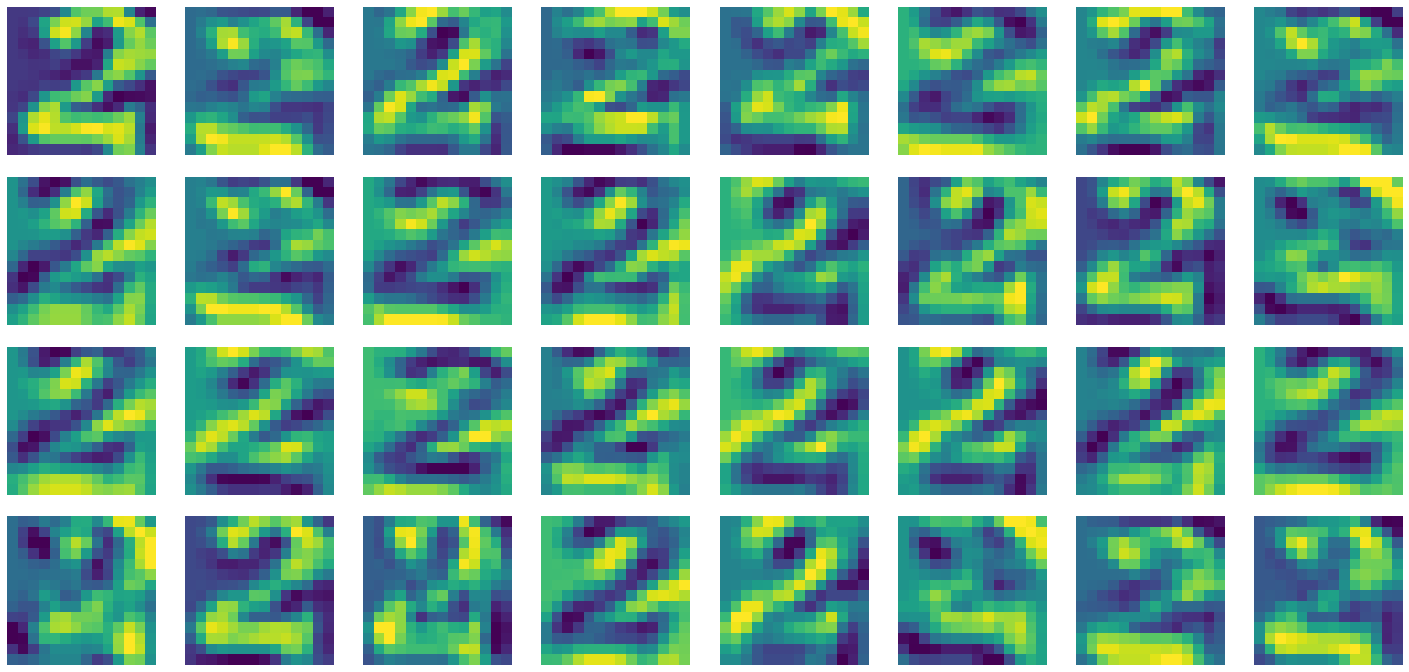}
	\includegraphics[height=3.45cm,width=0.5\linewidth]{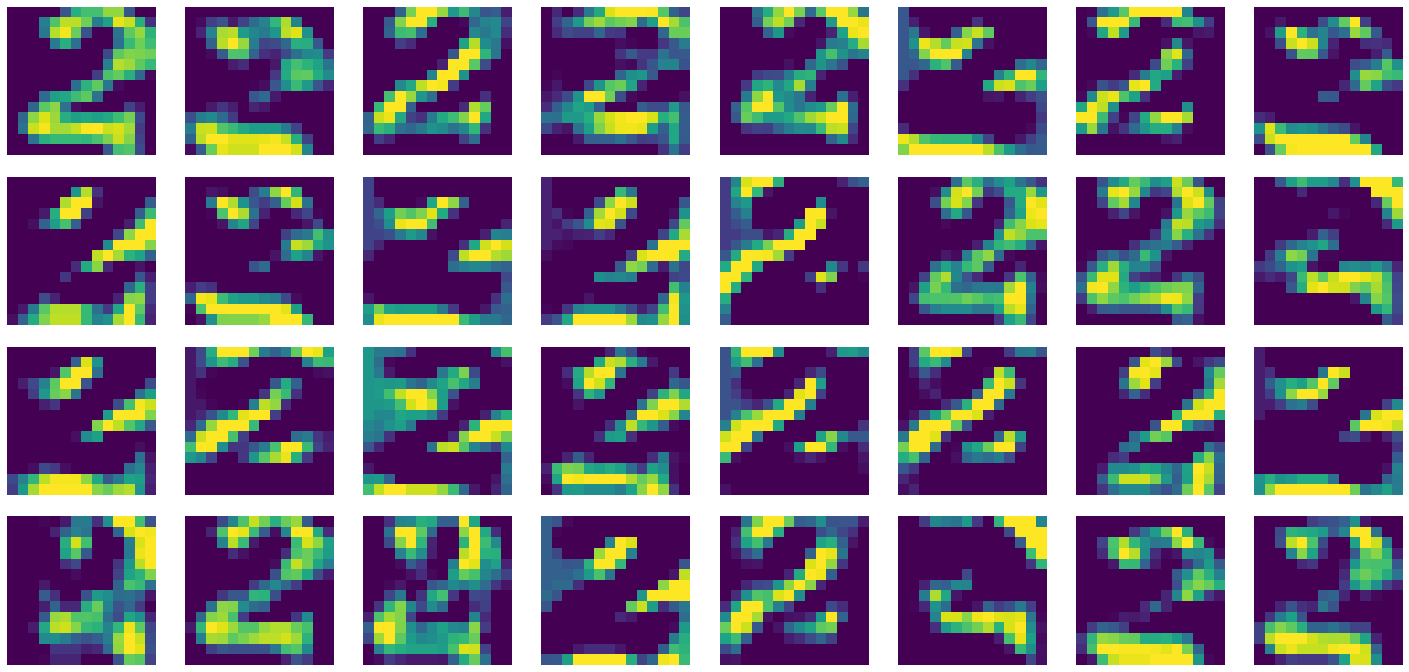}\includegraphics[height=3.45cm,width=0.5\linewidth]{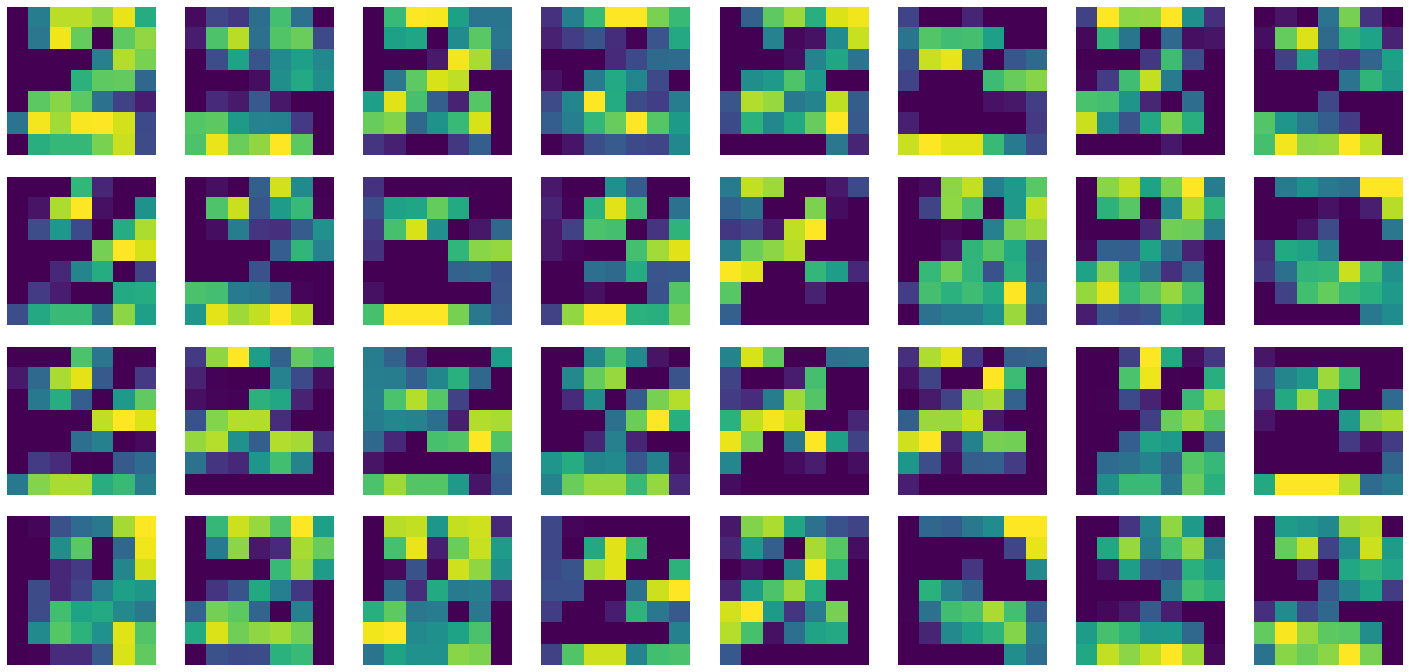}
	\includegraphics[height=3.45cm,width=1\linewidth]{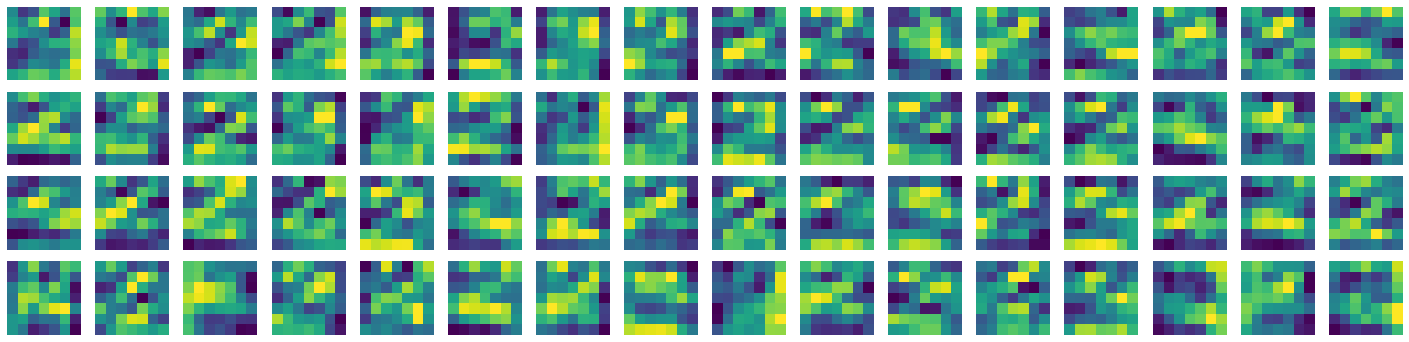}
	\includegraphics[height=3.45cm,width=1\linewidth]{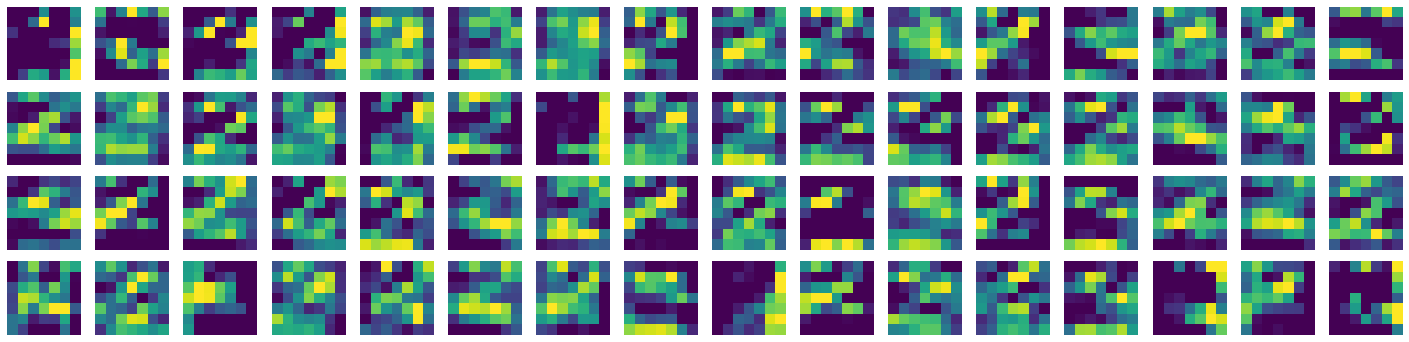}
	\includegraphics[height=3.45cm,width=1\linewidth]{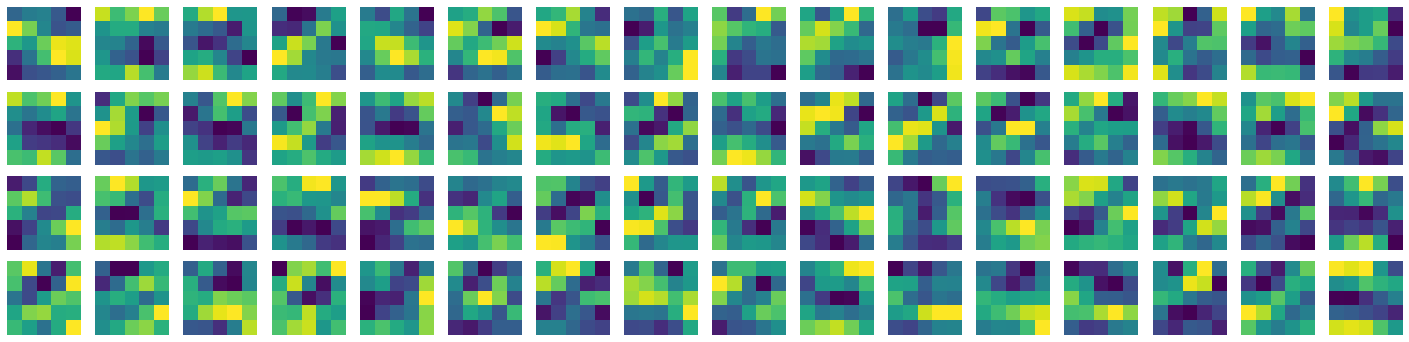}
	\includegraphics[height=3.45cm,width=1\linewidth]{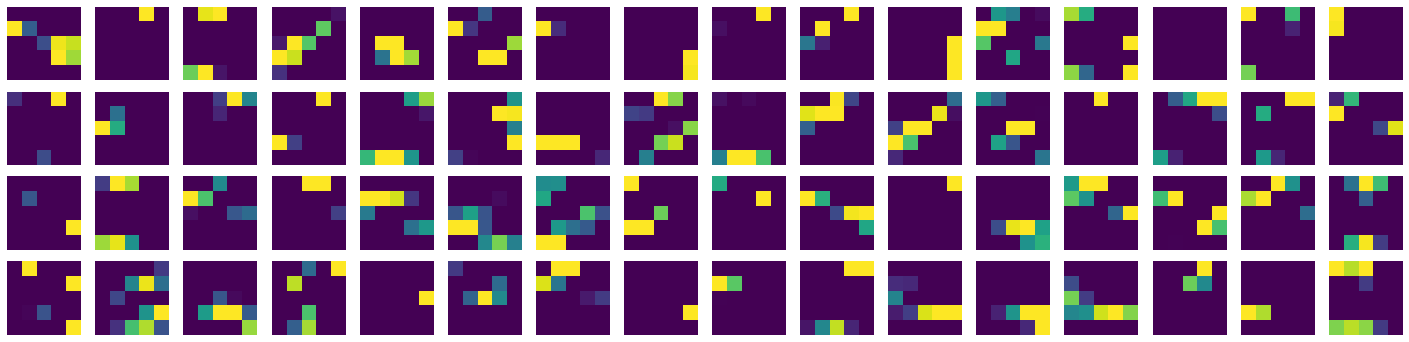}
	\includegraphics[height=3.45cm,width=1\linewidth]{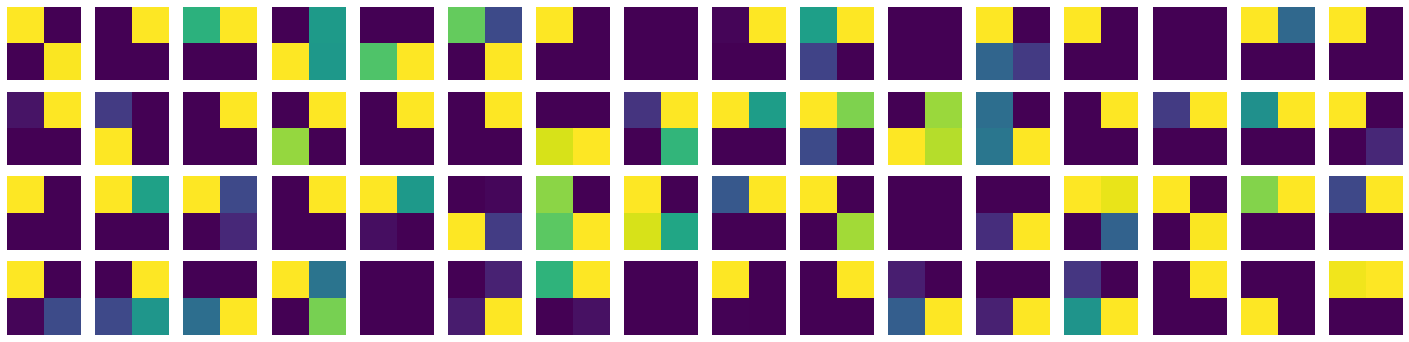}
	\caption{The visualizations of the features in a fully trained model when passed an image for letter 2.}
	\label{CNN_c4l_16x16_550_DL_2}
\end{figure}

\begin{figure}
	\centering
	\includegraphics[height=3.45cm,width=0.5\linewidth]{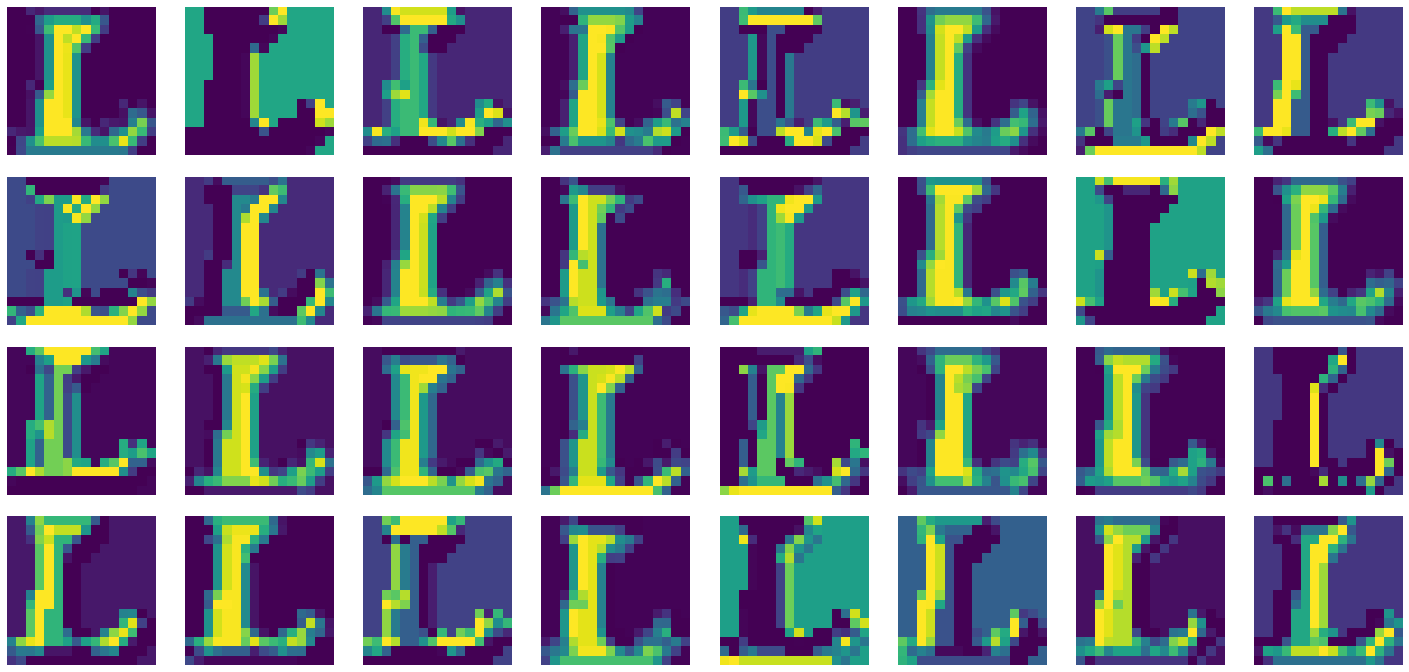}\includegraphics[height=3.45cm,width=0.5\linewidth]{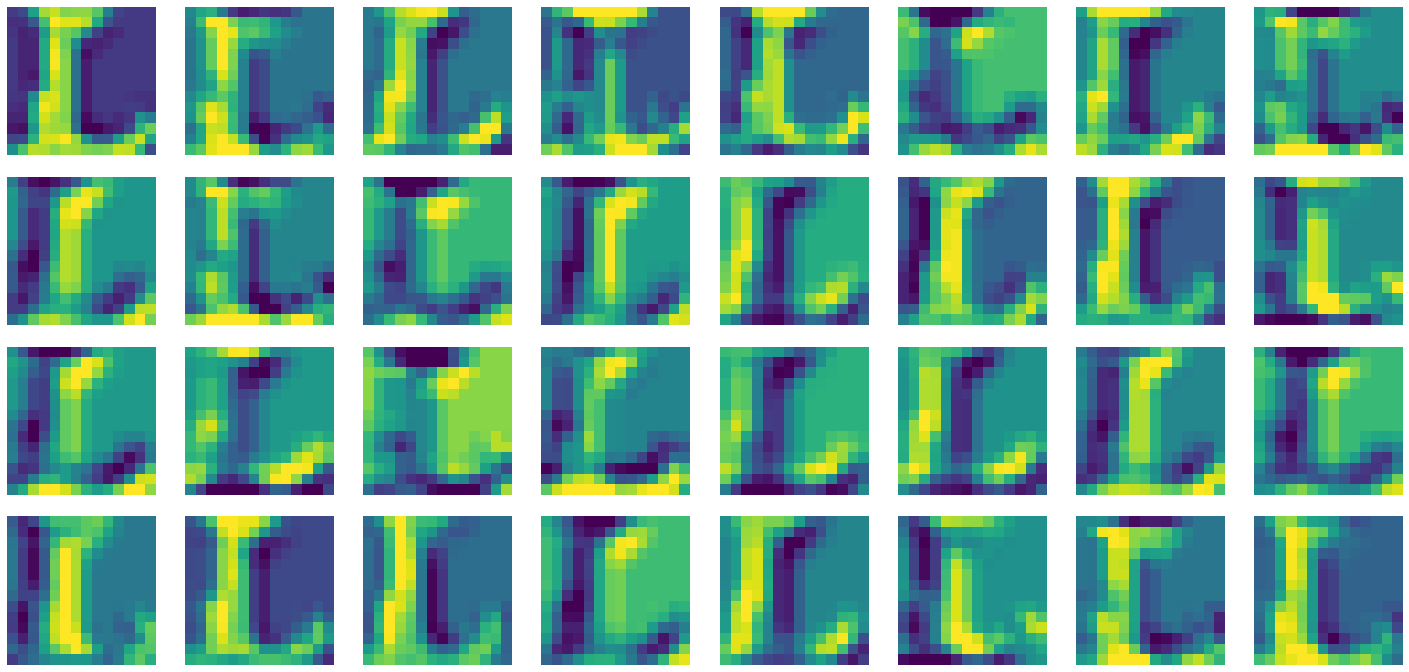}
	\includegraphics[height=3.45cm,width=0.5\linewidth]{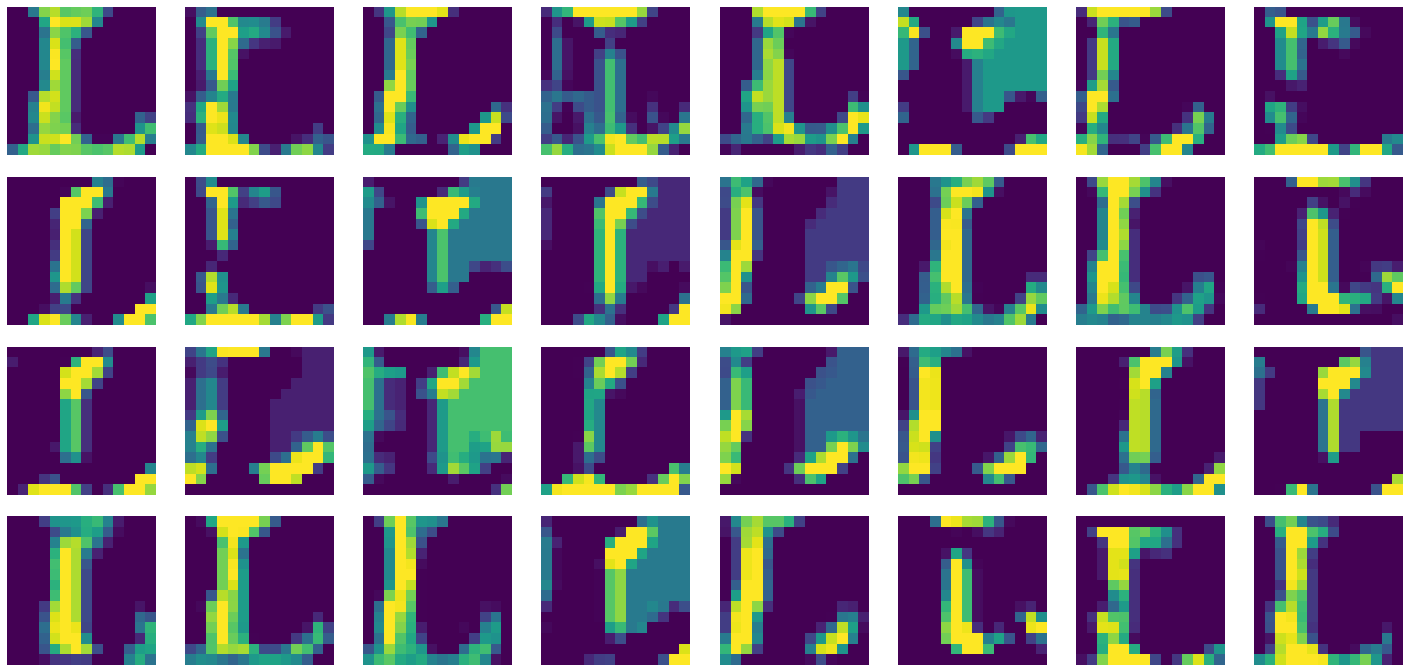}\includegraphics[height=3.45cm,width=0.5\linewidth]{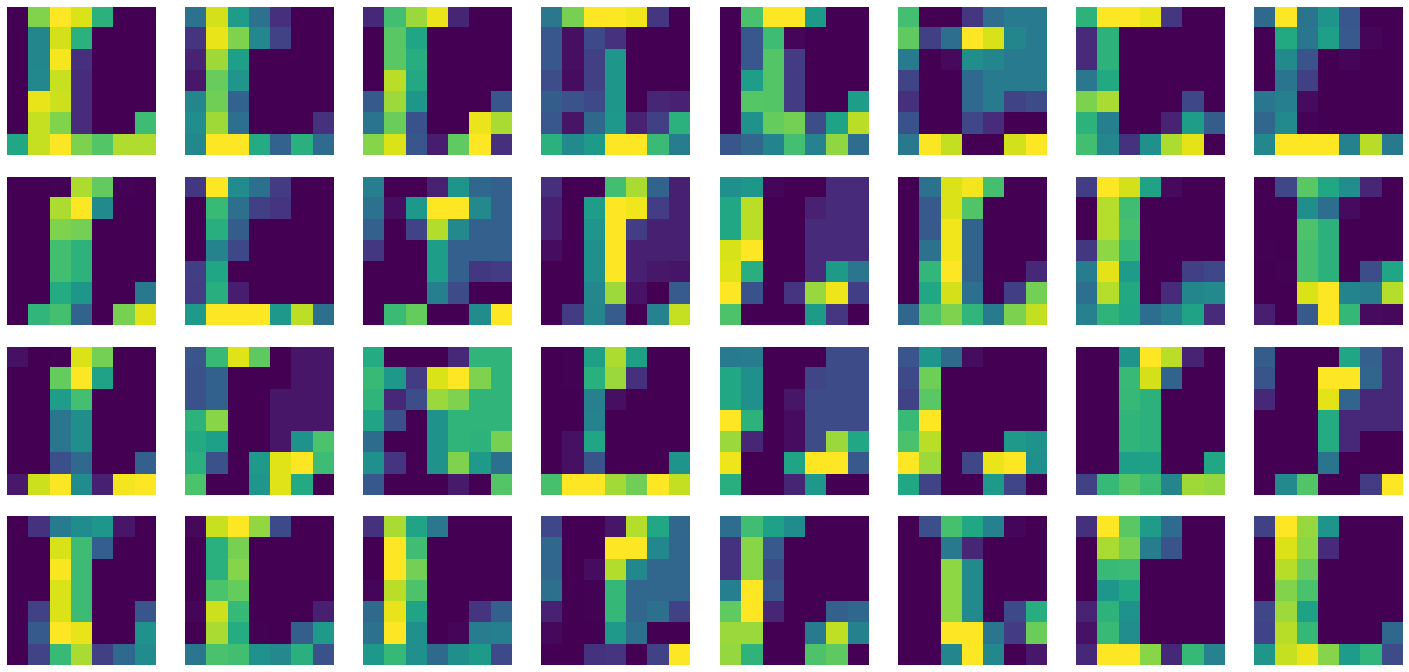}
	\includegraphics[height=3.45cm,width=1\linewidth]{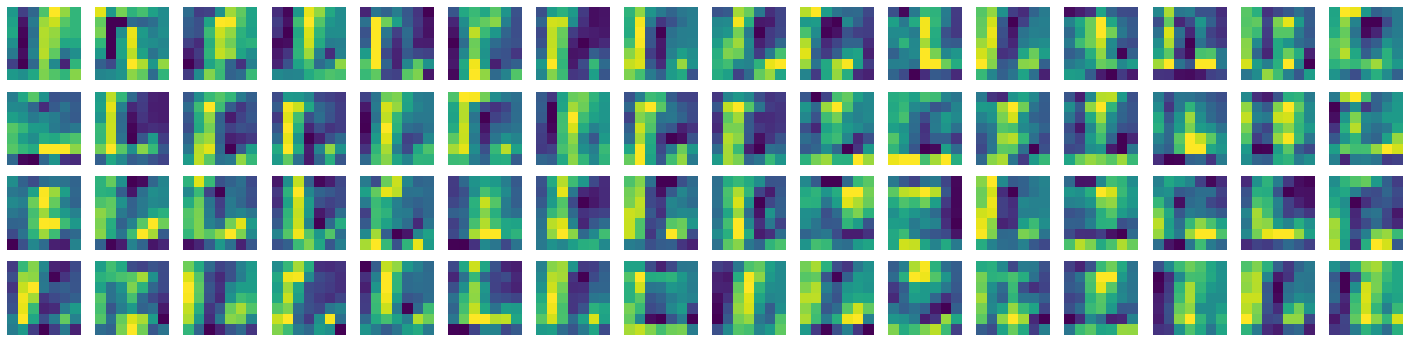}
	\includegraphics[height=3.45cm,width=1\linewidth]{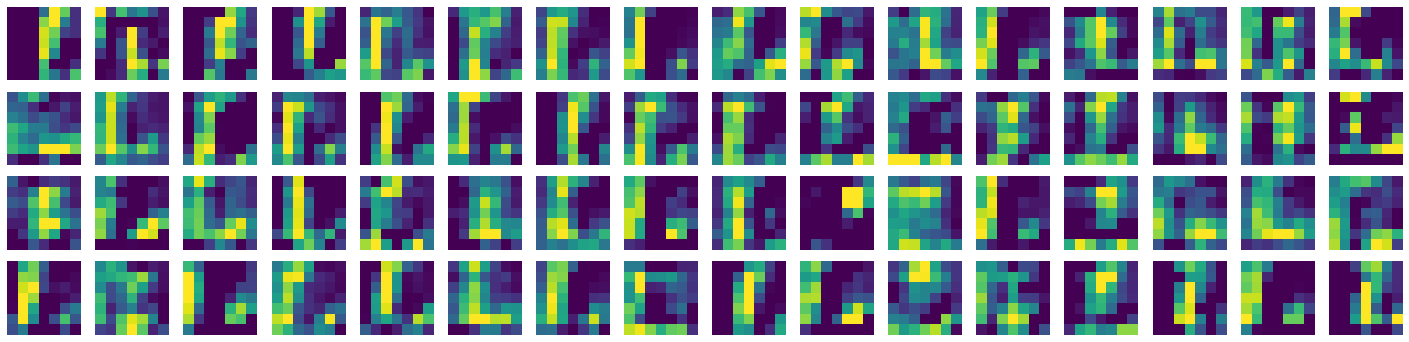}
	\includegraphics[height=3.45cm,width=1\linewidth]{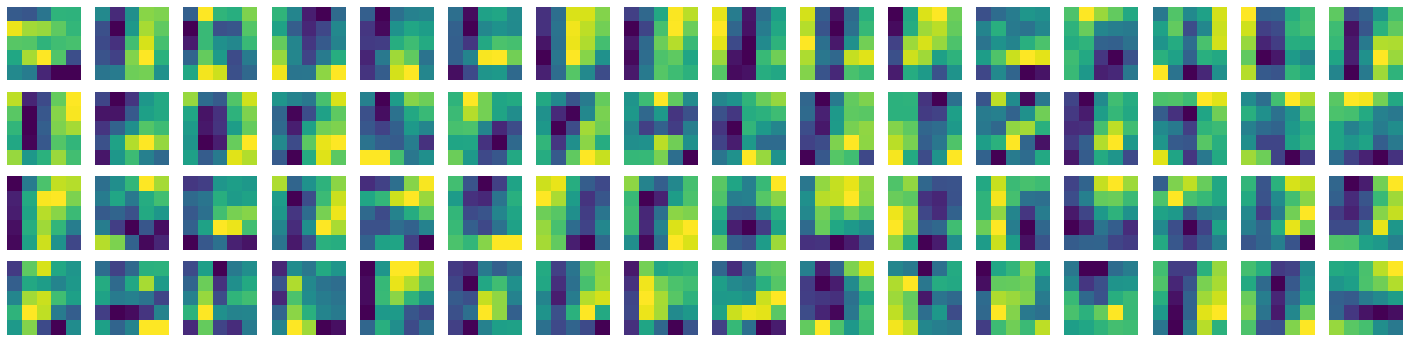}
	\includegraphics[height=3.45cm,width=1\linewidth]{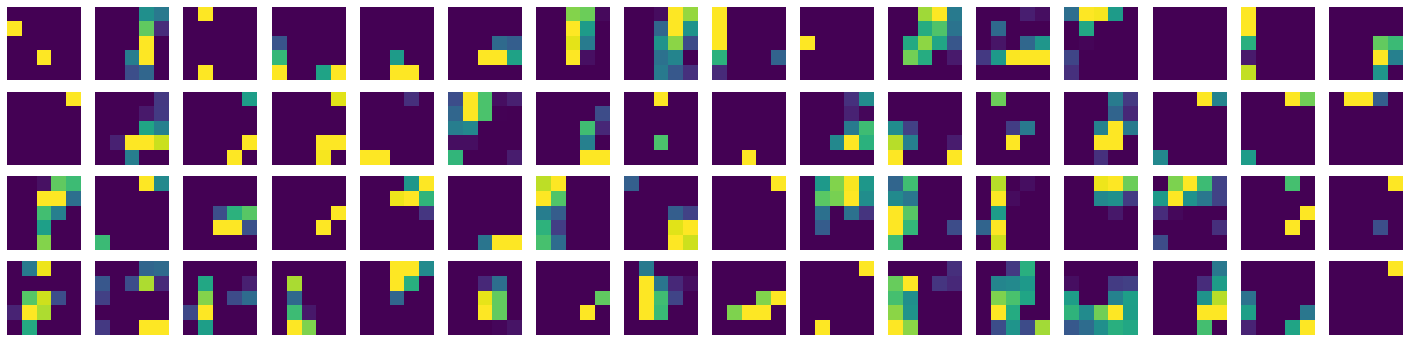}
	\includegraphics[height=3.45cm,width=1\linewidth]{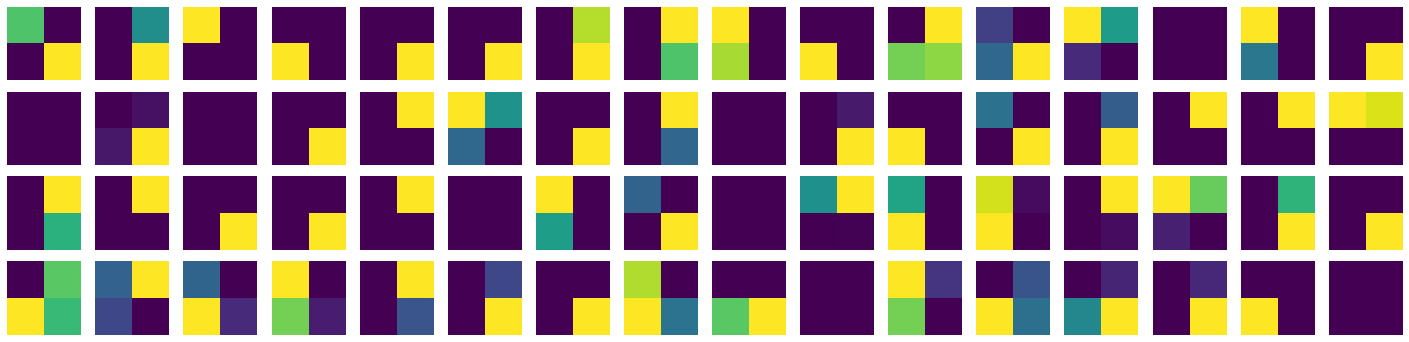}
	\caption{The visualizations of the features in a fully trained model when passed an image for letter  L.}
	\label{CNN_c4l_16x16_550_DL_L}
\end{figure}

\begin{figure}
	\centering
	\includegraphics[height=3.45cm,width=0.5\linewidth]{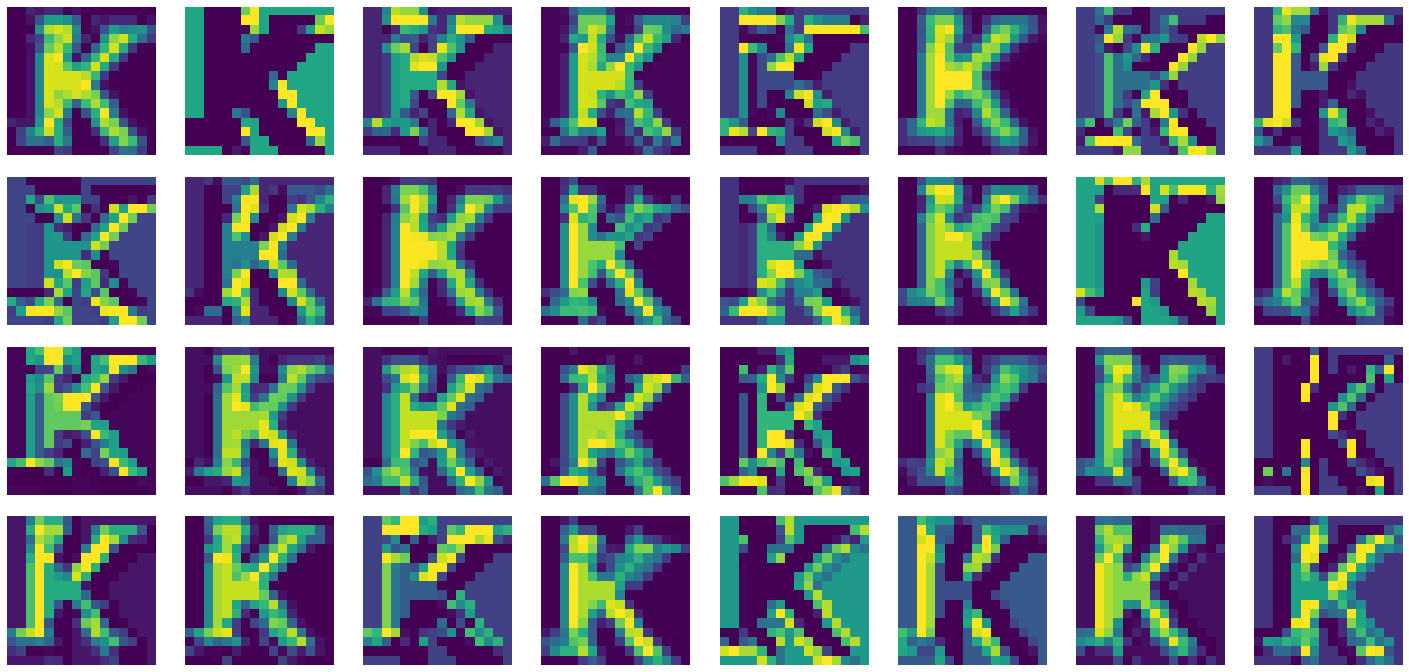}\includegraphics[height=3.45cm,width=0.5\linewidth]{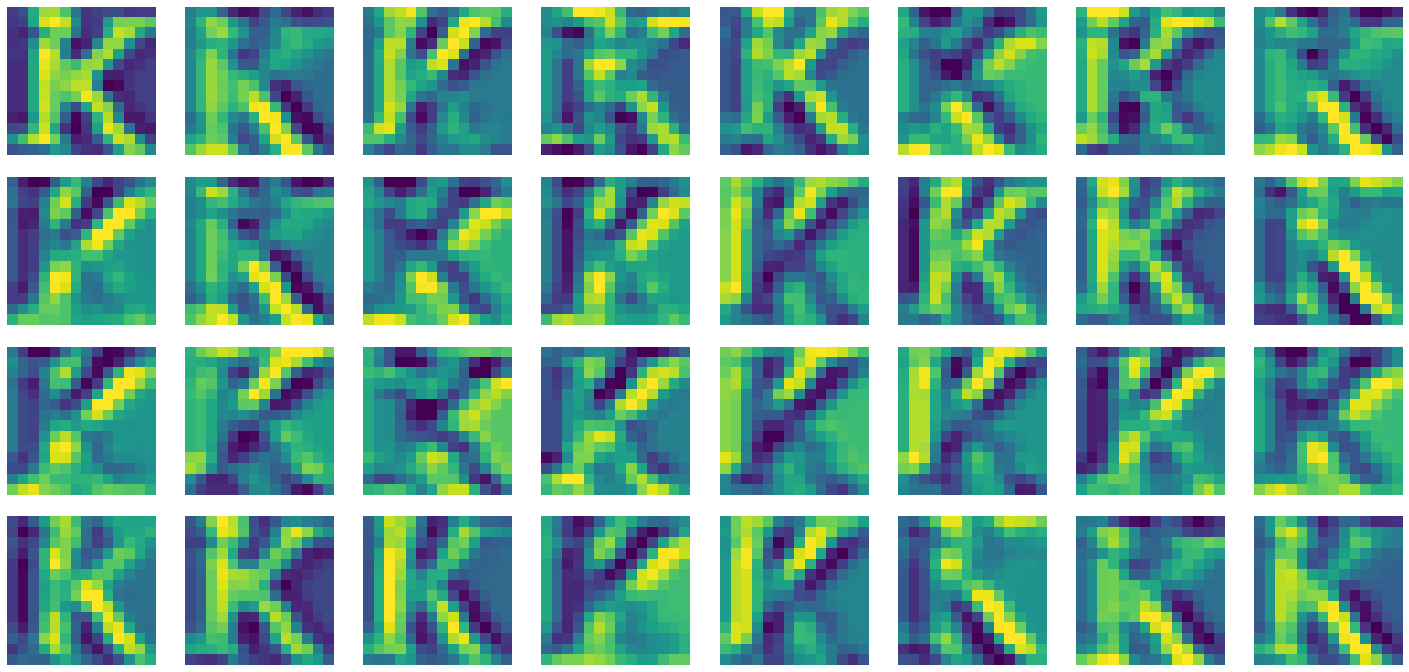}
	\includegraphics[height=3.45cm,width=0.5\linewidth]{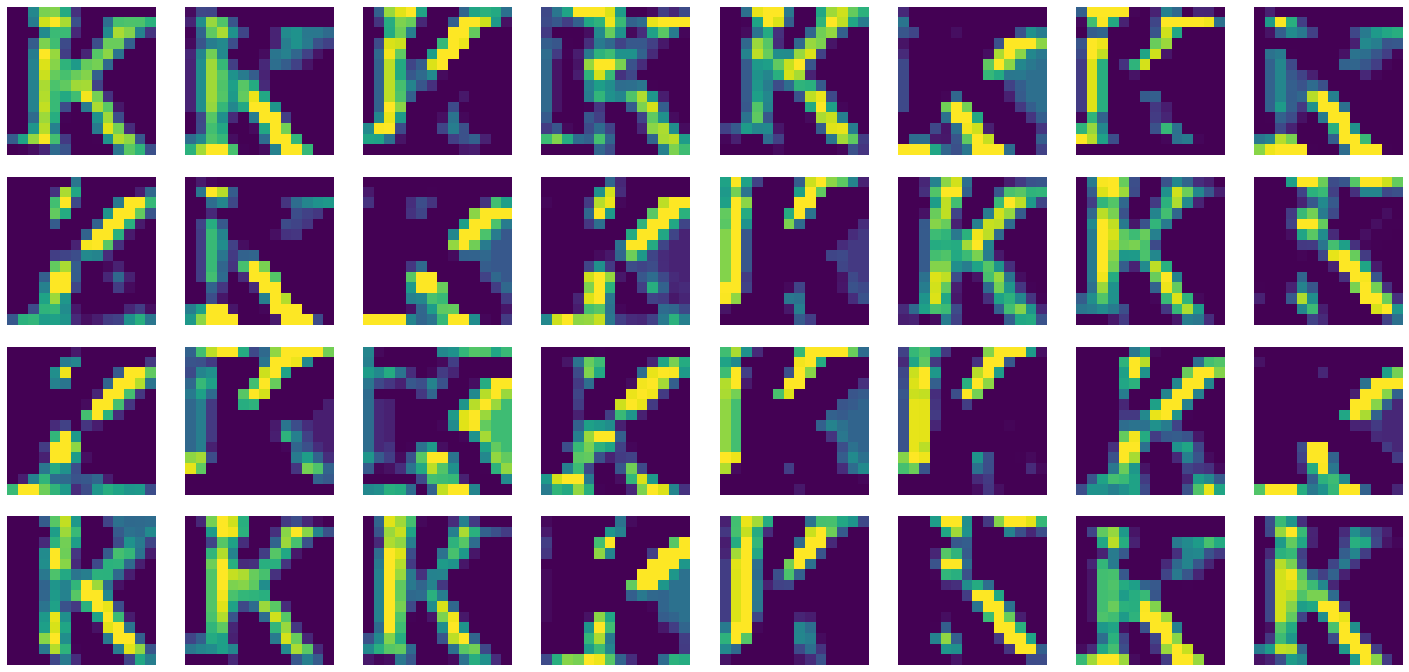}\includegraphics[height=3.45cm,width=0.5\linewidth]{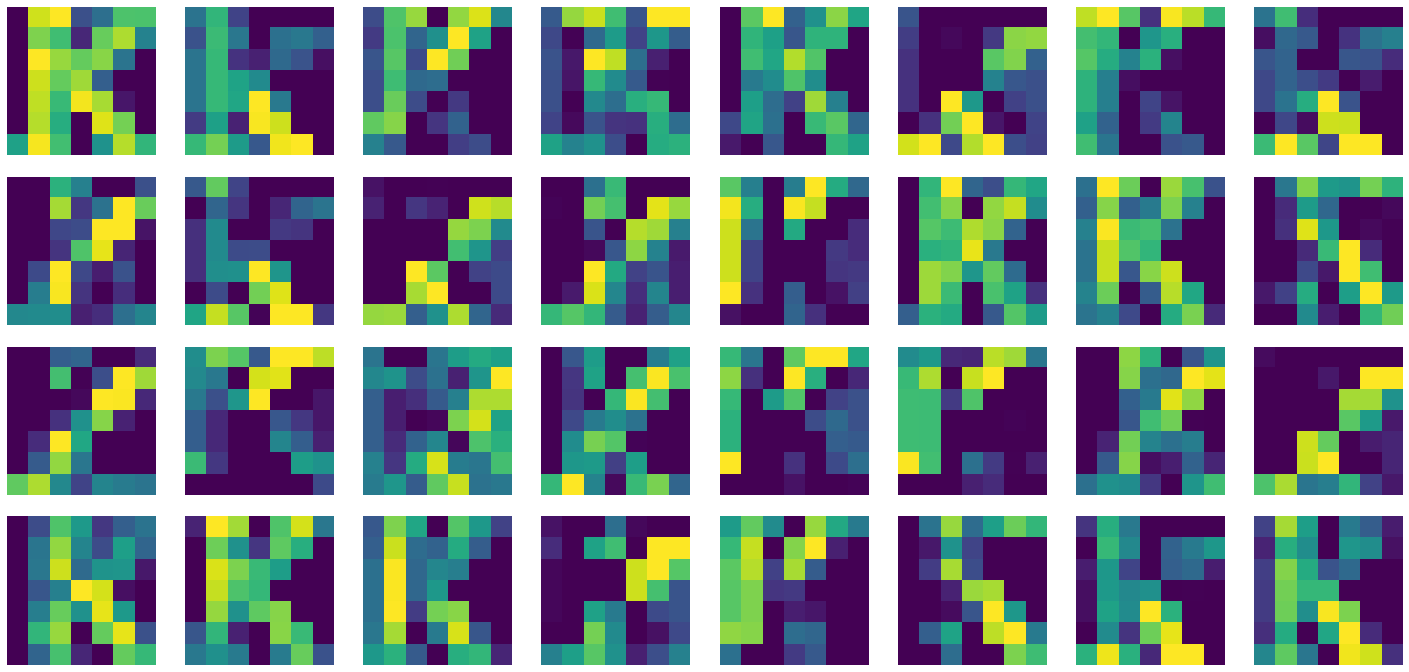}
	\includegraphics[height=3.45cm,width=1\linewidth]{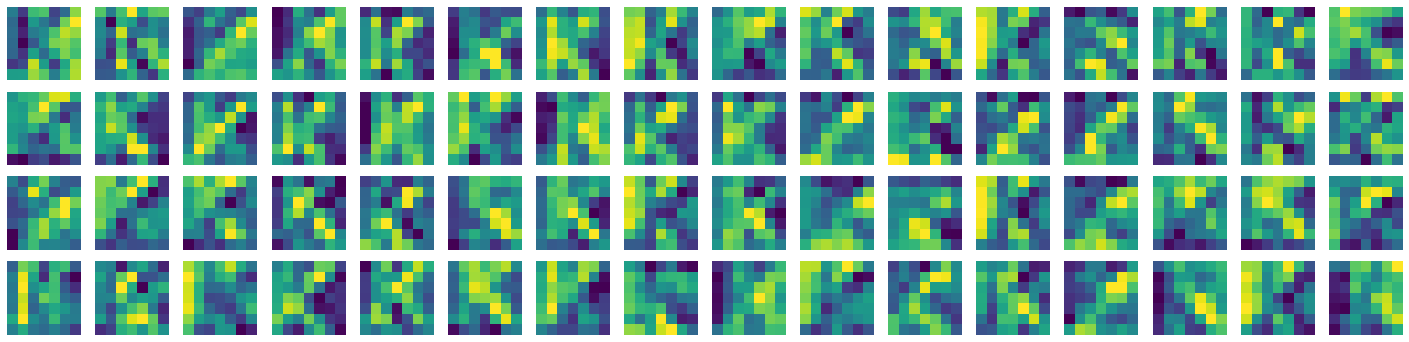}
	\includegraphics[height=3.45cm,width=1\linewidth]{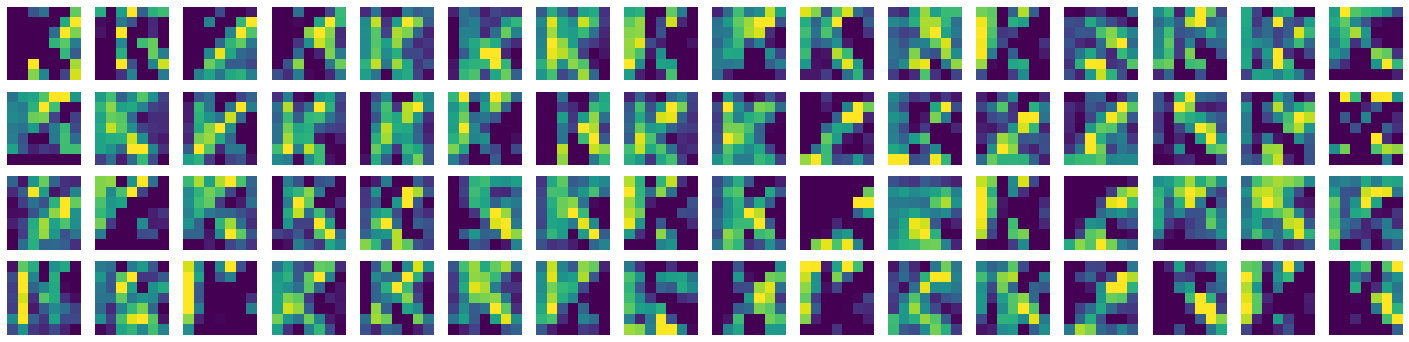}
	\includegraphics[height=3.45cm,width=1\linewidth]{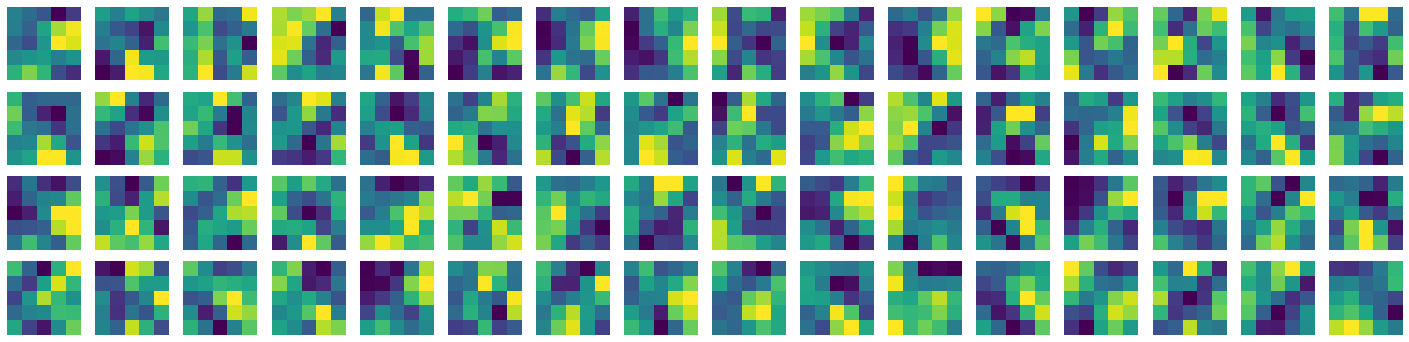}
	\includegraphics[height=3.45cm,width=1\linewidth]{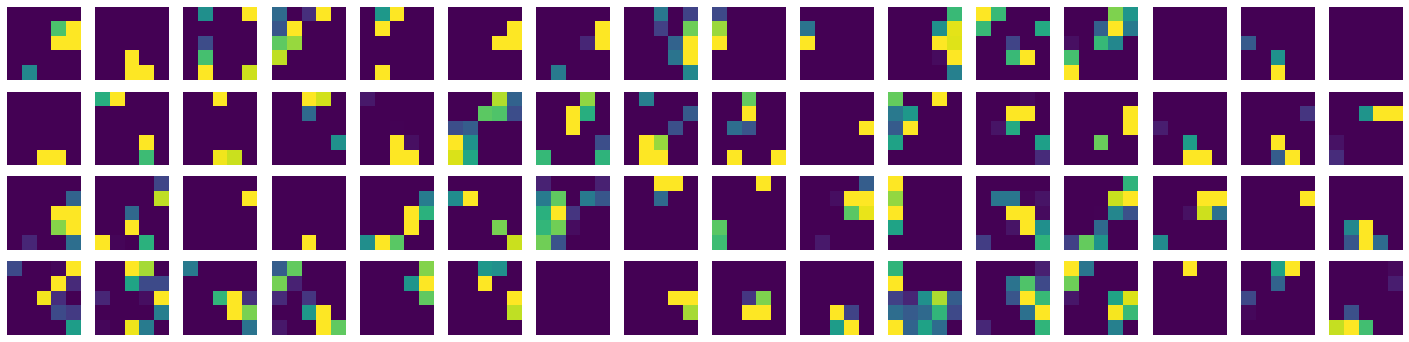}
	\includegraphics[height=3.45cm,width=1\linewidth]{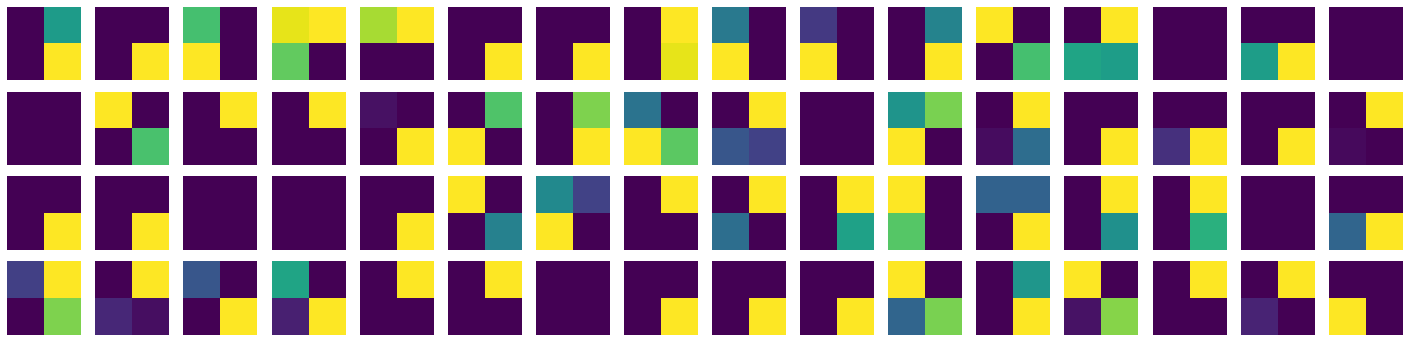}
	\caption{The visualizations of the features in a fully trained model when passed an image for letter K.}
	\label{CNN_c4l_16x16_550_DL_K}
\end{figure}

\begin{figure}
	\centering
	\includegraphics[height=3.45cm,width=0.5\linewidth]{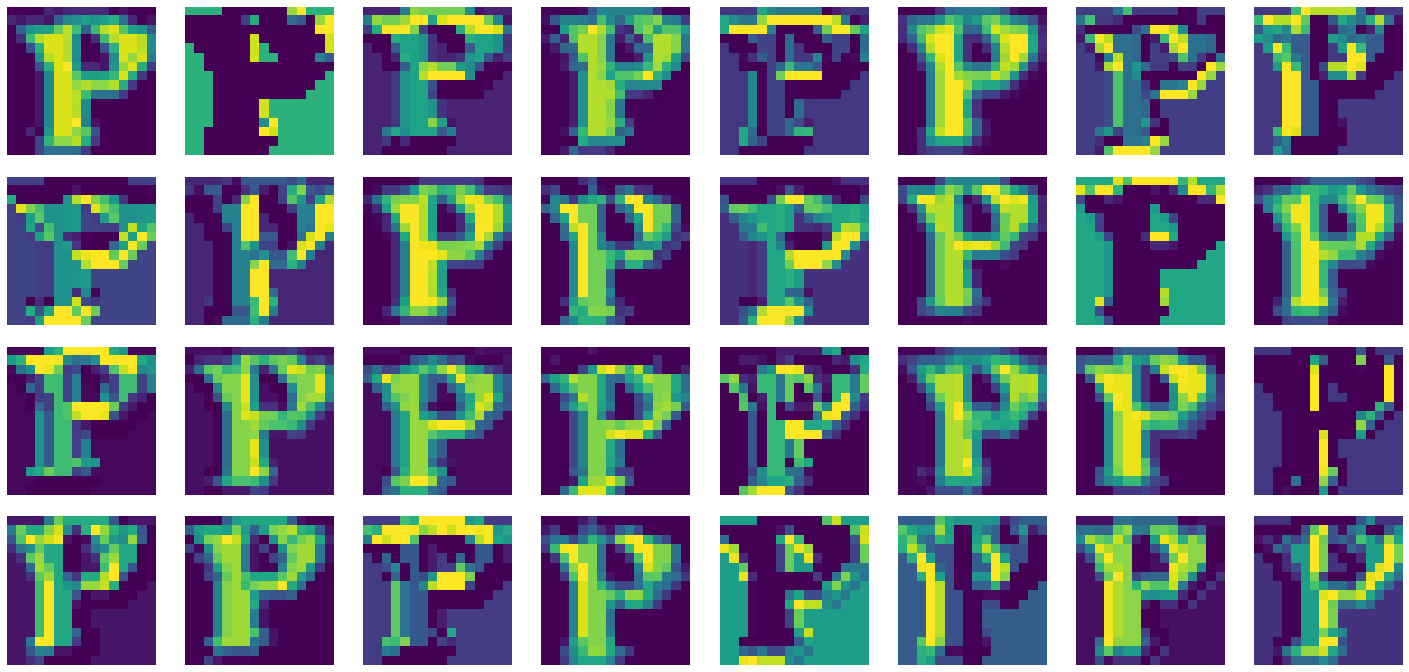}\includegraphics[height=3.45cm,width=0.5\linewidth]{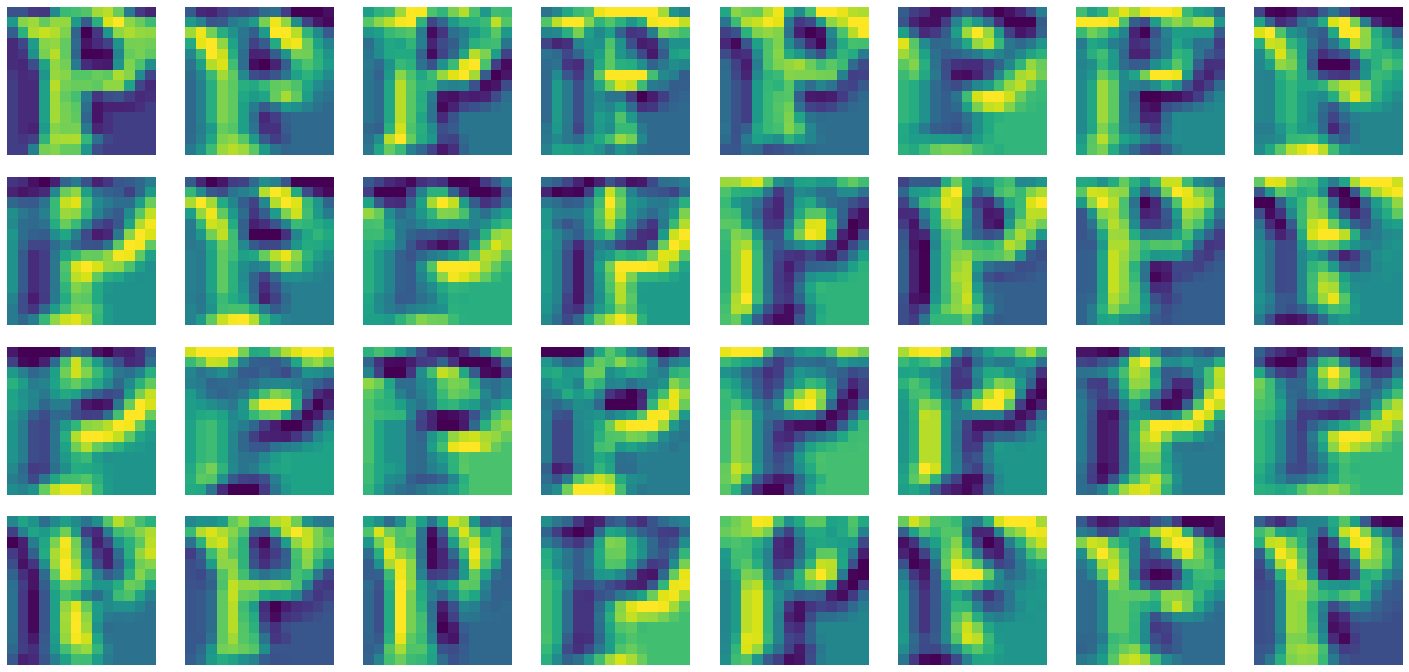}
	\includegraphics[height=3.45cm,width=0.5\linewidth]{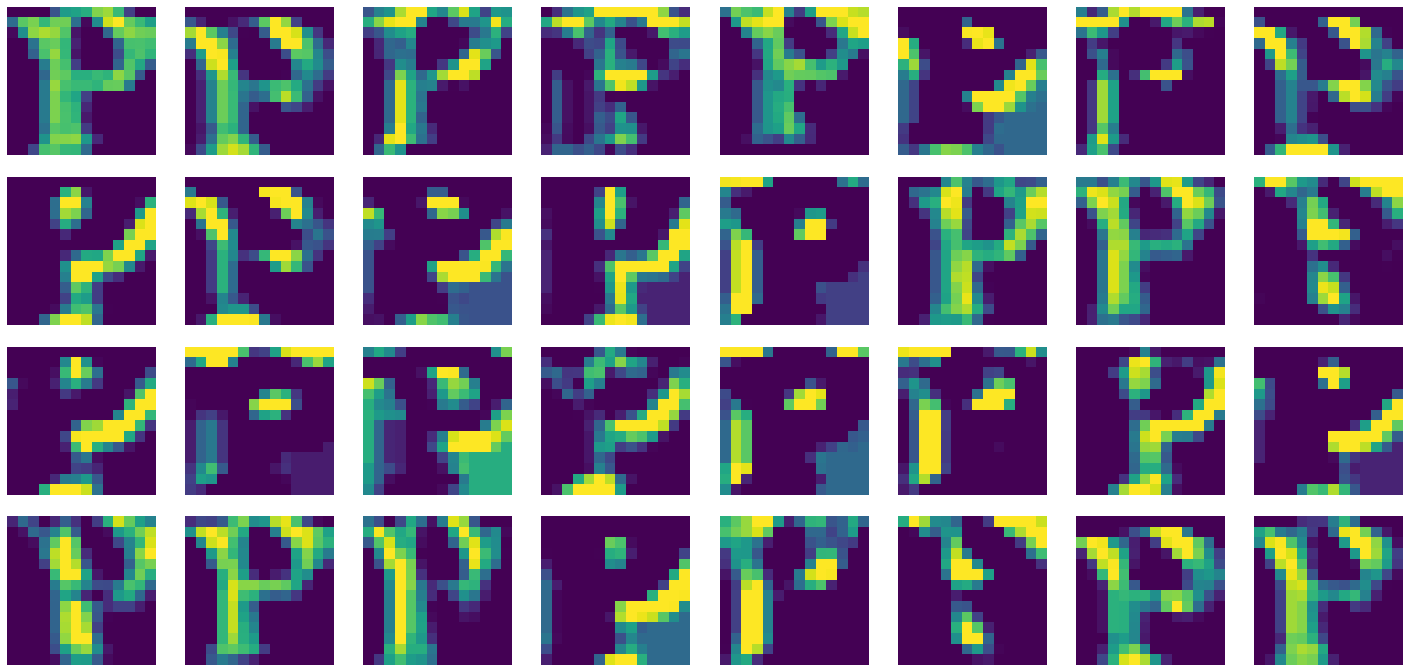}\includegraphics[height=3.45cm,width=0.5\linewidth]{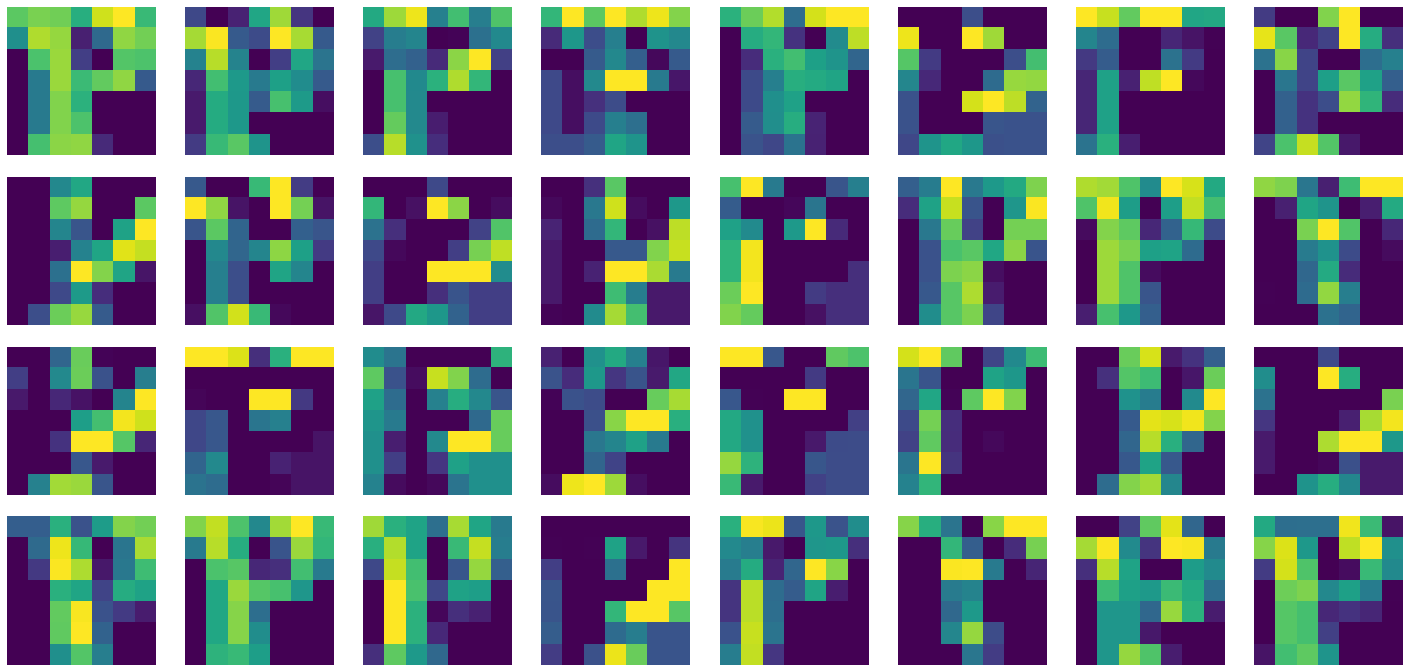}
	\includegraphics[height=3.45cm,width=1\linewidth]{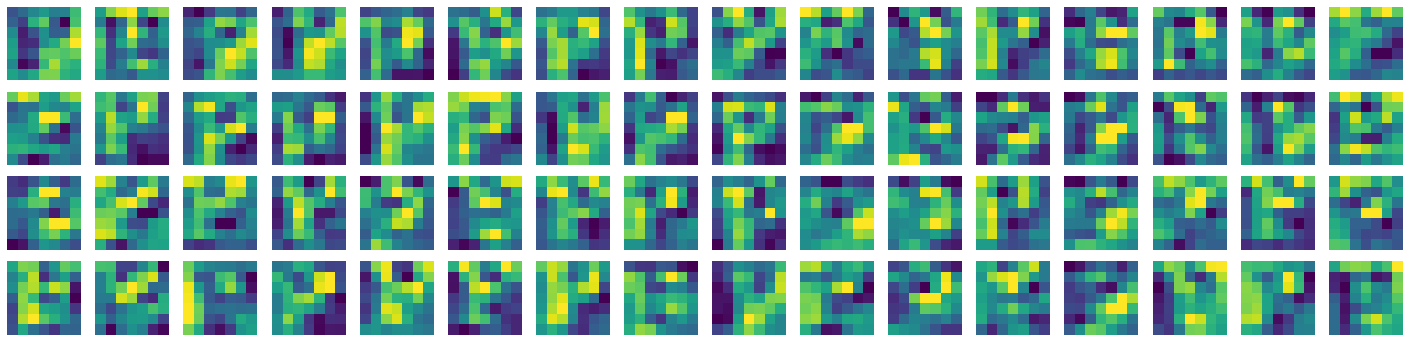}
	\includegraphics[height=3.45cm,width=1\linewidth]{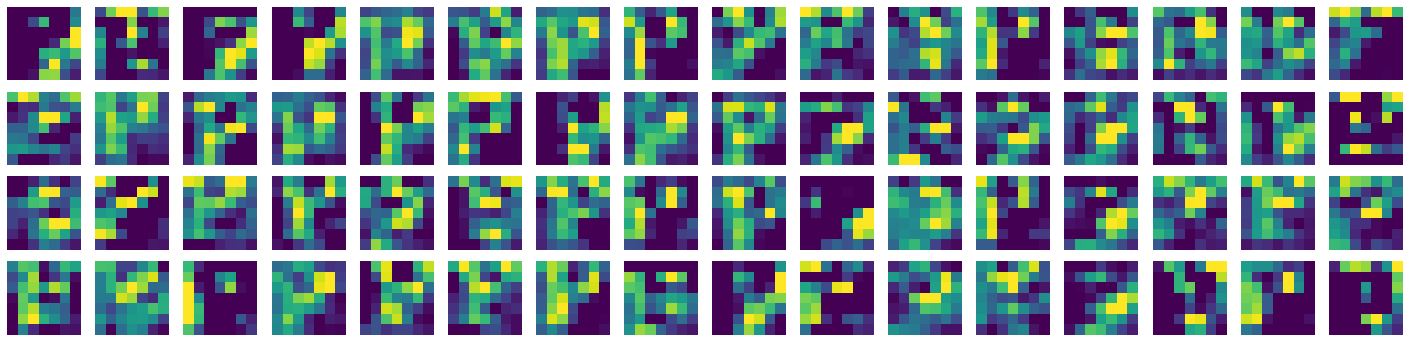}
	\includegraphics[height=3.45cm,width=1\linewidth]{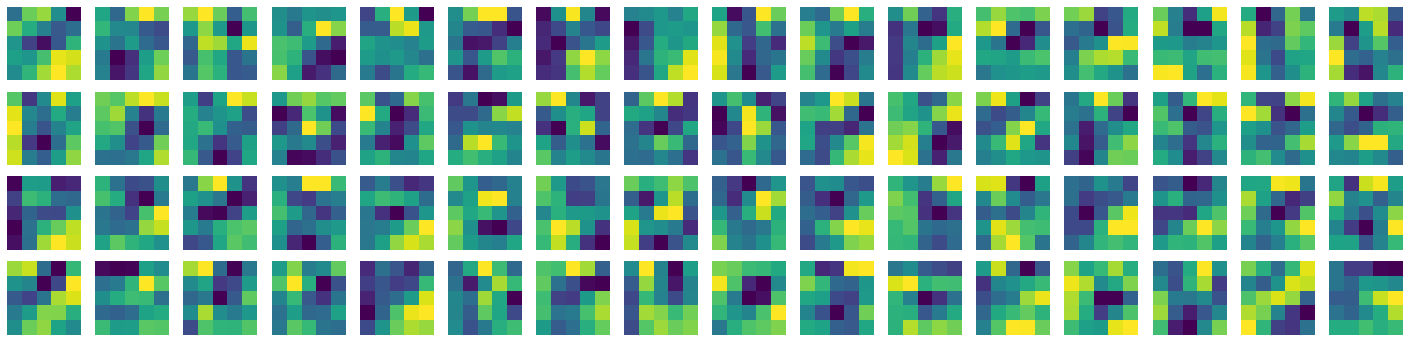}
	\includegraphics[height=3.45cm,width=1\linewidth]{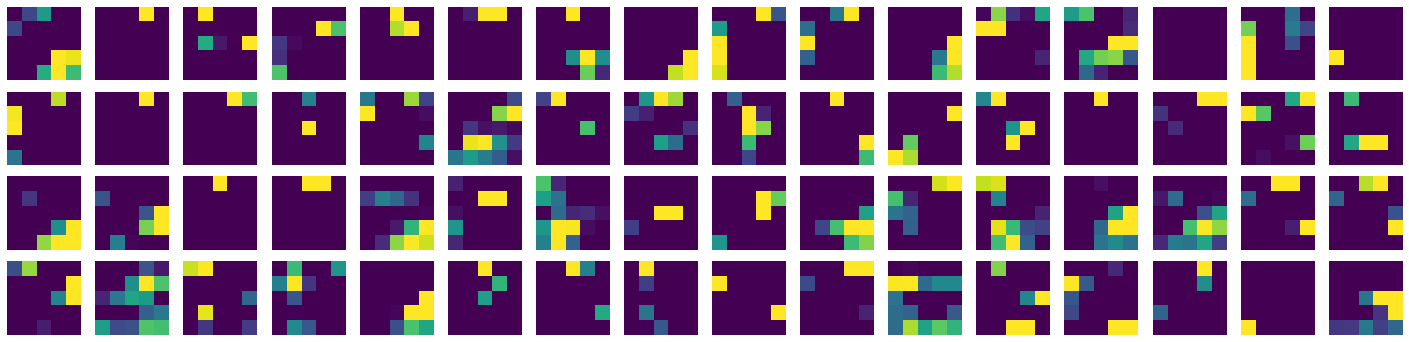}
	\includegraphics[height=3.45cm,width=1\linewidth]{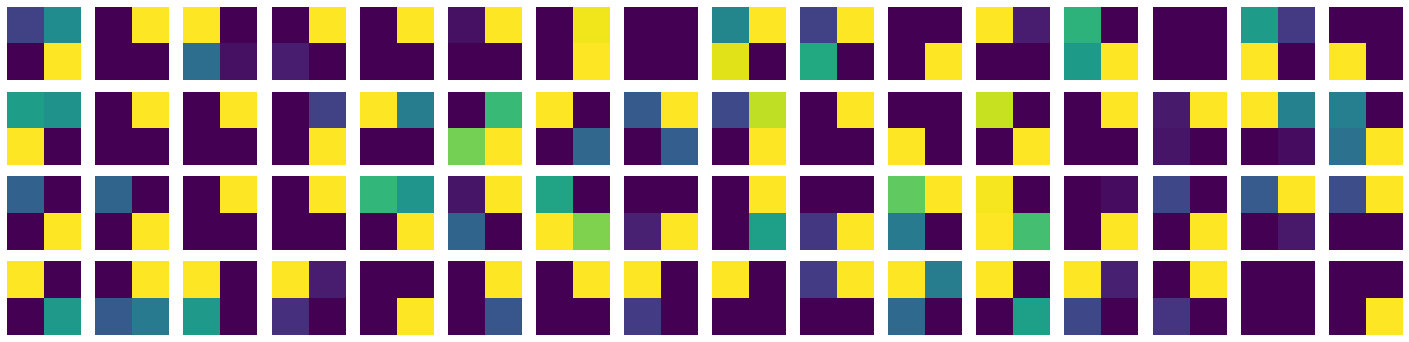}
	\caption{The visualizations of the features in a fully trained model when passed an image for letter P.}
	\label{CNN_c4l_16x16_550_DL_P}
\end{figure}

Visualizations generated from the layers of convolutional neural network when passed the images of character 2, L, K and P are shown in Figures \ref{CNN_c4l_16x16_550_DL_2}, \ref{CNN_c4l_16x16_550_DL_L}, \ref{CNN_c4l_16x16_550_DL_K} and \ref{CNN_c4l_16x16_550_DL_P}. These are the intermediate visualizations generated to finally arrive at the last layer of fully connected layer to determine which letter it actually belongs to.

\section{When using on real unseen data}

\begin{figure}
	\centering
	\includegraphics[width=0.45\linewidth]{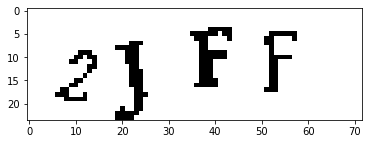}
	\includegraphics[width=0.45\linewidth]{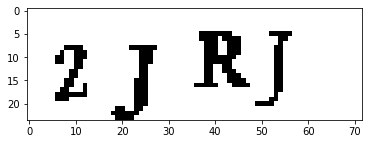}\\
	2JFF classified as 2AFF   \hspace{2cm}  2JRJ classified as 2KRJ \\
	\includegraphics[width=0.45\linewidth]{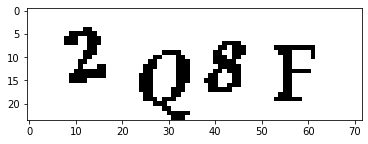}
	\includegraphics[width=0.45\linewidth]{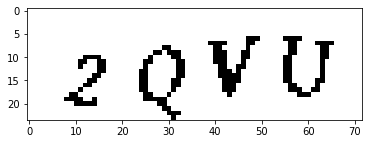}\\
	2Q8F classified as 2A8F \hspace{2cm} 2QVU classified as 2JVU
	\caption{The wrong classification of the model.}
	\label{CNN_c4l_16x16_550_err}
\end{figure}

When the model was downloaded from \index[\idxKeywordName]{Google Collaboratory} Google Collaboratory to local machine, it was tested on a sample of about 100 images, of which 96 were correctly classified, except these four images as shown in Figure \ref{CNN_c4l_16x16_550_err}.

After testing on 100 unseen CAPTCHA images, we found that these four were incorrectly labelled as shown in Figure \ref{CNN_c4l_16x16_550_err}. This is good since we are getting the validation accuracy on individual characters and it somehow mislabells \textbf{J} as \textbf{F} and \textbf{K}, \textbf{Q} as \textbf{J} and and \textbf{A}. The accuracy on individual character will be 99.99\% which is great.

\begin{figure}
	\centering
	\includegraphics[width=1\linewidth]{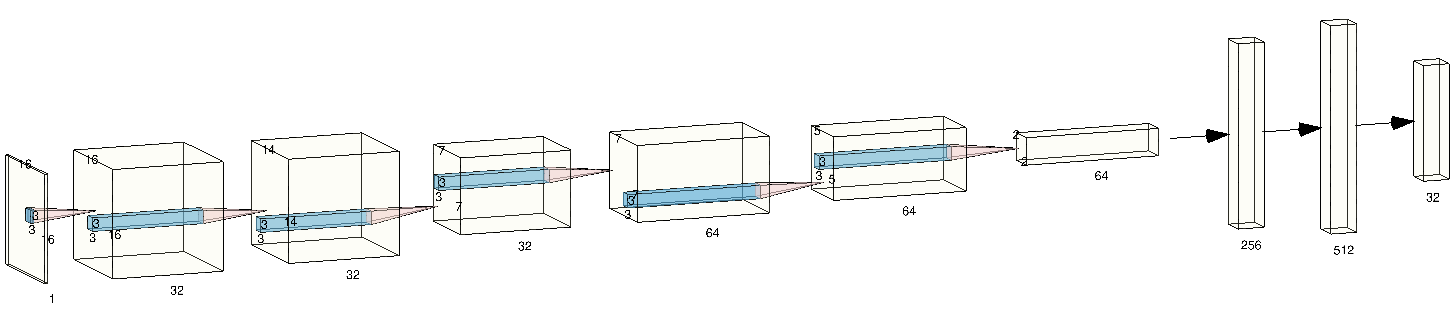}	
	\caption{The Alex net type architecture. Created using \url{https://alexlenail.me/NN-SVG/}.}
	\label{CNN_svg}
\end{figure}

\begin{figure}
	\centering
	\includegraphics[width=0.75\linewidth]{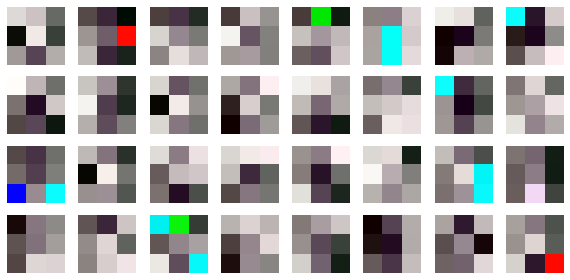}
	\caption{The 3x3 kernels (filter) learned by the model (\ref{model_segmented_c4l_16x16_550}) for layer 1.}
	\label{c4l_each-filter-learned}
\end{figure}

\begin{figure}
	\centering
	\includegraphics[width=0.75\linewidth]{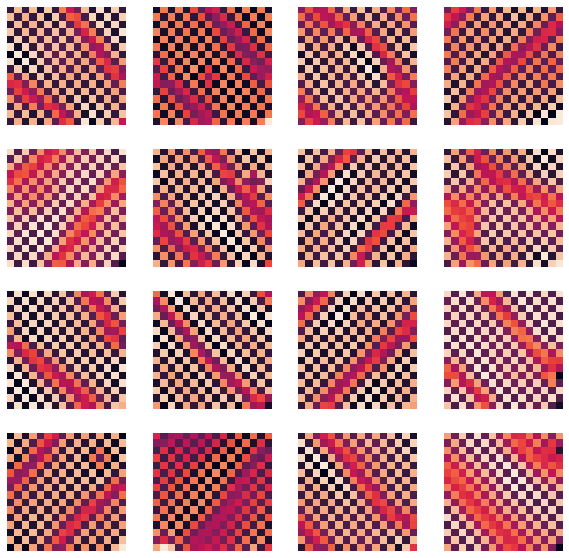}
	\caption{The weights learned by the model (\ref{model_segmented_c4l_16x16_550}) for layers 1 to 16.}
	\label{c4l_each-weights-learned}
\end{figure}

In Figure \ref{c4l_each-filter-learned} we see the 3x3 kernels that are being learned by the layer 1 of the network for the Model \ref{model_segmented_c4l_16x16_550}. The visualization of the weights learned by the same model can be seen in Figure \ref{c4l_each-weights-learned}.

\chapter{Using resized\_JAM CAPTCHAs dataset to predict JAM data (99.53 \% accuracy)}

\section{Using k-NN for resized\_JAM CAPTCHA dataset}

\begin{figure}
	\centering
	\includegraphics[width=0.8\linewidth]{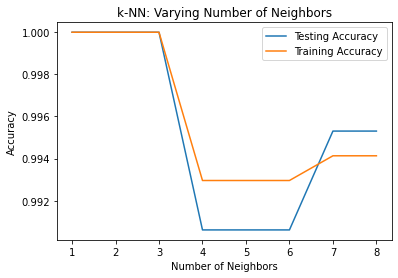}
	\caption{Varying the value of $k$ to get the best testing accuracy.}
	\label{k_fold_jam}
\end{figure}

\begin{figure}
	\centering
	\includegraphics[width=0.85\linewidth]{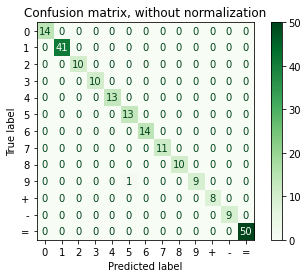}
	\caption{The confusion matrix of the resized\_JAM dataset.}
	\label{confusion_matrix_jam}
\end{figure}

We used the \index[\idxKeywordName]{$k$-Nearest Neighbor } $k$-Nearest Neighbor (k-NN) algorithm for prediction of resized\_JAM dataset. Using a $k$ of value $7$ we got 99.53\% testing accuracy, which is almost state of the art. We varied the value of $k$ to get the best accuracy as shown in Figure \ref{k_fold_jam}. The \index[\idxKeywordName]{confusion matrix} confusion matrix for the same is shown in Figure \ref{confusion_matrix_jam}. The only confusion here is 9 is misclassified as 5 in some data which might be due to noise. This shows that even a simple and computationally efficient model (when the data is less), can give a high accuracy which is almost the state of the art. 

This even shows that the IITK JAM website needs a better version of CAPTCHA since this is very easy to break using KNN. When there are choices whether to choose a complex model or a simple model, people should choose a simple model which gives accuracy comparable to the complex model.

\chapter{Using CAPTCHA-version-2 dataset (90.102 \% accuracy)}

\section{Using AlexNet type Model }
\begin{figure}
	\centering
	\includegraphics[width=0.75\linewidth]{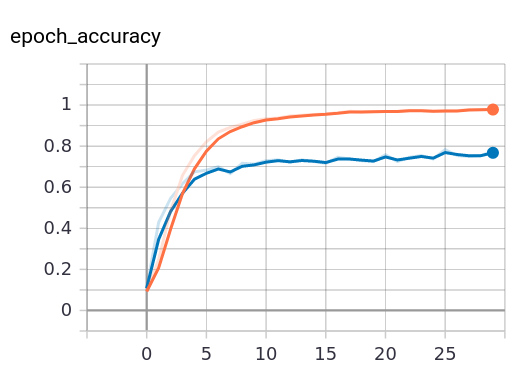}
	\caption{Accuracy obtained from the AlexNet type model (\ref{model_faded}).}
	\label{acc_cv2_model_AlexNet}
\end{figure}

\begin{figure}
	\centering
	\includegraphics[width=0.75\linewidth]{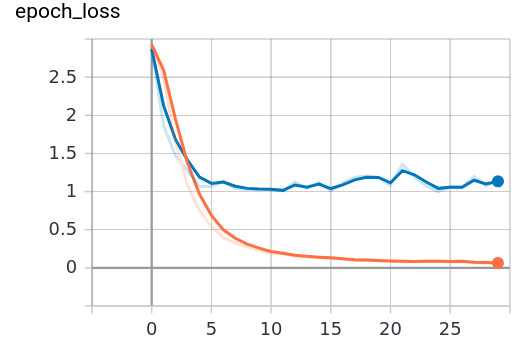}
	\caption{Loss obtained from the AlexNet type model (\ref{model_faded}).}
	\label{loss_cv2_model_AlexNet}
\end{figure}

The sample images of CAPTCHA-version-2 dataset is shown in Figure \ref{captcha-v2}. It is clear that there is a great challenge in classifying the labels of such a dataset. The main reason is presence of certain amount of noise in the dataset in terms of clutter and scratch. Firstly, we used an Alex Net type model as shown in Figure \ref{model_cv2_summary_alexNet}. We trained about 30 epochs and each epoch took about 1 second. This means it took 30 second to complete the whole training, which is pretty fast. The model used is similar to \ref{model_faded}, which can detect even those CAPTCHAs which are occluded by noise. The loss and accuracy for the training is shown in Figure \ref{loss_cv2_model_AlexNet} and Figure \ref{acc_cv2_model_AlexNet} respectively. The accuracy obtained during training was about 97.87\% and the accuracy obtained during testing was about 76.54 \% which was pretty useless. The summary of the epochs can be found in Figure \ref{model_cv2Alex_epochs}. The model can be found from this url (\url{https://jimut123.github.io/blogs/CAPTCHA/models/captcha_v2_alexNet.h5}).


\begin{figure}
	\centering
	\includegraphics[width=0.75\linewidth]{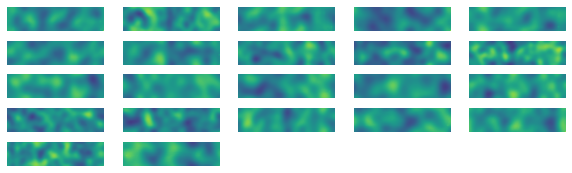}
	\caption{The weights learned by the model (\ref{model_faded}) for layers 1 to 16.}
	\label{cv2-weights-learned}
\end{figure}

\begin{figure} 
	\centering
	{%
		\lstset{frame=single,basicstyle=\scriptsize,style=myModelSummaryStyle}
		\centering
		\begin{lstlisting}
		_________________________________________________________________
		Layer (type)                 Output Shape              Param #   
		==================================================
		input_1 (InputLayer)         [(None, 50, 200, 1)]      0         
		_________________________________________________________________
		conv2d (Conv2D)              (None, 48, 198, 32)       320       
		_________________________________________________________________
		max_pooling2d (MaxPooling2D) (None, 24, 99, 32)        0         
		_________________________________________________________________
		conv2d_1 (Conv2D)            (None, 22, 97, 64)        18496     
		_________________________________________________________________
		max_pooling2d_1 (MaxPooling2 (None, 11, 48, 64)        0         
		_________________________________________________________________
		conv2d_2 (Conv2D)            (None, 9, 46, 64)         36928     
		_________________________________________________________________
		max_pooling2d_2 (MaxPooling2 (None, 4, 23, 64)         0         
		_________________________________________________________________
		flatten (Flatten)            (None, 5888)              0         
		_________________________________________________________________
		dense (Dense)                (None, 1024)              6030336   
		_________________________________________________________________
		dropout (Dropout)            (None, 1024)              0         
		_________________________________________________________________
		dense_1 (Dense)              (None, 95)                97375     
		_________________________________________________________________
		reshape (Reshape)            (None, 5, 19)             0         
		==================================================
		Total params: 6,183,455
		Trainable params: 6,183,455
		Non-trainable params: 0
		_________________________________________________________________

		\end{lstlisting}
	}
\caption{Summary of the AlexNet type model for classifying the labels of CAPTCHA-version-2 dataset.}
\label{model_cv2_summary_alexNet}
\end{figure}

\begin{figure} 
	\centering
	{%
		\lstset{frame=single,basicstyle=\scriptsize,style=myModelSummaryStyle}
		\centering
		\begin{lstlisting}
		
			Epoch 30/30
			120/120 [==============================] - 1s 12ms/step - loss: 0.0613 - accuracy: 0.9787 - val_loss: 1.1524 - val_accuracy: 0.7750
			
		\end{lstlisting}
	}
	\caption{Summary of epochs for the model (\ref{model_cv2_summary_alexNet}).}
	\label{model_cv2Alex_epochs}
\end{figure}

\begin{figure}
	\centering
	\includegraphics[width=0.75\linewidth]{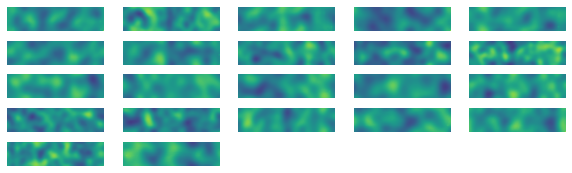}
	\caption{The weights learned by the model (\ref{tensorboard_modelcv2}) for layers 1 to 16.}
	\label{cv2-weights-learned}
\end{figure}

\section{Using a better Model }

\begin{figure} 
	\centering
	{%
		\lstset{frame=single,basicstyle=\scriptsize,style=myModelSummaryStyle}
		\centering
		\begin{lstlisting}
			__________________________________________________________________________________________________
			Layer (type)                    Output Shape         Param #     Connected to                     
			==================================================
			input_1 (InputLayer)            (None, 50, 200, 1)   0                                            
			__________________________________________________________________________________________________
			conv2d_1 (Conv2D)               (None, 50, 200, 16)  160         input_1[0][0]                    
			__________________________________________________________________________________________________
			max_pooling2d_1 (MaxPooling2D)  (None, 25, 100, 16)  0           conv2d_1[0][0]                   
			__________________________________________________________________________________________________
			conv2d_2 (Conv2D)               (None, 25, 100, 32)  4640        max_pooling2d_1[0][0]            
			__________________________________________________________________________________________________
			max_pooling2d_2 (MaxPooling2D)  (None, 13, 50, 32)   0           conv2d_2[0][0]                   
			__________________________________________________________________________________________________
			conv2d_3 (Conv2D)               (None, 13, 50, 32)   9248        max_pooling2d_2[0][0]            
			__________________________________________________________________________________________________
			batch_normalization_1 (BatchNor (None, 13, 50, 32)   128         conv2d_3[0][0]                   
			__________________________________________________________________________________________________
			max_pooling2d_3 (MaxPooling2D)  (None, 7, 25, 32)    0           batch_normalization_1[0][0]      
			__________________________________________________________________________________________________
			flatten_1 (Flatten)             (None, 5600)         0           max_pooling2d_3[0][0]            
			__________________________________________________________________________________________________
			dense_1 (Dense)                 (None, 64)           358464      flatten_1[0][0]                  
			__________________________________________________________________________________________________
			dense_3 (Dense)                 (None, 64)           358464      flatten_1[0][0]                  
			__________________________________________________________________________________________________
			dense_5 (Dense)                 (None, 64)           358464      flatten_1[0][0]                  
			__________________________________________________________________________________________________
			dense_7 (Dense)                 (None, 64)           358464      flatten_1[0][0]                  
			__________________________________________________________________________________________________
			dense_9 (Dense)                 (None, 64)           358464      flatten_1[0][0]                  
			__________________________________________________________________________________________________
			dropout_1 (Dropout)             (None, 64)           0           dense_1[0][0]                    
			__________________________________________________________________________________________________
			dropout_2 (Dropout)             (None, 64)           0           dense_3[0][0]                    
			__________________________________________________________________________________________________
			dropout_3 (Dropout)             (None, 64)           0           dense_5[0][0]                    
			__________________________________________________________________________________________________
			dropout_4 (Dropout)             (None, 64)           0           dense_7[0][0]                    
			__________________________________________________________________________________________________
			dropout_5 (Dropout)             (None, 64)           0           dense_9[0][0]                    
			__________________________________________________________________________________________________
			dense_2 (Dense)                 (None, 36)           2340        dropout_1[0][0]                  
			__________________________________________________________________________________________________
			dense_4 (Dense)                 (None, 36)           2340        dropout_2[0][0]                  
			__________________________________________________________________________________________________
			dense_6 (Dense)                 (None, 36)           2340        dropout_3[0][0]                  
			__________________________________________________________________________________________________
			dense_8 (Dense)                 (None, 36)           2340        dropout_4[0][0]                  
			__________________________________________________________________________________________________
			dense_10 (Dense)                (None, 36)           2340        dropout_5[0][0]                  
			==================================================
			Total params: 1,818,196
			Trainable params: 1,818,132
			Non-trainable params: 64
			__________________________________________________________________________________________________
		\end{lstlisting}
	}
	\caption{Summary of the model for classifying the labels of CAPTCHA-version-2 dataset.}
	\label{model_cv2_summary}
\end{figure}

\begin{figure} 
	\centering
	\fbox{\includegraphics[height=17cm,width=0.85\linewidth]{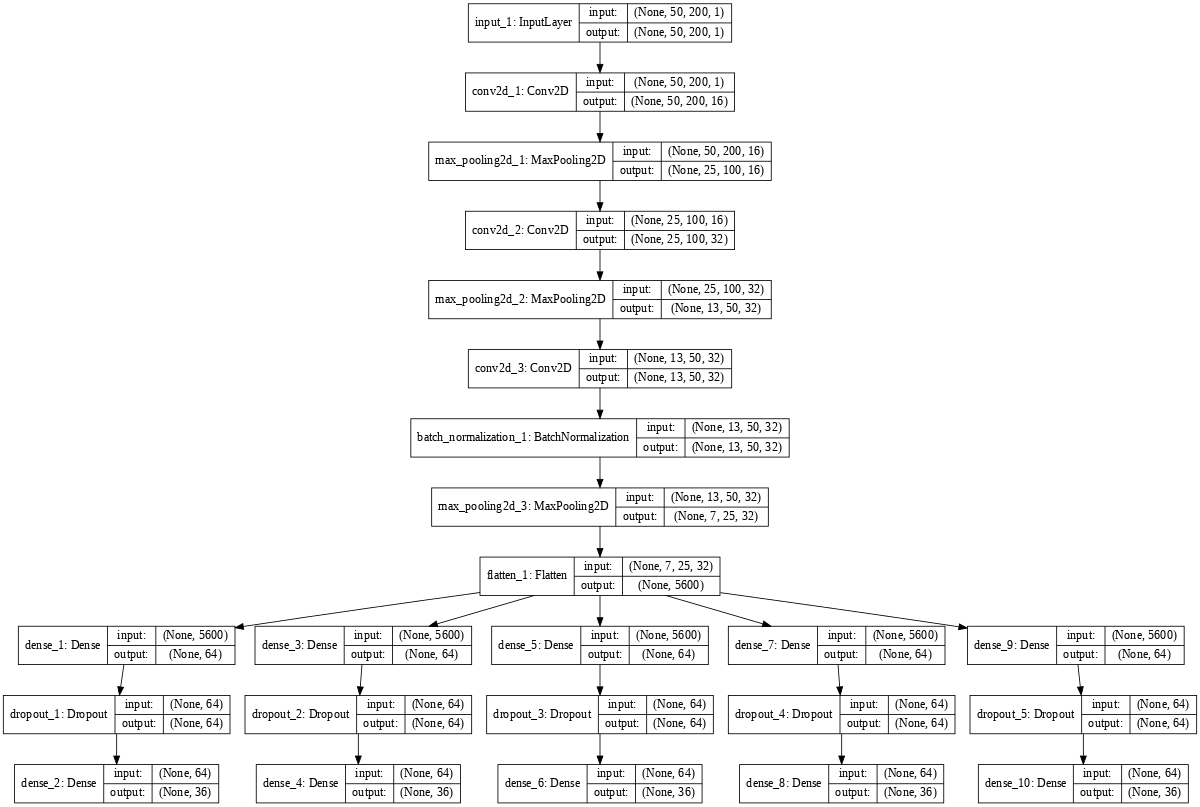}}	
	\caption{Model for training CAPTCHA-version-2 dataset.}
	\label{model_cv2}
\end{figure}

\begin{figure} 
	\centering
	{%
		\lstset{frame=single,basicstyle=\scriptsize,style=myModelSummaryStyle}
		\centering
		\begin{lstlisting}
	
		Epoch 30/30
		776/776 [==============================] - 1s 693us/step - loss: 1.2762 - dense_2_loss: 0.1197 - dense_4_loss: 0.1451 - dense_6_loss: 0.3858 - dense_8_loss: 0.3225 - dense_10_loss: 0.3031 - dense_2_acc: 0.9601 - dense_4_acc: 0.9549 - dense_6_acc: 0.8389 - dense_8_acc: 0.8802 - dense_10_acc: 0.8892 - val_loss: 2.0466 - val_dense_2_loss: 0.0376 - val_dense_4_loss: 0.2783 - val_dense_6_loss: 0.4414 - val_dense_8_loss: 0.8160 - val_dense_10_loss: 0.4733 - val_dense_2_acc: 0.9897 - val_dense_4_acc: 0.9433 - val_dense_6_acc: 0.8608 - val_dense_8_acc: 0.8299 - val_dense_10_acc: 0.8814
		\end{lstlisting}
	}
	\caption{Summary of epochs for the model (\ref{model_cv2}).}
\label{model_cv2_epochs}
\end{figure}

\begin{figure}
	\centering
	\includegraphics[width=1\linewidth]{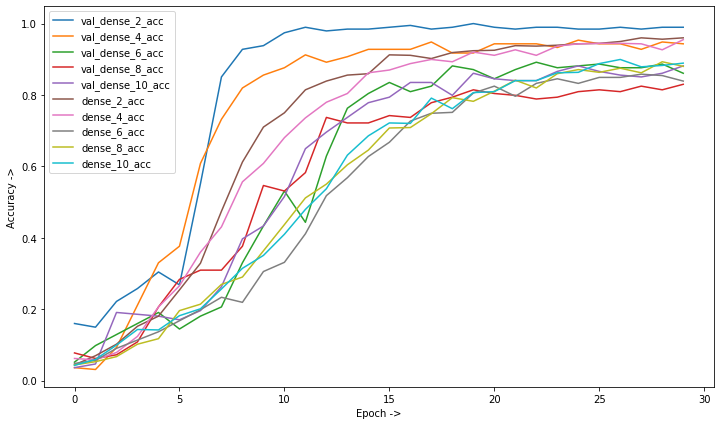}
	\caption{Accuracy of the model (\ref{model_cv2}).}
	\label{acc_cv2_model}
\end{figure}
\begin{figure}
	\centering
	\includegraphics[width=1\linewidth]{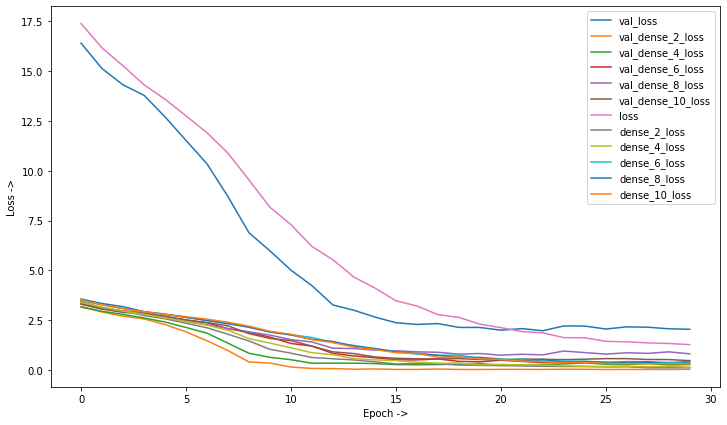}
	\caption{Loss of the model (\ref{model_cv2}).}
	\label{loss_cv2_model}
\end{figure}
\begin{figure}
	\centering
	\includegraphics[height=16cm,width=1\linewidth]{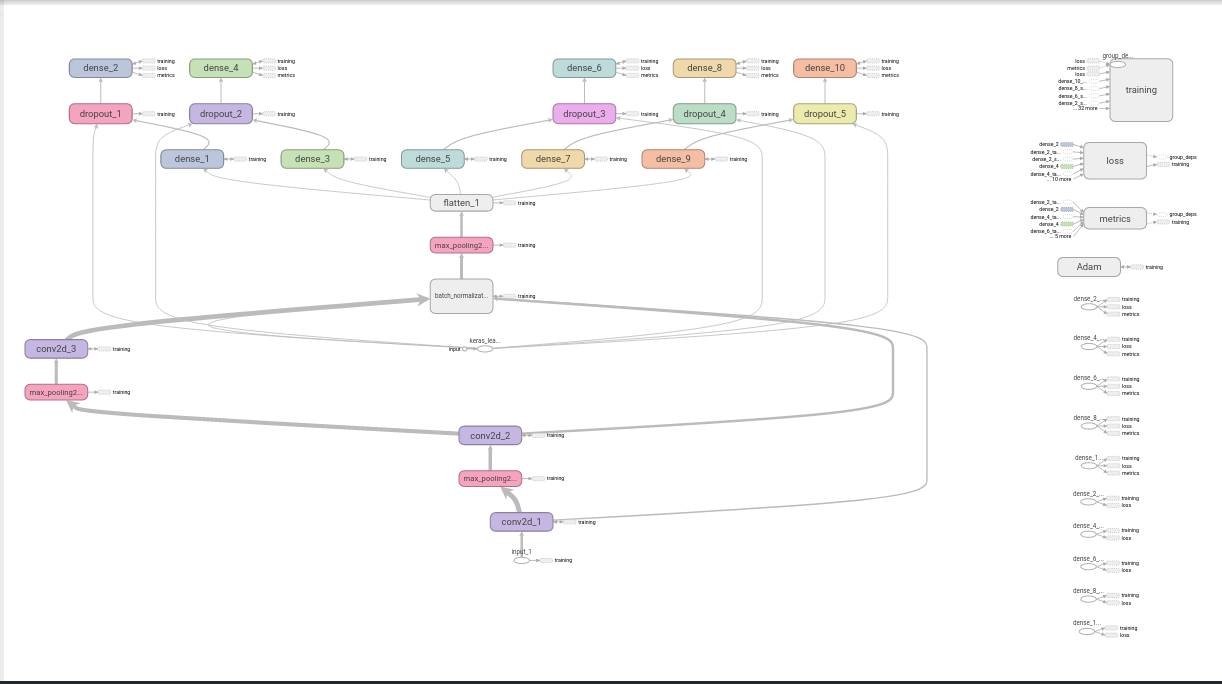}
	\caption{Tensorboard Visualization of the structure of the model (\ref{model_cv2}).}
	\label{tensorboard_modelcv2}
\end{figure}

\begin{figure}
	\centering
	\includegraphics[height=3cm,width=0.495\linewidth]{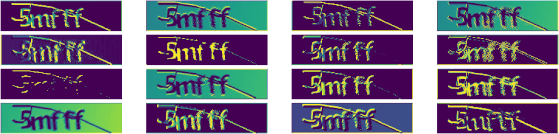}\hspace{4px}\includegraphics[height=3cm,width=0.495\linewidth]{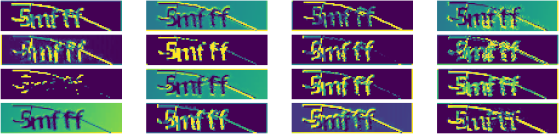}\vspace{4px}
	\includegraphics[height=3cm,width=1\linewidth]{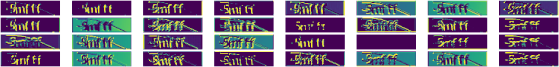}
	\includegraphics[height=3cm,width=1\linewidth]{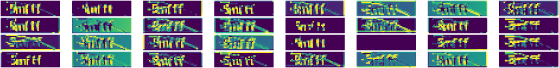}\vspace{4px}
	\includegraphics[height=3cm,width=1\linewidth]{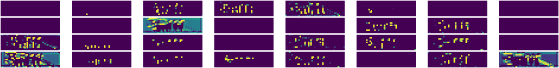}\vspace{4px}
	\includegraphics[height=3cm,width=1\linewidth]{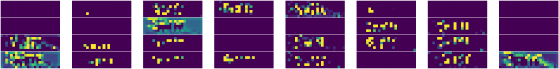}

	\caption{The visualizations of the features in a fully trained model for Model \ref{tensorboard_modelcv2} when passed a CAPTCHA containing the text \textbf{5mfff}.}
	\label{cv2_layers_model1_5mfff}
\end{figure}

\begin{figure}
	\centering
	\includegraphics[height=3cm,width=0.495\linewidth]{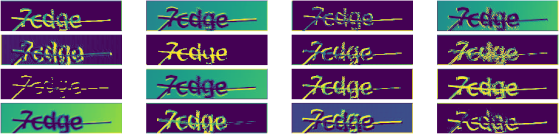}\hspace{4px}\includegraphics[height=3cm,width=0.495\linewidth]{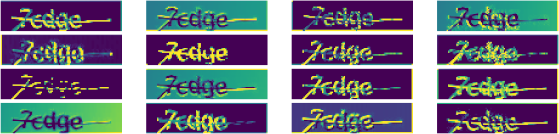}\vspace{4px}
	\includegraphics[height=3cm,width=1\linewidth]{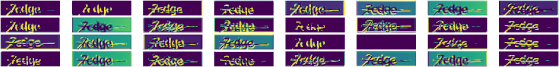}
	\includegraphics[height=3cm,width=1\linewidth]{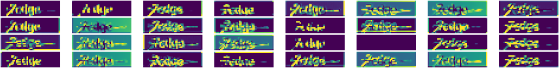}\vspace{4px}
	\includegraphics[height=3cm,width=1\linewidth]{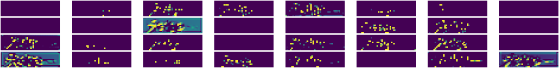}\vspace{4px}
	\includegraphics[height=3cm,width=1\linewidth]{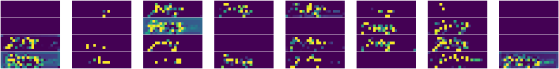}
	
	\caption{The visualizations of the features in a fully trained model for Model \ref{tensorboard_modelcv2} when passed a CAPTCHA containing the text \textbf{7cdge}.}
	\label{cv2_layers_model1_7cdge}
\end{figure}

\begin{figure}
	\centering
	\includegraphics[height=3cm,width=0.495\linewidth]{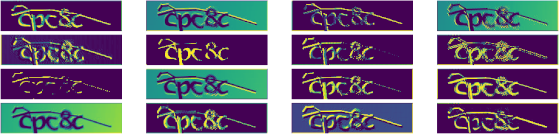}\hspace{4px}\includegraphics[height=3cm,width=0.495\linewidth]{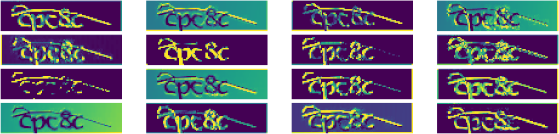}\vspace{4px}
	\includegraphics[height=3cm,width=1\linewidth]{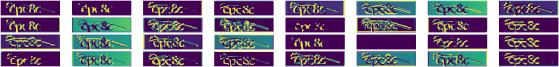}
	\includegraphics[height=3cm,width=1\linewidth]{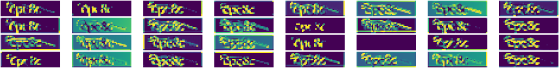}\vspace{4px}
	\includegraphics[height=3cm,width=1\linewidth]{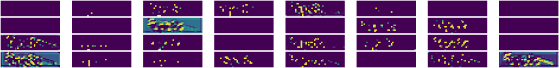}\vspace{4px}
	\includegraphics[height=3cm,width=1\linewidth]{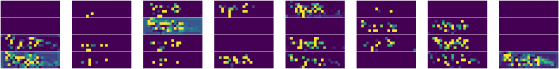}
	
	\caption{The visualizations of the features in a fully trained model for Model \ref{tensorboard_modelcv2} when passed a CAPTCHA containing the text \textbf{cpc8c}.}
	\label{cv2_layers_model1_cpc8c}
\end{figure}

\begin{figure}
	\centering
	\includegraphics[width=1\linewidth]{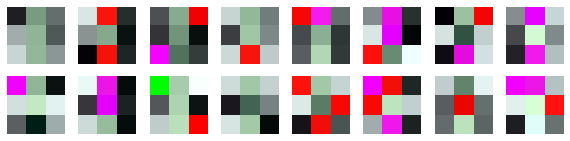}
	\caption{The 3x3 kernels (filter) learned by the model (\ref{tensorboard_modelcv2}) for layer 1.}
	\label{cv2-filter-learned}
\end{figure}

In Figure \ref{cv2-filter-learned} we see the 3x3 kernels that are being learned by the layer 1 of the network for the Model \ref{tensorboard_modelcv2}. The visualization of the weights learned by the same model can be seen in Figure \ref{cv2-weights-learned}.

The summary of the structure of the model is shown in Figure \ref{model_cv2_summary}. The graph diagram of the model is shown in Figure \ref{model_cv2}. The trained model can be retrieved from this url (\url{https://jimut123.github.io/blogs/CAPTCHA/models/captcha_v2_1_drive.h5}). This is clear that it is a multi-label classification which classifies 36 labels including the lower-case ASCII alphabets (a-z) and the number of digits from 0-9. The model was trained in 30 epochs, the final epochs' summary is shown in Figure \ref{model_cv2_epochs}. It took about 1 second per epochs after the first epoch which took 8 second. The total training time of the model was about 38 seconds which is pretty fast.

The graph of the accuracy and the loss are shown in Figure \ref{acc_cv2_model} and \ref{loss_cv2_model}. The tensor board visualization of the model graph is shown in Figure \ref{tensorboard_modelcv2}. The intermediate visualization of the convolution layer when passed through the model is shown in Figure \ref{cv2_layers_model1_5mfff} for the letter \textbf{5mfff}, \ref{cv2_layers_model1_7cdge} for the letter \textbf{7cdge}, and Figure \ref{cv2_layers_model1_cpc8c} for the letter \textbf{cpc8c}.

\chapter{Using 1L-labelled-CAPTCHAs dataset (99.67 \% accuracy)}

%\section{Model for 1L-labelled-CAPTCHAs dataset}

\begin{figure}
	\centering
	\includegraphics[width=1\linewidth]{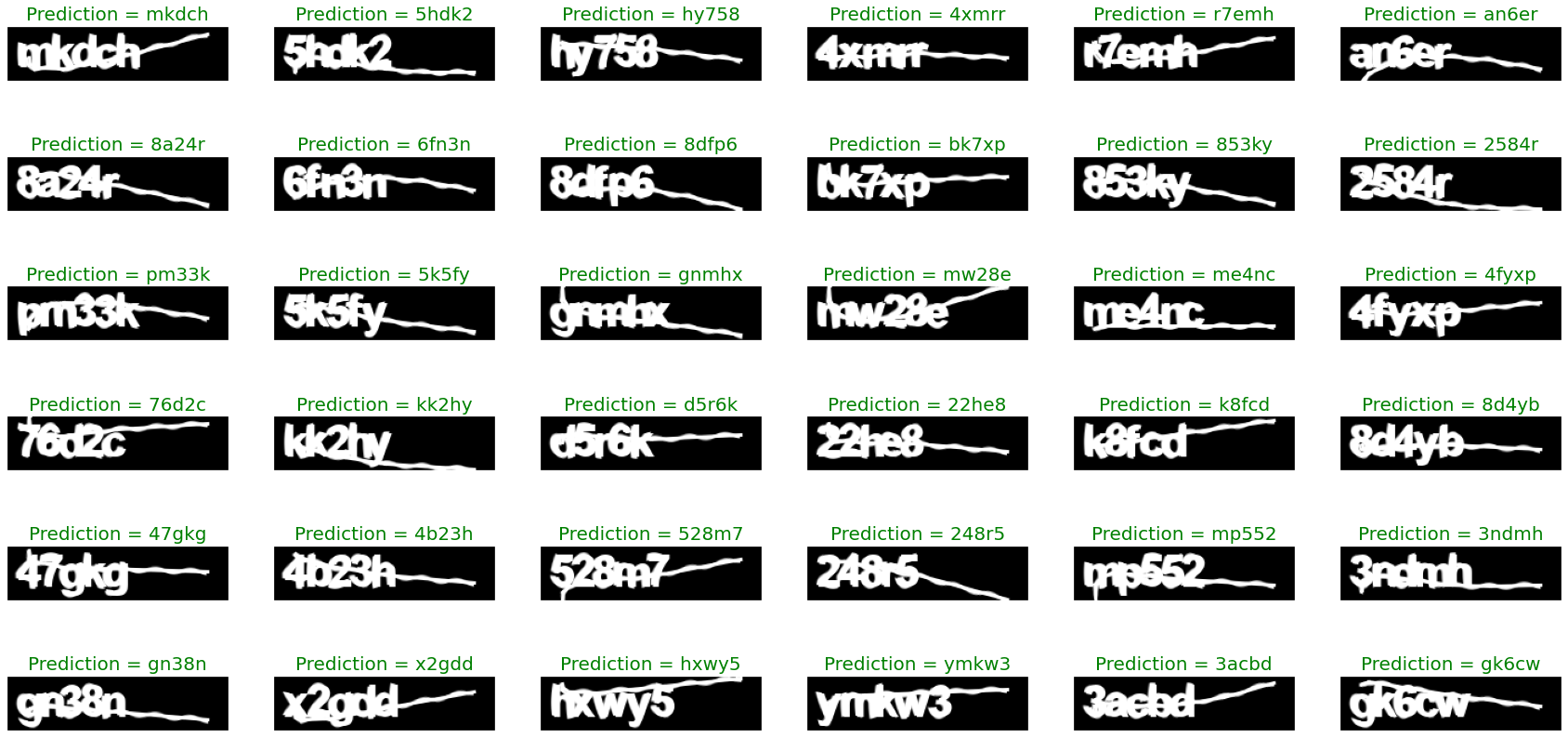}
	\caption{Predictions with the model (\ref{model_1Lc_1}).}
	\label{1Lc-predictions}
\end{figure}

\begin{figure} 
	\centering
	\fbox{\includegraphics[height=24cm,width=1\linewidth]{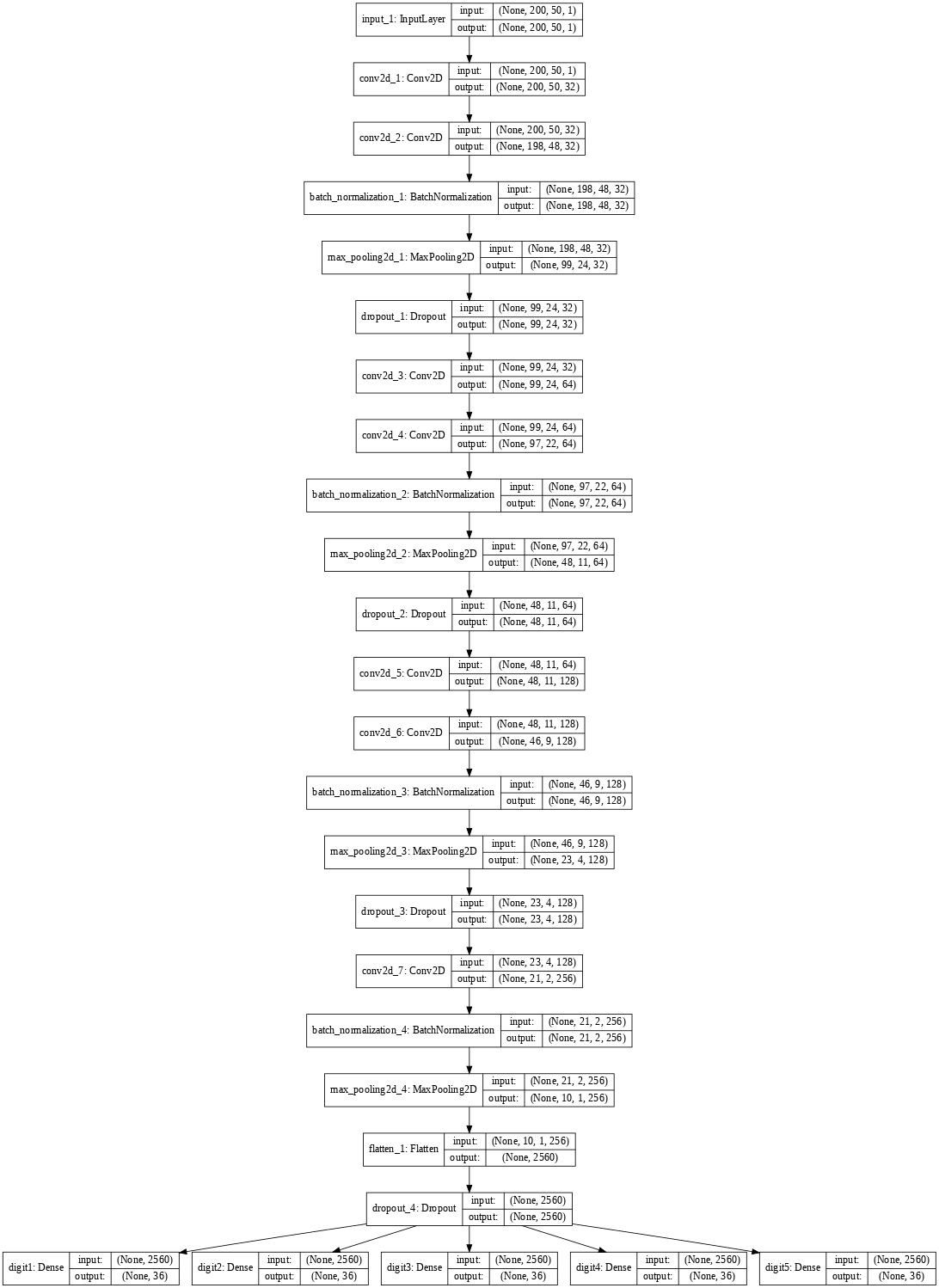}}	
	\caption{CIFAR 10 like model for identifying individual characters.}
	\label{model_1Lc_1}
\end{figure}

\begin{figure}
	\centering
	\includegraphics[width=1\linewidth]{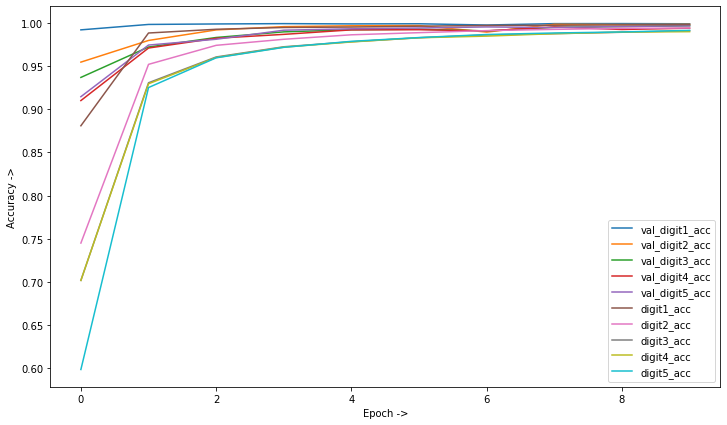}
	\caption{Training Accuracy for the model (\ref{model_1Lc_1}).}
	\label{1Lc-train-acc}
\end{figure}

\begin{figure}
	\centering
	\includegraphics[width=1\linewidth]{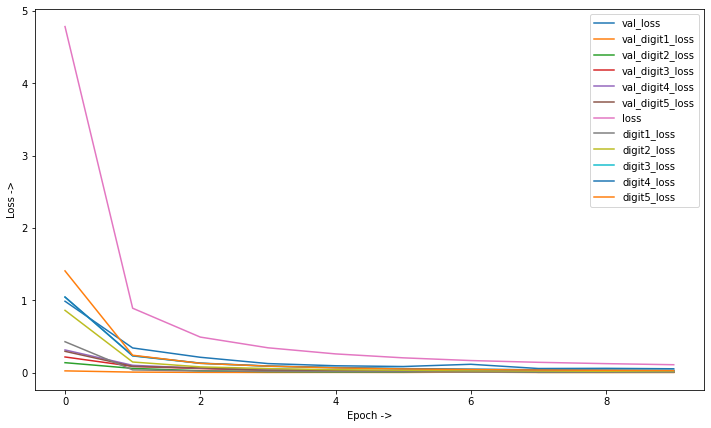}
	\caption{Training Loss for the model (\ref{model_1Lc_1}).}
	\label{1Lc-train-loss}
\end{figure}

\begin{figure}
	\centering
	\includegraphics[height=4cm,width=0.985\linewidth]{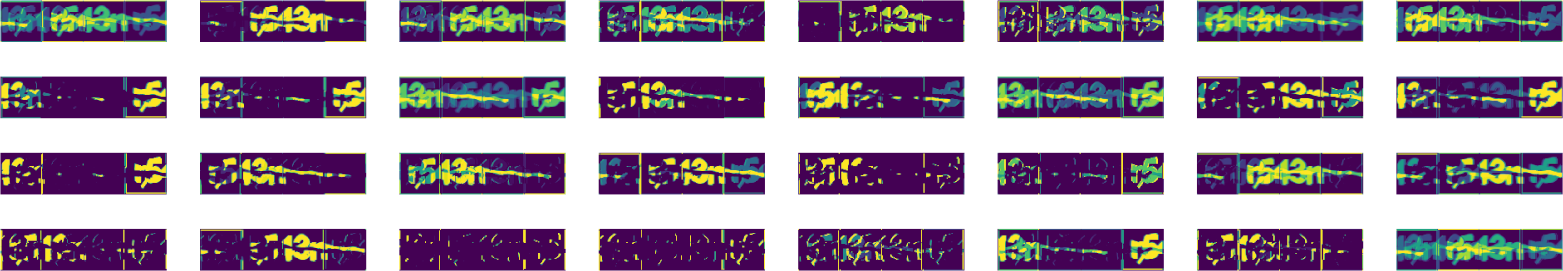}\vspace{2px}
	\includegraphics[height=4.6cm,width=0.5\linewidth]{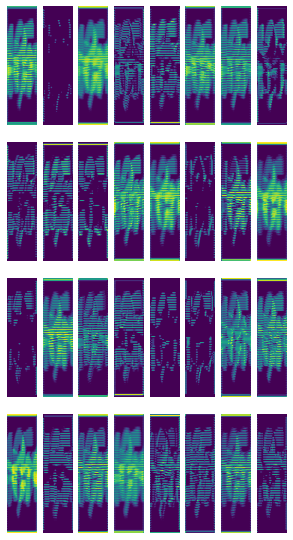}\includegraphics[height=4.6cm,width=0.5\linewidth]{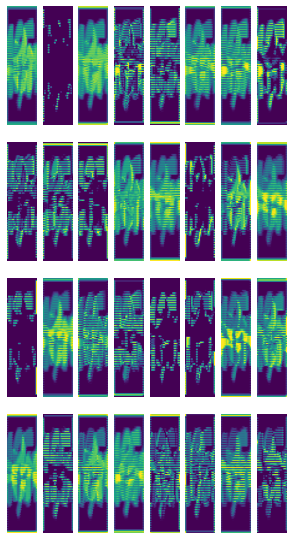}
	\includegraphics[height=3.9cm,width=0.5\linewidth]{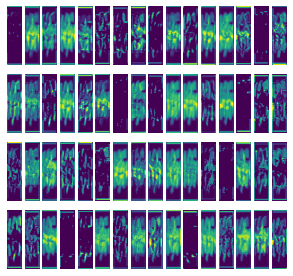}\includegraphics[height=3.9cm,width=0.5\linewidth]{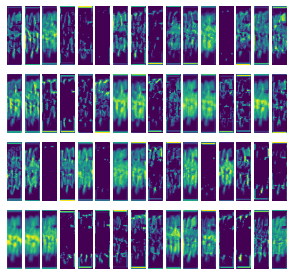}
	\includegraphics[height=3.9cm,width=0.5\linewidth]{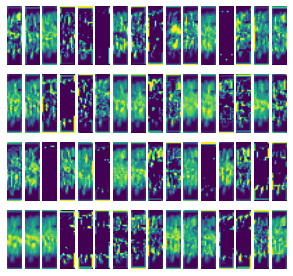}\includegraphics[height=4cm,width=0.5\linewidth]{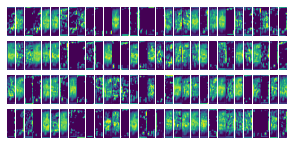}
	\includegraphics[height=3.9cm,width=0.5\linewidth]{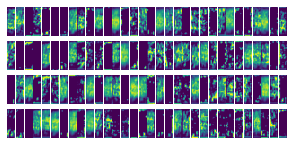}\includegraphics[height=3.9cm,width=0.5\linewidth]{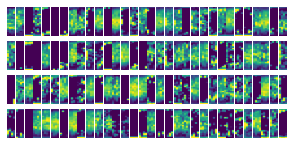}
	\includegraphics[height=3cm,width=0.5\linewidth]{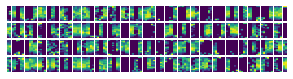}\includegraphics[height=3cm,width=0.5\linewidth]{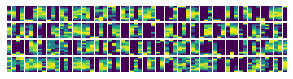}
	\caption{The visualizations of the features in a fully trained model (\ref{model_1Lc_1}) when passed a CAPTCHA containing the text r543n.}
	\label{1Lc-layers-viz}
\end{figure}

The summary of the model for 1L-labelled-CAPTCHAs dataset is shown in Figure \ref{model_1Lc_model}. The summary for the epochs for the model is shown in Figure \ref{model_1Lc_epochs}. It was trained on 30 epochs and each step of the epochs took about 71 seconds. This means that the total time taken for all of the computaion for 30 epochs is about 35.5 minutes. This is relatively slow as compared to the previous computation times as there are about 100000 images and the model has to go through each of the images for calculating the gradients. The images contained are of size 50x200 and is of \index[\idxKeywordName]{binary images} binary images (i.e., intensity value of 0 and 255). This is a multi-label classification which has a total of 36 labels containing the lower-case ASCII letters (a-z) and the digits 0-9. Since, this has a total of 5 labels, there are 5 nodes that are coming out of the final layer, which helps in classifying the labels.  The Keras visualization of the model is shown in Figure \ref{model_1Lc_1}. The training accuracy of the model is shown in Figure \ref{1Lc-train-acc} and the training loss is shown in Figure \ref{1Lc-train-loss}. The visualization of the intermediate layers of the learned model when a CAPTCHA containing the digit \textbf{r543n} is passed is shown in Figure \ref{1Lc-layers-viz}. The CAPTCHA gave a total accuracy of 99.67\% and it operates in human level. When passed through unseen data, we get predictions as shown in Figure \ref{1Lc-predictions}. The trained model can be retrieved from this url (\url{https://jimut123.github.io/blogs/CAPTCHA/models/images-1L-processed.h5}). We see that all of the CAPTCHA is correctly classified incspite of the clutter in the background.

\begin{figure} 
	\centering
	{%
		\lstset{frame=single,basicstyle=\scriptsize,style=myModelSummaryStyle}
		\centering
		\begin{lstlisting}
		Epoch 10/10
		72000/72000 [==============================] - 71s 991us/step - loss: 0.1094 - digit1_loss: 0.0055 - digit2_loss: 0.0171 - digit3_loss: 0.0281 - digit4_loss: 0.0317 - digit5_loss: 0.0270 - digit1_acc: 0.9982 - digit2_acc: 0.9942 - digit3_acc: 0.9911 - digit4_acc: 0.9900 - digit5_acc: 0.9914 - val_loss: 0.0532 - val_digit1_loss: 0.0023 - val_digit2_loss: 0.0049 - val_digit3_loss: 0.0095 - val_digit4_loss: 0.0212 - val_digit5_loss: 0.0154 - val_digit1_acc: 0.9991 - val_digit2_acc: 0.9984 - val_digit3_acc: 0.9968 - val_digit4_acc: 0.9940 - val_digit5_acc: 0.9967
		\end{lstlisting}
	}
	\caption{Summary of the epochs after training the model (\ref{model_1Lc_1}). }
	\label{model_1Lc_epochs}
\end{figure}

\begin{figure}
	\centering
	\includegraphics[width=0.75\linewidth]{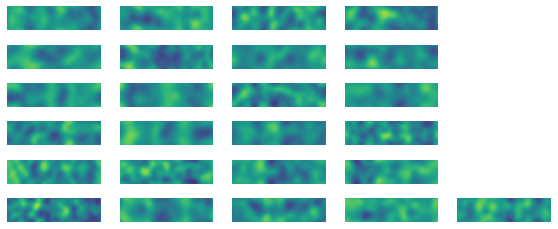}
	\caption{The weights learned by the model (\ref{model_1Lc_1}) for layers 1 to 26.}
	\label{1Lc-weights-learned}
\end{figure}

\begin{figure} 
	\centering
	{%
		\lstset{frame=single,basicstyle=\scriptsize,style=myModelSummaryStyle}
		\centering
		\begin{lstlisting}
			__________________________________________________________________________________________________
			Layer (type)                    Output Shape         Param #     Connected to                     
			==================================================
			input_1 (InputLayer)            (None, 200, 50, 1)   0                                            
			__________________________________________________________________________________________________
			conv2d_1 (Conv2D)               (None, 200, 50, 32)  320         input_1[0][0]                    
			__________________________________________________________________________________________________
			conv2d_2 (Conv2D)               (None, 198, 48, 32)  9248        conv2d_1[0][0]                   
			__________________________________________________________________________________________________
			batch_normalization_1 (BatchNor (None, 198, 48, 32)  128         conv2d_2[0][0]                   
			__________________________________________________________________________________________________
			max_pooling2d_1 (MaxPooling2D)  (None, 99, 24, 32)   0           batch_normalization_1[0][0]      
			__________________________________________________________________________________________________
			dropout_1 (Dropout)             (None, 99, 24, 32)   0           max_pooling2d_1[0][0]            
			__________________________________________________________________________________________________
			conv2d_3 (Conv2D)               (None, 99, 24, 64)   18496       dropout_1[0][0]                  
			__________________________________________________________________________________________________
			conv2d_4 (Conv2D)               (None, 97, 22, 64)   36928       conv2d_3[0][0]                   
			__________________________________________________________________________________________________
			batch_normalization_2 (BatchNor (None, 97, 22, 64)   256         conv2d_4[0][0]                   
			__________________________________________________________________________________________________
			max_pooling2d_2 (MaxPooling2D)  (None, 48, 11, 64)   0           batch_normalization_2[0][0]      
			__________________________________________________________________________________________________
			dropout_2 (Dropout)             (None, 48, 11, 64)   0           max_pooling2d_2[0][0]            
			__________________________________________________________________________________________________
			conv2d_5 (Conv2D)               (None, 48, 11, 128)  73856       dropout_2[0][0]                  
			__________________________________________________________________________________________________
			conv2d_6 (Conv2D)               (None, 46, 9, 128)   147584      conv2d_5[0][0]                   
			__________________________________________________________________________________________________
			batch_normalization_3 (BatchNor (None, 46, 9, 128)   512         conv2d_6[0][0]                   
			__________________________________________________________________________________________________
			max_pooling2d_3 (MaxPooling2D)  (None, 23, 4, 128)   0           batch_normalization_3[0][0]      
			__________________________________________________________________________________________________
			dropout_3 (Dropout)             (None, 23, 4, 128)   0           max_pooling2d_3[0][0]            
			__________________________________________________________________________________________________
			conv2d_7 (Conv2D)               (None, 21, 2, 256)   295168      dropout_3[0][0]                  
			__________________________________________________________________________________________________
			batch_normalization_4 (BatchNor (None, 21, 2, 256)   1024        conv2d_7[0][0]                   
			__________________________________________________________________________________________________
			max_pooling2d_4 (MaxPooling2D)  (None, 10, 1, 256)   0           batch_normalization_4[0][0]      
			__________________________________________________________________________________________________
			flatten_1 (Flatten)             (None, 2560)         0           max_pooling2d_4[0][0]            
			__________________________________________________________________________________________________
			dropout_4 (Dropout)             (None, 2560)         0           flatten_1[0][0]                  
			__________________________________________________________________________________________________
			digit1 (Dense)                  (None, 36)           92196       dropout_4[0][0]                  
			__________________________________________________________________________________________________
			digit2 (Dense)                  (None, 36)           92196       dropout_4[0][0]                  
			__________________________________________________________________________________________________
			digit3 (Dense)                  (None, 36)           92196       dropout_4[0][0]                  
			__________________________________________________________________________________________________
			digit4 (Dense)                  (None, 36)           92196       dropout_4[0][0]                  
			__________________________________________________________________________________________________
			digit5 (Dense)                  (None, 36)           92196       dropout_4[0][0]                  
			==================================================
			Total params: 1,044,500
			Trainable params: 1,043,540
			Non-trainable params: 960
			__________________________________________________________________________________________________
			
			
		\end{lstlisting}
	}

	\caption{Model (\ref{model_1Lc_1}) for identifying 1L-labelled-CAPTCHA dataset.}
	\label{model_1Lc_model}
\end{figure}

In Figure \ref{1Lc-filter-learned} we see the 3x3 kernels that are being learned by the layer 1 of the network for the Model \ref{model_1Lc_1}. The visualization of the weights learned by the same model can be seen in Figure \ref{1Lc-weights-learned}.

\begin{figure}
	\centering
	\includegraphics[width=0.95\linewidth]{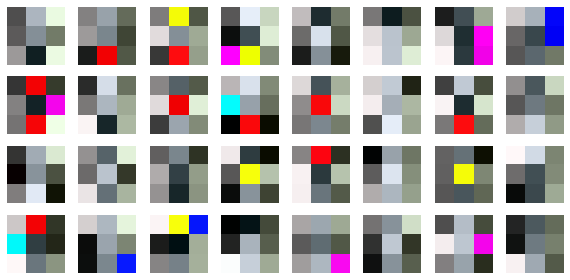}
	\caption{The 3x3 kernels (filter) learned by the model (\ref{model_1Lc_1}) for layer 1.}
	\label{1Lc-filter-learned}
\end{figure}

\chapter{Using faded CAPTCHA dataset (99.44 \% accuracy)}

\section{AlexNet type model for faded CAPTCHA dataset}

We have used an Alex-Net type model for the classification of the CAPTCHA. We haven't performed any image operations and preserved the original size of the CAPTCHA data. The fundamental feature of the model which makes it different from others is it have a dropout layer before the last output in multilabel classification. This makes it robust to noise, since dropouts helps in training ensemble of Neural Networks. The summary of the model can be shown below.

{%
	\centering
	\begin{lstlisting}
	_________________________________________________________________
	Layer (type)                 Output Shape              Param #   
	=================================================================
	input_4 (InputLayer)         [(None, 100, 120, 3)]     0         
	_________________________________________________________________
	conv2d (Conv2D)              (None, 98, 118, 32)       896       
	_________________________________________________________________
	max_pooling2d (MaxPooling2D) (None, 49, 59, 32)        0         
	_________________________________________________________________
	conv2d_1 (Conv2D)            (None, 47, 57, 64)        18496     
	_________________________________________________________________
	max_pooling2d_1 (MaxPooling2 (None, 23, 28, 64)        0         
	_________________________________________________________________
	conv2d_2 (Conv2D)            (None, 21, 26, 64)        36928     
	_________________________________________________________________
	max_pooling2d_2 (MaxPooling2 (None, 10, 13, 64)        0         
	_________________________________________________________________
	flatten (Flatten)            (None, 8320)              0         
	_________________________________________________________________
	dense (Dense)                (None, 1024)              8520704   
	_________________________________________________________________
	dropout (Dropout)            (None, 1024)              0         
	_________________________________________________________________
	dense_1 (Dense)              (None, 40)                41000     
	_________________________________________________________________
	reshape (Reshape)            (None, 4, 10)             0         
	=================================================================
	Total params: 8,618,024
	Trainable params: 8,618,024
	Non-trainable params: 0
	_________________________________________________________________
		
	\end{lstlisting}
}

\begin{figure} 
	\centering
	\fbox{\includegraphics[width=0.75\linewidth]{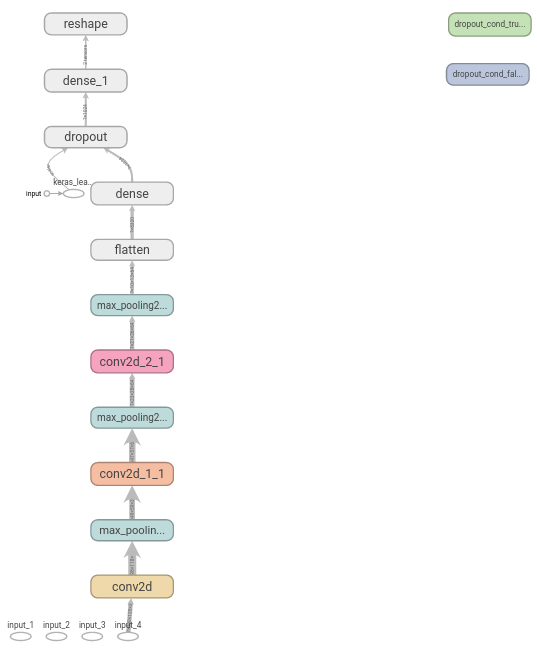}}	
	\caption{Alex-Net type model for multi-label classification.}
	\label{model_faded}
\end{figure}

\begin{figure}
	\centering
	\includegraphics[width=0.95\linewidth]{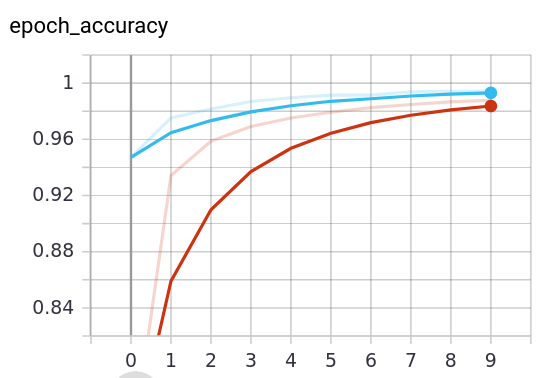}
	\includegraphics[width=0.95\linewidth]{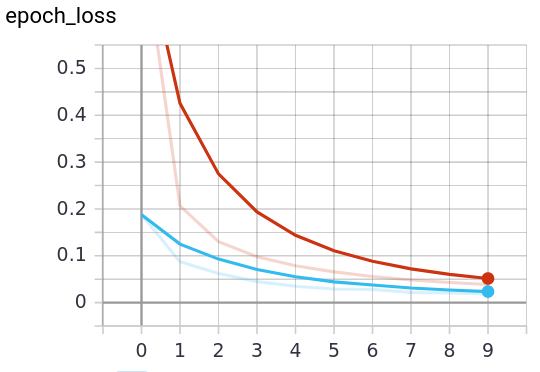}
	\caption{The accuracy and Loss obtained from the model when visualized via tensorboard. }
	\label{faded_loss_acc_tensorbd}
\end{figure}

The structure of the model when visualized in tensorboard can be found from figure \ref{model_faded}. It took 10 epochs and a time of about 923 second per epochs. So, it basically took about 2.56 hours to finish the training. We can clearly see that it gets an accuracy of about 98.78\% on the validation set and an accuracy of 99.44\% on the test set, which is comparable to human level operators given the complexity and noise of this type of CAPTCHA. The loss and accuracy obtained from the training of the model is shown in Figure \ref{faded_loss_acc_tensorbd}. 

\begin{figure} 
	\centering
	\begin{lstlisting}
	Epoch 10/10
	533/533 [==============================] - ETA: 0s - loss: 0.0386 - accuracy: 0.9878INFO:tensorflow:Assets written to: ./model_checkpoint/assets
	533/533 [==============================] - 923s 2s/step - loss: 0.0386 - accuracy: 0.9878 - val_loss: 0.0186 - val_accuracy: 0.9944
	\end{lstlisting}
\caption{epochs for faded CAPTCHA.}
\label{model_epochs_faded}
\end{figure}

\begin{figure}
	\centering
	\includegraphics[width=0.95\linewidth]{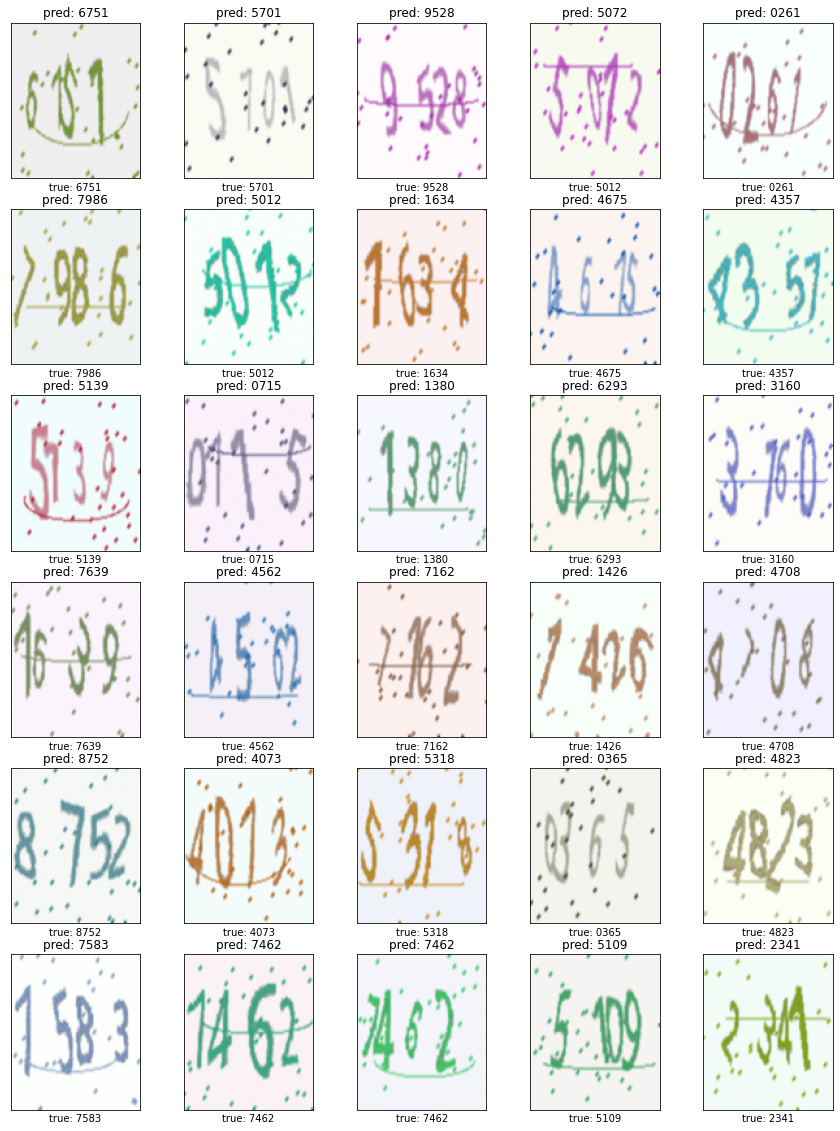}
	\caption{The prediction of the model with 100\% accuracy on unseen data.}
	\label{pred_faded}
\end{figure}

When we test the model on unseen data, we get about 100\% accuracy as shown in Figure \ref{pred_faded}. The model obtained can be found from here \url{https://jimut123.github.io/blogs/CAPTCHA/models/faded_captcha.h5}, which is about 98.7 MiB in size. From this we find that AlexNet have a capability of removing noise and generalizing well on unseen data.

\chapter{Using Circle CAPTCHA dataset (99.99 \% accuracy)}

\section{AlexNet type model for Circle CAPTCHA dataset}

We have used a similar type of model, from \ref{model_faded}, for preserving all the three channels of the CAPTCHA data. The summary of the model is shown below \ref{circle_captcha_summary}:

\begin{figure}
	\centering
	{%
		\lstset{frame=single,basicstyle=\scriptsize,style=myModelSummaryStyle}
		\centering
		\begin{lstlisting}
		_________________________________________________________________
		Layer (type)                 Output Shape              Param #   
		==================================================
		input_1 (InputLayer)         [(None, 35, 120, 3)]      0         
		_________________________________________________________________
		conv2d (Conv2D)              (None, 33, 118, 32)       896       
		_________________________________________________________________
		max_pooling2d (MaxPooling2D) (None, 16, 59, 32)        0         
		_________________________________________________________________
		conv2d_1 (Conv2D)            (None, 14, 57, 64)        18496     
		_________________________________________________________________
		max_pooling2d_1 (MaxPooling2 (None, 7, 28, 64)         0         
		_________________________________________________________________
		conv2d_2 (Conv2D)            (None, 5, 26, 64)         36928     
		_________________________________________________________________
		max_pooling2d_2 (MaxPooling2 (None, 2, 13, 64)         0         
		_________________________________________________________________
		flatten (Flatten)            (None, 1664)              0         
		_________________________________________________________________
		dense (Dense)                (None, 1024)              1704960   
		_________________________________________________________________
		dropout (Dropout)            (None, 1024)              0         
		_________________________________________________________________
		dense_1 (Dense)              (None, 216)               221400    
		_________________________________________________________________
		reshape (Reshape)            (None, 6, 36)             0         
		==================================================
		Total params: 1,982,680
		Trainable params: 1,982,680
		Non-trainable params: 0
		_________________________________________________________________
		\end{lstlisting}
	}
	\caption{Summary of the model for \ref{model_mc}.}
	\label{circle_captcha_summary}
\end{figure}

\begin{figure}
	\centering
	\includegraphics[height=20cm,width=0.5\linewidth]{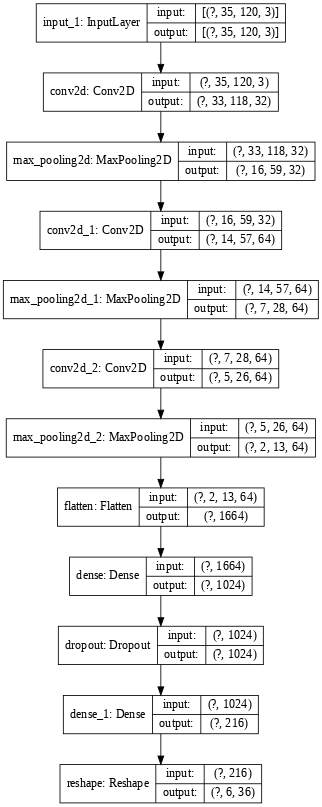}
	\caption{Keras model structure for Circle CAPTCHA.}
	\label{keras_model_circle_captcha}
\end{figure}

\begin{figure}
	\centering
	\includegraphics[width=0.5\linewidth]{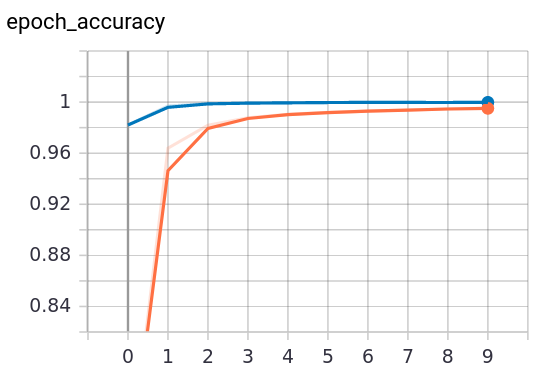}
	\caption{Accuracy of the model (\ref{keras_model_circle_captcha}).}
	\label{acc_cc_model}
\end{figure}

\begin{figure}
	\centering
	\includegraphics[width=0.5\linewidth]{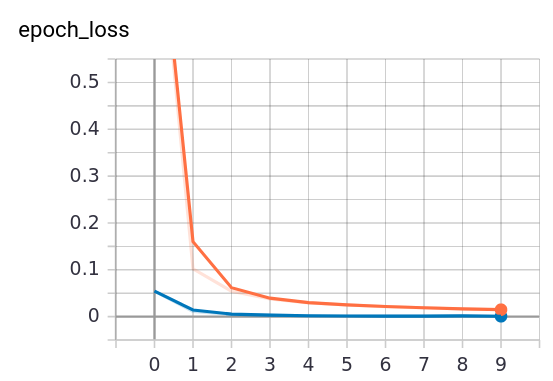}
	\caption{Loss of the model (\ref{keras_model_circle_captcha}).}
	\label{loss_cc_model}
\end{figure}

\begin{figure}
	\centering
	\includegraphics[width=1\linewidth]{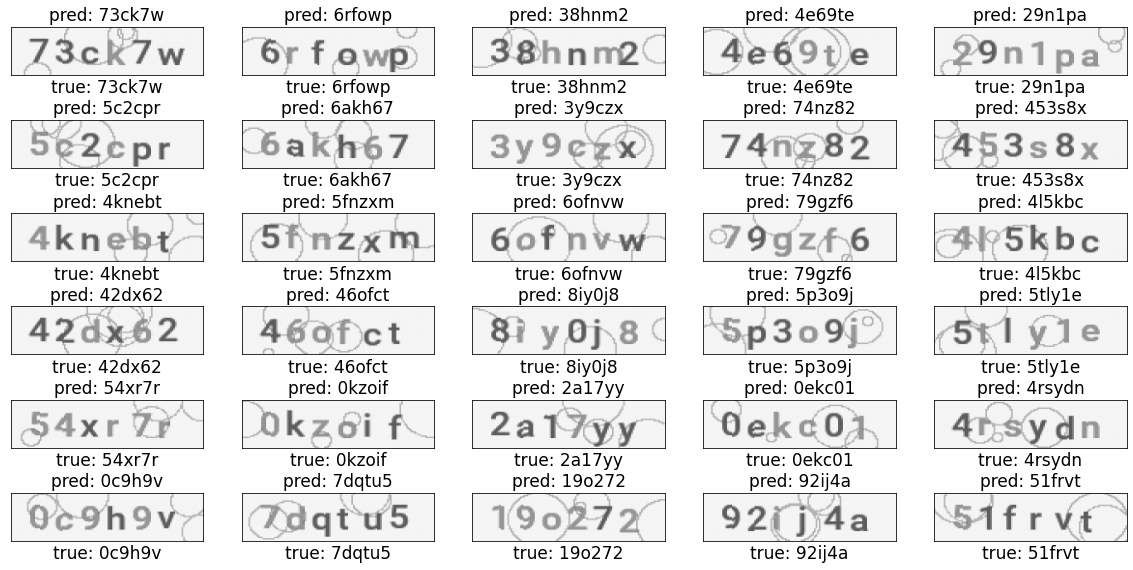}
	\caption{Predictions of the model (\ref{keras_model_circle_captcha}).}
	\label{cc_pred}
\end{figure}

The Model was trained for 10 epochs and it gave a validation accuracy of 99.99\% and 99.98\% on the test set. This is pretty good result since the model kind of learns from the little circular marks on the dataset that they are just noise. The model took about 650 second per epochs and a total of 1.80 hours for finishing the training for 10 epochs. The model can be found from this link (\url{https://jimut123.github.io/blogs/CAPTCHA/models/circle_captcha_10e.h5}). The size of the model is about 22.7 MiB. The structure of the model is shown in Figure \ref{keras_model_circle_captcha}, the visualization of the accuracy of the model is shown in Figure \ref{acc_cc_model}, and the loss for the same model is shown in Figure \ref{loss_cc_model}. The prediction of the model is shown in Figure \ref{cc_pred}, on unseen data, and it accurately classifies all the labels of the CAPTCHA.

\begin{figure} 
	\centering
	\begin{lstlisting}
	Epoch 10/10
	392/392 [==============================] - ETA: 0s - loss: 0.0147 - accuracy: 0.9951INFO:tensorflow:Assets written to: ./model_checkpoint/assets
	392/392 [==============================] - 650s 2s/step - loss: 0.0147 - accuracy: 0.9951 - val_loss: 6.1774e-04 - val_accuracy: 0.9999
	\end{lstlisting}
	\caption{epochs for circle CAPTCHA.}
	\label{model_epochs_circle_captcha}
\end{figure}


\chapter{Using Sphinx CAPTCHA dataset (99.62 \% accuracy)}

\section{AlexNet type model for faded CAPTCHA dataset}

We have used a similar type of model, from \ref{model_faded}, for preserving all the three channels of the CAPTCHA data. The summary of the model is shown below \ref{sphinx_captcha_summary}:

\begin{figure}
	\centering
	{%
		\lstset{frame=single,basicstyle=\scriptsize,style=myModelSummaryStyle}
		\centering
		\begin{lstlisting}
		_________________________________________________________________
		Layer (type)                 Output Shape              Param #   
		==================================================
		input_1 (InputLayer)         [(None, 40, 150, 3)]      0         
		_________________________________________________________________
		conv2d (Conv2D)              (None, 38, 148, 32)       896       
		_________________________________________________________________
		max_pooling2d (MaxPooling2D) (None, 19, 74, 32)        0         
		_________________________________________________________________
		conv2d_1 (Conv2D)            (None, 17, 72, 64)        18496     
		_________________________________________________________________
		max_pooling2d_1 (MaxPooling2 (None, 8, 36, 64)         0         
		_________________________________________________________________
		conv2d_2 (Conv2D)            (None, 6, 34, 64)         36928     
		_________________________________________________________________
		max_pooling2d_2 (MaxPooling2 (None, 3, 17, 64)         0         
		_________________________________________________________________
		flatten (Flatten)            (None, 3264)              0         
		_________________________________________________________________
		dense (Dense)                (None, 1024)              3343360   
		_________________________________________________________________
		dropout (Dropout)            (None, 1024)              0         
		_________________________________________________________________
		dense_1 (Dense)              (None, 76)                77900     
		_________________________________________________________________
		reshape (Reshape)            (None, 4, 19)             0         
		==================================================
		Total params: 3,477,580
		Trainable params: 3,477,580
		Non-trainable params: 0
		_________________________________________________________________
		\end{lstlisting}
	}
\caption{Summary of the model for \ref{model_mc}.}
\label{sphinx_captcha_summary}
\end{figure}

The model is trained for about 15 epochs. Each epoch took about 1293 second, which means it took about 5.3875 hours to train the whole model. The summary of the epochs is shown in \ref{model_epochs_sphinx}. The summary of the 15th epochs is about 99.62 \% accuracy on the validation set and test set, which means it is able to capture the characters from the noise very well. It is able to generalize the data pretty good.

\begin{figure} 
	\centering
	\begin{lstlisting}
	Epoch 15/15
	446/446 [==============================] - ETA: 0s - loss: 0.0503 - accuracy: 0.9833INFO:tensorflow:Assets written to: ./model_checkpoint/assets
	446/446 [==============================] - 1293s 3s/step - loss: 0.0503 - accuracy: 0.9833 - val_loss: 0.0125 - val_accuracy: 0.9962
	\end{lstlisting}
	\caption{epochs for sphinx CAPTCHA.}
	\label{model_epochs_sphinx}
\end{figure}

The training accuracy and loss can be shown in Figure \ref{acc_sphinx_model} and \ref{loss_sphinx_model} respectively. The prediction of the model on unseen data can be seen in Figure \ref{sphinx_pred}. This is like human level accuracy and this CAPTCHA is widely used over the internet. So, AlexNet type model can capture the necessary information very well and differentiate from noise very well. The whole CAPTCHA data is about 12.7 GB and the model is of size 39.9 MB. The model can be obtained from this link (\url{https://jimut123.github.io/blogs/CAPTCHA/models/sphinx_full_34_15e_9962.h5}).

\begin{figure}
	\centering
	\includegraphics[height=20cm,width=0.5\linewidth]{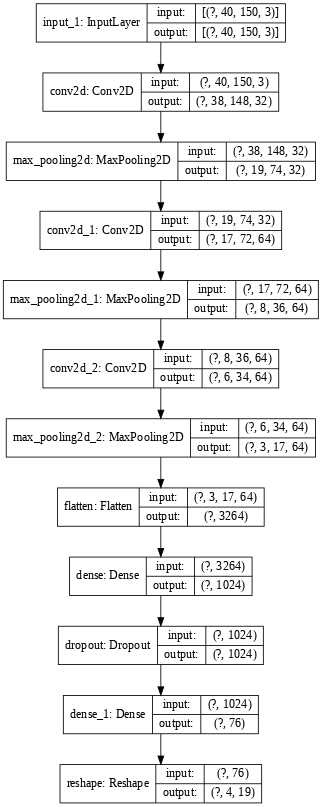}
	\caption{Tensorboard Visualization of the structure of the model (\ref{model_cv2}).}
	\label{keras_model_sphinx}
\end{figure}

\begin{figure}
	\centering
	\includegraphics[width=0.5\linewidth]{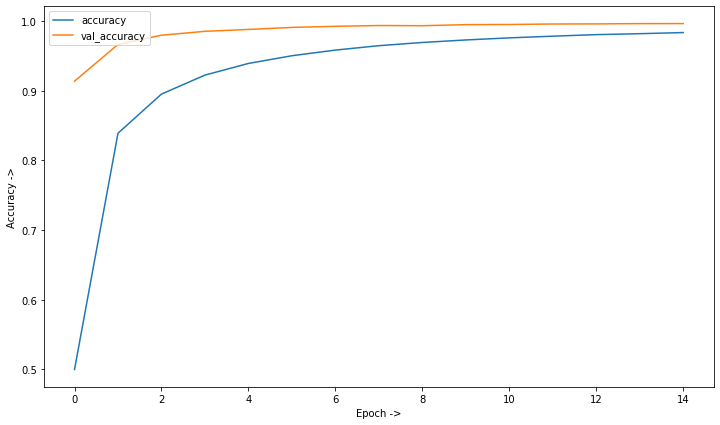}
	\caption{Accuracy of the model (\ref{keras_model_sphinx}).}
	\label{acc_sphinx_model}
\end{figure}
\begin{figure}
	\centering
	\includegraphics[width=0.5\linewidth]{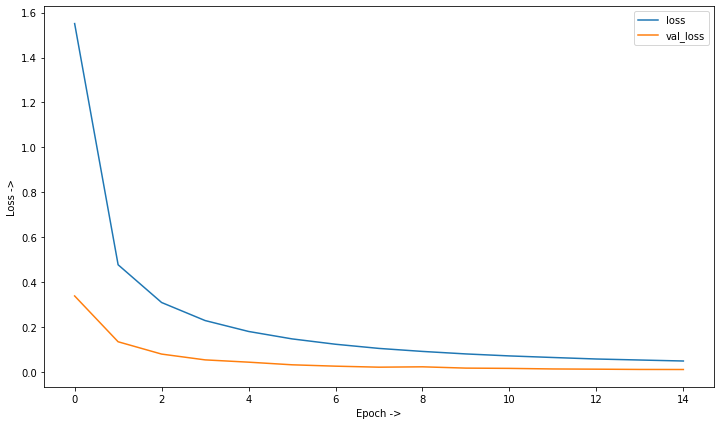}
	\caption{Loss of the model (\ref{keras_model_sphinx}).}
	\label{loss_sphinx_model}
\end{figure}

\begin{figure}
	\centering
	\includegraphics[width=1\linewidth]{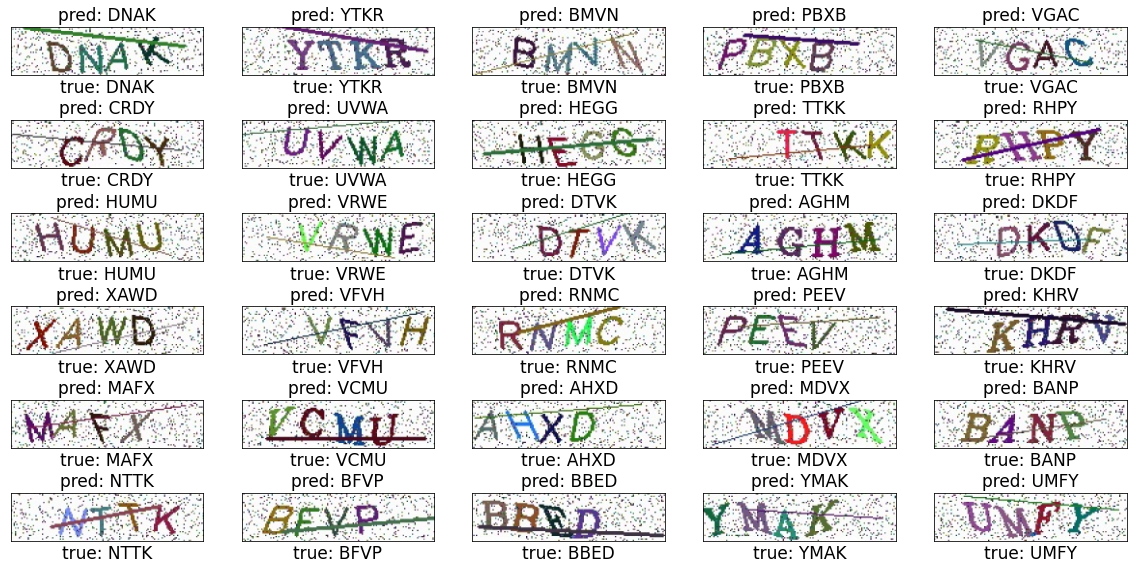}
	\caption{Predictions of the model (\ref{keras_model_sphinx}).}
	\label{sphinx_pred}
\end{figure}

\chapter{Using fish eye dataset (99.46 \% accuracy)}

\section{Using AlexNet type Model}

\begin{figure}
	\centering
	\includegraphics[height=20cm,width=0.5\linewidth]{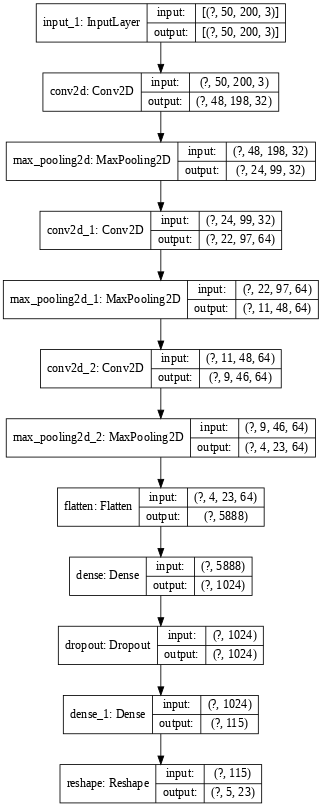}
	\caption{Tensorboard Visualization of the structure of the model (\ref{model_cv2}).}
	\label{keras_model_fe}
\end{figure}

Since there is a lot of clutter in the CAPTCHA dataset, we will use this AlexNet type model. The summary of the model is shown in Figure \ref{model_fe_summary_alexNet}. This is similar to the type of model shown in Figure \ref{model_faded}, except the fact that there are 5 outgoing labels from the final layer of the model. The dataset was trained using this model, and the accuracy obtained can be shown in Figure \ref{model_fe_epochs}. It took about 3902 second per epochs and there was a total of 5 epochs. The total training time was about 10.83 hours, which is too much. The plot of the model can be seen in Figure \ref{keras_model_fe}. The accuracy obtained on training's validation set was about 97.49 \% and the same for test dataset was 99.46
\% which was pretty great given that the model converges in only 5 epochs. Since the data size was too large, so we did a batch-based training where the training batch size was about 2048 and validation batch size was 512. The trained model can be obtained from the following link (\url{https://jimut123.github.io/blogs/CAPTCHA/models/fish_eye.h5}).

\begin{figure}
	\centering
	\includegraphics[width=0.5\linewidth]{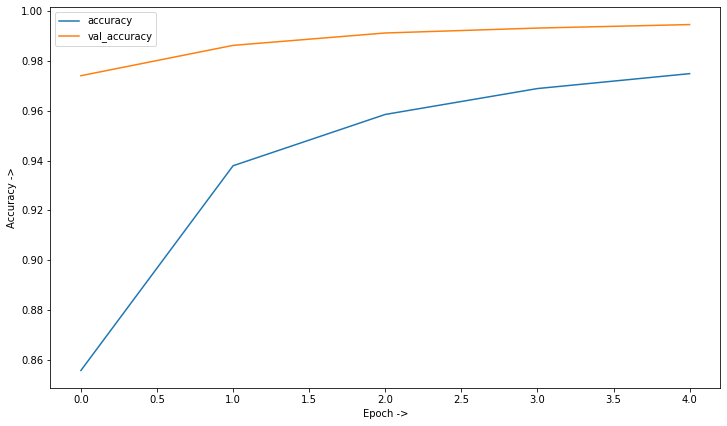}
	\caption{Accuracy of the model (\ref{keras_model_fe}).}
	\label{acc_fe_model}
\end{figure}

\begin{figure}
	\centering
	\includegraphics[width=0.5\linewidth]{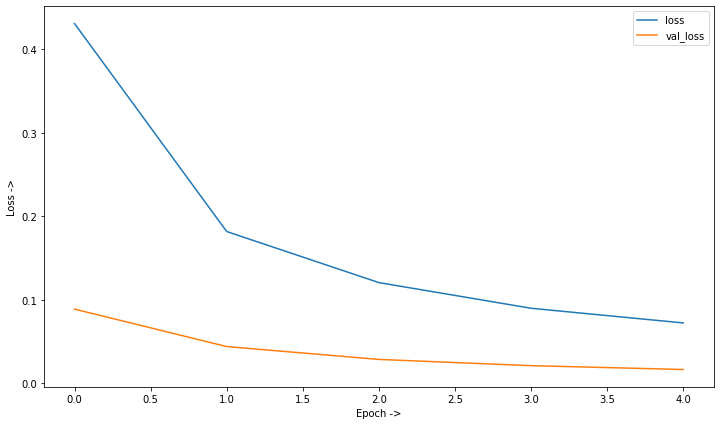}
	\caption{Loss of the model (\ref{keras_model_fe}).}
	\label{loss_fe_model}
\end{figure}

\begin{figure}
	\centering
	\includegraphics[width=1\linewidth]{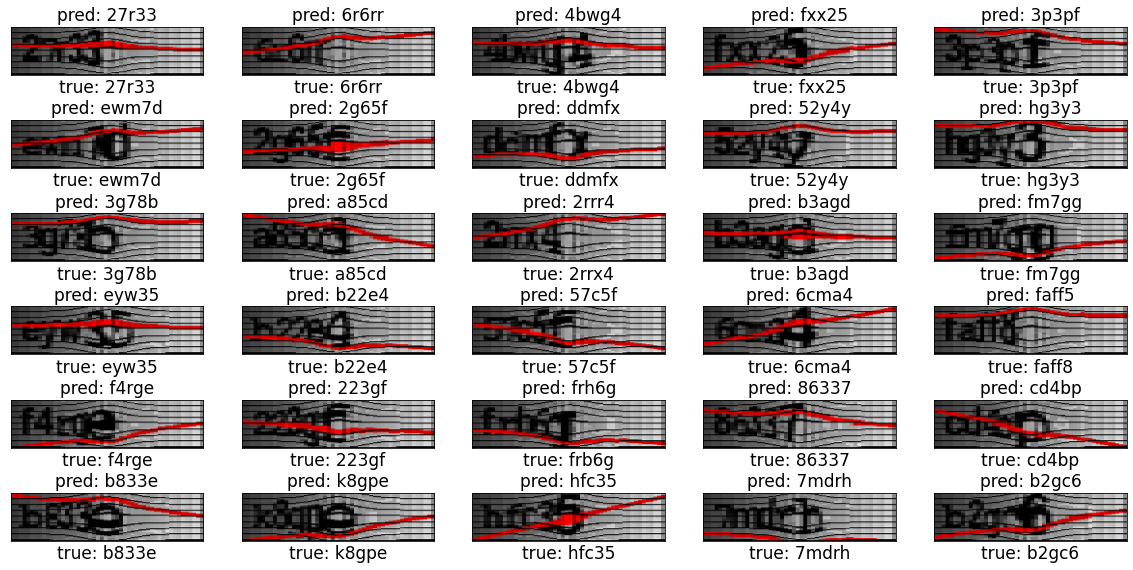}
	\caption{Predictions of the model (\ref{keras_model_fe}).}
	\label{fe_pred}
\end{figure}

The accuracy and the loss obtained by the model can be shown in Figure \ref{acc_fe_model} and Figure \ref{loss_fe_model} respectively. The predictions of the model on fully unseen data is shown in Figure \ref{fe_pred}, which shows that it predicts almost all of the CAPTCHA labels correctly and is able to generalize well on completely unseen data. The only wrong classifications it wrongly classified was \textbf{2rrx4} as \textbf{2rrr4}, \textbf{frb6g} as \textbf{frh6g}, and  \textbf{faff8} as \textbf{faff5} which is even difficult for humans to classify correctly since the letters were cluttered in occlusion.

\begin{figure} 
	\centering
	{%
		\lstset{frame=single,basicstyle=\scriptsize,style=myModelSummaryStyle}
		\centering
		\begin{lstlisting}
		_________________________________________________________________
		_________________________________________________________________
		Layer (type)                 Output Shape              Param #   
		==================================================
		input_1 (InputLayer)         [(None, 50, 200, 3)]      0         
		_________________________________________________________________
		conv2d (Conv2D)              (None, 48, 198, 32)       896       
		_________________________________________________________________
		max_pooling2d (MaxPooling2D) (None, 24, 99, 32)        0         
		_________________________________________________________________
		conv2d_1 (Conv2D)            (None, 22, 97, 64)        18496     
		_________________________________________________________________
		max_pooling2d_1 (MaxPooling2 (None, 11, 48, 64)        0         
		_________________________________________________________________
		conv2d_2 (Conv2D)            (None, 9, 46, 64)         36928     
		_________________________________________________________________
		max_pooling2d_2 (MaxPooling2 (None, 4, 23, 64)         0         
		_________________________________________________________________
		flatten (Flatten)            (None, 5888)              0         
		_________________________________________________________________
		dense (Dense)                (None, 1024)              6030336   
		_________________________________________________________________
		dropout (Dropout)            (None, 1024)              0         
		_________________________________________________________________
		dense_1 (Dense)              (None, 115)               117875    
		_________________________________________________________________
		reshape (Reshape)            (None, 5, 23)             0         
		==================================================
		Total params: 6,204,531
		Trainable params: 6,204,531
		Non-trainable params: 0
		_________________________________________________________________

		
		\end{lstlisting}
	}
	\caption{Summary of the AlexNet type model for classifying the labels of fish eye CAPTCHA dataset.}
	\label{model_fe_summary_alexNet}
\end{figure}

\begin{figure} 
	\centering
	{%
		\lstset{frame=single,basicstyle=\scriptsize,style=myModelSummaryStyle}
		\centering
		\begin{lstlisting}
		
		881/881 [==============================] - 3902s 4s/step - loss: 0.0725 - accuracy: 0.9749 - val_loss: 0.0167 - val_accuracy: 0.9946
		
		\end{lstlisting}
	}
	\caption{Summary of epochs for the model (\ref{model_cv2}).}
	\label{model_fe_epochs}
\end{figure}


\chapter{Using Mini CAPTCHA dataset (97.25 \% accuracy)}

\section{AlexNet type model for Mini CAPTCHA dataset}

We have used a similar type of model, from \ref{model_faded}, for preserving all the three channels of the CAPTCHA data. The summary of the model is shown below \ref{mini_captcha_summary}:

\begin{figure}
	\centering
	{%
		\lstset{frame=single,basicstyle=\scriptsize,style=myModelSummaryStyle}
		\centering
		\begin{lstlisting}
		_________________________________________________________________
		Layer (type)                 Output Shape              Param #   
		==================================================
		input_1 (InputLayer)         [(None, 100, 264, 3)]     0         
		_________________________________________________________________
		conv2d (Conv2D)              (None, 98, 262, 32)       896       
		_________________________________________________________________
		max_pooling2d (MaxPooling2D) (None, 49, 131, 32)       0         
		_________________________________________________________________
		conv2d_1 (Conv2D)            (None, 47, 129, 64)       18496     
		_________________________________________________________________
		max_pooling2d_1 (MaxPooling2 (None, 23, 64, 64)        0         
		_________________________________________________________________
		conv2d_2 (Conv2D)            (None, 21, 62, 64)        36928     
		_________________________________________________________________
		max_pooling2d_2 (MaxPooling2 (None, 10, 31, 64)        0         
		_________________________________________________________________
		flatten (Flatten)            (None, 19840)             0         
		_________________________________________________________________
		dense (Dense)                (None, 1024)              20317184  
		_________________________________________________________________
		dropout (Dropout)            (None, 1024)              0         
		_________________________________________________________________
		dense_1 (Dense)              (None, 248)               254200    
		_________________________________________________________________
		reshape (Reshape)            (None, 4, 62)             0         
		==================================================
		Total params: 20,627,704
		Trainable params: 20,627,704
		Non-trainable params: 0
		_________________________________________________________________
		\end{lstlisting}
	}
	\caption{Summary of the model for \ref{model_mc}.}
	\label{mini_captcha_summary}
\end{figure}

The Model was trained for 10 epochs and it gave a validation accuracy of 97.25\% and 97.09\% on the test set. This is pretty good result since the model learns a whole lot of class to classify and generalizes well on unseen data. The model took about 2070 second per epochs and a total of 5.75 hours for finishing the training for 10 epochs. The model can be found from this link (\url{https://drive.google.com/open?id=1w5BhaeSvc1LYQfUuvwBApeNbcx0vkOUP}). The size of the model is about 236 MiB. 

\begin{figure}
	\centering
	\includegraphics[height=20cm,width=0.5\linewidth]{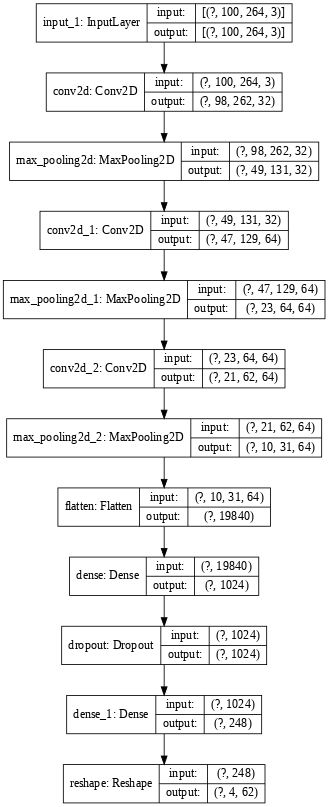}
	\caption{Keras model structure for Circle CAPTCHA.}
	\label{keras_model_mini_captcha}
\end{figure}

\begin{figure}
	\centering
	\includegraphics[width=0.5\linewidth]{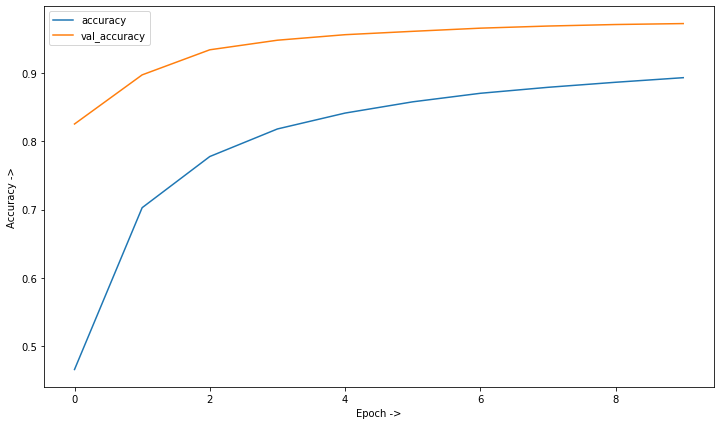}
	\caption{Accuracy of the model (\ref{keras_model_mini_captcha}).}
	\label{acc_mini_model}
\end{figure}

\begin{figure}
	\centering
	\includegraphics[width=0.5\linewidth]{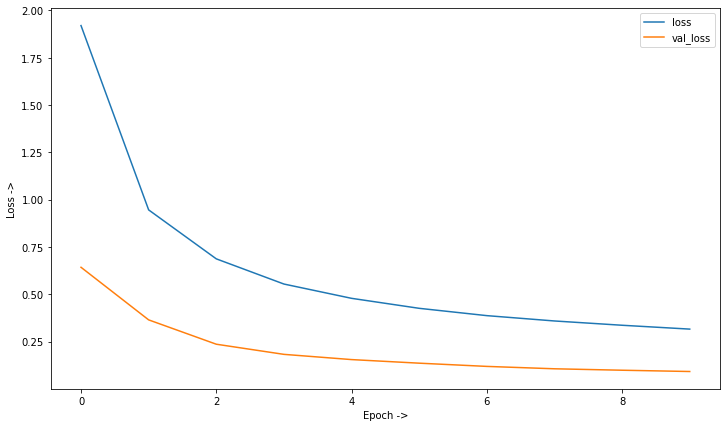}
	\caption{Loss of the model (\ref{keras_model_mini_captcha}).}
	\label{loss_mini_model}
\end{figure}

\begin{figure}
	\centering
	\includegraphics[width=1\linewidth]{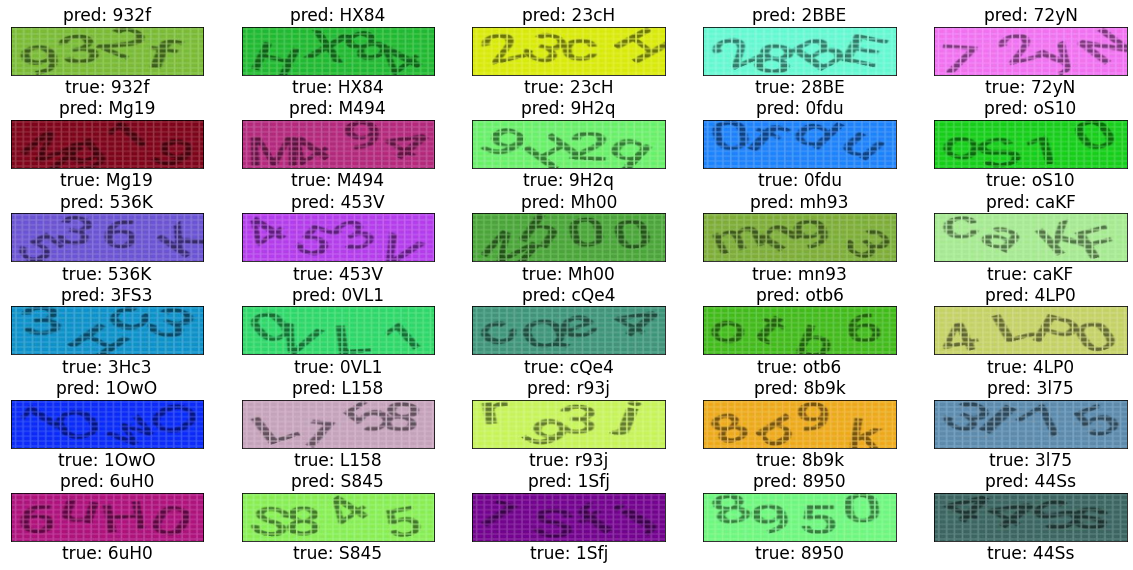}
	\caption{Predictions of the model (\ref{keras_model_mini_captcha}).}
	\label{mini_pred}
\end{figure}

The structure of the model is shown in Figure \ref{keras_model_mini_captcha}, the visualization of the accuracy of the model is shown in Figure \ref{acc_mini_model}, and the loss for the same model is shown in Figure \ref{loss_mini_model}. The prediction of the model is shown in Figure \ref{mini_pred}, on unseen data, and it accurately classifies all the labels of the CAPTCHA.

\begin{figure} 
	\centering
	\begin{lstlisting}
	Epoch 10/10
	1762/1762 [==============================] - ETA: 0s - loss: 0.3160 - accuracy: 0.8926INFO:tensorflow:Assets written to: ./model_checkpoint/assets
	1762/1762 [==============================] - 2070s 1s/step - loss: 0.3160 - accuracy: 0.8926 - val_loss: 0.0920 - val_accuracy: 0.9725
	\end{lstlisting}
	\caption{epochs for mini CAPTCHA.}
	\label{model_epochs_mini_captcha}
\end{figure}


\chapter{Using Multicolor CAPTCHA dataset  (95.69 \% accuracy)}

\section{Alex Net type model for Multicolor CAPTCHA dataset }

The initial version of the dataset contained a black background which was significantly reducing the information, by cluttering the convnet as shown in Figure \ref{multicolor}. We needed to crop the image before feeding into the convolutional neural network. We have trained the dataset using the model similar to Model \ref{model_faded}. The details of the dataset can be found on the appendix section and we have used only 50\% of the multicolor dataset for this purpose.

\begin{figure} 
	\centering
	\fbox{\includegraphics[height=10cm,width=0.5\linewidth]{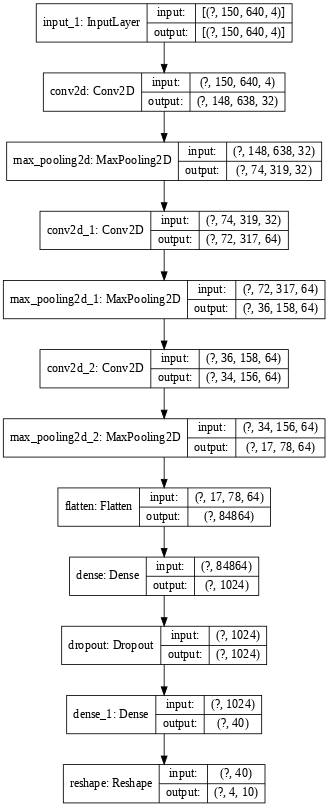}}
	\caption{Our model for predicting the labels of the multicolor CAPTCHA dataset.}
	\label{model_mc}
\end{figure}

The summary of the model is shown below \ref{multicolor_captcha}, and the summary of the epochs is shown \ref{mc_9_epochs}. The model is stopped early because of getting a low loss, and minimizing the loss as much as possible. It could great even better result but this is pretty doable for a hard CAPTCHA like this. We claim that the loss could even go down, resulting in increase of accuracy if we trained it for a little more time. The model for the same is shown in figure \ref{model_mc}.

\begin{figure}
	\centering
	{%
		\lstset{frame=single,basicstyle=\scriptsize,style=myModelSummaryStyle}
		\centering
		\begin{lstlisting}
		_________________________________________________________________
		Layer (type)                 Output Shape              Param #   
		==================================================
		input_1 (InputLayer)         [(None, 150, 640, 4)]     0         
		_________________________________________________________________
		conv2d (Conv2D)              (None, 148, 638, 32)      1184      
		_________________________________________________________________
		max_pooling2d (MaxPooling2D) (None, 74, 319, 32)       0         
		_________________________________________________________________
		conv2d_1 (Conv2D)            (None, 72, 317, 64)       18496     
		_________________________________________________________________
		max_pooling2d_1 (MaxPooling2 (None, 36, 158, 64)       0         
		_________________________________________________________________
		conv2d_2 (Conv2D)            (None, 34, 156, 64)       36928     
		_________________________________________________________________
		max_pooling2d_2 (MaxPooling2 (None, 17, 78, 64)        0         
		_________________________________________________________________
		flatten (Flatten)            (None, 84864)             0         
		_________________________________________________________________
		dense (Dense)                (None, 1024)              86901760  
		_________________________________________________________________
		dropout (Dropout)            (None, 1024)              0         
		_________________________________________________________________
		dense_1 (Dense)              (None, 40)                41000     
		_________________________________________________________________
		reshape (Reshape)            (None, 4, 10)             0         
		==================================================
		Total params: 86,999,368
		Trainable params: 86,999,368
		Non-trainable params: 0
		_________________________________________________________________
		\end{lstlisting}
	}
\caption{Summary of the model for \ref{model_mc}.}
\label{multicolor_captcha}
\end{figure}

\begin{figure}
	\centering
	{%
		\lstset{frame=single,basicstyle=\scriptsize,style=myModelSummaryStyle}
		\begin{lstlisting}
		Epoch 9/20
		3525/3525 [==============================] - ETA: 0s - loss: 0.2153 - accuracy: 0.9245INFO:tensorflow:Assets written to: ./model_checkpoint/assets
		3525/3525 [==============================] - 3464s 983ms/step - loss: 0.2153 - accuracy: 0.9245 - val_loss: 0.1315 - val_accuracy: 0.9569
		\end{lstlisting}
	}
	\caption{The summary of 9 epochs for the model \ref{model_mc}.}
	\label{mc_9_epochs}
\end{figure}

\begin{figure}
	\centering
	\includegraphics[width=0.75\linewidth]{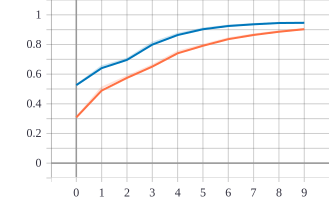}
	\caption{Plot for the training accuracy of our model.}
	\label{mc-train-acc}
\end{figure}

\begin{figure}
	\centering
	\includegraphics[width=0.75\linewidth]{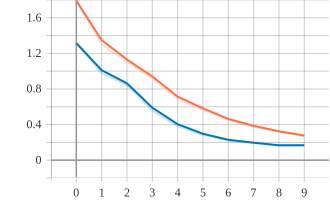}
	\caption{Plot for the training loss of our model.}
	\label{mc-train-loss}
\end{figure}

\begin{figure}
	\centering
	\includegraphics[width=0.75\linewidth]{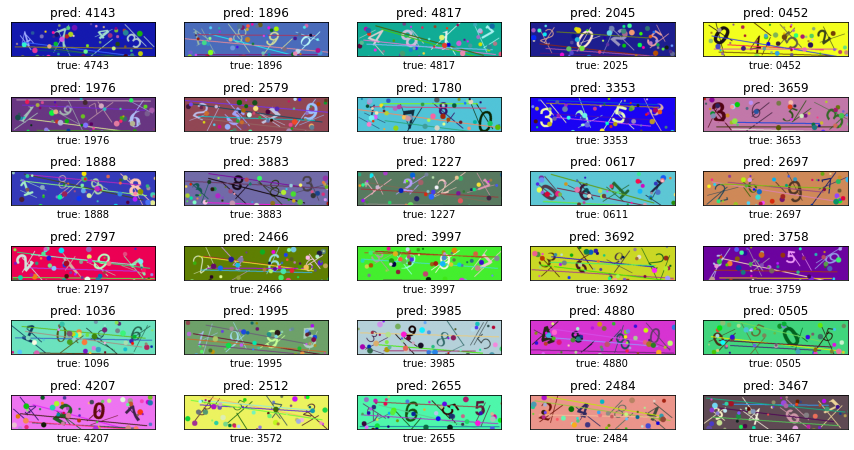}
	\caption{Prediction of the new model (\ref{model_mc}) on unseen data.}
	\label{pred_model_mc}
\end{figure}

The loss and accuracy plots for the mentioned model is shown in Figure \ref{mc-train-loss} and Figure \ref{mc-train-acc}. The model took about 3482 seconds for each of the epochs which is about 8.705 hours for the complete training for 9 epochs. The model generated (in .h5 format) can be found on this drive link \url{https://drive.google.com/open?id=1-2DW3CBkddRoUsBbaukf7OxR1MZpwVrn}. The size of the model is 996 MB, since is due to the fact that the model captures many details of the huge dataset. We see that the model trained generalizes well on the unseen data as shown in Figure \ref{pred_model_mc}. The accuracy of the model obtained is 95.69\% which is pretty great for an AlexNet type model. This is one of the hardest open source CAPTCHAs even for the humans because these contains a variety of colors, and those who suffer from night blindness, will find it difficult to predict the letters of this CAPTCHA accurately.

\chapter{Determining complexity of CAPTCHAs through Railway-CAPTCHA}

Our main objective in this section is to see how complexity of CAPTCHA increases with the length of the CAPTCHA for multilabel classification. Here we use a similar model to test all the CAPTCHAs with varying length, ranging from 3 to 7, with 100K sample of data (both test, validation, and training data included). We conjecture that the length of the CAPTCHA increases the complexity of CAPTCHA when it is hard to segment each letter and the CAPTCHA is noisy. A human can read a paragraph of simple letters with some noise with no hindrance, but it takes a lot of time for computers to do the same task. This is the reason why there is no perfect OCR for storing the documents of Guttenberg project to text digitally, which is still a problem for data retrieval. Whereas humans have seen too much of textual data in their entire life, so they can understand quickly what a paper-document is saying in different lighting conditions. This is the reason why there are still many sites which still rely on long length text CAPTCHAs to save their resources.

\section{Railway-CAPTCHA with 3 Letters}

\begin{figure}
	\centering
	{%
		\lstset{frame=single,basicstyle=\scriptsize,style=myModelSummaryStyle}
		\centering
		\begin{lstlisting}
		__________________________________________________________________________________________________
		Layer (type)                    Output Shape         Param #     Connected to                     
		==================================================

		input_1 (InputLayer)            (None, 200, 60, 1)   0                                            
		__________________________________________________________________________________________________
		conv2d_1 (Conv2D)               (None, 200, 60, 32)  320         input_1[0][0]                    
		__________________________________________________________________________________________________
		conv2d_2 (Conv2D)               (None, 198, 58, 32)  9248        conv2d_1[0][0]                   
		__________________________________________________________________________________________________
		batch_normalization_1 (BatchNor (None, 198, 58, 32)  128         conv2d_2[0][0]                   
		__________________________________________________________________________________________________
		max_pooling2d_1 (MaxPooling2D)  (None, 99, 29, 32)   0           batch_normalization_1[0][0]      
		__________________________________________________________________________________________________
		dropout_1 (Dropout)             (None, 99, 29, 32)   0           max_pooling2d_1[0][0]            
		__________________________________________________________________________________________________
		conv2d_3 (Conv2D)               (None, 99, 29, 64)   18496       dropout_1[0][0]                  
		__________________________________________________________________________________________________
		conv2d_4 (Conv2D)               (None, 97, 27, 64)   36928       conv2d_3[0][0]                   
		__________________________________________________________________________________________________
		batch_normalization_2 (BatchNor (None, 97, 27, 64)   256         conv2d_4[0][0]                   
		__________________________________________________________________________________________________
		max_pooling2d_2 (MaxPooling2D)  (None, 48, 13, 64)   0           batch_normalization_2[0][0]      
		__________________________________________________________________________________________________
		dropout_2 (Dropout)             (None, 48, 13, 64)   0           max_pooling2d_2[0][0]            
		__________________________________________________________________________________________________
		conv2d_5 (Conv2D)               (None, 48, 13, 128)  73856       dropout_2[0][0]                  
		__________________________________________________________________________________________________
		conv2d_6 (Conv2D)               (None, 46, 11, 128)  147584      conv2d_5[0][0]                   
		__________________________________________________________________________________________________
		batch_normalization_3 (BatchNor (None, 46, 11, 128)  512         conv2d_6[0][0]                   
		__________________________________________________________________________________________________
		max_pooling2d_3 (MaxPooling2D)  (None, 23, 5, 128)   0           batch_normalization_3[0][0]      
		__________________________________________________________________________________________________
		dropout_3 (Dropout)             (None, 23, 5, 128)   0           max_pooling2d_3[0][0]            
		__________________________________________________________________________________________________
		conv2d_7 (Conv2D)               (None, 21, 3, 256)   295168      dropout_3[0][0]                  
		__________________________________________________________________________________________________
		batch_normalization_4 (BatchNor (None, 21, 3, 256)   1024        conv2d_7[0][0]                   
		__________________________________________________________________________________________________
		max_pooling2d_4 (MaxPooling2D)  (None, 10, 1, 256)   0           batch_normalization_4[0][0]      
		__________________________________________________________________________________________________
		flatten_1 (Flatten)             (None, 2560)         0           max_pooling2d_4[0][0]            
		__________________________________________________________________________________________________
		dropout_4 (Dropout)             (None, 2560)         0           flatten_1[0][0]                  
		__________________________________________________________________________________________________
		digit1 (Dense)                  (None, 36)           87074       dropout_4[0][0]                  
		__________________________________________________________________________________________________
		digit2 (Dense)                  (None, 36)           87074       dropout_4[0][0]                  
		__________________________________________________________________________________________________
		digit3 (Dense)                  (None, 36)           87074       dropout_4[0][0]                  
		==================================================

		Total params: 844,742
		Trainable params: 843,782
		Non-trainable params: 960
		__________________________________________________________________________________________________
		\end{lstlisting}
	}
	\caption{Summary of the vanilla model (\ref{model_rc3L_1}) for railway 3 letter CAPTCHA dataset.}
	\label{model_rc3L_summary}
\end{figure}

\begin{figure}
	\centering
	{%
		\lstset{frame=single,basicstyle=\scriptsize,style=myModelSummaryStyle}
		\centering
		\begin{lstlisting}
			Epoch 30/30
			72000/72000 [==============================] - 81s 1ms/step - loss: 2.3520 - digit1_loss: 0.7551 - digit2_loss: 0.8290 - digit3_loss: 0.7678 - digit1_acc: 0.7729 - digit2_acc: 0.7508 - digit3_acc: 0.7699 - val_loss: 2.3121 - val_digit1_loss: 0.7538 - val_digit2_loss: 0.7967 - val_digit3_loss: 0.7616 - val_digit1_acc: 0.8068 - val_digit2_acc: 0.7862 - val_digit3_acc: 0.8023
		\end{lstlisting}
	}
	\caption{Summary of the epochs for the vanilla model (\ref{model_rc3L_1}) for railway 3 letter CAPTCHA dataset.}
\label{model_rc3L_epochs}
\end{figure}

\begin{figure} 
	\centering
	\fbox{\includegraphics[height=2cm,width=1\linewidth]{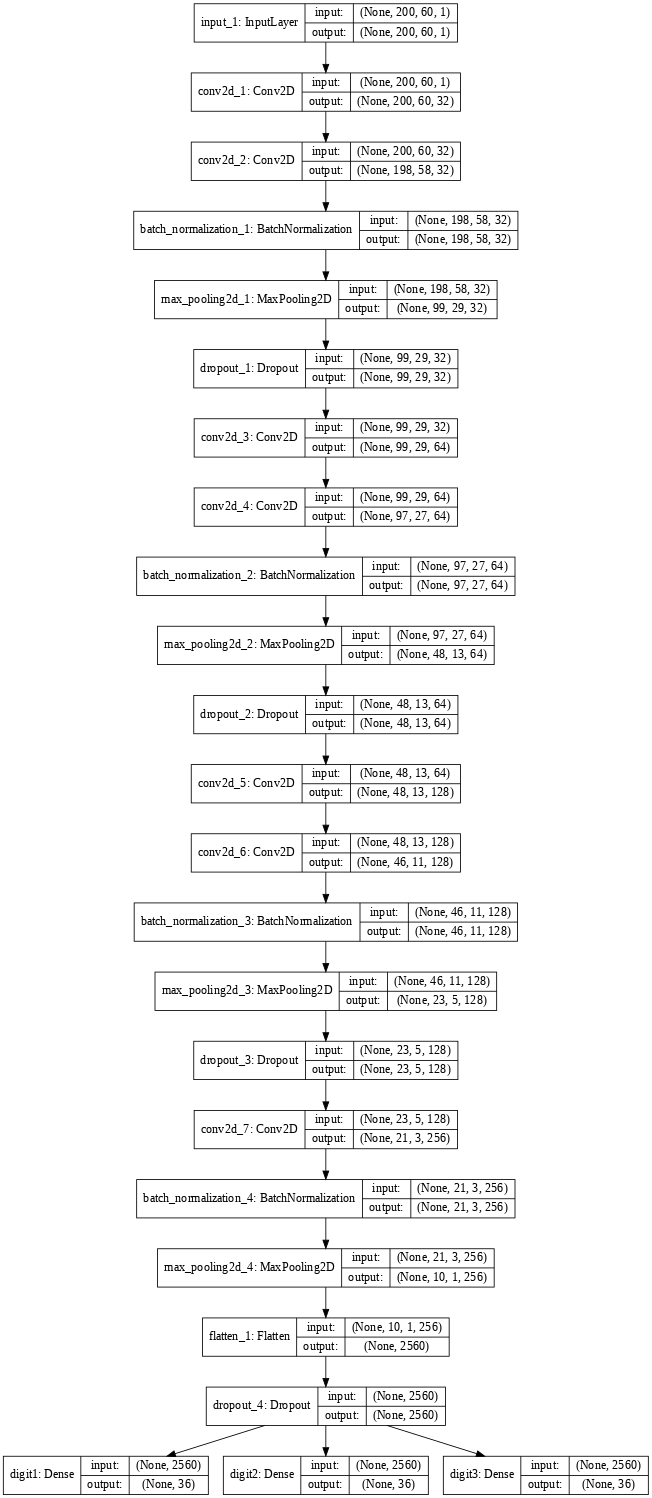}}	
	\caption{The vanilla model for railway 3 letter CAPTCHA dataset.}
	\label{model_rc3L_1}
\end{figure}

\begin{figure}
	\centering
	\includegraphics[width=1\linewidth]{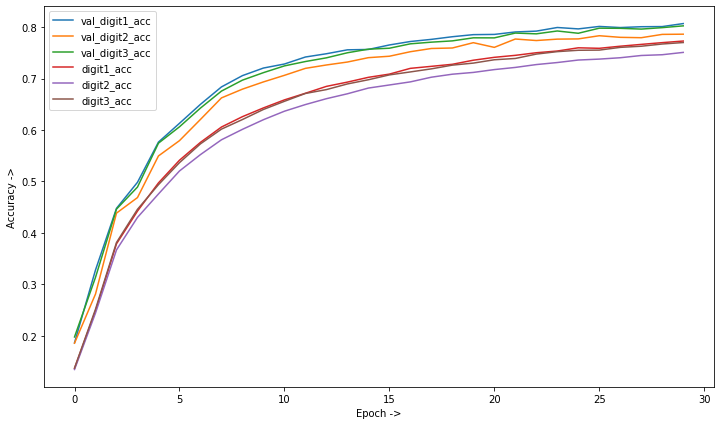}
	\caption{Training Accuracy for the vanilla model (\ref{model_rc3L_1}).}
	\label{rc3Lm1-train-acc}
\end{figure}

\begin{figure}
	\centering
	\includegraphics[width=1\linewidth]{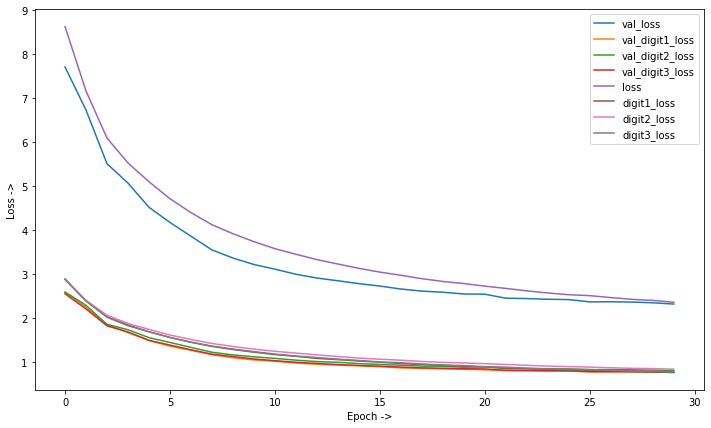}
	\caption{Training Loss for the vanilla model (\ref{model_rc3L_1}).}
	\label{rc3Lm1-train-loss}
\end{figure}

\begin{figure}
	\centering
	\fbox{\includegraphics[width=1\linewidth]{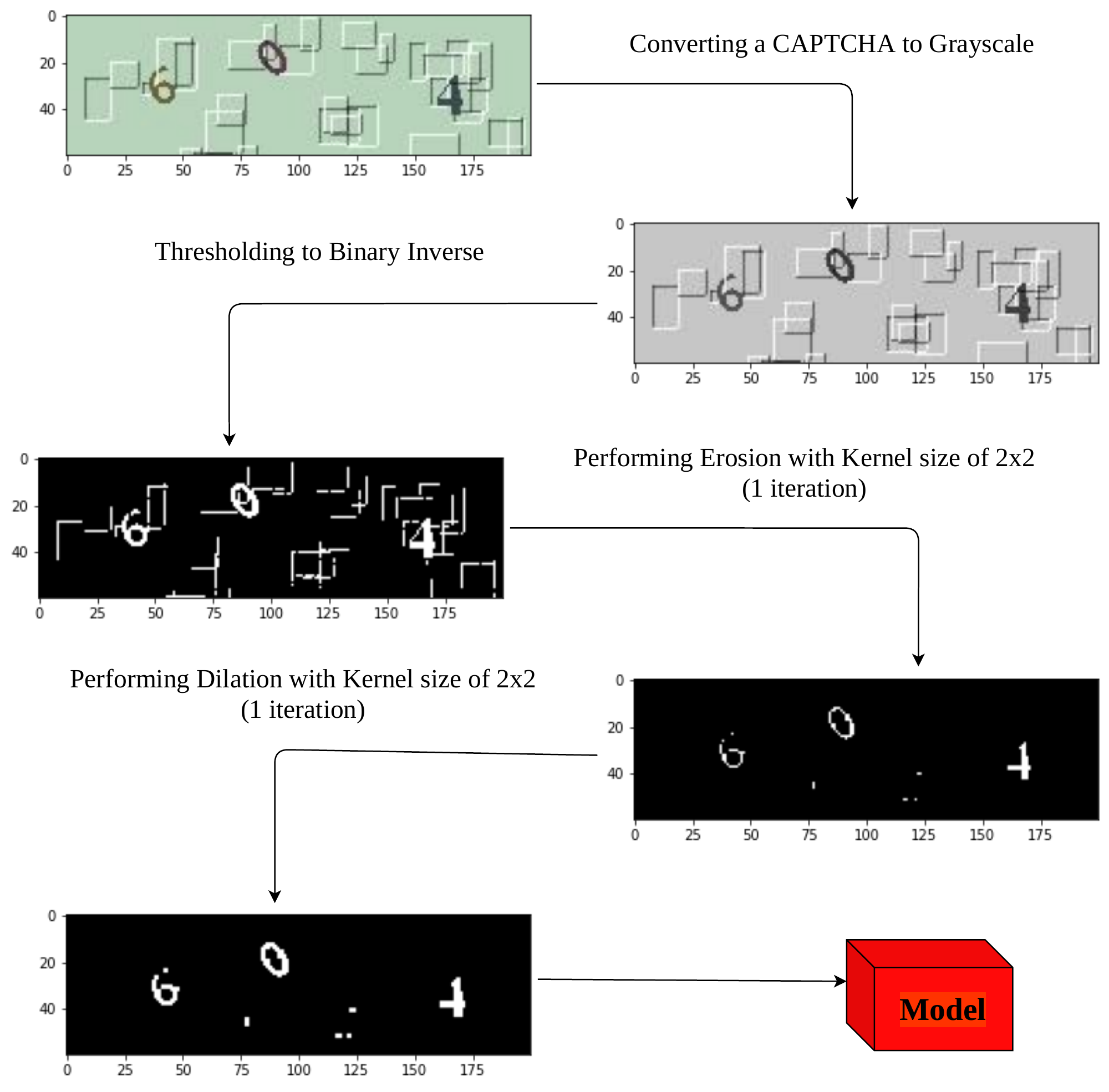}}
	\caption{The preprocessing done to improve accuracy before feeding it to the model (\ref{model_rc3L_1}).}
	\label{rc3_conv_diag}
\end{figure}

\begin{figure}
	\centering
	\includegraphics[width=0.3\linewidth]{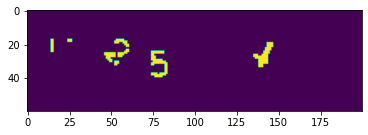}
	\caption{A CAPTCHA of 254 was given to the layer of the model (\ref{model_rc3L_1}) for visualizations.}
	\label{rc3Lm1_given_to_layer}
\end{figure}

\begin{figure}
	\centering
	\includegraphics[width=1\linewidth]{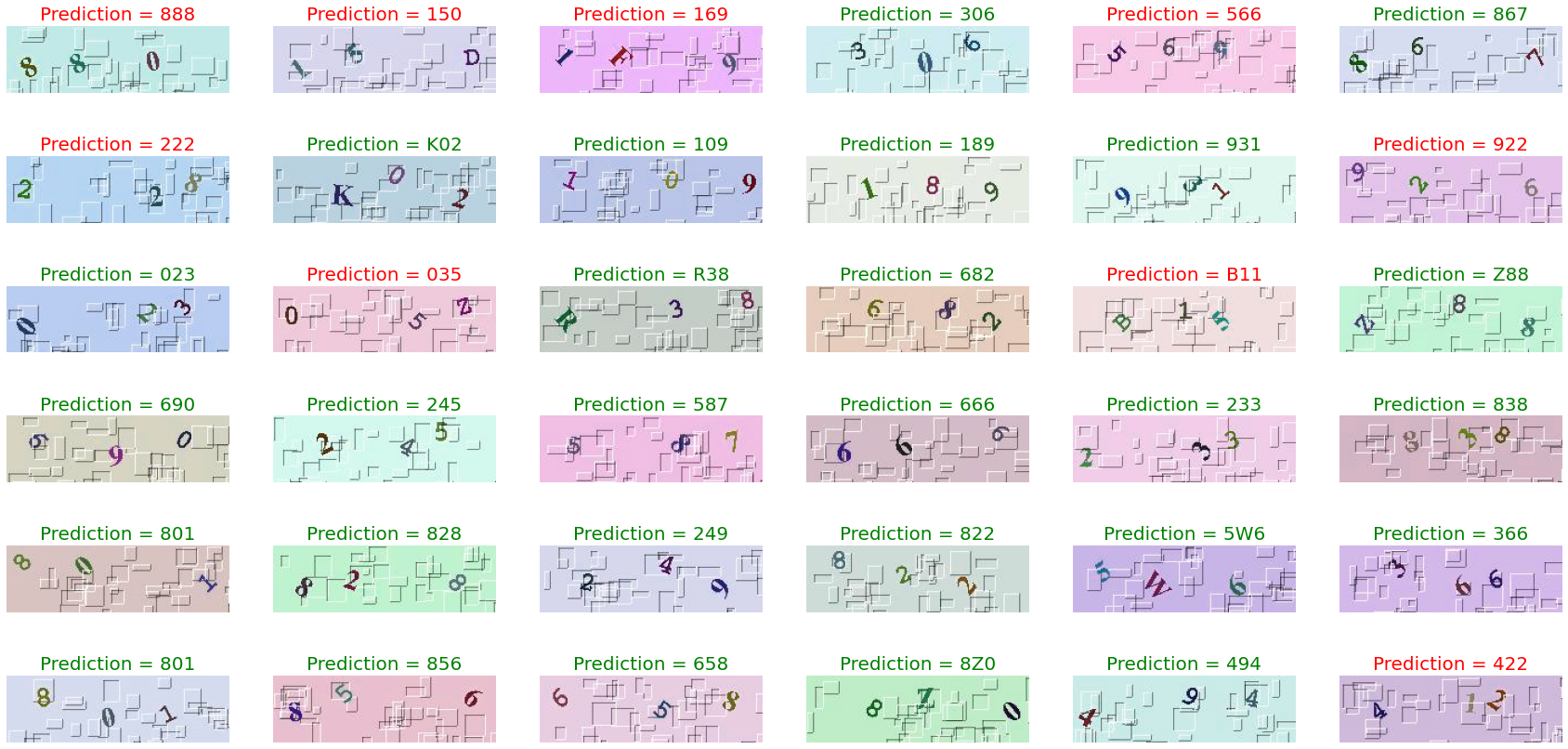}
	\caption{Predictions of the model (\ref{model_rc3L_1}).}
	\label{rc3_predict}
\end{figure}

We started our experiment with the open source synthetic CAPTCHA data generator by this CAPTCHA of 3 letters. We used our previous models and tried to improve on the existing models. The summary of the model is shown in Figure \ref{model_rc3L_summary}. This is generated by the Keras's $model.summary()$ method. The epochs of the model is shown in Figure \ref{model_rc3L_epochs}. It took about 81 second per epochs and there was a total of 30 epochs. The total time of training for the model was about 40.5 minutes, which is okay, since there is a total of 100K images of 3 letter CAPTCHA. This is again a multi-label classification which classified about 36 labels, i.e., all the 26 capital letters from A to Z and all the digits 0 to 9. The images are of 60x200 size and they are all preprocessed before passing to the naive version of the model. The preprocessing of the image is shown in Figure \ref{rc3_conv_diag}. The original images are of 3 channels, i.e., red, green and blue channels. It is then converted to grayscale and then applied binary inverse thresholding to make the image in binary format (i.e., intensity values of 255 and 0 only). Here all those pixels intensity which is less than 127 is set to 255 and all the pixels which are greater than 127 is set to 0. Then the resultant image formed, has some noise in them. This is because of the nature of the image to make the computer confused and make humans easy to decipher what code is written in them. To remove the noise, we perform erosion to make them thinner and then dilation to make them thicker. Then it is passed to the model for getting better results. The graph layout of the model when visualized with Keras is shown in Figure \ref{model_rc3L_1}. The training accuracy obtained for the model is shown in Figure \ref{rc3Lm1-train-acc} and the loss is shown in Figure \ref{rc3Lm1-train-loss}. The accuracy obtained from this model is about 80\% which is still low compared to the human level operator and other models that we built. The prediction of the model trained on some random unseen images is shown in Figure \ref{rc3_predict}. The trained model can be retrieved from this url (\url{https://jimut123.github.io/blogs/CAPTCHA/models/railway_captcha_3.h5}). We see that 9 out of 36 images are incorrect, which is still bad and we have a lot to improve on this model. The CAPTCHA image for this kind is a little bit trickier to train. There is various clutter and occlusion of the resultant binary image which makes the model confuse which label to classify and predict. We need to improve on the preprocessing step to make clear images which can help the model to learn better CAPTCHA and predict them correctly. The visualization of the layers of the model when passed through an image as shown in Figure \ref{rc3Lm1_given_to_layer} is shown in Figure \ref{model_rc3L_1}.

\begin{figure}
	\centering
	\includegraphics[height=4cm,width=0.985\linewidth]{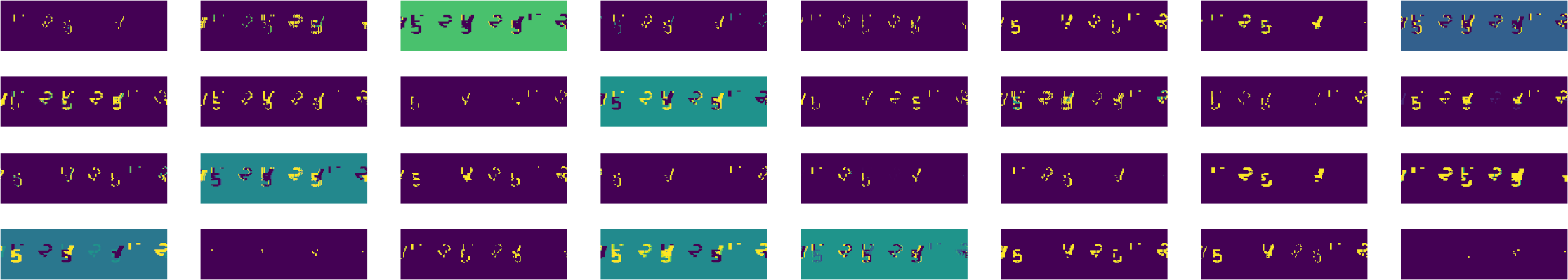}\vspace{4px}
	\includegraphics[height=4.6cm,width=0.5\linewidth]{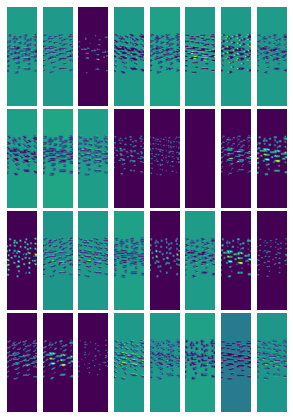}\includegraphics[height=4.6cm,width=0.5\linewidth]{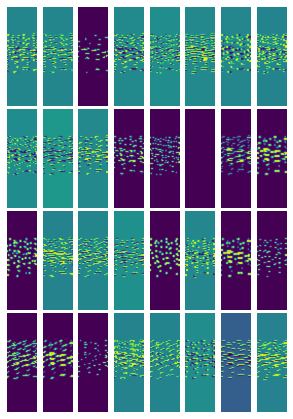}
	\includegraphics[height=3.7cm,width=0.5\linewidth]{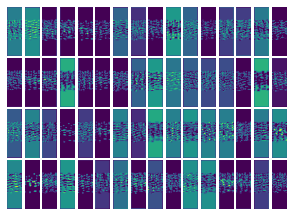}\includegraphics[height=3.7cm,width=0.5\linewidth]{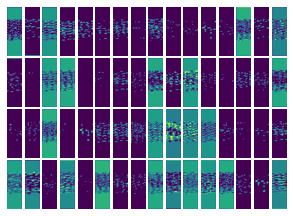}
	\includegraphics[height=3.7cm,width=0.5\linewidth]{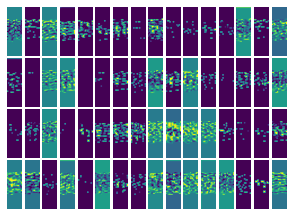}\includegraphics[height=3.7cm,width=0.5\linewidth]{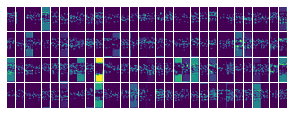}
	\includegraphics[height=3.9cm,width=0.5\linewidth]{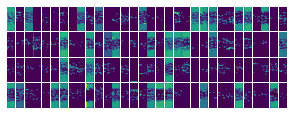}\includegraphics[height=3.9cm,width=0.5\linewidth]{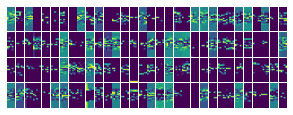}
	\includegraphics[height=3.5cm,width=0.5\linewidth]{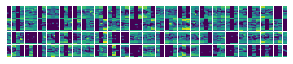}\includegraphics[height=3.5cm,width=0.5\linewidth]{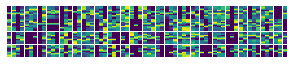}
	\caption{The visualizations of the features in a fully trained model \ref{model_rc3L_1} when passed a CAPTCHA containing the characters 254.}
	\label{rc3-layers-viz}
\end{figure}

\section{Railway-CAPTCHA with 4 Letters}

\begin{figure} 
	\centering
	\fbox{\includegraphics[height=24cm,width=1\linewidth]{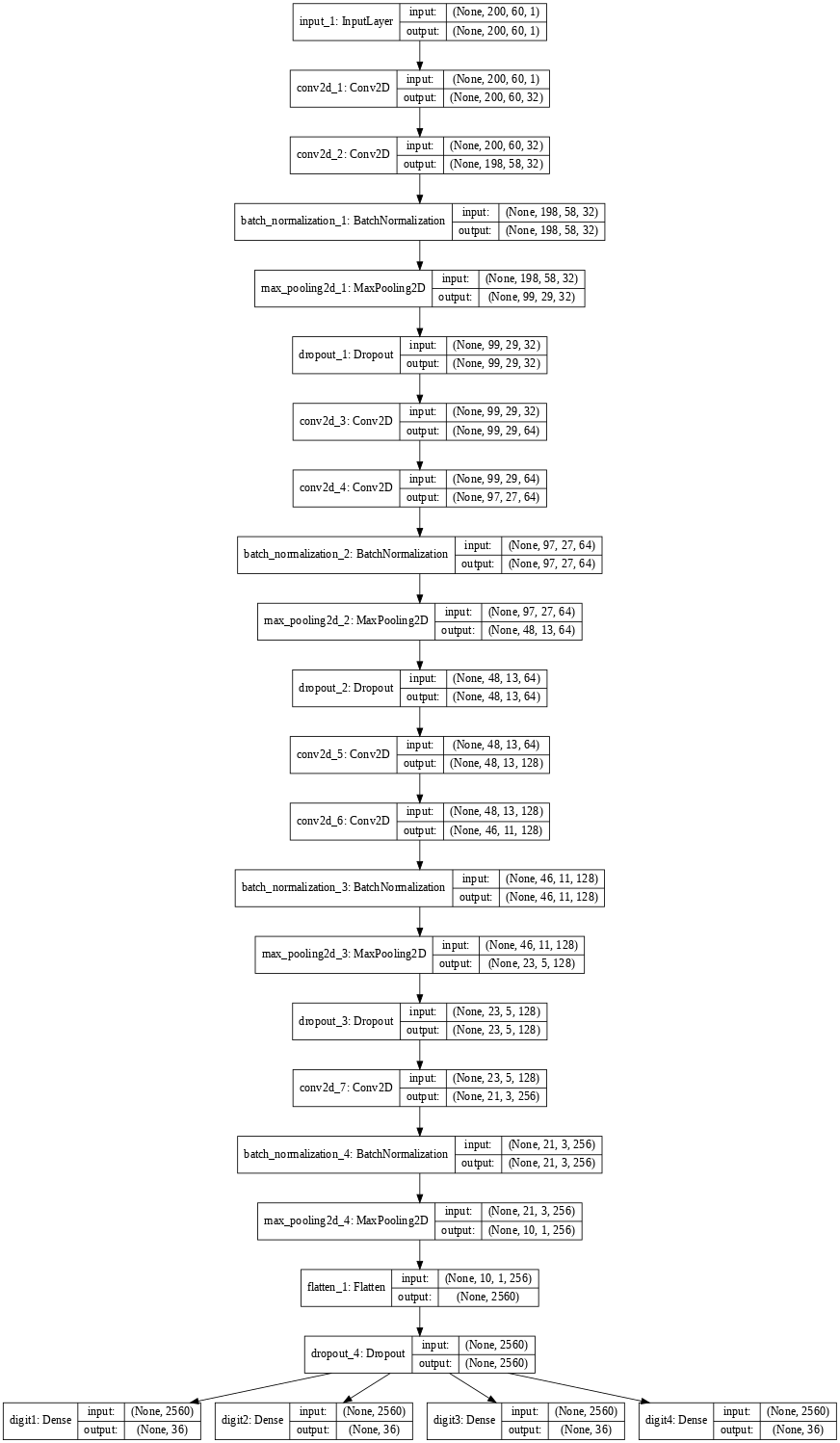}}	
	\caption{Our vanilla model for cracking 4-letter railway CAPTCHA.}
	\label{model_rc4L_model}
\end{figure}

\begin{figure}
	\centering
	\includegraphics[width=1\linewidth]{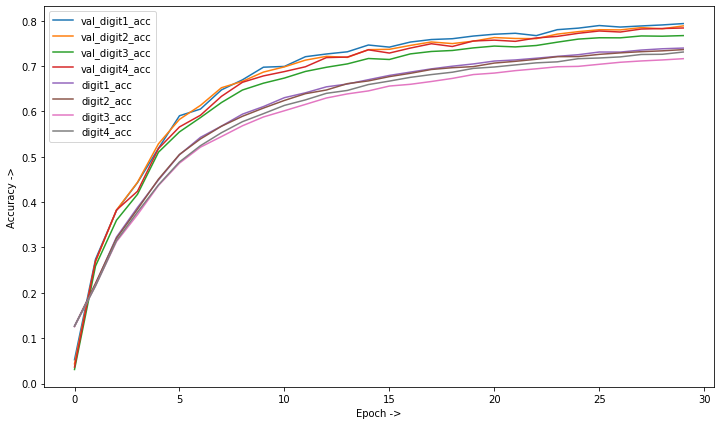}
	\caption{Training Accuracy for the model  (\ref{model_rc4L_model}).}
	\label{rc4Lm1-train-acc}
\end{figure}

\begin{figure}
	\centering
	\includegraphics[width=1\linewidth]{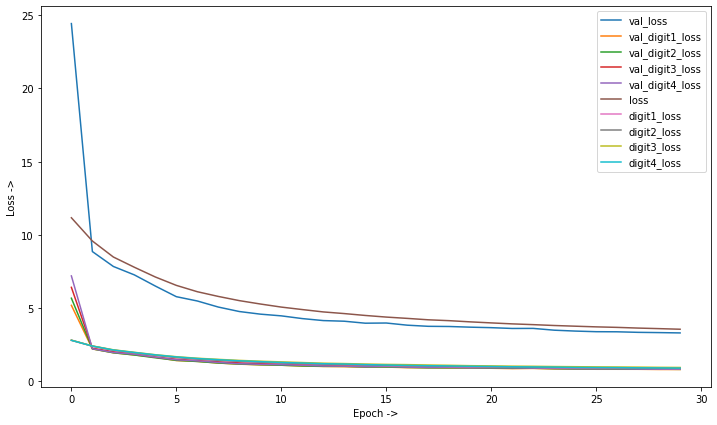}
	\caption{Training Loss for the model  (\ref{model_rc4L_model}).}
	\label{rc4Lm1-train-loss}
\end{figure}

\begin{figure}
	\centering
	\includegraphics[width=1\linewidth]{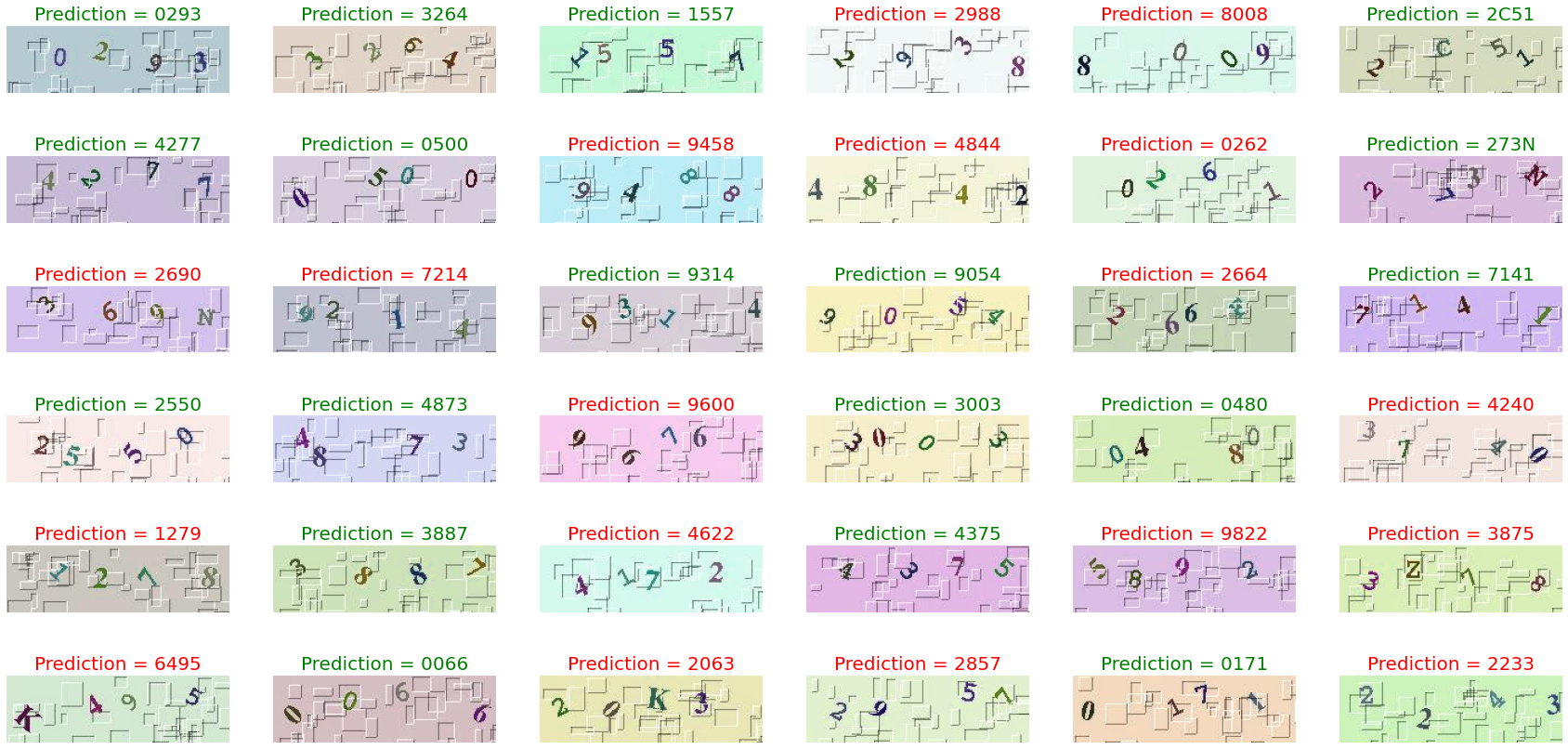}
	\caption{Predictions of the model (\ref{model_rc4L_model}).}
	\label{rc4_predict}
\end{figure}

We then checked how the same model performs with CAPTCHAs generated from the same dataset but this time with 4-character length labels. Here, the size and dimensions of the image are same as the previous one, the pre-processing done on the image is same too. We just changed some of the output nodes of the model and checked how this performed. The summary of the model is shown in Figure \ref{model_rc4L_1}. The summary of the epochs is shown in Figure \ref{epochs_rc4L_1}. It takes about 87 seconds to complete one epoch and a total of 43.5 minutes to finish the training. The graph layout of the model is shown in Figure \ref{model_rc4L_model}. The accuracy obtained is about 79\% and the plot is shown in Figure \ref{rc4Lm1-train-acc}. The trained model can be retrieved from this url (\url{https://jimut123.github.io/blogs/CAPTCHA/models/railway_captcha_4_imop.h5}). The plot for the loss for the same model is shown in Figure \ref{rc4Lm1-train-loss}. When the model is tested on real world unseen data, we get the prediction of \ref{rc4_predict}, which shows that the performance is not so well on real world data. The model predicts 18 images wrongly out of 36 images which is way more less than human level operator.

\begin{figure} 
	\centering
	{%
		\lstset{frame=single,basicstyle=\scriptsize,style=myModelSummaryStyle}
		\centering
		\begin{lstlisting}
			__________________________________________________________________________________________________
			Layer (type)                    Output Shape         Param #     Connected to                     
			==================================================
			input_1 (InputLayer)            (None, 200, 60, 1)   0                                            
			__________________________________________________________________________________________________
			conv2d_1 (Conv2D)               (None, 200, 60, 32)  320         input_1[0][0]                    
			__________________________________________________________________________________________________
			conv2d_2 (Conv2D)               (None, 198, 58, 32)  9248        conv2d_1[0][0]                   
			__________________________________________________________________________________________________
			batch_normalization_1 (BatchNor (None, 198, 58, 32)  128         conv2d_2[0][0]                   
			__________________________________________________________________________________________________
			max_pooling2d_1 (MaxPooling2D)  (None, 99, 29, 32)   0           batch_normalization_1[0][0]      
			__________________________________________________________________________________________________
			dropout_1 (Dropout)             (None, 99, 29, 32)   0           max_pooling2d_1[0][0]            
			__________________________________________________________________________________________________
			conv2d_3 (Conv2D)               (None, 99, 29, 64)   18496       dropout_1[0][0]                  
			__________________________________________________________________________________________________
			conv2d_4 (Conv2D)               (None, 97, 27, 64)   36928       conv2d_3[0][0]                   
			__________________________________________________________________________________________________
			batch_normalization_2 (BatchNor (None, 97, 27, 64)   256         conv2d_4[0][0]                   
			__________________________________________________________________________________________________
			max_pooling2d_2 (MaxPooling2D)  (None, 48, 13, 64)   0           batch_normalization_2[0][0]      
			__________________________________________________________________________________________________
			dropout_2 (Dropout)             (None, 48, 13, 64)   0           max_pooling2d_2[0][0]            
			__________________________________________________________________________________________________
			conv2d_5 (Conv2D)               (None, 48, 13, 128)  73856       dropout_2[0][0]                  
			__________________________________________________________________________________________________
			conv2d_6 (Conv2D)               (None, 46, 11, 128)  147584      conv2d_5[0][0]                   
			__________________________________________________________________________________________________
			batch_normalization_3 (BatchNor (None, 46, 11, 128)  512         conv2d_6[0][0]                   
			__________________________________________________________________________________________________
			max_pooling2d_3 (MaxPooling2D)  (None, 23, 5, 128)   0           batch_normalization_3[0][0]      
			__________________________________________________________________________________________________
			dropout_3 (Dropout)             (None, 23, 5, 128)   0           max_pooling2d_3[0][0]            
			__________________________________________________________________________________________________
			conv2d_7 (Conv2D)               (None, 21, 3, 256)   295168      dropout_3[0][0]                  
			__________________________________________________________________________________________________
			batch_normalization_4 (BatchNor (None, 21, 3, 256)   1024        conv2d_7[0][0]                   
			__________________________________________________________________________________________________
			max_pooling2d_4 (MaxPooling2D)  (None, 10, 1, 256)   0           batch_normalization_4[0][0]      
			__________________________________________________________________________________________________
			flatten_1 (Flatten)             (None, 2560)         0           max_pooling2d_4[0][0]            
			__________________________________________________________________________________________________
			dropout_4 (Dropout)             (None, 2560)         0           flatten_1[0][0]                  
			__________________________________________________________________________________________________
			digit1 (Dense)                  (None, 36)           87074       dropout_4[0][0]                  
			__________________________________________________________________________________________________
			digit2 (Dense)                  (None, 36)           87074       dropout_4[0][0]                  
			__________________________________________________________________________________________________
			digit3 (Dense)                  (None, 36)           87074       dropout_4[0][0]                  
			__________________________________________________________________________________________________
			digit4 (Dense)                  (None, 36)           87074       dropout_4[0][0]                  
			==================================================
			Total params: 931,816
			Trainable params: 930,856
			Non-trainable params: 960
			__________________________________________________________________________________________________
		\end{lstlisting}
	}
	\caption{Model \ref{model_rc4L_model} Summary for cracking 4-letter railway CAPTCHA. }
	\label{model_rc4L_1}
\end{figure}

\begin{figure} 
	\centering
	{%
		\lstset{frame=single,basicstyle=\scriptsize,style=myModelSummaryStyle}
		\centering
		\begin{lstlisting}
			Epoch 30/30
			72000/72000 [==============================] - 87s 1ms/step - loss: 3.5429 - digit1_loss: 0.8553 - digit2_loss: 0.8677 - digit3_loss: 0.9308 - digit4_loss: 0.8891 - digit1_acc: 0.7397 - digit2_acc: 0.7362 - digit3_acc: 0.7162 - digit4_acc: 0.7309 - val_loss: 3.2911 - val_digit1_loss: 0.7909 - val_digit2_loss: 0.8026 - val_digit3_loss: 0.8854 - val_digit4_loss: 0.8121 - val_digit1_acc: 0.7937 - val_digit2_acc: 0.7886 - val_digit3_acc: 0.7672 - val_digit4_acc: 0.7839
		\end{lstlisting}
	}
	\caption{Model \ref{model_rc4L_model} Summary.}
	\label{epochs_rc4L_1}
\end{figure}

\section{Railway-CAPTCHA with 5 Letters}

\begin{figure} 
	\centering
	\fbox{\includegraphics[height=24cm,width=1\linewidth]{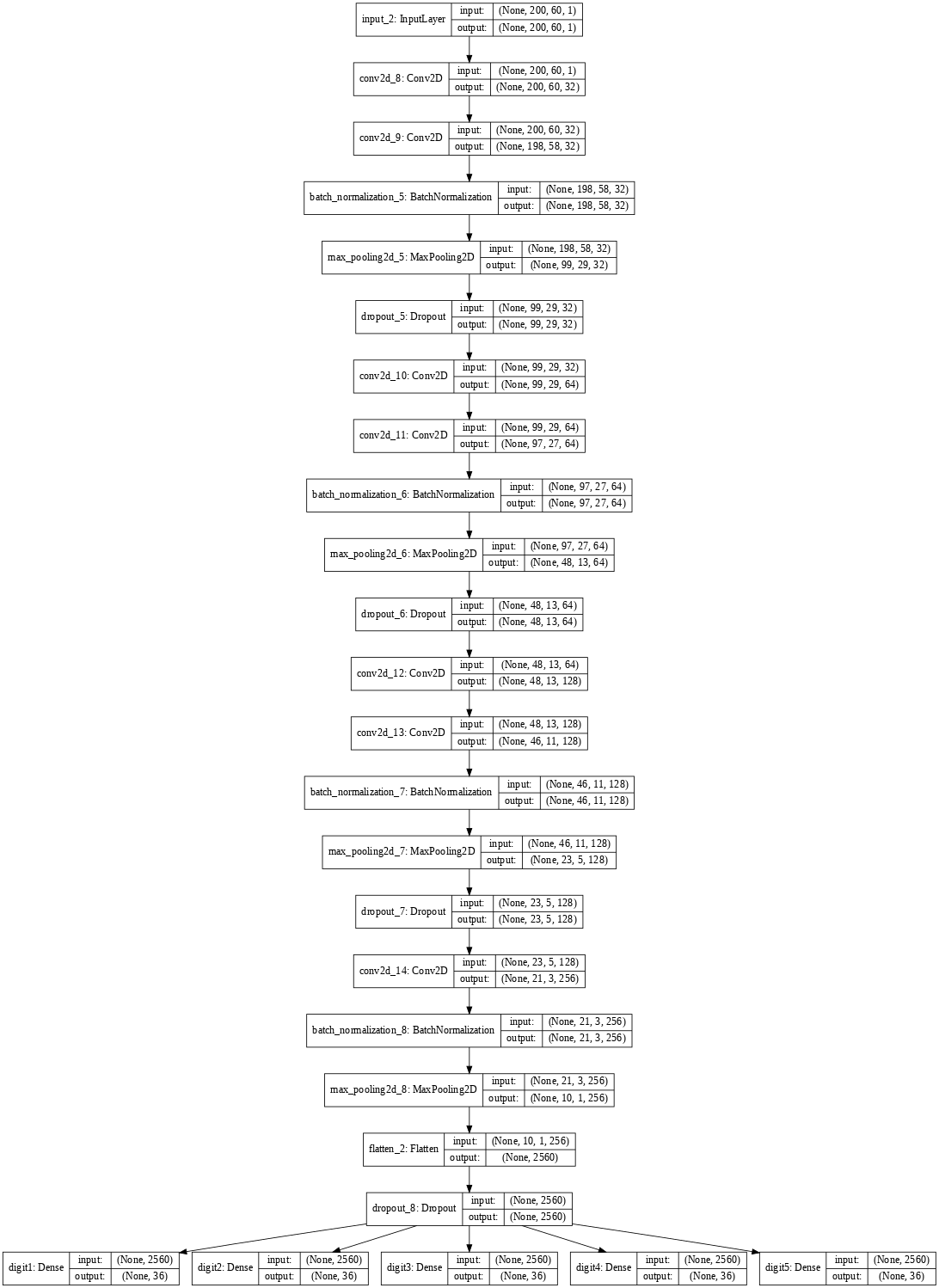}}	
	\caption{Our vanilla model for cracking 5-letter railway CAPTCHA.}
	\label{model_rc5L_1}
\end{figure}

\begin{figure}
	\centering
	\includegraphics[width=1\linewidth]{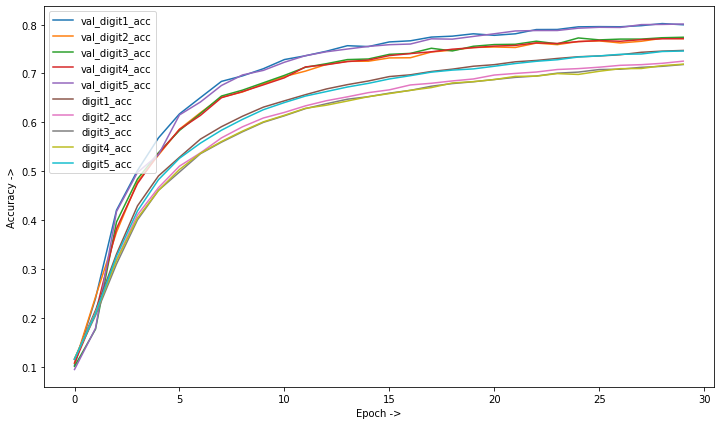}
	\caption{Training Accuracy for our vanilla model (\ref{model_rc5L_1}). }
	\label{rc5Lm1-train-acc}
\end{figure}

\begin{figure}
	\centering
	\includegraphics[width=1\linewidth]{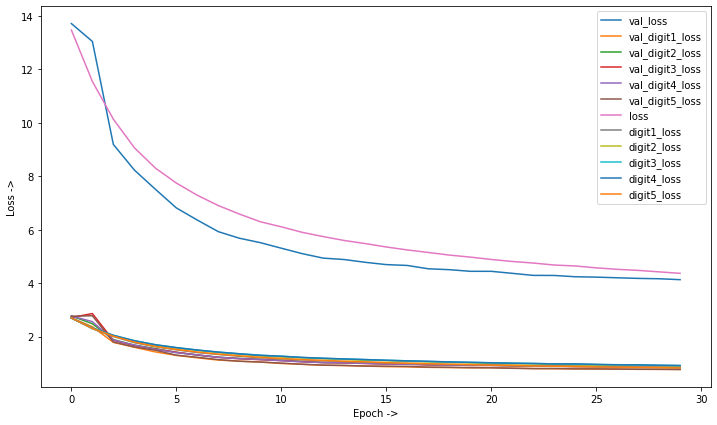}
	\caption{Training Loss for the vanilla model (\ref{model_rc5L_1}).}
	\label{rc5Lm1-train-loss}
\end{figure}

\begin{figure}
	\centering
	\includegraphics[width=1\linewidth]{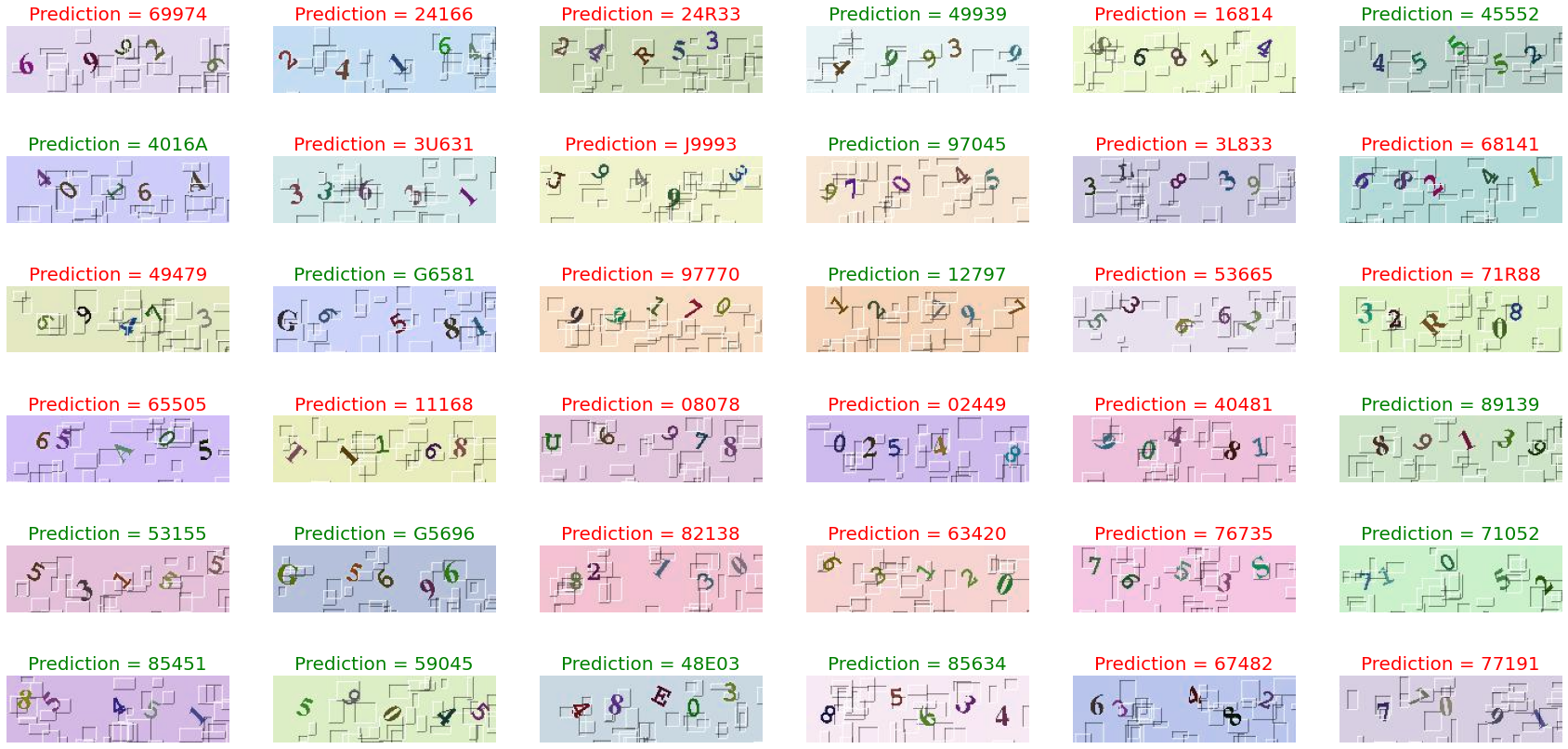}
	\caption{Predictions of the vanilla model (\ref{model_rc5L_1}).}
	\label{rc5_predict}
\end{figure}

We then checked how the same model performs with CAPTCHAs generated from the same dataset but this time with 5-character length labels. Here, the size and dimensions of the image are same as the previous one, the pre-processing done on the image is same too. We just changed some of the output nodes of the model and checked how this performed. The summary of the model is shown in Figure \ref{model_rc5L_summary}. The summary of the epochs is shown in Figure \ref{model_rc5L_epochs}. It takes about 83 seconds to complete one epoch and a total of 41.5 minutes to finish the training. The graph layout of the model is shown in Figure \ref{model_rc5L_1}. The accuracy obtained is about 77\% and the plot is shown in Figure \ref{rc5Lm1-train-acc}. The trained model can be retrieved from this url (\url{https://jimut123.github.io/blogs/CAPTCHA/models/railway_captcha_5_imop.h5}). The plot for the loss for the same model is shown in Figure \ref{rc5Lm1-train-loss}. When the model is tested on real world unseen data, we get the prediction of \ref{rc5_predict}, which shows that the performance is not so well on real world data. The model predicts 23 images wrongly out of 36 images which is way more less than human level operator.

\begin{figure}
	\centering

	{%
		\lstset{frame=single,basicstyle=\scriptsize,style=myModelSummaryStyle}
		\centering
		\begin{lstlisting}
			__________________________________________________________________________________________________
			Layer (type)                    Output Shape         Param #     Connected to                     
			==================================================
			input_1 (InputLayer)            (None, 200, 60, 1)   0                                            
			__________________________________________________________________________________________________
			conv2d_1 (Conv2D)               (None, 200, 60, 32)  320         input_1[0][0]                    
			__________________________________________________________________________________________________
			conv2d_2 (Conv2D)               (None, 198, 58, 32)  9248        conv2d_1[0][0]                   
			__________________________________________________________________________________________________
			batch_normalization_1 (BatchNor (None, 198, 58, 32)  128         conv2d_2[0][0]                   
			__________________________________________________________________________________________________
			max_pooling2d_1 (MaxPooling2D)  (None, 99, 29, 32)   0           batch_normalization_1[0][0]      
			__________________________________________________________________________________________________
			dropout_1 (Dropout)             (None, 99, 29, 32)   0           max_pooling2d_1[0][0]            
			__________________________________________________________________________________________________
			conv2d_3 (Conv2D)               (None, 99, 29, 64)   18496       dropout_1[0][0]                  
			__________________________________________________________________________________________________
			conv2d_4 (Conv2D)               (None, 97, 27, 64)   36928       conv2d_3[0][0]                   
			__________________________________________________________________________________________________
			batch_normalization_2 (BatchNor (None, 97, 27, 64)   256         conv2d_4[0][0]                   
			__________________________________________________________________________________________________
			max_pooling2d_2 (MaxPooling2D)  (None, 48, 13, 64)   0           batch_normalization_2[0][0]      
			__________________________________________________________________________________________________
			dropout_2 (Dropout)             (None, 48, 13, 64)   0           max_pooling2d_2[0][0]            
			__________________________________________________________________________________________________
			conv2d_5 (Conv2D)               (None, 48, 13, 128)  73856       dropout_2[0][0]                  
			__________________________________________________________________________________________________
			conv2d_6 (Conv2D)               (None, 46, 11, 128)  147584      conv2d_5[0][0]                   
			__________________________________________________________________________________________________
			batch_normalization_3 (BatchNor (None, 46, 11, 128)  512         conv2d_6[0][0]                   
			__________________________________________________________________________________________________
			max_pooling2d_3 (MaxPooling2D)  (None, 23, 5, 128)   0           batch_normalization_3[0][0]      
			__________________________________________________________________________________________________
			dropout_3 (Dropout)             (None, 23, 5, 128)   0           max_pooling2d_3[0][0]            
			__________________________________________________________________________________________________
			conv2d_7 (Conv2D)               (None, 21, 3, 256)   295168      dropout_3[0][0]                  
			__________________________________________________________________________________________________
			batch_normalization_4 (BatchNor (None, 21, 3, 256)   1024        conv2d_7[0][0]                   
			__________________________________________________________________________________________________
			max_pooling2d_4 (MaxPooling2D)  (None, 10, 1, 256)   0           batch_normalization_4[0][0]      
			__________________________________________________________________________________________________
			flatten_1 (Flatten)             (None, 2560)         0           max_pooling2d_4[0][0]            
			__________________________________________________________________________________________________
			dropout_4 (Dropout)             (None, 2560)         0           flatten_1[0][0]                  
			__________________________________________________________________________________________________
			digit1 (Dense)                  (None, 36)           87074       dropout_4[0][0]                  
			__________________________________________________________________________________________________
			digit2 (Dense)                  (None, 36)           87074       dropout_4[0][0]                  
			__________________________________________________________________________________________________
			digit3 (Dense)                  (None, 36)           87074       dropout_4[0][0]                  
			__________________________________________________________________________________________________
			digit4 (Dense)                  (None, 36)           87074       dropout_4[0][0]                  
			__________________________________________________________________________________________________
			digit5 (Dense)                  (None, 36)           87074       dropout_4[0][0]                  
			==================================================
			Total params: 1,018,890
			Trainable params: 1,017,930
			Non-trainable params: 960
			__________________________________________________________________________________________________
		\end{lstlisting}
	}
	\caption{Model (\ref{model_rc5L_1}) Summary.}
	\label{model_rc5L_summary}
\end{figure}

\begin{figure} 
	\centering
	{%
		\lstset{frame=single,basicstyle=\scriptsize,style=myModelSummaryStyle}
		\centering
		\begin{lstlisting}
			Epoch 30/30
			72000/72000 [==============================] - 83s 1ms/step - loss: 4.3662 - digit1_loss: 0.8213 - digit2_loss: 0.8910 - digit3_loss: 0.9141 - digit4_loss: 0.9161 - digit5_loss: 0.8237 - digit1_acc: 0.7472 - digit2_acc: 0.7252 - digit3_acc: 0.7183 - digit4_acc: 0.7191 - digit5_acc: 0.7460 - val_loss: 4.1281 - val_digit1_loss: 0.7711 - val_digit2_loss: 0.8651 - val_digit3_loss: 0.8587 - val_digit4_loss: 0.8666 - val_digit5_loss: 0.7666 - val_digit1_acc: 0.7996 - val_digit2_acc: 0.7713 - val_digit3_acc: 0.7741 - val_digit4_acc: 0.7713 - val_digit5_acc: 0.8012
		\end{lstlisting}
	}
	\caption{Summary for model's (\ref{model_rc5L_1}) epochs}
\label{model_rc5L_epochs}
\end{figure}

\begin{figure} 
	\centering
	{%
		\lstset{frame=single,basicstyle=\scriptsize,style=myModelSummaryStyle}
		\centering
		\begin{lstlisting}
		Epoch 30/30
		72000/72000 [==============================] - 95s 1ms/step - loss: 5.9938 - digit1_loss: 0.8645 - digit2_loss: 1.0304 - digit3_loss: 1.1001 - digit4_loss: 1.0711 - digit5_loss: 1.0102 - digit6_loss: 0.9174 - digit1_acc: 0.7351 - digit2_acc: 0.6786 - digit3_acc: 0.6537 - digit4_acc: 0.6624 - digit5_acc: 0.6869 - digit6_acc: 0.7162 - val_loss: 5.3350 - val_digit1_loss: 0.7786 - val_digit2_loss: 0.9233 - val_digit3_loss: 0.9596 - val_digit4_loss: 0.9521 - val_digit5_loss: 0.9080 - val_digit6_loss: 0.8133 - val_digit1_acc: 0.7951 - val_digit2_acc: 0.7558 - val_digit3_acc: 0.7511 - val_digit4_acc: 0.7475 - val_digit5_acc: 0.7544 - val_digit6_acc: 0.7829
		\end{lstlisting}
	}
	\caption{Summary for model's (\ref{model_rc6L_1}) epochs}
	\label{model_rc6L_epochs}
\end{figure}

\begin{figure}
	\centering
	\includegraphics[width=1\linewidth]{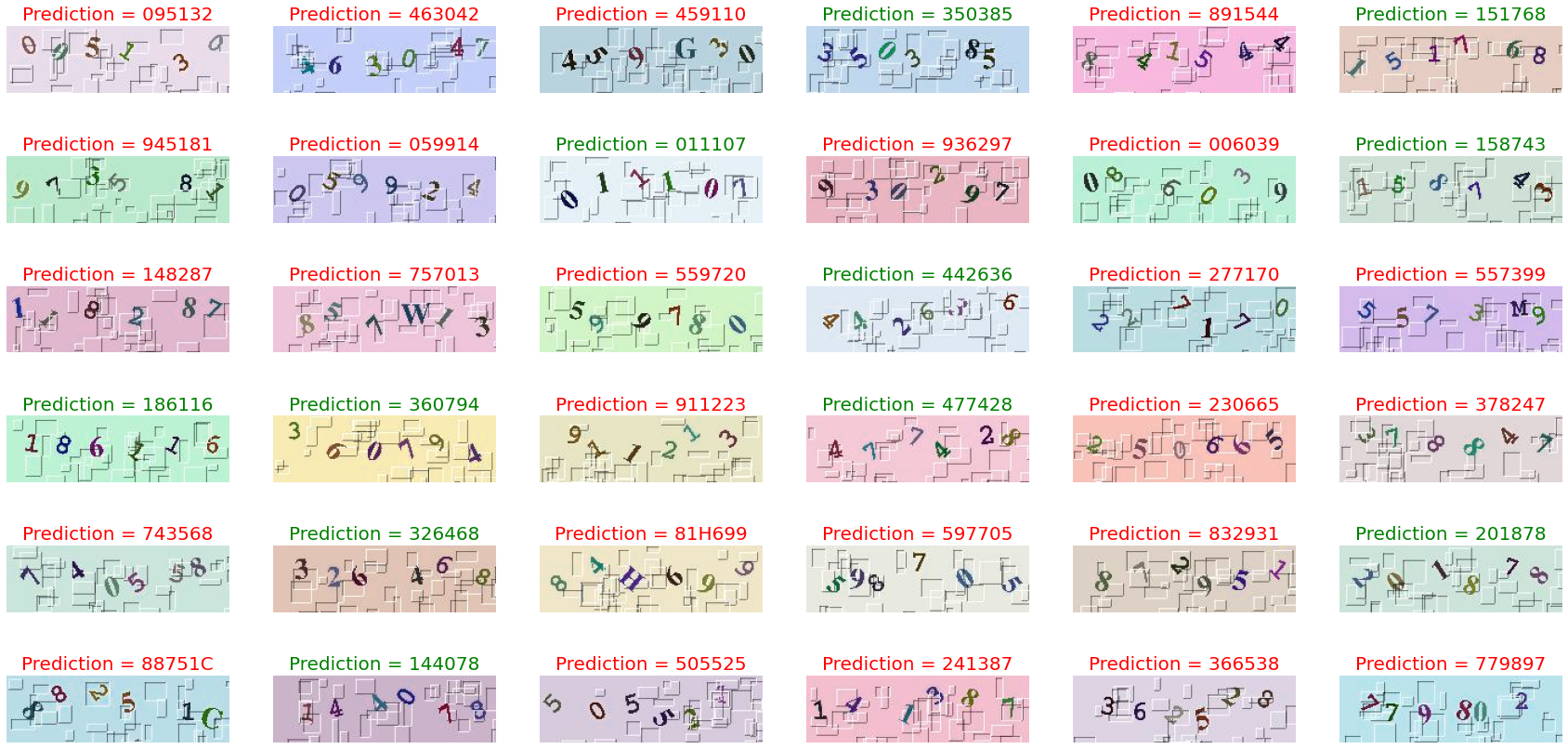}
	\caption{Predictions of the vanilla model (\ref{model_rc6L_1}).}
	\label{rc6_predict}
\end{figure}

\section{Railway-CAPTCHA with 6 Letters}

\begin{figure} 
	\centering
	\fbox{\includegraphics[height=24cm,width=1\linewidth]{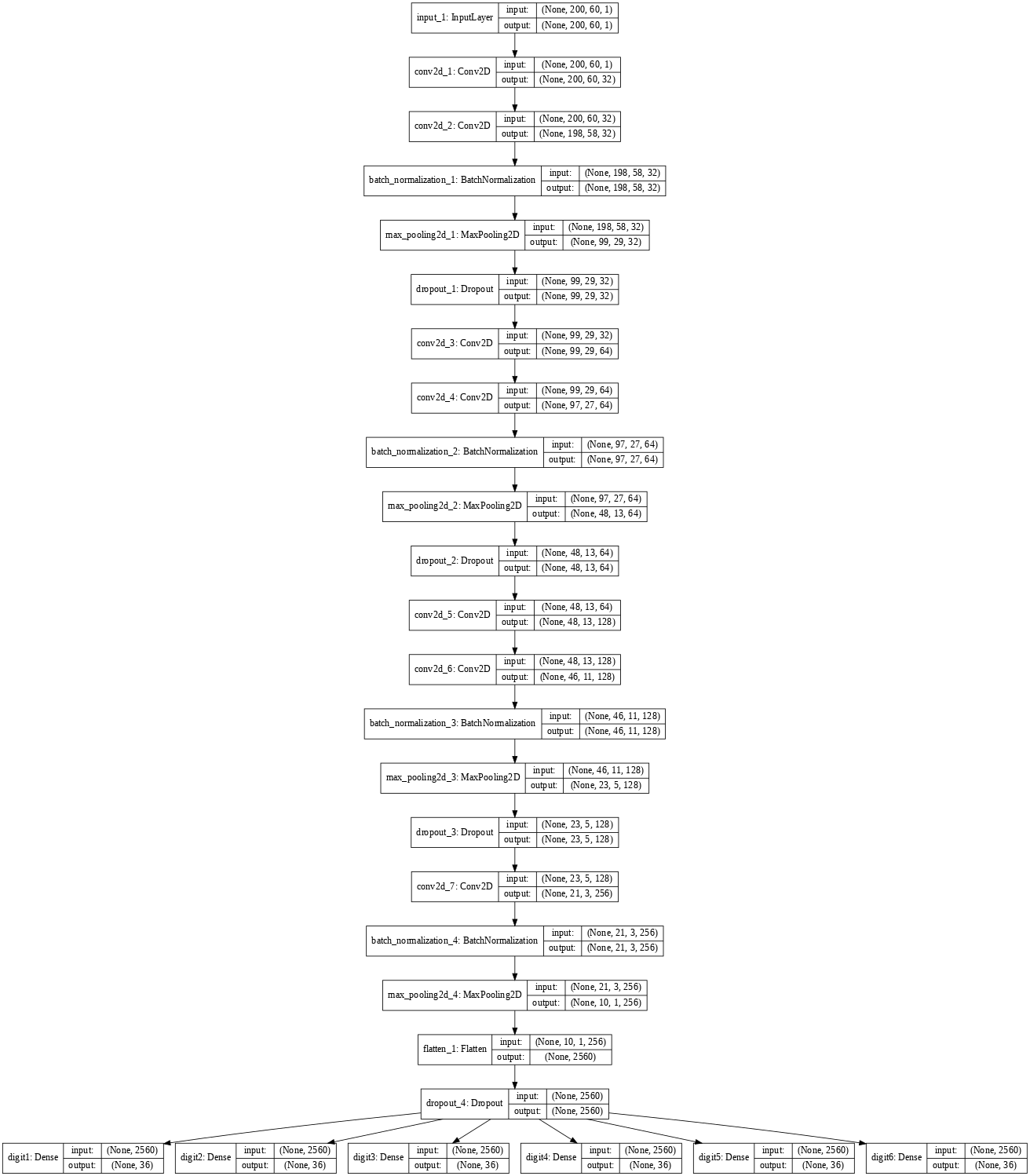}}	
	\caption{Our vanilla model for cracking 6-letter railway CAPTCHA.}
	\label{model_rc6L_1}
\end{figure}

\begin{figure}
	\centering
	\includegraphics[width=1\linewidth]{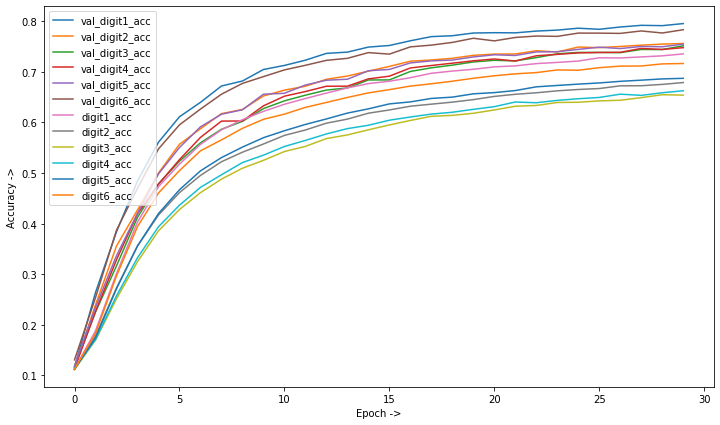}
	\caption{Training Accuracy for vanilla model (\ref{model_rc6L_1}).}
	\label{rc6Lm1-train-acc}
\end{figure}

\begin{figure}
	\centering
	\includegraphics[width=1\linewidth]{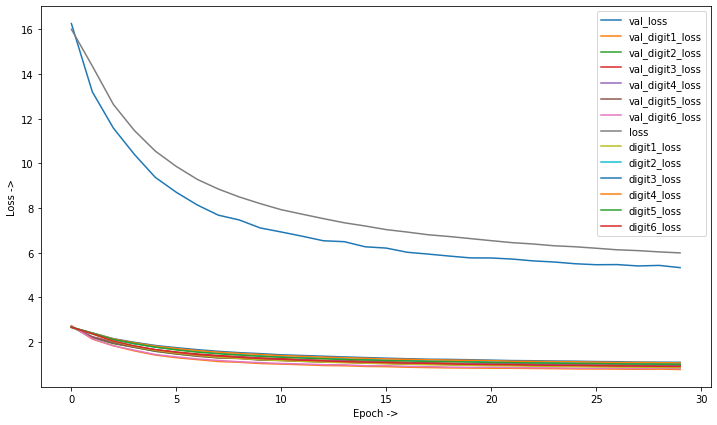}
	\caption{Training Loss for vanilla model (\ref{model_rc6L_1}).}
	\label{rc6Lm1-train-loss}
\end{figure}

\begin{figure}
	\centering
	\includegraphics[height=22cm,width=1\linewidth]{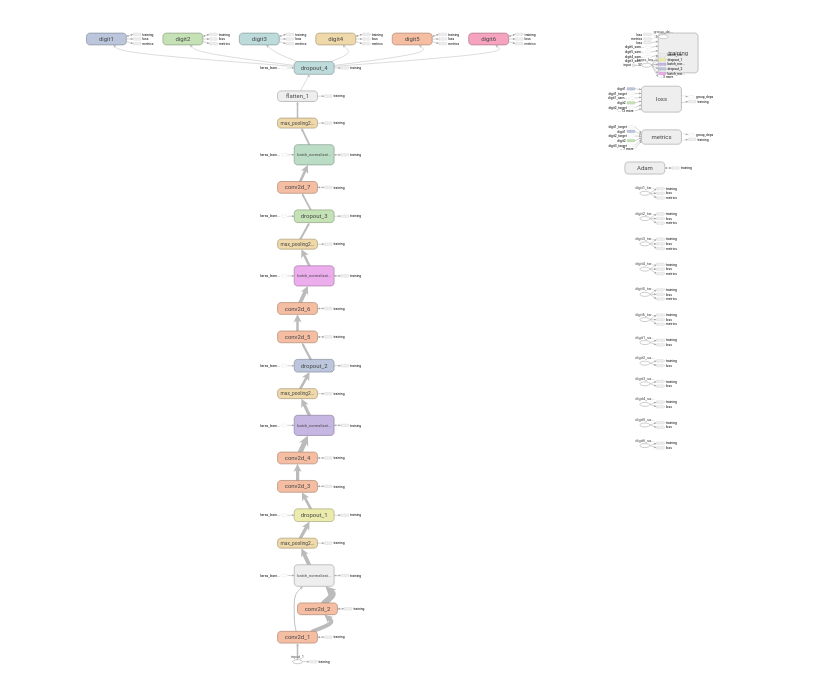}
	\caption{Model's graph as visualized in tensorboard (\ref{model_rc6L_1}).}
	\label{rc6_tb}
\end{figure}

We then checked how the same model performs with CAPTCHAs generated from the same dataset but this time with 6-character length labels. Here, the size and dimensions of the image are same as the previous one, the pre-processing done on the image is same too. We just changed some of the output nodes of the model and checked how this performed. The summary of the model is shown in Figure \ref{model_rc6L_summary}. The summary of the epochs is shown in Figure \ref{model_rc6L_epochs}. It takes about 83 seconds to complete one epoch and a total of 41.5 minutes to finish the training. The graph layout of the model is shown in Figure \ref{model_rc6L_1}. The same layout when visualized in tensorboard is shown in Figure \ref{rc6_tb}. The accuracy obtained is about 77\% and the plot is shown in Figure \ref{rc6Lm1-train-acc}. The trained model can be retrieved from this url (\url{https://jimut123.github.io/blogs/CAPTCHA/models/railway_captcha_6_imop.h5}). The plot for the loss for the same model is shown in Figure \ref{rc6Lm1-train-loss}. When the model is tested on real world unseen data, we get the prediction of \ref{rc6_predict}, which shows that the performance is not so well on real world data. The model predicts 25 images wrongly out of 36 images which is way less than human level operator.

\begin{figure}
	\centering

	{%
		\lstset{frame=single,basicstyle=\scriptsize,style=myModelSummaryStyle}
		\centering
		\begin{lstlisting}
			Layer (type)                    Output Shape         Param #     Connected to                     
			==================================================
			input_1 (InputLayer)            (None, 200, 60, 1)   0                                            
			__________________________________________________________________________________________________
			conv2d_1 (Conv2D)               (None, 200, 60, 32)  320         input_1[0][0]                    
			__________________________________________________________________________________________________
			conv2d_2 (Conv2D)               (None, 198, 58, 32)  9248        conv2d_1[0][0]                   
			__________________________________________________________________________________________________
			batch_normalization_1 (BatchNor (None, 198, 58, 32)  128         conv2d_2[0][0]                   
			__________________________________________________________________________________________________
			max_pooling2d_1 (MaxPooling2D)  (None, 99, 29, 32)   0           batch_normalization_1[0][0]      
			__________________________________________________________________________________________________
			dropout_1 (Dropout)             (None, 99, 29, 32)   0           max_pooling2d_1[0][0]            
			__________________________________________________________________________________________________
			conv2d_3 (Conv2D)               (None, 99, 29, 64)   18496       dropout_1[0][0]                  
			__________________________________________________________________________________________________
			conv2d_4 (Conv2D)               (None, 97, 27, 64)   36928       conv2d_3[0][0]                   
			__________________________________________________________________________________________________
			batch_normalization_2 (BatchNor (None, 97, 27, 64)   256         conv2d_4[0][0]                   
			__________________________________________________________________________________________________
			max_pooling2d_2 (MaxPooling2D)  (None, 48, 13, 64)   0           batch_normalization_2[0][0]      
			__________________________________________________________________________________________________
			dropout_2 (Dropout)             (None, 48, 13, 64)   0           max_pooling2d_2[0][0]            
			__________________________________________________________________________________________________
			conv2d_5 (Conv2D)               (None, 48, 13, 128)  73856       dropout_2[0][0]                  
			__________________________________________________________________________________________________
			conv2d_6 (Conv2D)               (None, 46, 11, 128)  147584      conv2d_5[0][0]                   
			__________________________________________________________________________________________________
			batch_normalization_3 (BatchNor (None, 46, 11, 128)  512         conv2d_6[0][0]                   
			__________________________________________________________________________________________________
			max_pooling2d_3 (MaxPooling2D)  (None, 23, 5, 128)   0           batch_normalization_3[0][0]      
			__________________________________________________________________________________________________
			dropout_3 (Dropout)             (None, 23, 5, 128)   0           max_pooling2d_3[0][0]            
			__________________________________________________________________________________________________
			conv2d_7 (Conv2D)               (None, 21, 3, 256)   295168      dropout_3[0][0]                  
			__________________________________________________________________________________________________
			batch_normalization_4 (BatchNor (None, 21, 3, 256)   1024        conv2d_7[0][0]                   
			__________________________________________________________________________________________________
			max_pooling2d_4 (MaxPooling2D)  (None, 10, 1, 256)   0           batch_normalization_4[0][0]      
			__________________________________________________________________________________________________
			flatten_1 (Flatten)             (None, 2560)         0           max_pooling2d_4[0][0]            
			__________________________________________________________________________________________________
			dropout_4 (Dropout)             (None, 2560)         0           flatten_1[0][0]                  
			__________________________________________________________________________________________________
			digit1 (Dense)                  (None, 36)           87074       dropout_4[0][0]                  
			__________________________________________________________________________________________________
			digit2 (Dense)                  (None, 36)           87074       dropout_4[0][0]                  
			__________________________________________________________________________________________________
			digit3 (Dense)                  (None, 36)           87074       dropout_4[0][0]                  
			__________________________________________________________________________________________________
			digit4 (Dense)                  (None, 36)           87074       dropout_4[0][0]                  
			__________________________________________________________________________________________________
			digit5 (Dense)                  (None, 36)           87074       dropout_4[0][0]                  
			__________________________________________________________________________________________________
			digit6 (Dense)                  (None, 36)           87074       dropout_4[0][0]                  
			==================================================
			Total params: 1,105,964
			Trainable params: 1,105,004
			Non-trainable params: 960
			__________________________________________________________________________________________________
		\end{lstlisting}
	}
	\caption{Model (\ref{model_rc6L_1}) Summary. }
	\label{model_rc6L_summary}
\end{figure}

\section{Railway-CAPTCHA with 7 Letters}

\begin{figure} 
	\centering
	\fbox{\includegraphics[height=24cm,width=1\linewidth]{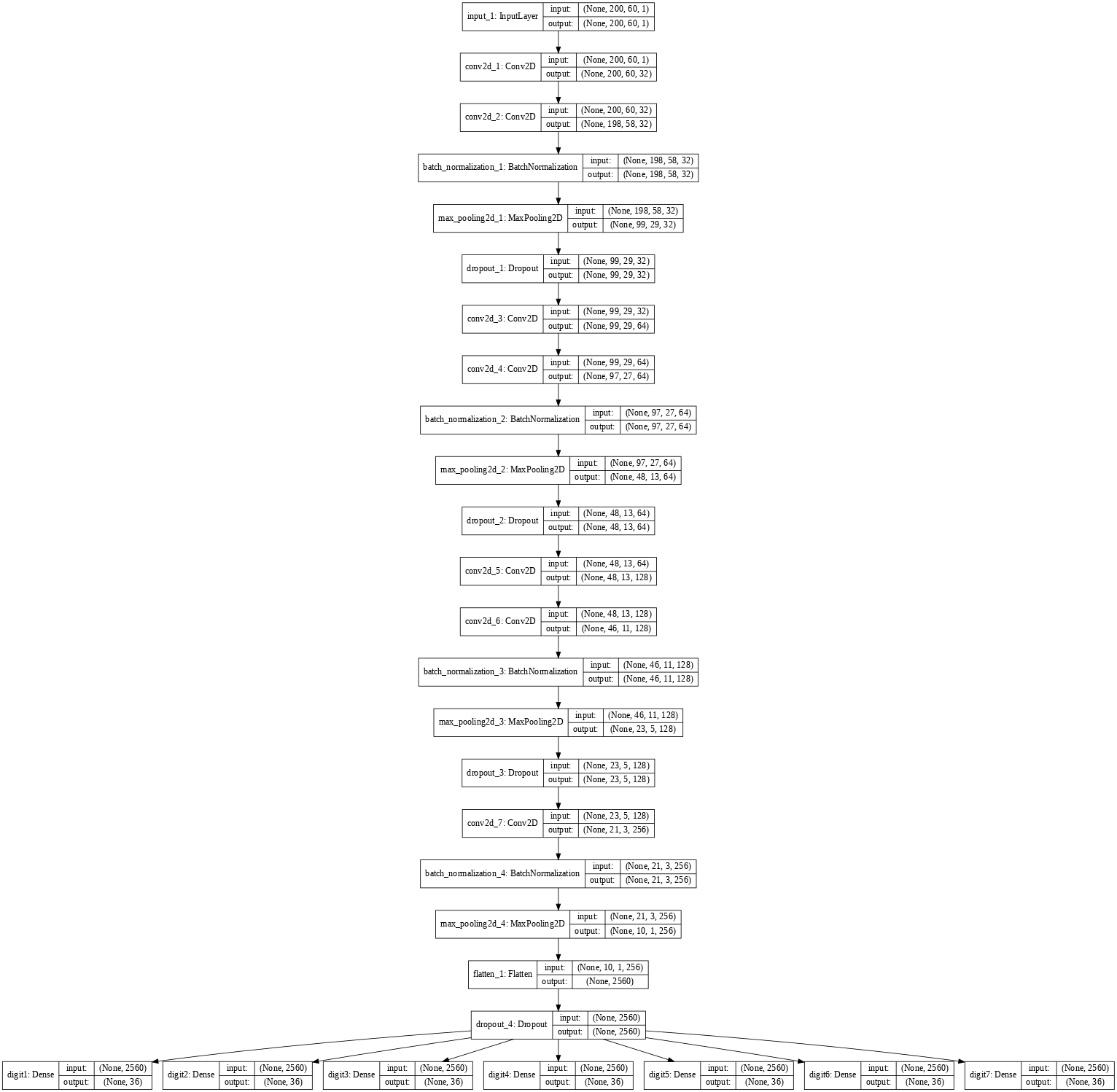}}	
	\caption{Our vanilla model for cracking 7-letter railway CAPTCHA.}
	\label{model_rc7L_1}
\end{figure}

\begin{figure}
	\centering
	\includegraphics[width=1\linewidth]{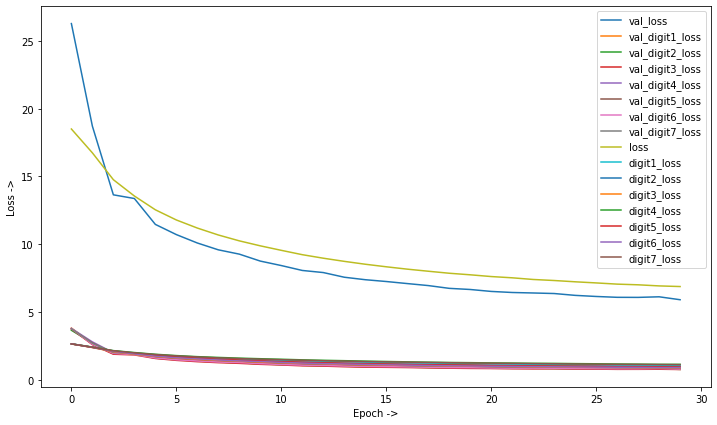}
	\caption{Training Loss for vanilla model (\ref{model_rc7L_1}).}
	\label{rc7Lm1-train-loss}
\end{figure}

\begin{figure}
	\centering
	\includegraphics[width=1\linewidth]{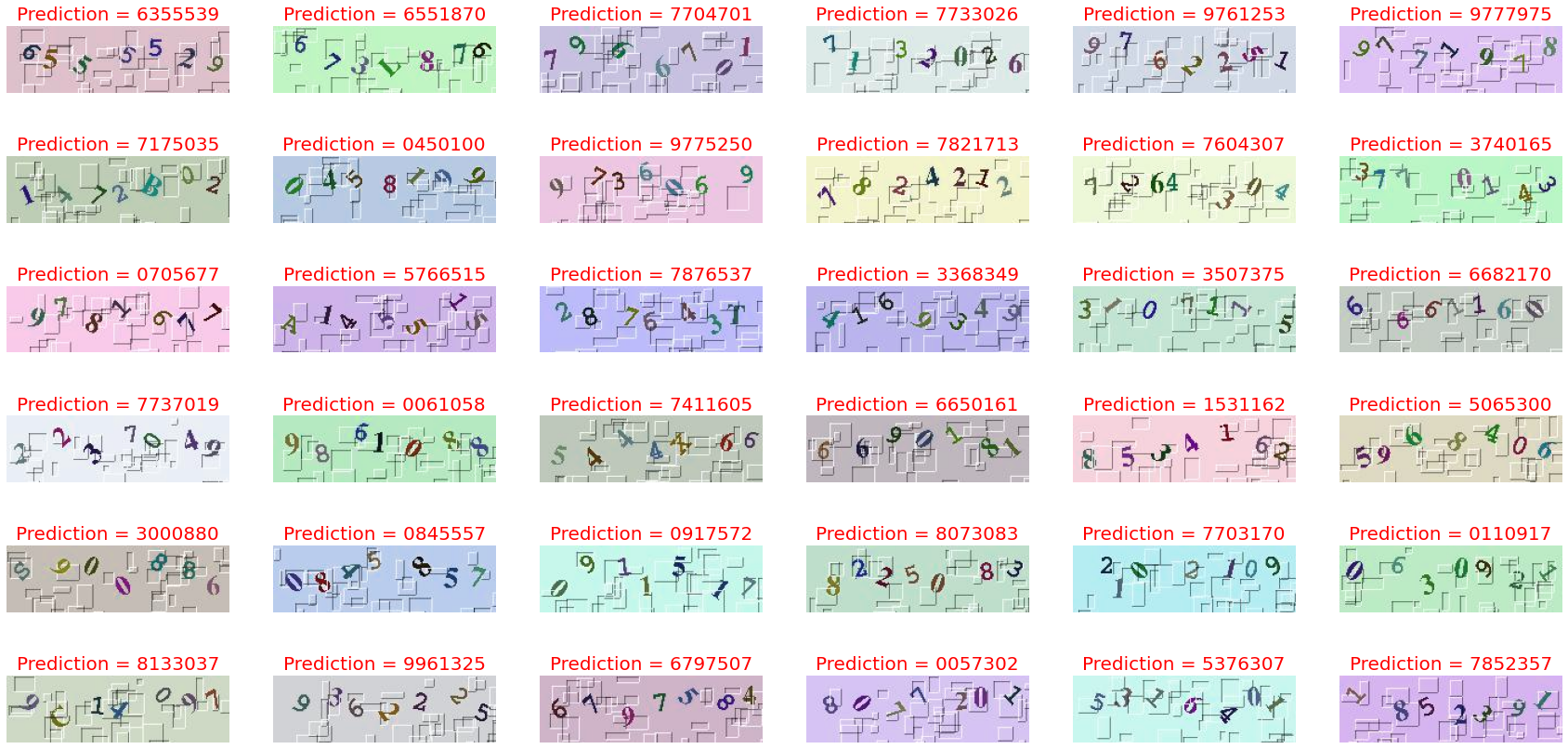}
	\caption{Predictions of the model (\ref{model_rc7L_1}).}
	\label{rc7_pred}
\end{figure}

We then checked how the same model performs with CAPTCHAs generated from the same dataset but this time with 7-character length labels. Here, the size and dimensions of the image are same as the previous one, the pre-processing done on the image is same too. We just changed some of the output nodes of the model and checked how this performed. The summary of the model is shown in Figure \ref{model_rc7L_summary}. The summary of the epochs is shown in Figure \ref{model_rc7L_epochs}. It takes about 91 seconds to complete one epoch and a total of 45.5 minutes to finish the training. The graph layout of the model is shown in Figure \ref{model_rc7L_1}. The accuracy obtained is about 75\% and the plot is shown in Figure \ref{rc6Lm1-train-acc}. The trained model can be retrieved from this url (\url{https://jimut123.github.io/blogs/CAPTCHA/models/railway_captcha_7_imop.h5}). The plot for the loss for the same model is shown in Figure \ref{rc7Lm1-train-loss}. When the model is tested on real world unseen data, we get the prediction of \ref{rc7_pred}, which shows that the performance is not at all well on real world data. The model predicts 36 images wrongly out of 36 images which is naive.

\begin{figure}
	\centering
	
	{%
		\lstset{frame=single,basicstyle=\scriptsize,style=myModelSummaryStyle}
		\centering
		\begin{lstlisting}
		__________________________________________________________________________________________________
		Layer (type)                    Output Shape         Param #     Connected to                     
		==================================================
		input_1 (InputLayer)            (None, 200, 60, 1)   0                                            
		__________________________________________________________________________________________________
		conv2d_1 (Conv2D)               (None, 200, 60, 32)  320         input_1[0][0]                    
		__________________________________________________________________________________________________
		conv2d_2 (Conv2D)               (None, 198, 58, 32)  9248        conv2d_1[0][0]                   
		__________________________________________________________________________________________________
		batch_normalization_1 (BatchNor (None, 198, 58, 32)  128         conv2d_2[0][0]                   
		__________________________________________________________________________________________________
		max_pooling2d_1 (MaxPooling2D)  (None, 99, 29, 32)   0           batch_normalization_1[0][0]      
		__________________________________________________________________________________________________
		dropout_1 (Dropout)             (None, 99, 29, 32)   0           max_pooling2d_1[0][0]            
		__________________________________________________________________________________________________
		conv2d_3 (Conv2D)               (None, 99, 29, 64)   18496       dropout_1[0][0]                  
		__________________________________________________________________________________________________
		conv2d_4 (Conv2D)               (None, 97, 27, 64)   36928       conv2d_3[0][0]                   
		__________________________________________________________________________________________________
		batch_normalization_2 (BatchNor (None, 97, 27, 64)   256         conv2d_4[0][0]                   
		__________________________________________________________________________________________________
		max_pooling2d_2 (MaxPooling2D)  (None, 48, 13, 64)   0           batch_normalization_2[0][0]      
		__________________________________________________________________________________________________
		dropout_2 (Dropout)             (None, 48, 13, 64)   0           max_pooling2d_2[0][0]            
		__________________________________________________________________________________________________
		conv2d_5 (Conv2D)               (None, 48, 13, 128)  73856       dropout_2[0][0]                  
		__________________________________________________________________________________________________
		conv2d_6 (Conv2D)               (None, 46, 11, 128)  147584      conv2d_5[0][0]                   
		__________________________________________________________________________________________________
		batch_normalization_3 (BatchNor (None, 46, 11, 128)  512         conv2d_6[0][0]                   
		__________________________________________________________________________________________________
		max_pooling2d_3 (MaxPooling2D)  (None, 23, 5, 128)   0           batch_normalization_3[0][0]      
		__________________________________________________________________________________________________
		dropout_3 (Dropout)             (None, 23, 5, 128)   0           max_pooling2d_3[0][0]            
		__________________________________________________________________________________________________
		conv2d_7 (Conv2D)               (None, 21, 3, 256)   295168      dropout_3[0][0]                  
		__________________________________________________________________________________________________
		batch_normalization_4 (BatchNor (None, 21, 3, 256)   1024        conv2d_7[0][0]                   
		__________________________________________________________________________________________________
		max_pooling2d_4 (MaxPooling2D)  (None, 10, 1, 256)   0           batch_normalization_4[0][0]      
		__________________________________________________________________________________________________
		flatten_1 (Flatten)             (None, 2560)         0           max_pooling2d_4[0][0]            
		__________________________________________________________________________________________________
		dropout_4 (Dropout)             (None, 2560)         0           flatten_1[0][0]                  
		__________________________________________________________________________________________________
		digit1 (Dense)                  (None, 36)           92196       dropout_4[0][0]                  
		__________________________________________________________________________________________________
		digit2 (Dense)                  (None, 36)           92196       dropout_4[0][0]                  
		__________________________________________________________________________________________________
		digit3 (Dense)                  (None, 36)           92196       dropout_4[0][0]                  
		__________________________________________________________________________________________________
		digit4 (Dense)                  (None, 36)           92196       dropout_4[0][0]                  
		__________________________________________________________________________________________________
		digit5 (Dense)                  (None, 36)           92196       dropout_4[0][0]                  
		__________________________________________________________________________________________________
		digit6 (Dense)                  (None, 36)           92196       dropout_4[0][0]                  
		__________________________________________________________________________________________________
		digit7 (Dense)                  (None, 36)           92196       dropout_4[0][0]                  
		==================================================
		Total params: 1,228,892
		Trainable params: 1,227,932
		Non-trainable params: 960
		__________________________________________________________________________________________________
		\end{lstlisting}
	}
	\caption{Model (\ref{model_rc7L_1}) Summary. }
	\label{model_rc7L_summary}
\end{figure}

\begin{figure} 
	\centering
	{%
		\lstset{frame=single,basicstyle=\scriptsize,style=myModelSummaryStyle}
		\centering
		\begin{lstlisting}
		Epoch 25/30
		72000/72000 [==============================] - 91s 1ms/step - loss: 7.2219 - digit1_loss: 1.1113 - digit2_loss: 0.9235 - digit3_loss: 0.9135 - digit4_loss: 1.1867 - digit5_loss: 0.9920 - digit6_loss: 0.9389 - digit7_loss: 1.1559 - digit1_acc: 0.6421 - digit2_acc: 0.7140 - digit3_acc: 0.7157 - digit4_acc: 0.6182 - digit5_acc: 0.6900 - digit6_acc: 0.7096 - digit7_acc: 0.6248 - val_loss: 6.2208 - val_digit1_loss: 0.9165 - val_digit2_loss: 0.8048 - val_digit3_loss: 0.7847 - val_digit4_loss: 1.0185 - val_digit5_loss: 0.8535 - val_digit6_loss: 0.8239 - val_digit7_loss: 1.0190 - val_digit1_acc: 0.7541 - val_digit2_acc: 0.7892 - val_digit3_acc: 0.7947 - val_digit4_acc: 0.7192 - val_digit5_acc: 0.7763 - val_digit6_acc: 0.7780 - val_digit7_acc: 0.7152
		\end{lstlisting}
	}
	\caption{Summary for model's (\ref{model_rc7L_1}) epochs}
	\label{model_rc7L_epochs}
\end{figure}

\begin{figure}
	\centering
	\includegraphics[width=1\linewidth]{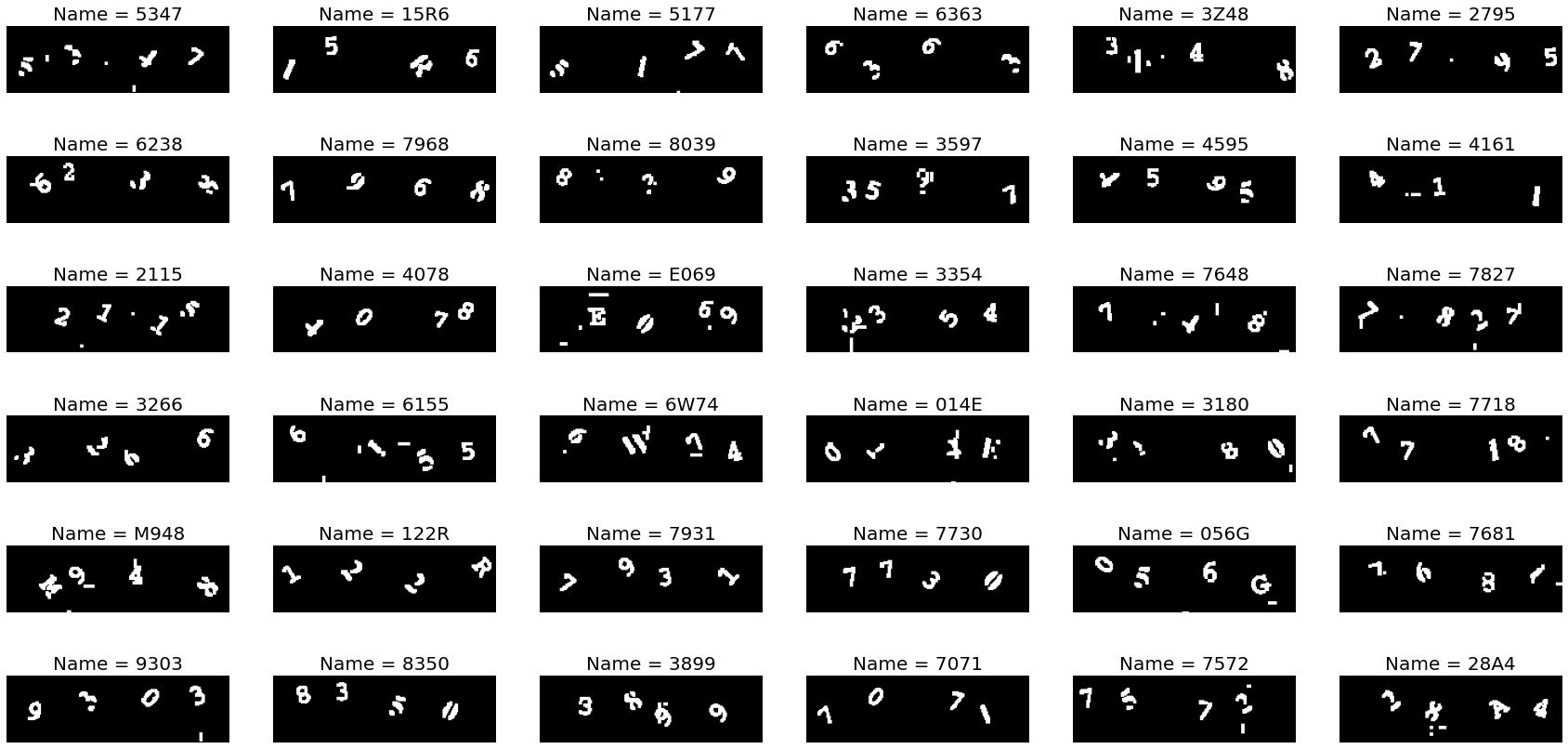}
	\caption{A subset of samples given to the model (\ref{model_rc4L_1})}
	\label{rc4_input_feed}
\end{figure}

\begin{figure}
	\centering
	\includegraphics[width=1\linewidth]{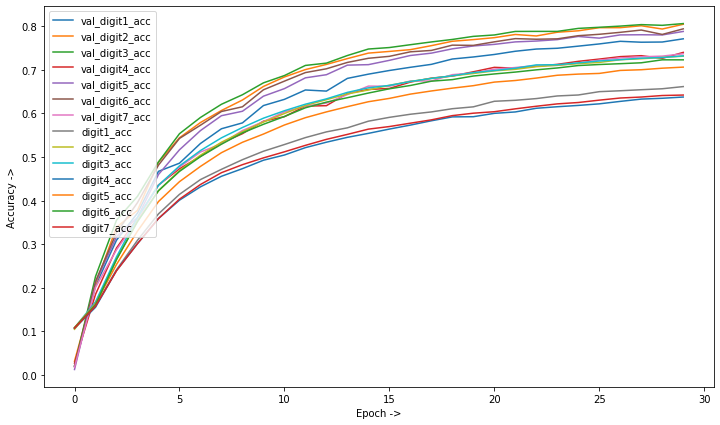}
	\caption{Training Accuracy for vanilla model (\ref{model_rc7L_1}).}
	\label{rc7Lm1-train-acc}
\end{figure}

\section{What possibly went wrong?}

We can see that the sample of images when preprocessed, before passing to the model has a lot of clutter in them as shown in Figure \ref{rc4_input_feed}. From studies \cite{Su_2019} we know that even a single pixel can change the output of the prediction by a vast amount. There were more to the existing problem of clutter and occlusion. When we saw the distribution of the dataset's character labels, we found that though it was generated randomly, it followed a certain trend. The labels for the dataset for 3 letters have characters 0-9 ten times more often than the other letter ranging from A to Z. Similarly, for other datasets. This might be due to the anomaly in clock speed and internal problem of pipelining and parallel execution of CPU of the machine in which it was generated. This leads to the bias for not learning the A-Z characters as good as 0-9 letters. So, we need to generate a brute force algorithm for generation of uniformly distributed character data. We first thought of brute forcing the labels and generating data like the output produced by the famous software crunch for brute forcing passwords, but it is too naive and the number of labels would increase exponentially with the number of characters in the labels. We can think of an algorithm which will initialize the distribution and select the labels randomly deducting each label generated from the initial list of the labels.

\begin{figure}[h]
	\centering
	\fbox{\includegraphics[width=0.85\linewidth]{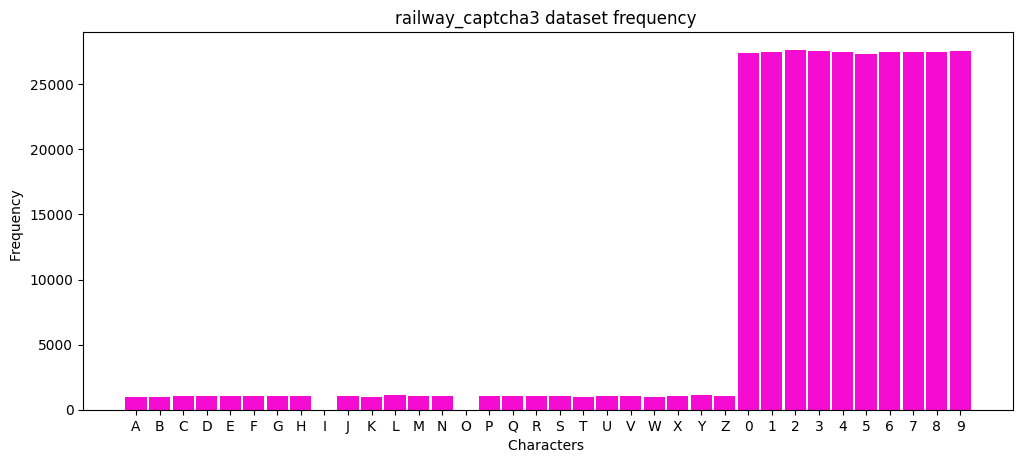}}\vspace{6px}\hspace{2px}
	\fbox{\includegraphics[width=0.85\linewidth]{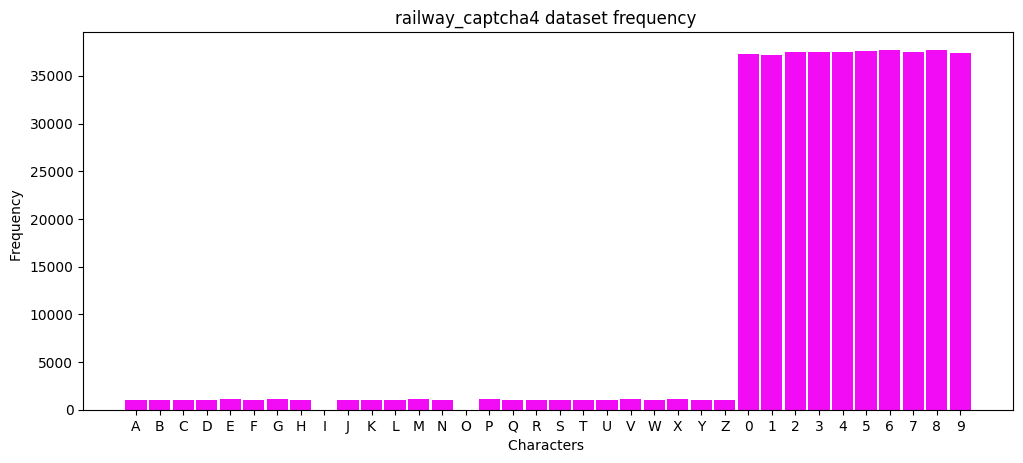}}\vspace{6px}
	\fbox{\includegraphics[width=0.85\linewidth]{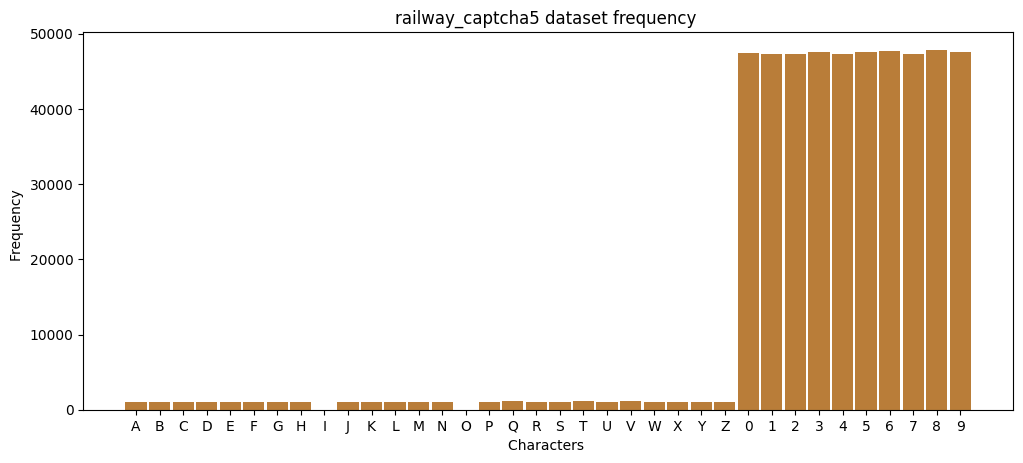}}\vspace{6px}\hspace{2px}	
	\caption{Distribution of the data for the various labels of the Railway CAPTCHA dataset for (3,4 and 5 letter images).}
	\label{rc_datadist_1}
\end{figure}
\begin{figure}[h]
	\centering
	\fbox{\includegraphics[width=0.85\linewidth]{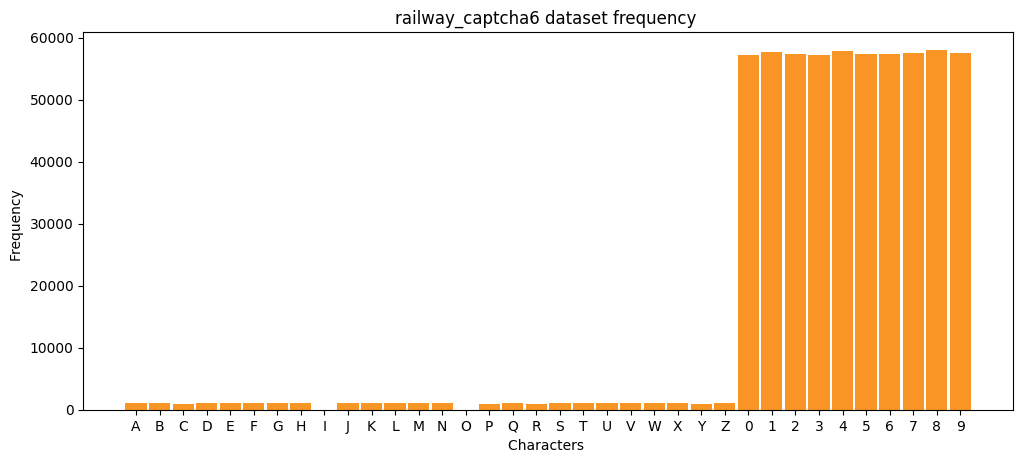}}\vspace{6px}
	\fbox{\includegraphics[width=0.85\linewidth]{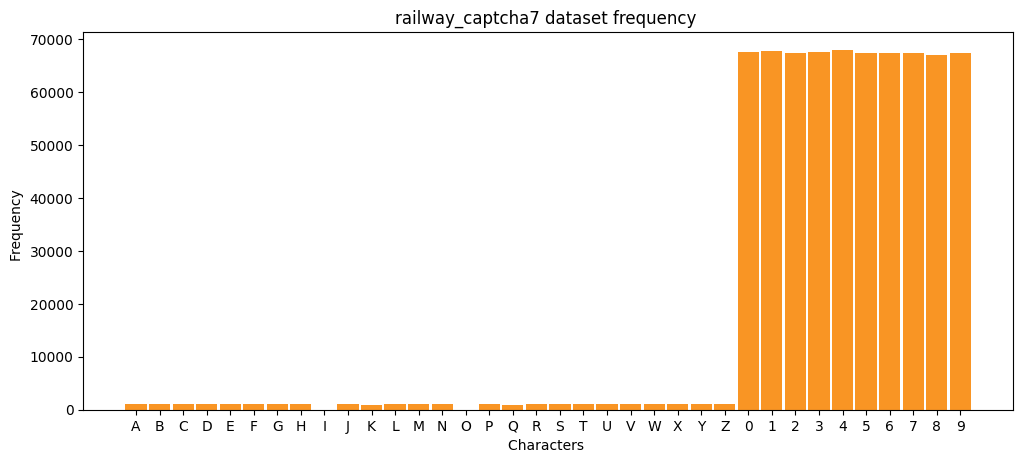}}
	\caption{Distribution of the data for the various labels of the Railway CAPTCHA dataset (for 6 and 7 letter images).}
	\label{rc_datadist_2}
\end{figure}

\section{What can we conclude from this?}

\begin{figure}[]
	\centering
	\includegraphics[width=0.75\linewidth]{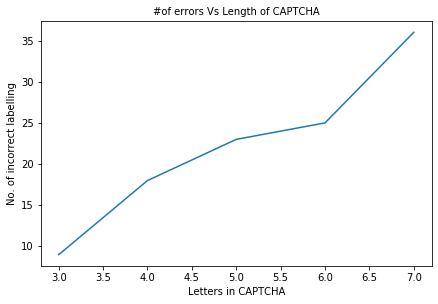}
	\caption{The complexity of the model increases with the length of the CAPTCHA, data from table (\ref{table_length_complexity}).}
	\label{length_complexity}
\end{figure}

It is sure from the behavior of the model that as the number of letters in the CAPTCHA increases, the complexity of the CAPTCHA increases. This is good for designing better CAPTCHAs since for humans, it is very easy task to solve long paragraphs of simple writing than to solve few sentences of obfuscated writing. The overall accuracy of the model might decrease significantly as shown in Table \ref{table_length_complexity} and Figure \ref{length_complexity}. The figure and the information in the table shows that the complexity of the model increases linearly with the increase in number of letters in CAPTCHA. In a multi label classification this reduces drastically, since the complexity of the model would also increase. In case of multi-class classification where each character is segmented and predicted, it will also become difficult for the machine learning algorithm to solve and decipher large texts of huge data. Since the CAPTCHA is said to be correct when each and every letter of the CAPTCHA is correctly classified, in any case each segmented character receives an accuracy of classification of about 95 \% then if there are about 10 letter then the total accuracy of correct classification becomes (95)\textsuperscript{10} which is 59.87\%. This shows that the complexity for CAPTCHAs to the computers can be increased significantly. So, we conjecture that long length CAPTCHA are better solution to hard games which are not easy to understand at first glance, and takes a reasonable amount of time in determining the moves for genuine users.

\begin{table}[]
	\centering
	\begin{tabular}{|c|c|c|}
		\hline
		\textit{\textbf{No. of letters}} & \textit{\textbf{No. of incorrect labelling}} & \textit{\textbf{Accuracy (in \%)}} \\ \hline
		\textit{3} & \textit{9}  & \textit{80} \\ \hline
		\textit{4} & \textit{18} & \textit{79} \\ \hline
		\textit{5} & \textit{23} & \textit{77} \\ \hline
		\textit{6} & \textit{25} & \textit{77} \\ \hline
		\textit{7} & \textit{36} & \textit{75} \\ \hline
	\end{tabular}
	\caption{Table showing that the length of the CAPTCHA is directly proportional to the complexity.}
	\label{table_length_complexity}
\end{table}

\chapter{Conclusions}

This investigation resulted in successfully breaking most of the text CAPTCHA's with human level accuracy. This means that the text CAPTCHA's are not so much reliable to predict whether a user is a human or not. We have also shown that if we increase the size of the length of labels in the CAPTCHA, then it might be difficult for computers to break them with human level accuracy. When we want to have state of the art accuracy in deciphering complex CAPTCHAs with multi-label classification, then it is necessary to have a large number of data \cite{Sontag98vcdimension}. If the number of parameters in the neural network is $P$, then it is always a good rule of thumb to have $P^2$ amount of data points. We have also seen that when a simple model can give state of the art accuracy as compared to a complex model, we will stick to a simple model since that will give a good generalization of the data without overfitting much (by Occam's razor). It is also good to visualize what we are trying to achieve at every step to get to the point where we can do better, to give us direction towards achieving good accuracy with models.

 We have also seen that in case of training deep neural networks, data augmentation is necessary since it might not be the case that we are getting so much amount of data, and performing some basic image processing tasks to augment the data always helps. It is also known that there is no convergence proof for deep neural networks, so it is astonishing that they perform so well when the amount of data increases and sometimes performs better than humans. The successful completion of this project will enhance the way raw text image is classified for data retrieval from web search queries. Converting images containing text to actual text can give rise to efficient algorithm for web search queries, where it is difficult to search for images by certain text when they are unlabeled. 
 
 It is also seen that dropouts helps in training ensemble of networks which helps in reducing noise and generalizing well on unseen data. The main motive of the project was to create a generic CAPTCHA solver, and the AlexNet type model with multilabel classification and a dropout before the final layer helps to generalize well on the unseen data, which gives almost human level accuracy. We have also seen that when the data is less, we need to feed the whole of the dataset with augmentations, if necessary, we need to give 3 channels if there is a lot of noise in the image data. This will help the network learn the important features all by itself.
 
 It is well known that text CAPTCHA is getting away and some new types of creative CAPTCHAs are coming day by day. We have seen that there has been gaming CAPTCHA, puzzle CAPTCHA, voice CAPTCHA as shown in  Figure \ref{CAPTCHA_types1}. These types of CAPTCHAs are more challenging to solve than text CAPTCHA and particular puzzle needs particular type of architecture to solve. Google reCAPTCHA v3 uses advanced cookie system to check if the user is a bot or not. These types of techniques to determine the authenticity of users are emerging which induces less pain to solve and crack CAPTCHA every day. So, indeed text-based CAPTCHA is a week type of CAPTCHA.
 
 \section{Future Work}
 
 This project, though is the beginning of obfuscated Optical Character Recognition, can be well suited in future for the digital conversion of large corpus of Handwritten datasets. One example of such a dataset can be IAM Handwriting Database. Though the claim of future work will be highly dramatic like classifying a huge corpus of text by segmentation procedures and converting each digital image found on internet to its corresponding text which will help in information retrieval and segregating data in their own form. It can also be used to segment huge database of handwritten documents which are not yet converted to digital text due to lack of human resources. Another typical future work will be to find text in unconstrained environments which will help to detect license plate with high accuracy. It may also be used for converting handwritten test papers to digit copy for preservation, and maybe some AI to auto check all the test papers in some huge test. Converting handwritten text to digital text is one of the most challenging tasks in computer vision since the data needs to be generalized on a wide variety of lighting conditions at different scales and a lot of different types of handwriting. This makes the problem even harder, since it is not possible to get all the handwriting styles for such a wide diversity of people.

%%%%%%%%%%%%%%%%%%%%%%%%%%%%%%%%%%%
% For index creation
\printindex[\idxKeywordName]

%%%%%%%%%%%%%%%%%%%%%%%%%%%%%%%%%%%%%%%

\appendix 
\chapter{Appendix}

\section{Softwares Used}

We have extensively used Keras, which is open source library capable of running on top of tensorflow. We have used tensorflow as backend. We have used matplotlib and plotly for generating graphs and plots used in this documentation. We have used Google's Colaboratory's GPU power extensively. We have created the projects and dataset in such a way that they can be easily run in cloud. There were various animations that were generated during this project. We have used figshare to host the data, so that they can be easily fetched from cloud without any token or certificates. We have also used \url{https://alexlenail.me/NN-SVG/} to create the models in 3D. The last picture in the thesis is captured by me.

\section{Dataset and Model generated}

Here, we will share the datasets and model that were generated and created by us during the course of this project.

\textbf{Captcha (4 letter)} : dataset can be found here \url{https://jimut123.github.io/blogs/CAPTCHA/data/captcha_4_letter.tar.gz} and the model for this dataset can be obtained from here \url{https://jimut123.github.io/blogs/CAPTCHA/models/captcha_4_letter_our_model.h5}.

\textbf{c4l\_16x16\_550}: dataset can be found here \url{https://jimut123.github.io/blogs/CAPTCHA/data/c4l-16x16_550.tar.gz} and the model for this dataset can be obtained from here \url{https://jimut123.github.io/blogs/CAPTCHA/models/CNN_c4l-16x16_550.h5}.

\textbf{resized\_JAM}: dataset can be found here \url{https://jimut123.github.io/blogs/CAPTCHA/data/resized_JAM.tar.gz}. The model used was k-NN and it can be easily formed without any computationally expensive training.

\textbf{CAPTCHA-version-2}: dataset can be found here \url{https://jimut123.github.io/blogs/CAPTCHA/data/captcha_v2.tar.gz} and the model for this dataset can be obtained from here \url{https://jimut123.github.io/blogs/CAPTCHA/models/captcha_v2_alexNet.h5} and here \url{https://jimut123.github.io/blogs/CAPTCHA/models/captcha_v2_1_drive.h5}.

\textbf{100000-labeled-captchas}: dataset can be found here \url{https://dx.doi.org/10.6084/m9.figshare.12046881.v1} and the model for this dataset can be obtained from here \url{https://jimut123.github.io/blogs/CAPTCHA/models/images-1L-processed.h5}.

\textbf{Faded CAPTCHA data}. The faded CAPTCHA dataset is comprised of about 3.7 GiB data files in total. It is partitioned into folders and uploaded to figshare. The dataset can be found from these links:

\url{https://dx.doi.org/10.6084/m9.figshare.12115656},\\
\url{https://dx.doi.org/10.6084/m9.figshare.12115671},\\
\url{https://dx.doi.org/10.6084/m9.figshare.12115689},\\
\url{https://dx.doi.org/10.6084/m9.figshare.12118302},\\
\url{https://dx.doi.org/10.6084/m9.figshare.12118371},\\
\url{https://dx.doi.org/10.6084/m9.figshare.12118482},\\
\url{https://dx.doi.org/10.6084/m9.figshare.12118836},\\
\url{https://dx.doi.org/10.6084/m9.figshare.12122853},\\
\url{https://dx.doi.org/10.6084/m9.figshare.12122934}, and \\
\url{https://dx.doi.org/10.6084/m9.figshare.12122967}. The model obtained from the dataset can be found from here \url{https://jimut123.github.io/blogs/CAPTCHA/models/faded_captcha.h5}.

\textbf{Railway CAPTCHA} \\
\textbf{Railway CAPTCHA 3 Letters}: dataset can be found from \url{https://dx.doi.org/10.6084/m9.figshare.12045249.v1}, and the model for this dataset can be obtained from here \url{https://jimut123.github.io/blogs/CAPTCHA/models/railway_captcha_3.h5}.

\textbf{Railway CAPTCHA 4 Letters}: dataset can be found from \url{https://dx.doi.org/10.6084/m9.figshare.12045288.v1}, and the model for this dataset can be obtained from here \url{https://jimut123.github.io/blogs/CAPTCHA/models/railway_captcha_4.h5}.

\textbf{Railway CAPTCHA 5 Letters}: dataset can be found from
\url{https://dx.doi.org/10.6084/m9.figshare.12045294.v1}, and the model for this dataset can be obtained from here \url{https://jimut123.github.io/blogs/CAPTCHA/models/railway_captcha_5.h5}.

\textbf{Railway CAPTCHA 6 Letters}: dataset can be found from
\url{https://dx.doi.org/10.6084/m9.figshare.12045657.v1}, and the model for this dataset can be obtained from here \url{https://jimut123.github.io/blogs/CAPTCHA/models/railway_captcha_6.h5}.

\textbf{Railway CAPTCHA 7 Letters}: dataset can be found from \url{https://dx.doi.org/10.6084/m9.figshare.12053937.v1}, and the model for this dataset can be obtained from here \url{https://jimut123.github.io/blogs/CAPTCHA/models/railway_captcha_7.h5}.

\textbf{Circle CAPTCHA}: dataset can be found here \url{https://figshare.com/articles/captcha_gen/12286766}  and the model for this dataset can be obtained from here (\url{https://jimut123.github.io/blogs/CAPTCHA/models/circle_captcha_10e.h5}).

\textbf{Sphinx CAPTCHA (4 letters)}: There are 34 pieces of tar.gz files that are found over figshare and zenodo (due to some problem during the CoviD-19 crisis in figshare). The whole dataset can be found from here:  

\url{https://ndownloader.figshare.com/files/22336164},\\
\url{https://ndownloader.figshare.com/files/22344408},\\
\url{https://ndownloader.figshare.com/files/22366674},\\
\url{https://ndownloader.figshare.com/files/22367007},\\
\url{https://ndownloader.figshare.com/files/22388370},\\
\url{https://ndownloader.figshare.com/files/22388556},\\
\url{https://ndownloader.figshare.com/files/22399503},\\
\url{https://ndownloader.figshare.com/files/22399554},\\
\url{https://ndownloader.figshare.com/files/22414872},\\
\url{https://zenodo.org/record/3782676/files/sphinx_010.tar.gz?download=1},\\
\url{https://ndownloader.figshare.com/files/22477673},\\
\url{https://ndownloader.figshare.com/files/22477823},\\
\url{https://zenodo.org/record/3782690/files/sphinx_013.tar.gz?download=1},\\
\url{https://zenodo.org/record/3782704/files/sphinx_014.tar.gz?download=1},\\
\url{https://zenodo.org/record/3782733/files/sphinx_015.tar.gz?download=1},\\
\url{https://zenodo.org/record/3783674/files/sphinx_016.tar.gz?download=1},\\
\url{https://zenodo.org/record/3783691/files/sphinx_017.tar.gz?download=1},\\
\url{https://zenodo.org/record/3785093/files/sphinx_018.tar.gz?download=1},\\
\url{https://ndownloader.figshare.com/files/22533749},\\
\url{https://ndownloader.figshare.com/files/22533821},\\
\url{https://ndownloader.figshare.com/files/22533965},\\
\url{https://ndownloader.figshare.com/files/22534058},\\
\url{https://ndownloader.figshare.com/files/22542704},\\
\url{https://ndownloader.figshare.com/files/22542782},\\
\url{https://ndownloader.figshare.com/files/22542851},\\
\url{https://ndownloader.figshare.com/files/22542944},\\
\url{https://ndownloader.figshare.com/files/22543832},\\
\url{https://ndownloader.figshare.com/files/22543850},\\
\url{https://ndownloader.figshare.com/files/22543979},\\
\url{https://ndownloader.figshare.com/files/22584218},\\
\url{https://ndownloader.figshare.com/files/22585160},\\
\url{https://ndownloader.figshare.com/files/22621634},\\
\url{https://ndownloader.figshare.com/files/22621694}, and\\
\url{https://ndownloader.figshare.com/files/22624358}.
The total dataset comprise of about 12.2 GB 
and the model for this dataset can be obtained from here 
(\url{https://jimut123.github.io/blogs/CAPTCHA/models/sphinx_full_34_15e_9962.h5}).

\textbf{Fish eye CAPTCHA}: dataset can be found from \url{https://figshare.com/articles/Fish_Eye/12235946}. The dataset contains 8 pieces of zip folders which can be further extracted to obtain the actual dataset.  The trained model can be obtained from the following link (\url{https://jimut123.github.io/blogs/CAPTCHA/models/fish_eye.h5}).

\textbf{Mini CAPTCHA}: dataset can be found from \url{https://figshare.com/articles/Mini_Captcha/12286697}. The dataset contains 8 pieces of zip folders which can be further extracted to obtain the actual dataset.  The trained model can be obtained from the following link (\url{https://drive.google.com/open?id=1w5BhaeSvc1LYQfUuvwBApeNbcx0vkOUP}).

\textbf{Multicolor CAPTCHA}: dataset can be found from 
\url{https://ndownloader.figshare.com/files/22307958},\\
\url{https://ndownloader.figshare.com/files/22318224},\\
\url{https://ndownloader.figshare.com/files/22335879},\\
\url{https://ndownloader.figshare.com/files/22344393},\\
\url{https://ndownloader.figshare.com/files/22346499},\\
\url{https://ndownloader.figshare.com/files/22397457},\\
\url{https://ndownloader.figshare.com/files/22414434},\\
\url{https://ndownloader.figshare.com/files/22422816},\\
\url{https://ndownloader.figshare.com/files/22451057}, and \\
\url{https://ndownloader.figshare.com/files/22467227}. These contains 10 pieces of tar.gz folders which amount to about 12.7 GiB of tar.gz file. The model that was trained for 10 epochs can be found from this drive link \url{https://drive.google.com/open?id=1-2DW3CBkddRoUsBbaukf7OxR1MZpwVrn}. The size of the model is 996 MB.

%\textbf{}: dataset can be found here   and the model for this dataset can be obtained from here

\section{Accuracy obtained}

The accuracy obtained and their respective CAPTCHA name are described in this section.

\begin{table}
	\centering
	\begin{tabular}{|l|l|l|}
		\hline
		Dataset &
		Note &
		Accuracy \\ \hline
		captcha\_4\_letter &
		\begin{tabular}[c]{@{}l@{}}This is the cleaned \\ version of the dataset obtained \\ from Kaggle, Size = 10.4 MiB, \\ 9955 files of 72x24.\end{tabular} &
		99.81 \% \\ \hline
		c4l-16x16\_550 &
		\begin{tabular}[c]{@{}l@{}}Extracted characters from \\ Captcha\_4\_letter dataset, \\ Size = 1.4 MiB, \\ 17600 files of 16x16.\end{tabular} &
		99.91 \% \\ \hline
		JAM, resized\_JAM &
		\begin{tabular}[c]{@{}l@{}}JAM has 308 files of 110x40,\\ Size = 101.7KiB.\\ resized\_JAM has 1065 files,\\ 65.7KiB\end{tabular} &
		99.53 \% \\ \hline
		CAPTCHA-version-2 &
		\begin{tabular}[c]{@{}l@{}}1070 files of 200x50,\\ Size = 21.5 Mib.\end{tabular} &
		90.102\% \\ \hline
		
		Circle CAPTCHA &
		\begin{tabular}[c]{@{}l@{}}222K files of 35x120,\\ Size = 437.58 Mib.\end{tabular} &
		99.99\% \\ \hline
		
		Sphinx CAPTCHA &
		\begin{tabular}[c]{@{}l@{}}1 Million files of 437.58,\\ Size = 12.2 GiB.\end{tabular} &
		99.62\% \\ \hline
		
		Fish-eye CAPTCHA &
		\begin{tabular}[c]{@{}l@{}}2 Million files of 50x200,\\ Size = 9.84 GB.\end{tabular} &
		99.46\% \\ \hline
		
		Mini CAPTCHA &
		\begin{tabular}[c]{@{}l@{}}1 Million files of 368x159x3,\\ Size = 5.58 GB.\end{tabular} &
		97.25\% \\ \hline
		
		Multicolor CAPTCHA &
		\begin{tabular}[c]{@{}l@{}}1 Million files of 375x113x3,\\ (used only 50\% of the dataset )\\ Size = 21.5 Mib.\end{tabular} &
		95.69\% \\ \hline
		
		faded &
		\begin{tabular}[c]{@{}l@{}} Size = 3.7 GiB, \\ 604800 files of 100x120x3 .\end{tabular} &
		99.44 \% \\ \hline
		1L-labeled-captchas &
		\begin{tabular}[c]{@{}l@{}}109,053 files of 200x50. \\ Size = 215.1 MiB\end{tabular} &
		99.67\% \\ \hline
		Railway CAPTCHA  &
		\begin{tabular}[c]{@{}l@{}}100K files of 200x60, Sizes =\\ 241.8 MiB, 253.8 MiB, 265.7MiB,\\ 277.4 MiB, 289.1 MiB\end{tabular} &
		\begin{tabular}[c]{@{}l@{}}80 \%\\ 79 \%\\ 77 \%\\ 77 \%\\ 75 \%\end{tabular} \\ \hline
	\end{tabular}
\end{table}

\bibliography{bigone}  % change this to the name of your bib-file

\bibliographystyle{latexeu}

\clearpage

%\newpage

\includepdf[scale=1,pages=1]{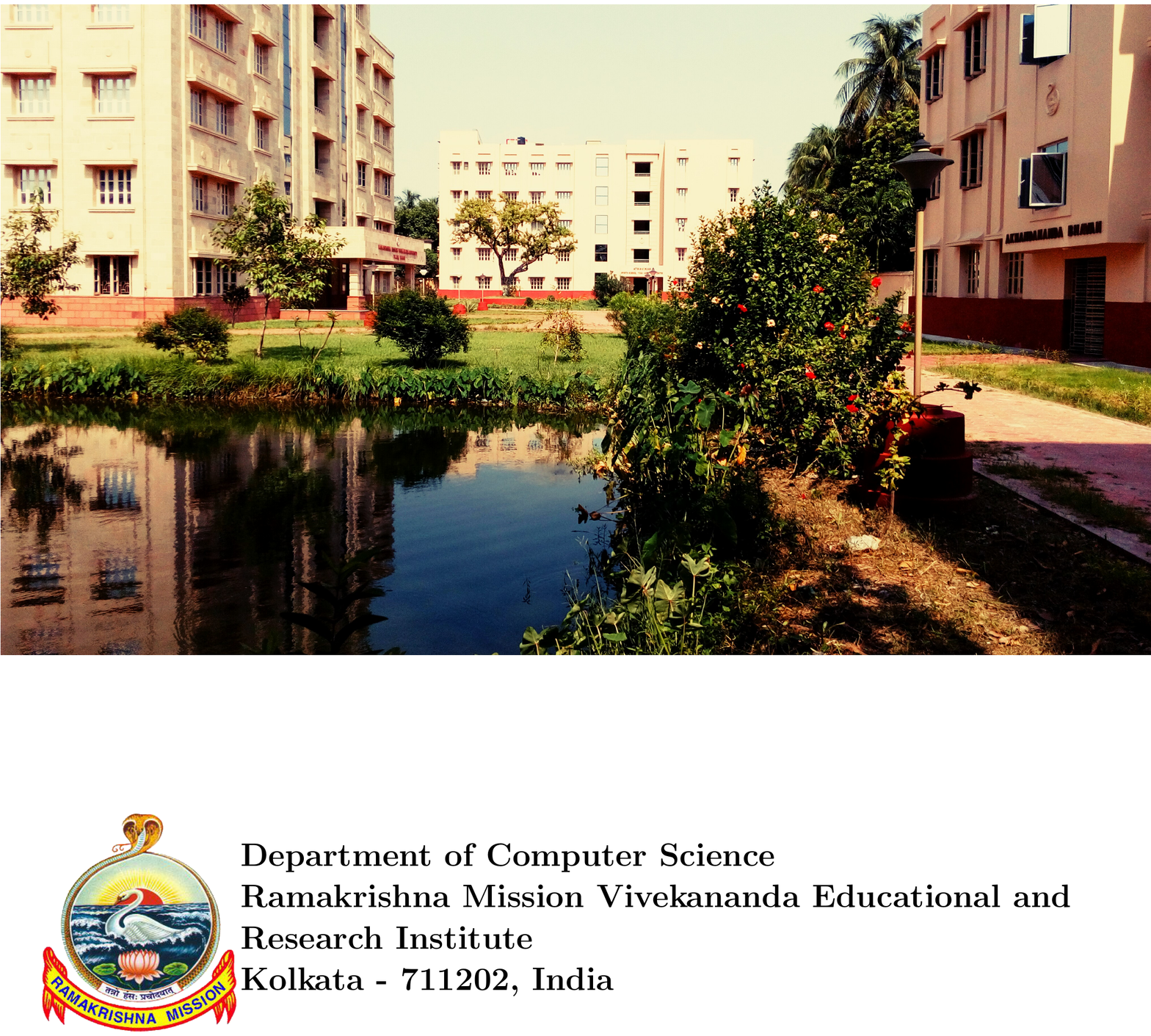}

\end{document}